\documentclass[10pt, a4paper, copyright, gdm]{google}

\usepackage{float}
\usepackage{makecell}        
\usepackage{fontawesome5}  
\usepackage{xspace}
\usepackage{multicol}
\usepackage{multirow}
\usepackage{longtable}
\usepackage{subcaption}
\usepackage{import}
\usepackage[usestackEOL]{stackengine}
\usepackage{framed}

\usepackage{tcolorbox}
\newtcolorbox{codebox}{colback=gray!1!white,colframe=gray!30,boxrule=1pt}

\definecolor{vlmblue}{RGB}{30, 90, 202}          
\definecolor{vlmblueDark}{RGB}{37, 99, 235}       
\definecolor{tablegreen}{RGB}{22, 163, 74}        
\definecolor{tablered}{RGB}{220, 38, 38}          
\definecolor{tableamber}{RGB}{245, 158, 11}       
\definecolor{chatgptcolor}{RGB}{120, 166, 90}   
\definecolor{geminicolor}{RGB}{139, 124, 190}   
\definecolor{claudecolor}{RGB}{236, 181, 118}   
\definecolor{otherblue}{RGB}{180, 190, 200}      
\definecolor{lightgray}{RGB}{249, 250, 251}       
\definecolor{subtlegray}{RGB}{241, 245, 249}      
\definecolor{tablebg}{RGB}{252, 252, 253}         
\definecolor{questiongray}{RGB}{107, 114, 128}    
\definecolor{answergreen}{RGB}{22, 163, 74}       

\definecolor{headerbg}{RGB}{25, 25, 25}           
\definecolor{gridgray}{gray}{0.85}                 
\definecolor{goodgreen}{RGB}{22, 163, 74}         
\definecolor{badred}{RGB}{220, 38, 38}            
\definecolor{cclnotreachedcolor}{RGB}{220, 53, 69} 

\definecolor{TargetGreen}{RGB}{22, 163, 74}  

\newcolumntype{C}{>{\centering\arraybackslash}X}

\newcommand{\failedimage}[1][3cm]{%
  \fcolorbox{tablered}{white}{%
    \parbox[c][#1][c]{#1}{%
      \centering
      \includegraphics[width=0.4#1]{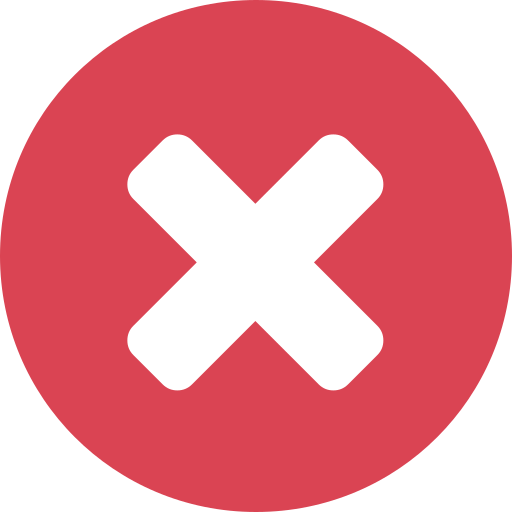}\\[-0.2em]
      \textcolor{tablered}{\footnotesize Failed}%
    }%
  }%
}

\newtcolorbox{ctabox}{
  colback=vlmblue!5!white,
  colframe=vlmblue,
  boxrule=1pt,
  arc=3pt,
  left=10pt,
  right=10pt,
  top=6pt,
  bottom=6pt,
  before skip=0.3em,
  after skip=0.5em
}

\newcommand{\prominentlink}[2]{%
  \textcolor{vlmblueDark}{\textbf{\href{#1}{#2}}}%
}

\newcommand{\ctasection}[1]{%
  \begin{ctabox}
    \centering
    \small
    #1
  \end{ctabox}%
}

\usepackage[authoryear, sort&compress, round]{natbib}
\usepackage{listings}
\usepackage{booktabs}
\usepackage{multirow}
\usepackage[usestackEOL]{stackengine}
\usepackage{threeparttable}
\usepackage{adjustbox}
\usepackage{subcaption}
\usepackage{inconsolata}
\usepackage{minted}
\setminted{
    fontfamily=tt,
    fontsize=\scriptsize,
    baselinestretch=1.1,
    framesep=12pt,
    xleftmargin=12pt,     
    xrightmargin=12pt,    
    breaklines=true,
}

\usepackage{siunitx}
\usepackage[dvipsnames]{xcolor}
\usepackage{tcolorbox}
\newcommand{\question}[1]{\frenchspacing\textbf{Q:} \textit{\textcolor{questiongray}{#1}}}
\newcommand{\answer}[1]{\frenchspacing\textbf{A:} \textit{\textcolor{vlmblue}{#1}}}

\uselogo{} 

\title{\textbf{Orion}: A Unified Visual Agent for Multimodal Perception, Advanced Visual Reasoning and Execution}

\correspondingauthor{research@vlm.run}

\renewcommand{\today}{}
\author{VLM Run Research}

\let\cite\citep

\begin{abstract}
We introduce \textcolor{vlmblue}{\textbf{\href{https://vlm.run/orion}{Orion}}}, a unified visual agent that integrates vision-based reasoning with tool-augmented execution to achieve powerful, precise, multi-step visual intelligence across images, video, and documents. Unlike traditional vision-language models that generate descriptive outputs, \textbf{Orion} orchestrates over a suite of specialized computer vision tools—including precise object detection, keypoint localization, panoptic segmentation, Optical Character Recognition (OCR), and geometric analysis—to execute complex multi-step visual workflows. The system achieves competitive performance across MMMU, MMBench, DocVQA, and MMLongBench while extending monolithic VLM capabilities to production-grade visual intelligence. Through its agentic tool-augmented approach, \textbf{Orion} enables autonomous visual reasoning that bridges neural perception with symbolic execution, marking the transition from passive visual understanding to active, tool-driven visual intelligence.
\end{abstract}

\begin{document}

\maketitle

\vspace{1em}
\ctasection{%
  \faRocket\ Try \textbf{Orion} for free at \prominentlink{https://chat.vlm.run}{chat.vlm.run} \quad $\bullet$ \quad \faBook\ Learn more at \prominentlink{https://vlm.run/orion}{vlm.run/orion}%
}


\begin{figure}[H]
    \centering
    \includegraphics[width=0.7\textwidth]{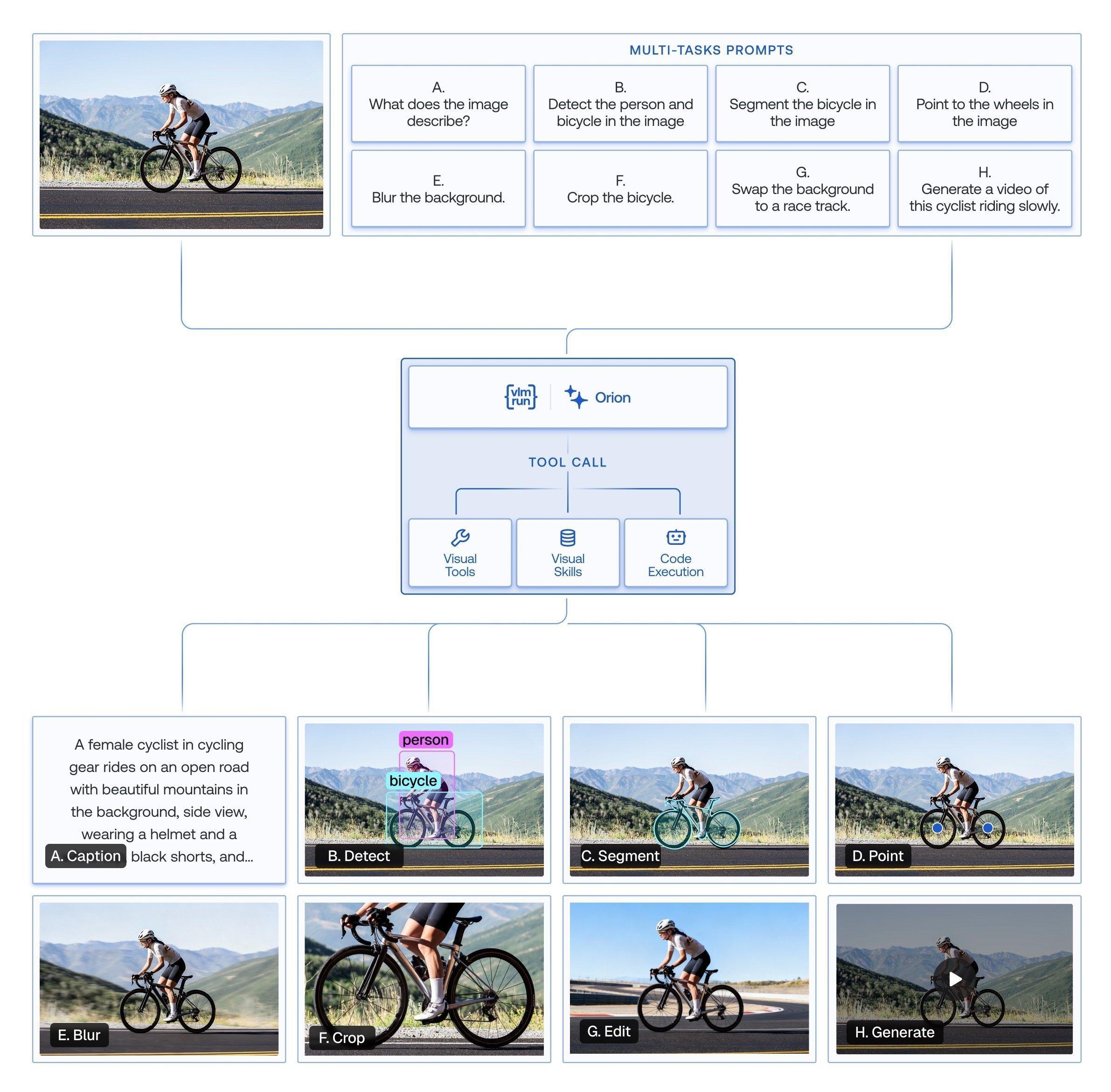}
    \caption{\textbf{Orion's} Image Understanding, Reasoning, and Execution Capabilities: In the illustration above, we showcase 8 of the many capabilities of \textbf{Orion} - including captioning, detection, segmentation, pointing, tool-calling, image-generation and video-generation – all orchestrated from the original image as the input.}
    \label{fig:architecture}
\end{figure}

\section{Introduction}
Today's frontier Vision-Language Models (VLMs)---GPT-5, Claude 4.5, Gemini 2.5 Pro \cite{comanici2025gemini25pushingfrontier}, and Qwen3-VL \cite{qwen3team2025qwen3}---can describe images, answer questions, and perform general-purpose reasoning over visual content. Building upon foundational vision-language models such as CLIP \cite{radford2021learning}, LLaVA \cite{li2023llava}, and BLIP-2 \cite{li2023blip2}, these systems have demonstrated remarkable capabilities in multimodal understanding. Yet they remain fundamentally limited: they operate as monolithic inference engines that generate descriptive or conversational outputs but cannot \textit{act} on visual data with precision, determinism, or compositional control. They lack the fine-grained tool integration, multi-step planning, and structured validation required for production-grade visual workflows.

\textbf{Orion} is a visual agent framework that can take in any modality and generate any modality, using an agentic framework with multiple tool calling capabilities geared towards visual AI tasks to generate state-of-the-art results. The framework's core innovation lies in its ability to seamlessly process and produce diverse data types---images, videos, documents, audio, and text---within a unified agentic architecture. This multi-modal flexibility enables \textbf{Orion} to handle complex visual reasoning tasks that require understanding, transformation, and generation across different data formats, from extracting structured information from documents to generating visual content from textual descriptions.

The system integrates the cognitive capabilities of large VLMs with the precision of tens of hyper-specialized computer vision tools---including Optical Character Recognition (OCR) for text extraction, detection for object identification, segmentation for pixel-level analysis, pointing for spatial localization, diffusion for image generation, and geometric analysis---orchestrated through an agentic architecture inspired by ReAct-style agent frameworks \cite{yao2023react} and tool-augmented language models \cite{schick2023toolformer,patil2023gorilla}. The agentic framework enables \textbf{Orion} to dynamically reason about task requirements, select appropriate tools from its extensive library, and compose them into sophisticated multi-step workflows. Unlike baseline VLMs that generate single-step responses, \textbf{Orion} dynamically constructs and executes multi-step workflows from natural language instructions, achieving higher degree of task success through intelligent tool orchestration and iterative refinement. 

\newcommand{\greencheck}{\textcolor{tablegreen}{\textbf{\Large$\checkmark$}}}
\newcommand{\redcross}{\textcolor{tablered}{\textbf{\Large$\times$}}}
\newcommand{\amberpartial}{\textcolor{tableamber}{\textbf{\Large$\sim$}}}

\begin{table}[H]
\centering
\caption{Comprehensive capability comparison showing unique features of \textbf{Orion}. Unlike monolithic Large-Language Models, \textbf{Orion} delivers comprehensive capabilities across all modalities and tasks. \textbf{Specialized Skills} refers to tasks such as object localization, segmentation, image generation/editing, or geometric tools typically found in specialized computer vision applications. In the table below, the symbols represent the following: \greencheck ~ Full support, \redcross ~ No support, \amberpartial ~ Partial support.}
\label{tab:capabilities}
\small
\renewcommand{\arraystretch}{1.6}
\setlength{\tabcolsep}{0.9em}
\setlength{\arrayrulewidth}{0.5pt}
\begin{tabular}{@{}>{\raggedright\arraybackslash}p{1.8cm}@{\hspace{0.15cm}}>{\raggedright\arraybackslash}p{2.8cm}@{\hspace{0.15cm}}>{\centering\arraybackslash}p{1.3cm}>{\centering\arraybackslash}p{1.3cm}>{\centering\arraybackslash}p{1.3cm}>{\centering\arraybackslash}p{1.7cm}>{\centering\arraybackslash}p{1.5cm}}
\toprule[1pt]
\textbf{Modality} & \textbf{Task} & \textcolor{vlmblue}{\textbf{Orion}} & \textcolor{chatgptcolor}{\textbf{GPT-5}} & \textcolor{geminicolor}{\textbf{Gemini 2.5}} & \textcolor{claudecolor}{\textbf{Claude Sonnet 4.5}} & \textcolor{otherblue}{\parbox{1.8cm}{\centering\textbf{Qwen3-VL 235B}}}\\
\midrule[0.5pt]
\multirow{5}{*}{\parbox{2.5cm}{\raggedright\textbf{Image /\\ Video}}} 
 & Understanding & \greencheck & \greencheck & \greencheck & \greencheck & \greencheck \\
 & Reasoning & \greencheck & \redcross & \redcross & \redcross & \greencheck \\
 & Structured Outputs & \greencheck & \greencheck & \greencheck & \greencheck & \greencheck \\
 & \parbox{2.2cm}{\raggedright Tool-Calling} & \greencheck & \redcross & \redcross & \redcross & \amberpartial \\
 & Specialized Skills & \greencheck & \redcross & \amberpartial & \amberpartial & \redcross \\
\midrule[0.5pt]
\multirow{5}{*}{\textbf{Document}} 
 & Understanding & \greencheck & \greencheck & \greencheck & \greencheck & \greencheck \\
 & Reasoning & \greencheck & \greencheck & \greencheck & \greencheck & \redcross \\
 & Structured Outputs & \greencheck & \greencheck & \greencheck & \greencheck & \greencheck \\
 & \parbox{2.2cm}{\raggedright Tool-Calling} & \greencheck & \amberpartial & \amberpartial & \amberpartial & \redcross \\
 & Specialized Skills & \greencheck & \greencheck & \amberpartial & \greencheck & \redcross \\
\bottomrule[1pt]
\end{tabular}%
\end{table}

The framework's multiple tool calling capabilities enable it to tackle a wide spectrum of visual AI tasks that were previously intractable for monolithic models. By combining specialized tools for document understanding, image analysis, video processing, and cross-modal reasoning, \textbf{Orion} can perform complex operations such as extracting structured data from invoices, analyzing medical imaging, tracking objects across video sequences, and generating contextually appropriate visual content. The agentic nature of the framework allows it to adapt its approach based on the specific requirements of each task, selecting the optimal sequence of tools and adjusting its strategy based on intermediate results.

This marks the transition from passive visual understanding to \textbf{autonomous, tool-augmented visual intelligence}---a new category of agentic AI platform capable of combining deep neural perception, symbolic reasoning, and structured automation within a unified system. \textbf{Orion} does not merely enhance existing multimodal models; it redefines what visual AI systems can achieve in complex, open-ended, and production-critical environments. The framework's ability to generate state-of-the-art results across diverse visual tasks stems from its unique combination of large-scale neural reasoning, specialized tool precision, and intelligent orchestration, enabling it to outperform both general-purpose VLMs and task-specific systems.

Comprehensive evaluation across 46 diverse visual tasks---ranging from object detection and segmentation to complex multi-step visual reasoning, document analysis, and image generation---demonstrates that \textbf{Orion} consistently outperforms frontier VLMs including GPT-5 \cite{achiam2023gpt4}, Claude 4.5, Gemini 2.5 Pro \cite{comanici2025gemini25pushingfrontier}, and Qwen3-VL \cite{qwen3team2025qwen3}. Evaluations span established benchmarks including MMMU \cite{yue2024mmmu}, MMBench \cite{liu2024mmbench}, and DocVQA \cite{docvqa2020}. As shown in Figure~\ref{fig:orion_evaluation}, \textbf{Orion} achieves superior performance across all evaluated categories, establishing it as the leading system for production-grade visual tasks that require precision, multi-step reasoning, and specialized tool integration.

\newpage

\section{System Overview}
\label{sec:system_overview}
\subsection{System Architecture}



        
\begin{figure}[H]
    \centering
    \includegraphics[width=0.9\textwidth]{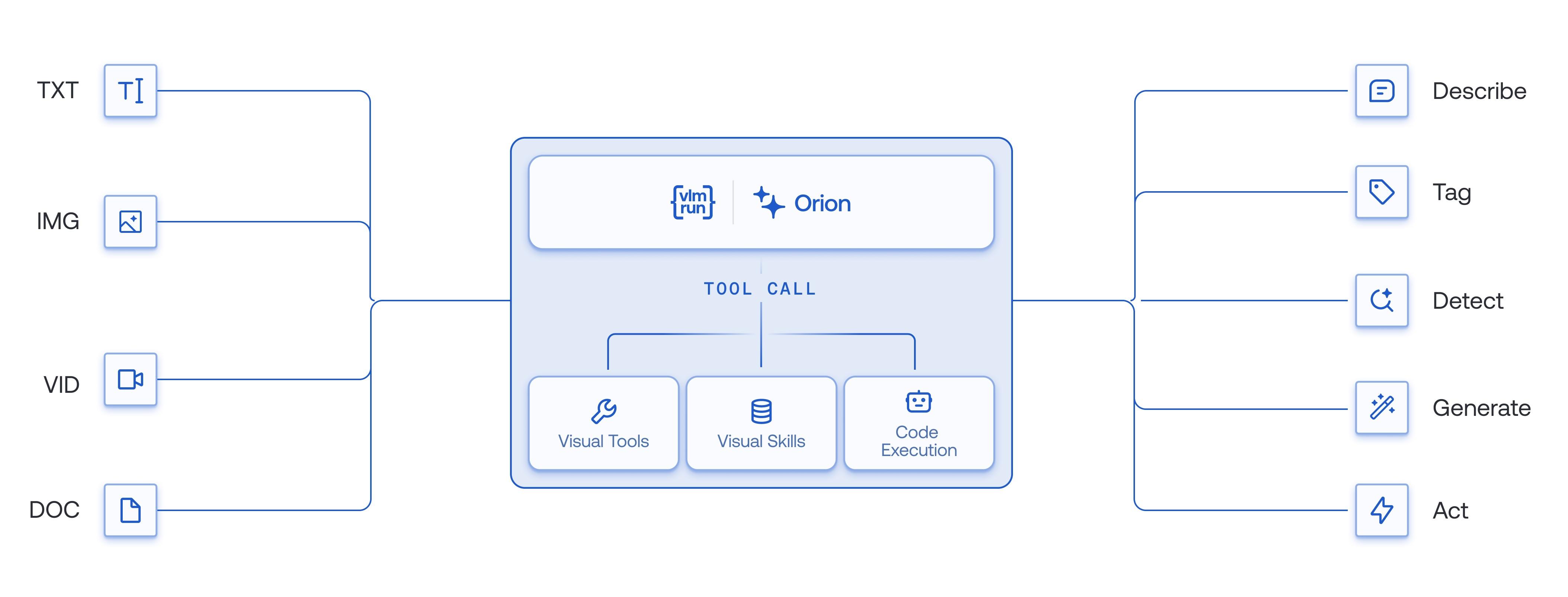}
    \caption{System architecture of \textbf{Orion}, illustrating the visual agent and its interaction with various visual tools, skills and code execution environments. \textbf{Orion} supports text, image, video and document inputs to answer questions (in multiple steps if possible) that may involve captioning, tagging, detection, generation or external tool-calling. See capabilities~\ref{sec:capabilities} section for more details.}
    \label{fig:system_architecture}
\end{figure}

\subsubsection{Unified Orchestration with Planning, Execution and Reflection}

\textbf{Orion} implements a ReAct-style agent controller \cite{yao2023react} that combines a large-language-model (LLM) based planner with a multi-modal tool-calling execution engine and a vision-language (VLM) based verifier. The core controller operates through three distinct phases (Plan-Execute-Reflect) in a closed-loop:


\begin{itemize}[leftmargin=*]
    \item \textbf{Plan} – \textbf{Orion} analyzes the user's request to identify task intent and generates a structured execution plan, leveraging chain-of-thought reasoning \cite{wei2022chain} to decompose complex visual tasks. This planning phase determines the sequence of tool invocations, establishes data dependencies between operations, and evaluates the optimal choice of tools and execution order. The controller reasons about which visual processing capabilities are required and how they should be orchestrated to achieve the desired outcome.

    \item \textbf{Execute} – The planned sequence of tools is executed according to the established execution plan. Tools are invoked either sequentially or in parallel, depending on dependency constraints and computational cost considerations. This flexible execution model allows \textbf{Orion} to maximize throughput while respecting data dependencies between operations. The execution engine integrates with multi-modal tool-calling capabilities including GPU-accelerated tool invocation to enable complex, multi-step visual reasoning and structured output validation. 

    \item \textbf{Reflect} – Intermediate and final outputs are evaluated against expected types and values to ensure correctness and consistency. This reflection phase identifies potential inconsistencies, validates visual output quality with respect to the inputs, and determines whether to retry failed operations, refine intermediate results, or finalize the completion request. The reflection mechanism enables \textbf{Orion} to course-correct and iteratively improve output quality throughout the execution process.
\end{itemize}


\subsubsection{Multi-Modal Tool Execution}

\textbf{Orion} provides first-class visual reasoning and execution capabilities through a comprehensive suite of specialized tools organized into four functional categories. The tool library comprises a suite of distinct tools spanning document understanding, image analysis, video processing, and cross-modal reasoning tasks. 

\textbf{Image Tools} – These tools include object detection for identifying and localizing visual entities, face recognition for identity verification, segmentation for pixel-level classification, and visual question answering for image-based reasoning tasks.

\textbf{Document Tools} – These tools include Optical Character Recognition (OCR) for text extraction, layout detection for structural analysis, table extraction for tabular data parsing, form understanding for structured document processing, and keyword search for content retrieval within documents.

\textbf{Video Tools} – These tools include long-form visual transcription for video-to-text captioning, scene segmentation for temporal boundary detection, object tracking for motion analysis, and action recognition for activity classification.

\textbf{Mixed-modality Tools} – This category encompasses cross-modal retrieval for finding related content across modalities, visual redaction for privacy-preserving content masking, and region-level content extraction for targeted information retrieval.

Each tool exposes standardized input/output interfaces that enable the controller to interpret and react to tool outputs dynamically. This standardization allows interleaved reasoning and re-interpretation of visual outputs, incorporating results from one tool as inputs to subsequent execution steps. The uniform interface design ensures composability and enables the controller to construct complex multi-step workflows without manual intervention.


\subsubsection{Multi-Modal Context and Session Management}

The \textbf{Orion} system handles diverse input modalities—including images, videos, documents, audio, and text, making it especially challenging to manage context without loss of information within the agent's context window. The system implements adaptive context management that selectively retrieves only the context relevant to the current instruction, rather than maintaining the entire conversation history. This approach enables efficient processing in multi-turn conversations extending beyond 25 turns while maintaining coherent session state across complex, long-running interactions.

\subsubsection{Reflection with Visual Judges}

The \textbf{Orion} system employs VLM-as-a-Judge~\cite{lee2024prometheusvisionvisionlanguagemodeljudge} models to visually evaluate tool outputs against expected schemas, identify potential inconsistencies, and determine whether to retry, refine, or finalize the workflow. This reflection mechanism enables closed-loop control over the agent's execution ensuring that intermediate and final results meet structural and semantic requirements of the original instruction.

\textit{Conditional Execution \& Closed Loop Control}: The reflection framework enables conditional invocation of VLM-as-a-Judge tools to evaluate the outputs of other tools within the execution plan. By incorporating these evaluation checkpoints, the system iteratively refines outputs and course-corrects the overall visual reasoning and execution trajectory. This feedback loop allows the controller to detect errors early, trigger corrective actions, and progressively improve result quality without manual intervention, enhancing both robustness and reliability across complex multi-step workflows.

 
\subsection{Key System Features}

\textbf{Orion} provides several distinctive capabilities enabling robust, flexible, and scalable multi-modal reasoning and execution in a single unified API.

\textbf{Natively Multi-Modal} – \textbf{Orion} is truly multi-modal, with native support for handling images, videos, documents, audio, and text within the same conversation. All modalities are processed consistently, enabling seamless cross-modal reasoning and execution without requiring separate workflows or manual format conversions.


\textbf{Dynamic Routing} – \textbf{Orion} implements dynamic routing of complex instructions and tasks to more sophisticated models and tools as needed. When the controller detects that a task exceeds the current agent's capabilities, it can dynamically escalate to more powerful models or specialized tools, with graceful fallbacks ensuring robustness when preferred resources are unavailable.

\textbf{Dual-Mode Agent Design} – \textbf{Orion} supports a \textit{dual-mode} agent design, operating in both \textit{chat} and \textit{structured-output} modes to serve conversational and API-driven use cases. Chat mode provides natural-language interaction for exploratory workflows (e.g., visual chat - \href{https://chat.vlm.run}{https://chat.vlm.run}), while structured mode delivers type-safe, schema-validated outputs for deterministic execution with high reliability.

\textbf{Closed-loop Reflection} – \textbf{Orion} incorporates closed-loop reflection to improve output quality and correctness. By evaluating intermediate and final results against expected schemas and semantic requirements, it iteratively refines outputs, detects errors early, and course-corrects execution trajectories without manual intervention.

\subsection{Key API Features}
\label{sec:key-api-features}

\textbf{Orion} exposes a comprehensive API surface designed to support diverse integration patterns and use cases, from interactive conversational interfaces to programmatic batch processing pipelines. See our docs page for more details:  \href{https://docs.vlm.run/agents/introduction}{https://docs.vlm.run/agents/introduction}.

\textbf{Multi-Modal Input Support} – The API provides native support for passing images, videos, documents, audio, and text within the same conversation. Content can be provided via file identifiers or direct URLs, enabling flexible integration with existing storage systems and content delivery networks. This unified input mechanism eliminates separate endpoints or pre-processing steps for different modalities.

\begin{center}
\begin{minipage}{0.7\textwidth}
\begin{codebox}
\centering
\begin{minted}[breaklines, linenos, tabsize=2]{python}
from vlmrun.client import VLMRun

# Initialize the client
client = VLMRun(base_url="https://agent.vlm.run/v1", api_key="...")

# Example: Chat completion with multi-modal inputs
image_file = client.files.upload("logo.jpg")
video_file = client.files.upload("racing.mp4")
messages = [
    {"role": "user", "content": [
        {"type": "text", "text": "Place this logo in the racing ad video?"},
        {"type": "input_file", "file_id": image_file.id},
        {"type": "input_file", "file_id": video_file.id}
    ]}
]

# Perform chat completion
chat_completion = client.agent.completions.create(
    model="vlmrun-orion-1:auto",
    messages=messages,
)
print(chat_completion.choices[0].message.content)
\end{minted}
\end{codebox}
\end{minipage}
\end{center}

\textbf{Streaming Chat Completion Interface} – The streaming interface supports real-time interaction by delivering responses incrementally as they are generated. This enables agents to reason, act, and respond incrementally, providing immediate feedback to users and allowing downstream systems to begin processing results before the entire response is complete. The streaming mechanism reduces perceived latency and enhances the interactive experience for conversational use cases. This can be done by simply adding \texttt{stream=True} to the \texttt{client.agent.completsion.create} call above. 

\textbf{OpenAI Compatibility} – \textbf{Orion} implements a fully OpenAI-compatible chat completions API, ensuring seamless integration with existing tools, libraries, and workflows built around the OpenAI ecosystem. The API includes streaming support for real-time interaction, enabling low-latency responses and progressive result delivery.

\begin{center}
\begin{minipage}{0.8\textwidth}
\begin{codebox}
\centering
\begin{minted}[breaklines, linenos, tabsize=2]{python}
import openai
from vlmrun.common.image import encode_image

# Initialize the OpenAI client
client = openai.OpenAI(base_url="https://agent.vlm.run/v1/openai", api_key="...")

# Example: Chat completion with an image input
messages = [
    {"role": "user", "content": [
        {"type": "text", "text": "What's in this image?"},
        {"type": "image_url", "image_url": {"url": encode_image(image), "detail": "auto"}},
    ]}
]

# Perform chat completion
chat_completion = client.chat.completions.create(
    model="vlmrun-orion-1:auto",
    messages=messages,
)
print(chat_completion.choices[0].message.content)
\end{minted}
\end{codebox}
\end{minipage}
\end{center}

\textbf{Structured Outputs API} – \textbf{Orion} provides structured output capabilities with type-safe output validation. Results conform to predefined schemas and are validated against expected types and constraints before delivery. This ensures that downstream systems receive well-formed, predictable data structures suitable for programmatic processing, reducing integration complexity and improving reliability in automated workflows. See section \ref{sec:structured-outputs} for more details.

\begin{center}
\begin{minipage}{0.8\textwidth}
\begin{codebox}
\begin{minted}[breaklines, linenos, tabsize=2]{python}
from vlmrun.client import VLMRun
from vlmrun.common.image import encode_image

# Initialize the client
client = VLMRun(base_url="https://agent.vlm.run/v1", api_key="...")

# Example: Chat completion with an image input
messages = [
    {"role": "user", "content": [
        {"type": "text", "text": "What's in this image?"},
        {"type": "image_url", "image_url": {"url": encode_image(image), "detail": "auto"}},
    ]}
]

# Example response format
class ImageCaption(BaseModel):
    caption: str = Field(..., description="Brief caption of the image")
    tags: list[str] = Field(..., description="List of relevant tags (ad, news, sports)")


# Perform chat completion with JSON schema
response_format = {"type": "json_schema", "schema": ImageCaption.model_json_schema()}
chat_completion = client.agent.completions.create(
    model="vlmrun-orion-1:auto",
    messages=messages,
    response_format=response_format,
)
content: str = chat_completion.choices[0].message.content
caption: ImageCaption = ImageCaption.model_validate_json(content)
\end{minted}
\end{codebox}
\end{minipage}
\end{center}


 
\subsection{Extensions and Future Work}

\textbf{Orion} is designed with extensibility in mind, and several planned enhancements will further expand its capabilities and flexibility.

\textbf{Tool-Calling Support} – Future versions will provide the ability to call external or user-defined tools and services, enabling seamless integration of custom processing modules into the execution plan. This capability allows users to extend \textbf{Orion} with domain-specific algorithms, proprietary APIs, or specialized processing pipelines without modifying core system components.

\textbf{Code-Generation and Execution} – Planned support for automatic generation of computer-vision tools on-the-fly will enable dynamic creation and execution of custom processing modules within the execution plan. This capability allows \textbf{Orion} to synthesize specialized tools tailored to specific task requirements, reducing the need for pre-built modules and enabling more flexible problem-solving approaches.

\textbf{Deeper LLM Integration} – The architecture will evolve to support a hot-swappable model architecture compatible with all leading frontier model architectures. This includes open-source model families such as LLaMA \cite{liu2023llama,touvron2023llama2} and Qwen \cite{qwen3team2025qwen3,liu2024qwen}, as well as frontier model families including OpenAI GPT \cite{achiam2023gpt4}, Anthropic Claude, and Google Gemini \cite{comanici2025gemini25pushingfrontier}. This flexibility enables users to select models based on performance requirements, cost constraints, deployment environments, and specific task characteristics.

\newpage
\section{Capabilities}
\label{sec:capabilities}
\subsection{Image Understanding, Reasoning and Tool-Use}
\noindent
\textbf{\textbf{Orion}} provides a comprehensive suite of image understanding and manipulation capabilities, enabling fine-grained visual reasoning and tool-augmented execution workflows. The capabilities are organized as follows:

\subsubsection{Image Captioning}
Generate dense, contextually rich captions and semantic tags for images, supporting both global scene descriptions and region-specific annotations. 
\begin{figure}[H]
\centering
\begin{subfigure}[t]{0.45\textwidth}
    \begin{minipage}{\textwidth}
    \raggedright\small\question{What is in this image?}\vspace{0.5em}
    \end{minipage}
    \centering
    \includegraphics[height=100px,keepaspectratio]{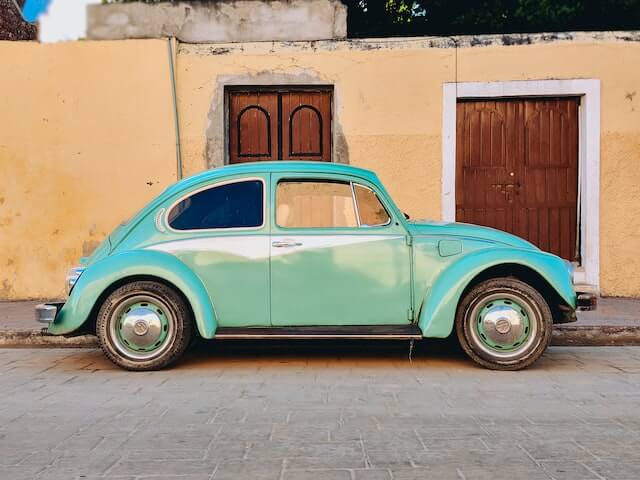}
    \caption{Vehicle Scene Captioning\vspace{0.5em}\\
    \begin{minipage}{\textwidth}
    \raggedright\small\answer{A classic, light turquoise Volkswagen Beetle with chrome accents is parked on a cobblestone street, set against a warm yellow stucco wall with rustic brown wooden doors and windows. Tags: car, volkswagen, beetle, street, cobblestone, wooden, doors, windows}
    \end{minipage}}
\end{subfigure}
\hfill
    \begin{subfigure}[t]{0.45\textwidth}
    \begin{minipage}{\textwidth}
    \raggedright\small\question{Densely caption this image in 2 sentences. }\vspace{0.5em}
    \end{minipage}
    \centering
    \includegraphics[height=100px,keepaspectratio]{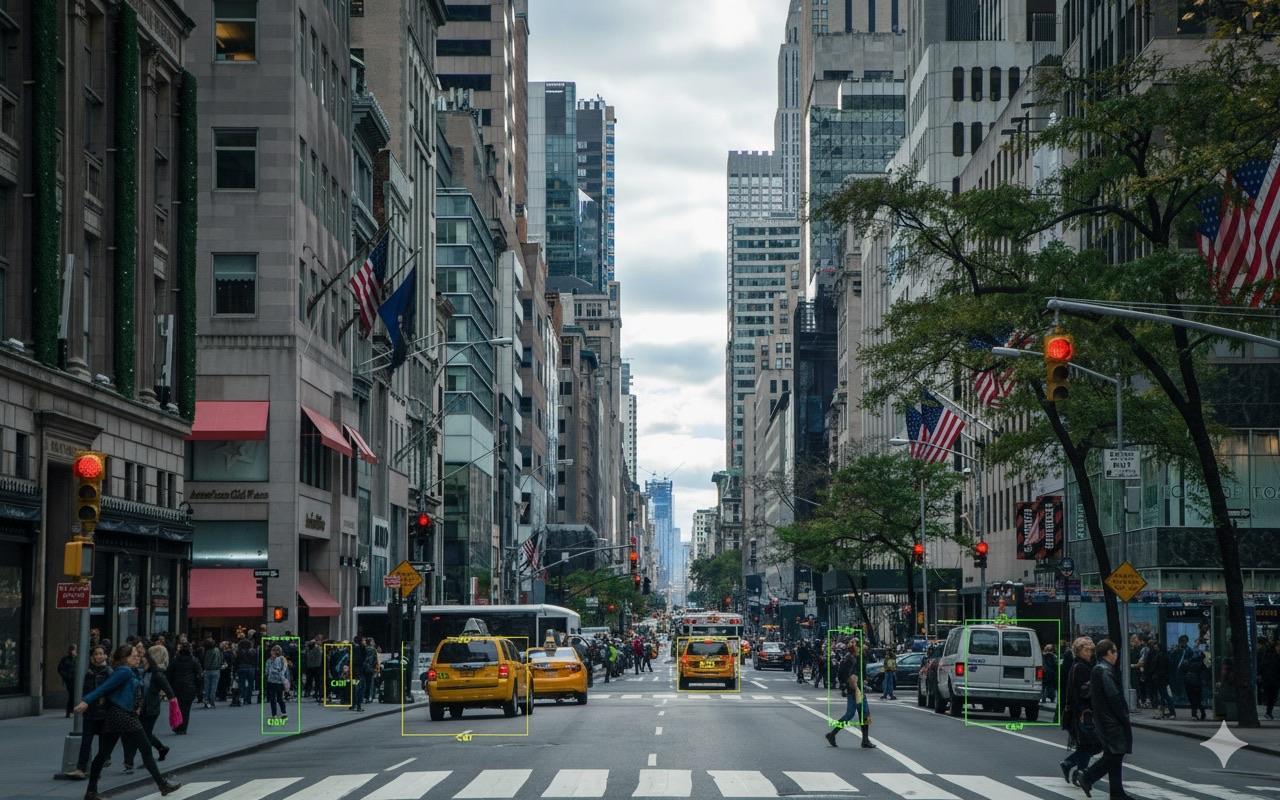}
    \raggedleft
    \caption{Dense captioning in Urban Environments\vspace{0.5em}\\
    \begin{minipage}{\textwidth}
    \raggedright\small\answer{A busy city street is depicted with tall, imposing buildings lining both sides under an overcast sky, as numerous yellow taxis, vans, and other vehicles occupy the multi-lane road. Pedestrians actively cross marked crosswalks and walk along the sidewalks, while American flags adorn several buildings and red traffic lights are visible throughout the dense urban environment.}
    \end{minipage}}
\end{subfigure}
\caption{Image captioning examples showing dense, rich description generation for scene understanding.}
\label{fig:captioning_example}
\end{figure}

\subsubsection{Visual Question Answering (VQA) with Grounding}
Perform open-ended visual question answering with spatial grounding, allowing users to query arbitrary visual content and receive precise, contextually grounded responses. \textbf{Orion} supports both free-form natural language queries and structured interrogation~\ref{sec:key-api-features} of visual elements.
\begin{figure}[H]
\centering
\begin{subfigure}[t]{0.45\textwidth}
    \begin{minipage}{\textwidth}
    \raggedright\small\question{Can you visualize the items in the image that I need to avoid if I'm allergic to seafood?}\vspace{0.5em}
    \end{minipage}
    \centering
    \includegraphics[height=100px]{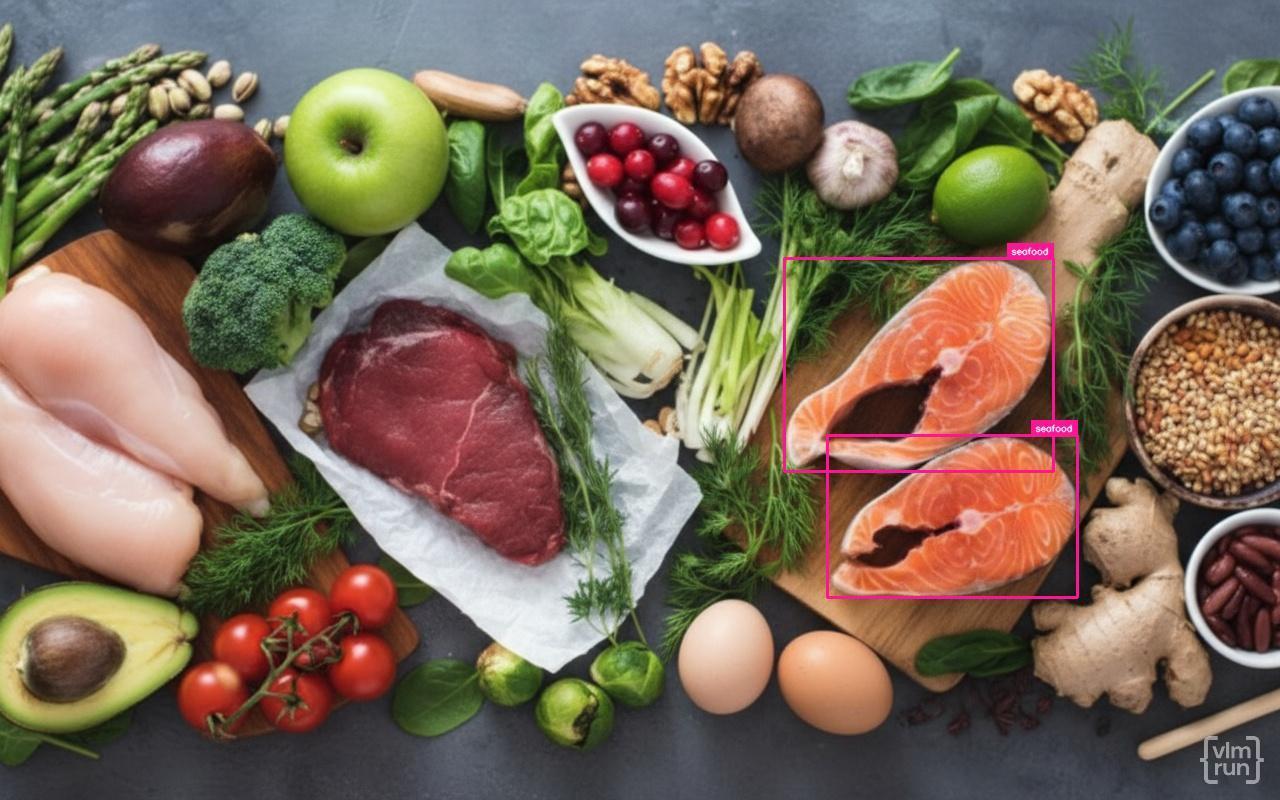}
    \caption{Food Allergy Visualization\vspace{0.5em}\\
    \begin{minipage}{\textwidth}
    \raggedright\small\answer{If you are allergic to seafood, you should avoid the items highlighted in the image. I have marked the seafood items for you.}
    \end{minipage}}
\end{subfigure}
\hfill
    \begin{subfigure}[t]{0.45\textwidth}
    \begin{minipage}{\textwidth}
    \raggedright\small\question{Detect the light sources in the image}\vspace{0.5em}
    \end{minipage}
    \centering
    \includegraphics[height=100px]{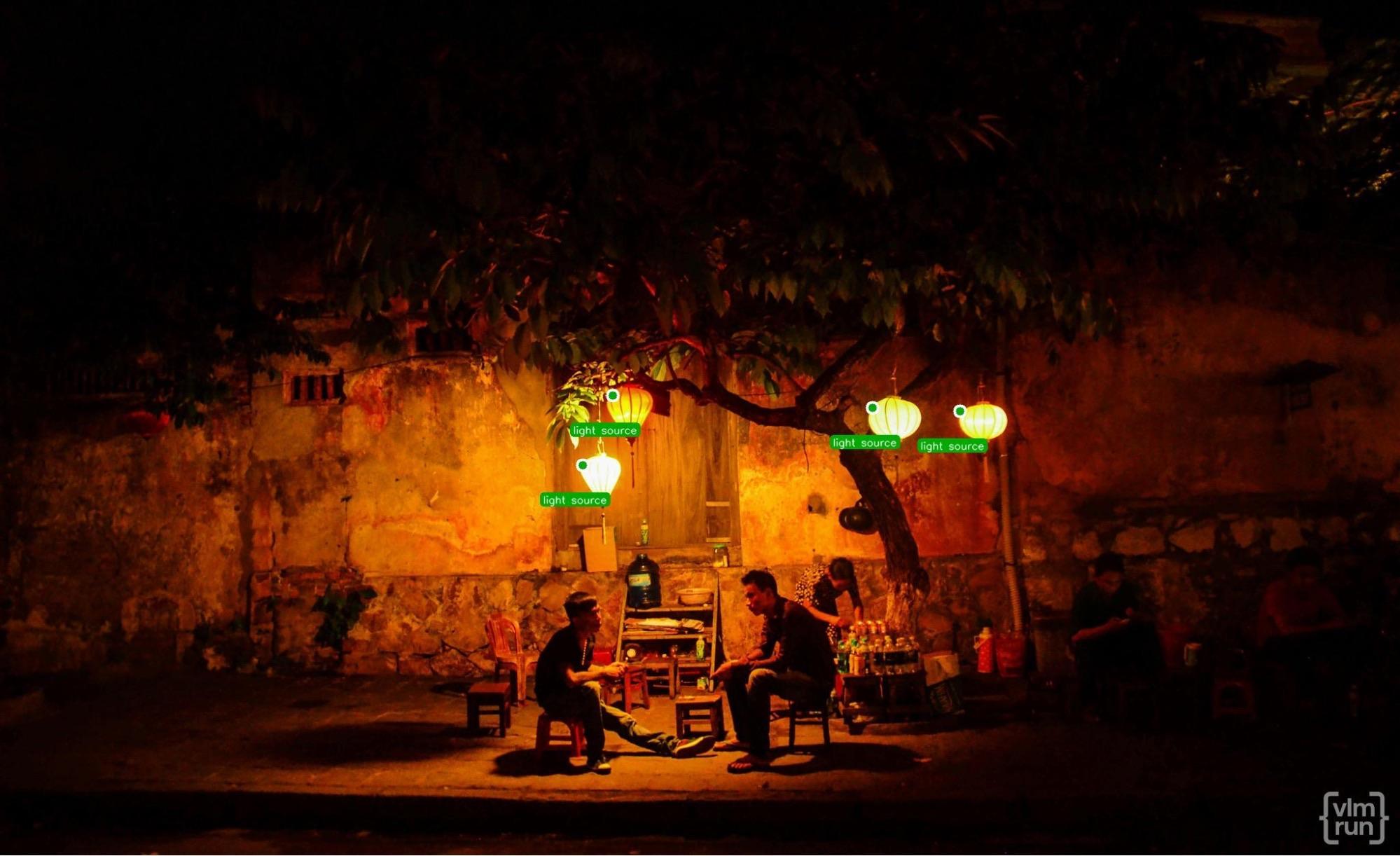}
    \caption{Light Source Detection\vspace{0.5em}\\
    \begin{minipage}{\textwidth}
    \raggedright\small\answer{I have detected and visualized all light sources in the image, identifying both natural and artificial lighting elements with their precise locations and intensities.}
    \end{minipage}}
\end{subfigure}

\caption{Visual question answering examples demonstrating open-ended queries with spatial grounding and contextual responses.}
\label{fig:vqa_example}
\end{figure}

\subsubsection{Object, Person, and Face Detection}
Perform comprehensive detection capabilities spanning multiple object categories, including general objects, faces, persons, logos, and landmarks. Provides bounding box localization, confidence scores, and class labels for downstream reasoning and analysis. Detection bounding boxes are returned in normalized xywh format.

\begin{figure}[H]
    \centering
    \begin{subfigure}{0.3\textwidth}
        \raggedright\small\question{Detect the objects in the image.}\vspace{0.5em}
        \centering
        \includegraphics[width=\textwidth]{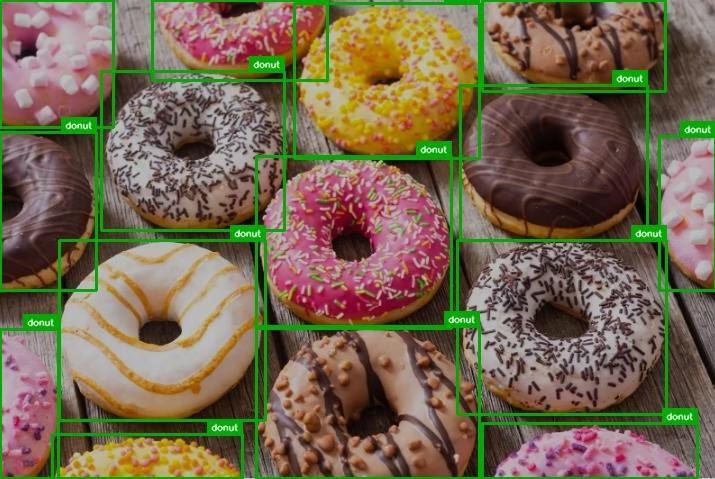}
        \caption{Object Detection}
    \end{subfigure}
    \hfill
    \begin{subfigure}{0.3\textwidth}
        \raggedright\small\question{Detect the people in the image.}\vspace{0.5em}
        \centering
        \includegraphics[width=\textwidth]{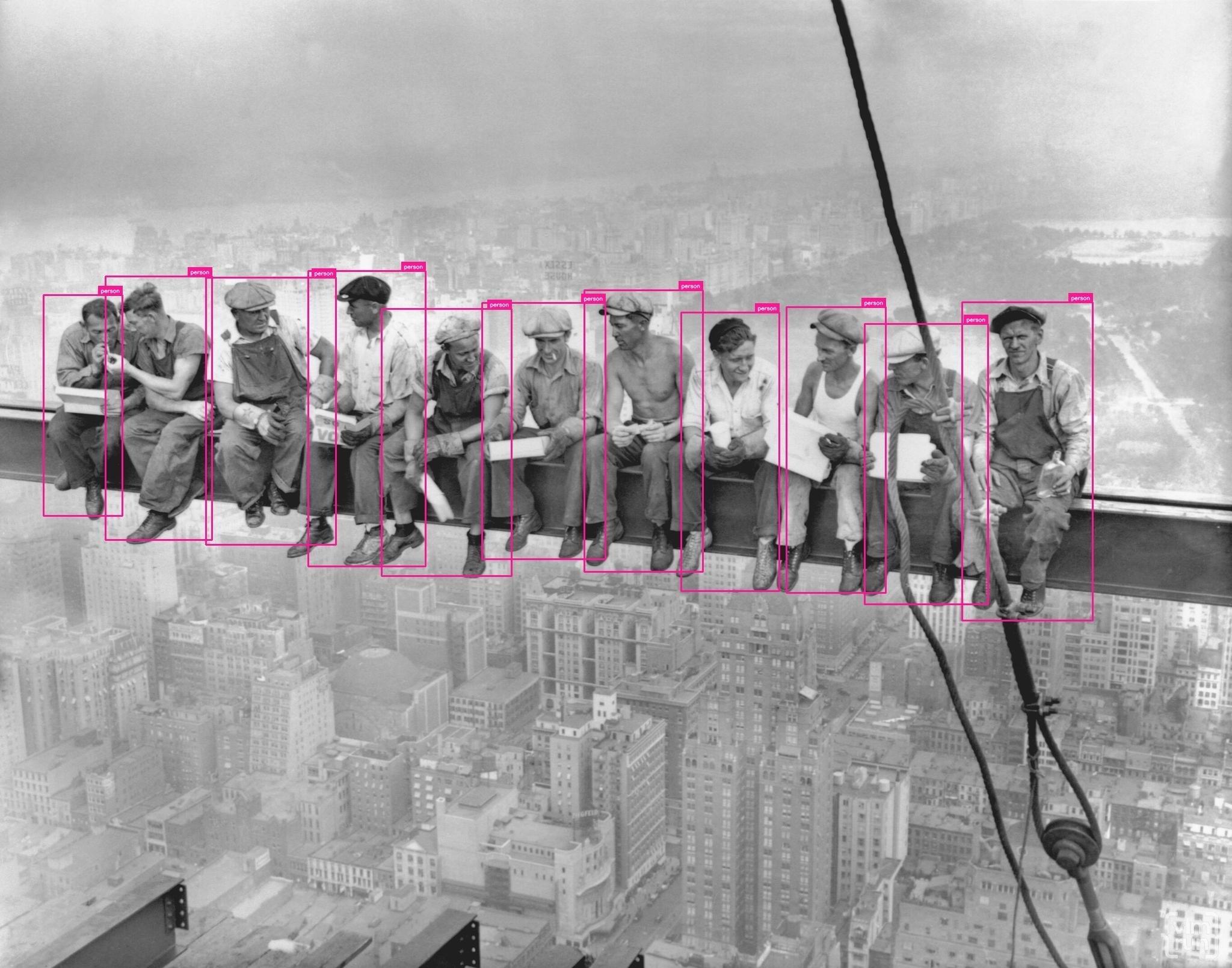}
        \caption{Person Detection}
    \end{subfigure}
    \hfill
    \begin{subfigure}{0.3\textwidth}
        \raggedright\small\question{Detect the faces in the image.}\vspace{0.5em}
        \centering
        \includegraphics[width=\textwidth]{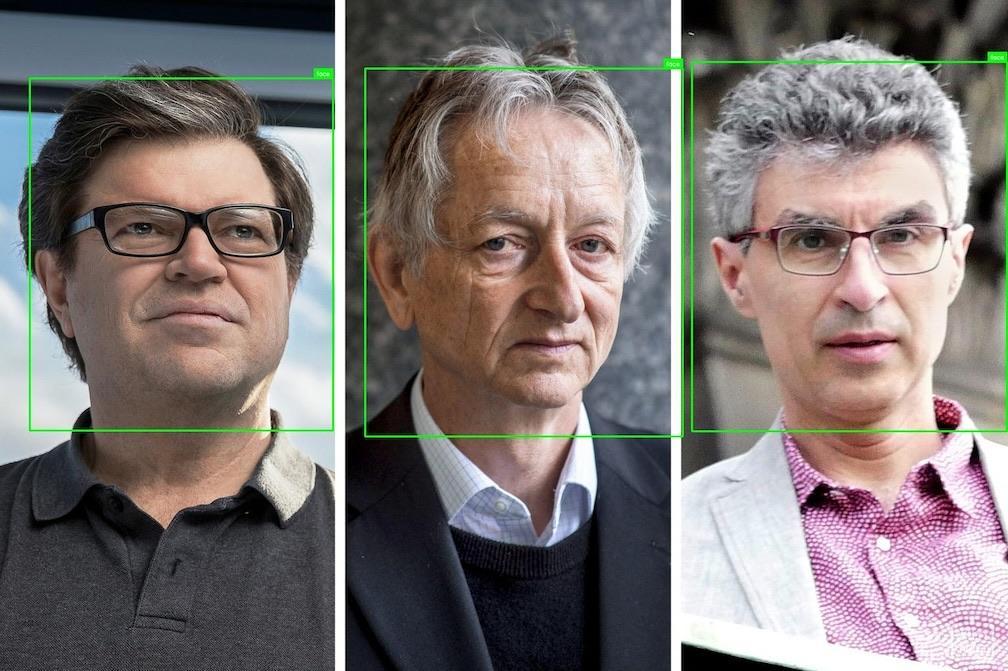}
        \caption{Face Detection}
    \end{subfigure}

    \caption{Examples of object, person, and face detection capabilities showing bounding box localization with confidence scores.}
    \label{fig:detection_examples}
\end{figure}

\subsubsection{Image Segmentation}
Supports both semantic segmentation (pixel-level classification of scene elements) and instance segmentation (individual object delineation), enabling precise spatial understanding and region-based analysis for complex visual scenes. Segmentation masks are returned as uint8 images of the same size as the input image, with unique labels (0, 1, 2, ..) for each segment instance.

\begin{figure}[H]
    \centering
    \begin{subfigure}{0.48\textwidth}
        \raggedright\small\question{Segment the car in the image}\vspace{0.5em}
        \centering
        \includegraphics[height=100px]{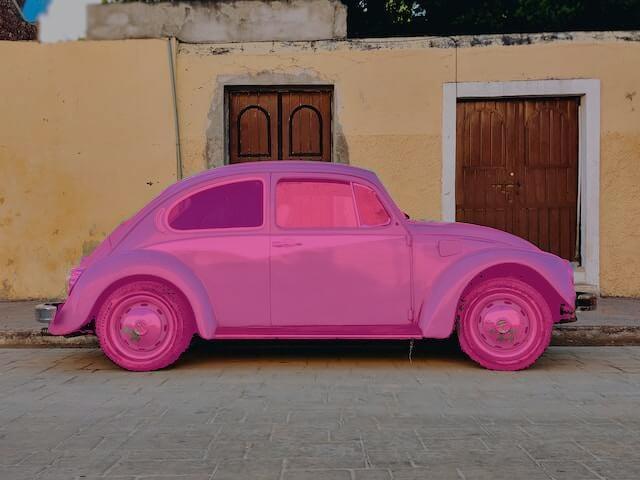}
        \caption{Vehicle Segmentation}
    \end{subfigure}
    \hfill
    \begin{subfigure}{0.48\textwidth}
        \raggedright\small\question{Segment all the faces in the image}\vspace{0.5em}
        \centering
        \includegraphics[height=100px]{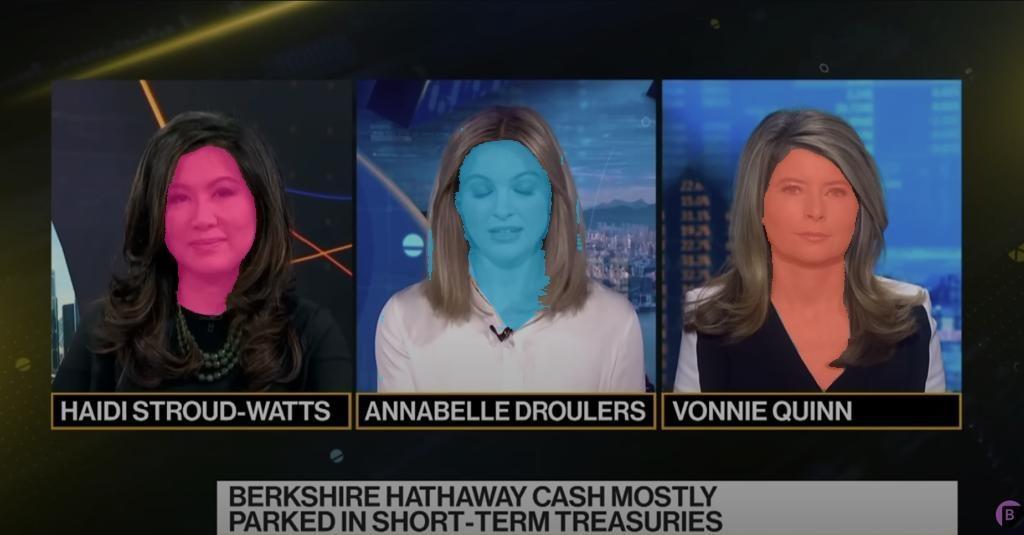}
        \caption{Face Segmentation}
    \end{subfigure}
    
    \vspace{1em}
    
    \begin{subfigure}{0.48\textwidth}
        \raggedright\small\question{Segment out the turtles in the image}\vspace{0.5em}
        \centering
        \includegraphics[height=100px]{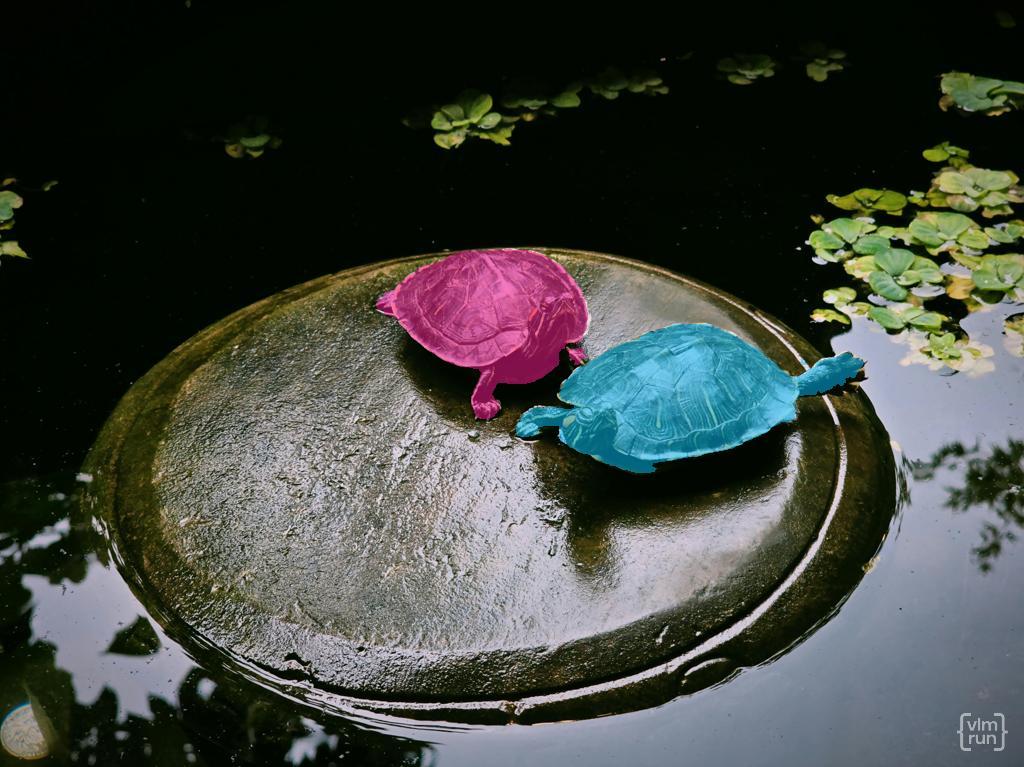}
        \caption{Instance Segmentation}
    \end{subfigure}
    \hfill
    \begin{subfigure}{0.48\textwidth}
        \raggedright\small\question{Segment all the furniture in the living room}\vspace{0.5em}
        \centering
        \includegraphics[height=100px]{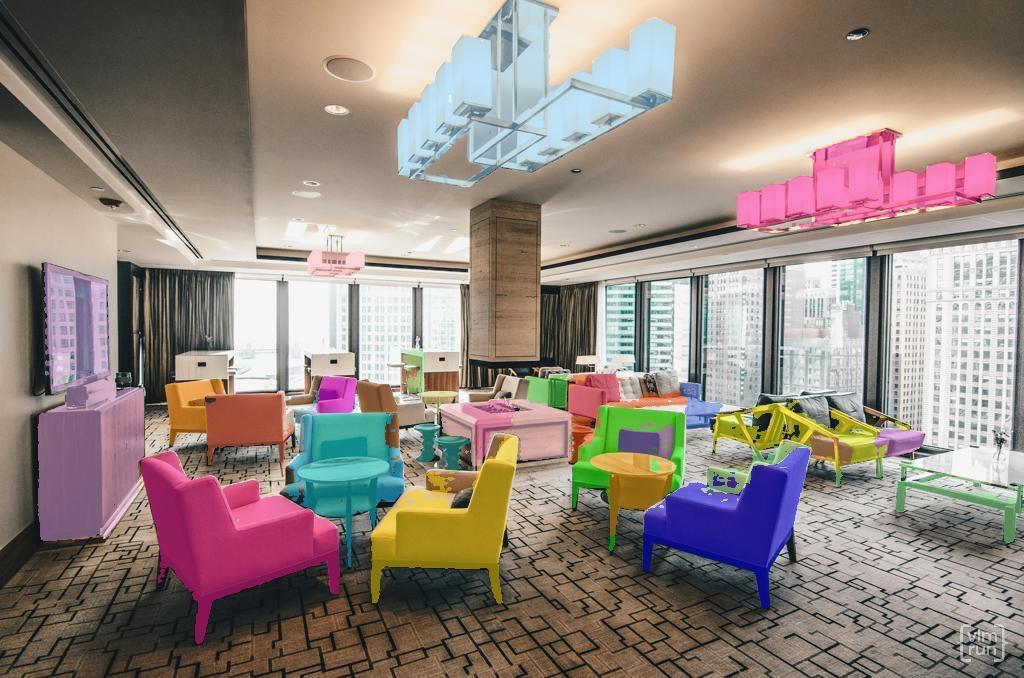}
        \caption{Furniture Segmentation}
    \end{subfigure}

    \caption{Semantic and instance segmentation examples with pixel-level masks.}
    \label{fig:segmentation_examples}
\end{figure}

\subsubsection{Spatial Localization and Pointing}
Performs keypoint detection, classification, and 2D localization tasks, including object counting, saliency detection, and fine-grained spatial reasoning. \textbf{Orion} enables precise identification of visual anchors and regions of interest within images. Keypoint detections are returned in normalized xy format.

\begin{figure}[H]
    \centering
    \begin{subfigure}{0.48\textwidth}
        \raggedright\small\question{Point to all the donuts in the image and localize them.}\vspace{0.5em}
        \centering
        \includegraphics[height=90px]{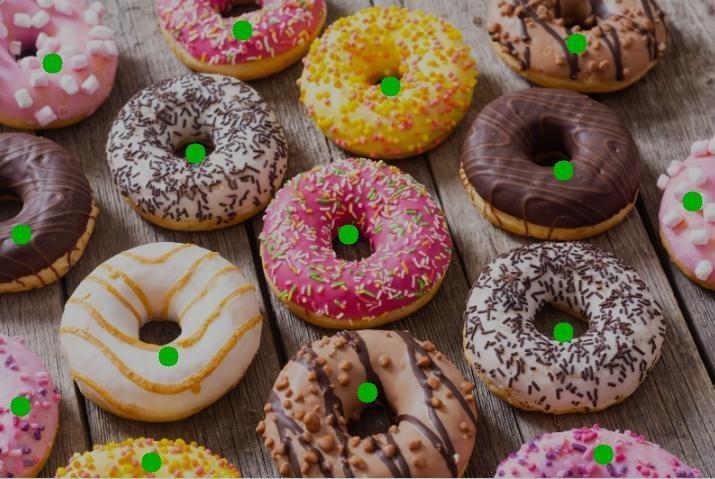}
        \caption{Object Keypoint Detection}
    \end{subfigure}
    \hfill
    \begin{subfigure}{0.48\textwidth}
        \raggedright\small\question{Point to all the people in the image and localize them.}\vspace{0.5em}
        \centering
        \includegraphics[height=90px]{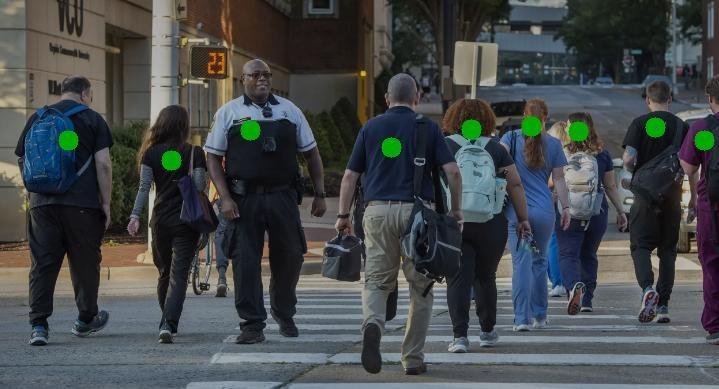}
        \caption{Person Keypoint Detection}
    \end{subfigure}
    
    \vspace{0.5em}
    
    \begin{subfigure}{0.48\textwidth}
        \raggedright\small\question{Point to all the faces in the image and localize them.}\vspace{0.5em}
        \centering
        \includegraphics[height=90px]{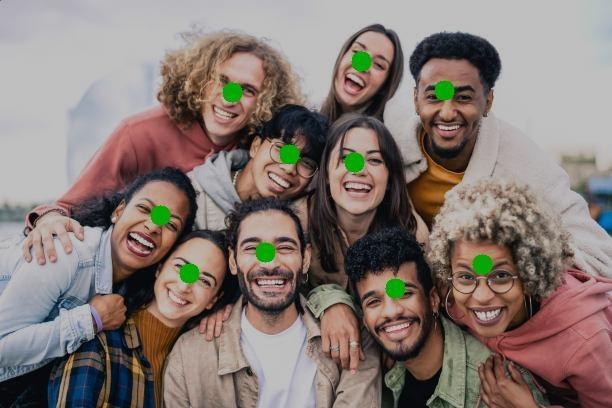}
        \caption{Face Keypoint Detection}
    \end{subfigure}
    \hfill
    \begin{subfigure}{0.48\textwidth}
        \raggedright\small\question{Point to all the eyes in the image.}\vspace{0.5em}
        \centering
        \includegraphics[height=90px]{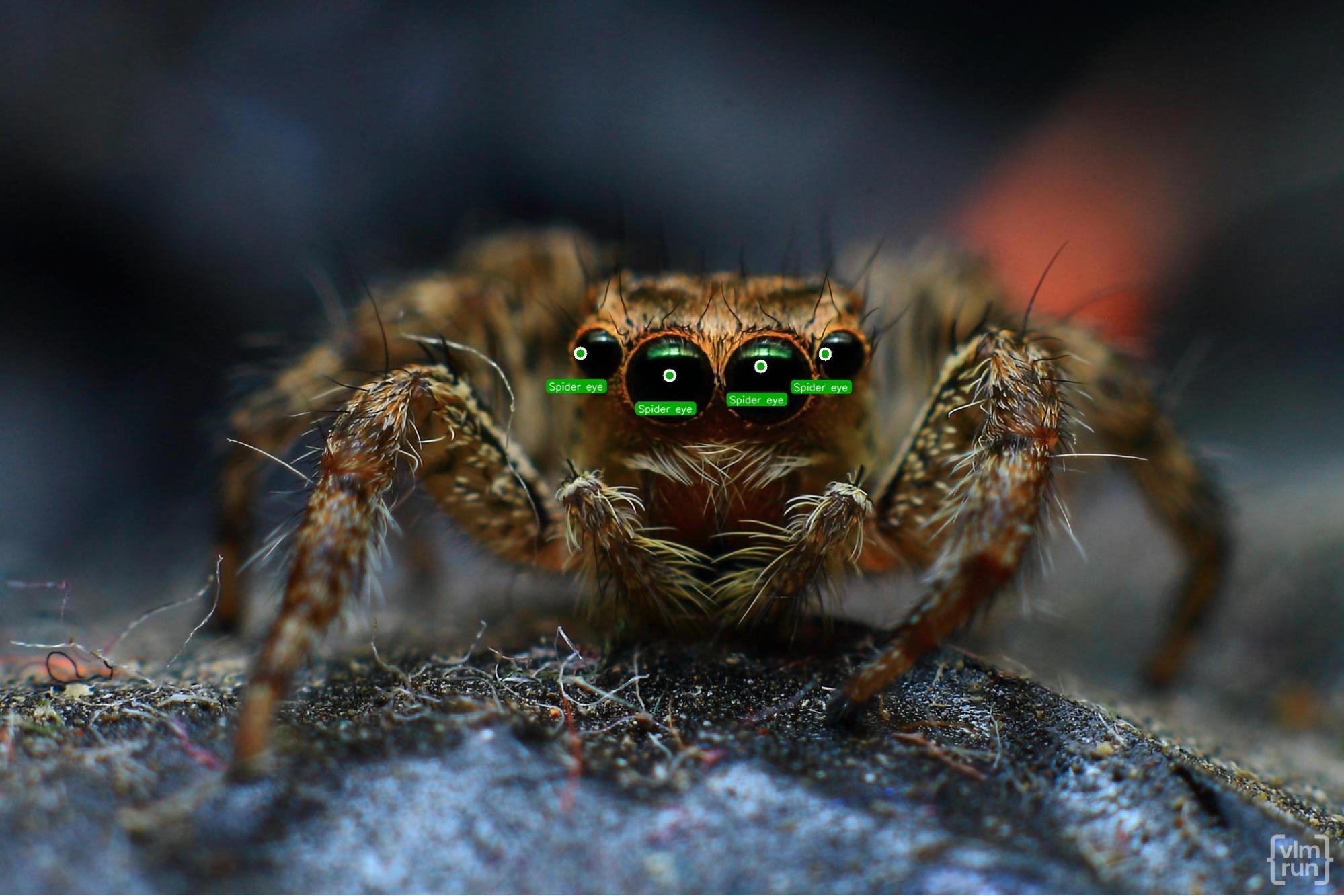}
        \caption{Eye Keypoint Detection}
    \end{subfigure}

    \caption{Keypoint detection and localization across different object categories.}
    \label{fig:keypoint_examples}
\end{figure}

\subsubsection{Image Generation and Editing}
Generate, edit, remix and transform images leveraging image-generation tools. We currently support the following capabilities under \textbf{Orion}, with more coming soon:
\begin{itemize}
    \item \textbf{Text-to-Image}: Generate high-quality images directly from natural language prompts, supporting diverse artistic styles and semantic specifications.
    \item \textbf{Image-to-Image}: Edit existing images based on reference images and textual instructions, enabling controlled visual transformations and content-aware modifications.
    \item \textbf{Image Inpainting}: Reconstruct or replace specified regions within images using contextual cues and user-provided prompts, supporting seamless content removal and insertion.
    \item \textbf{Style Transfer}: Transfer the artistic or visual style of one image onto another, preserving content structure while adapting stylistic attributes.
\end{itemize}

Notable, each of the above image generation capabilities are not restricted to single image inputs, but can support multiple inputs or inputs derived from outputs of other tools allowing for true compositionality. 

\begin{figure}[H]
    \centering
    \begin{subfigure}{0.24\textwidth}
        \begin{minipage}{\textwidth}
        \raggedright\small\question{Generate an image of a cat flying through a beautiful sky with clouds.}\vspace{0.5em}
        \end{minipage}
        \centering
        \includegraphics[height=80px]{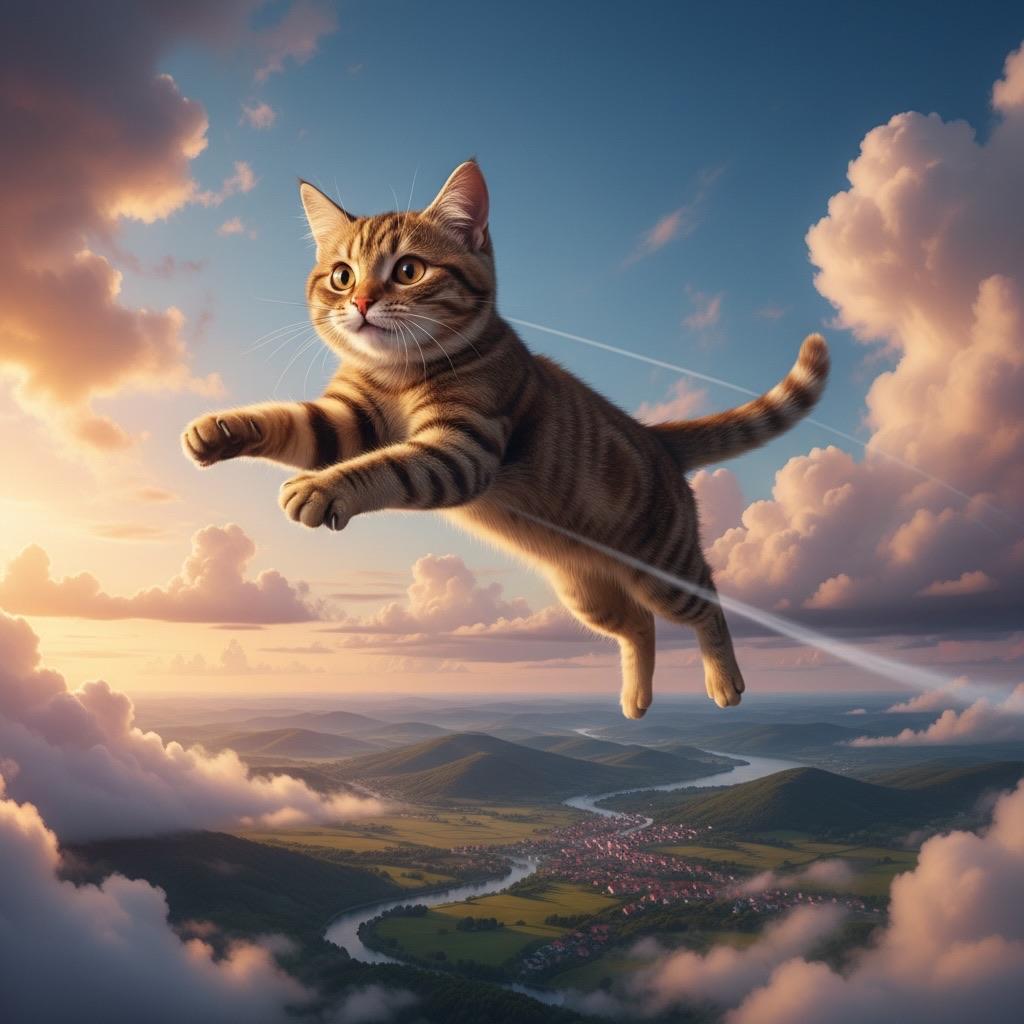}
        \caption{Text-to-Image}
    \end{subfigure}
    \hfill
    \begin{subfigure}{0.24\textwidth}
        \begin{minipage}{\textwidth}
        \raggedright\small\question{Generate the image of my dog flying through the sky with superman cape, with majestic background. My cat is on the left watching him.}\vspace{0.5em}
        \includegraphics[height=35px]{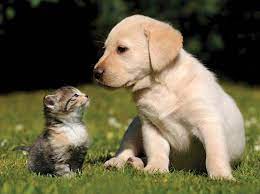}
        \end{minipage}
        \centering
        \includegraphics[height=80px]{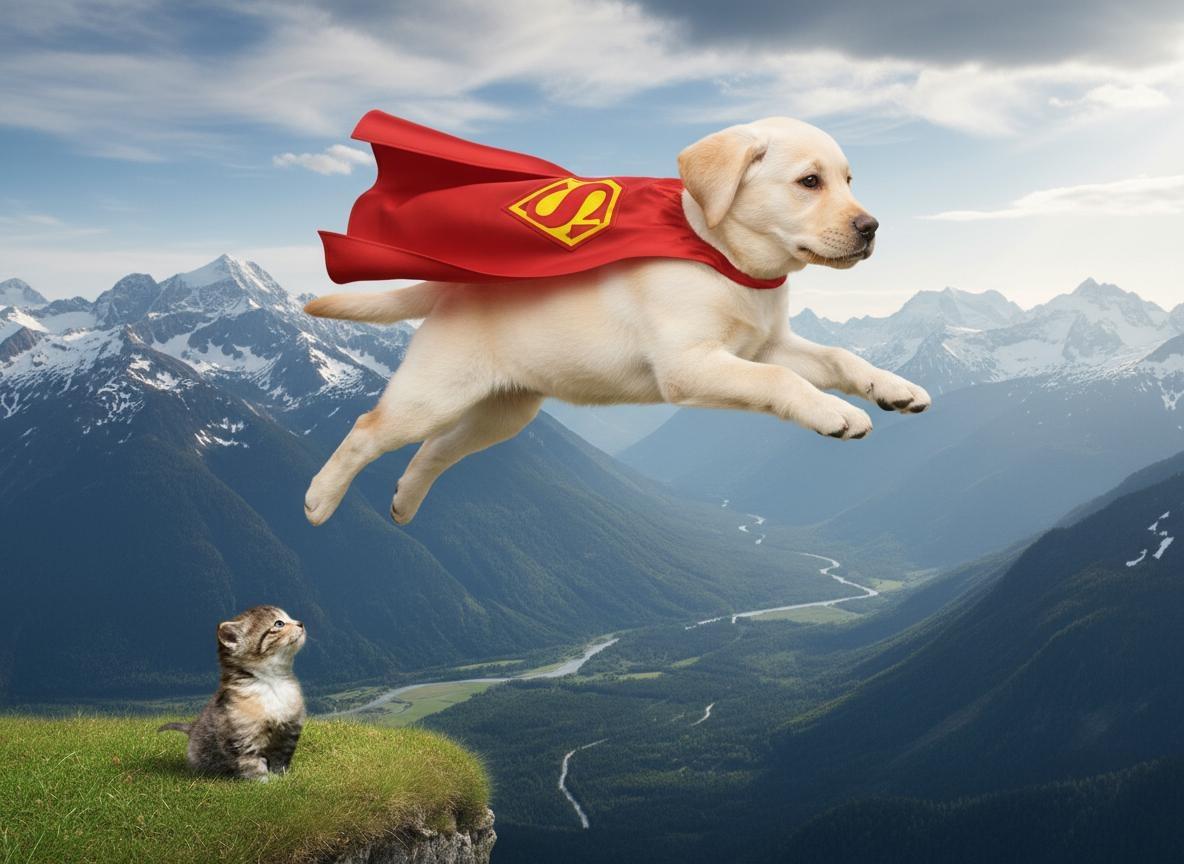}
        \caption{Image-to-Image}
    \end{subfigure}
    \hfill
    \begin{subfigure}{0.24\textwidth}
        \begin{minipage}{\textwidth}
        \raggedright\small\question{Make this dog drive the car in the image.}\vspace{0.5em}
        \end{minipage}
        \centering
        \includegraphics[height=35px]{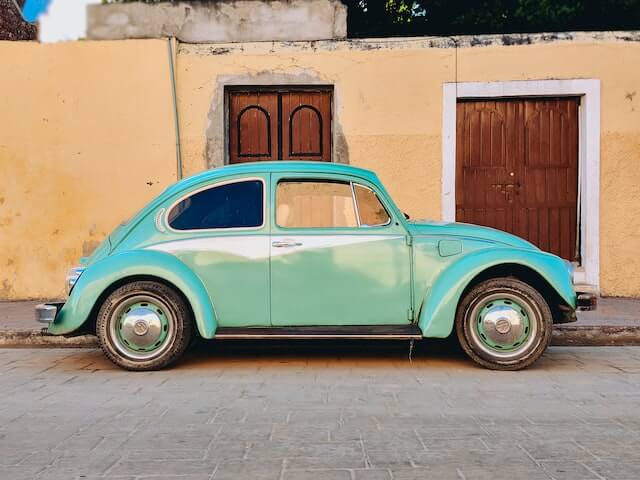}
        \includegraphics[height=35px]{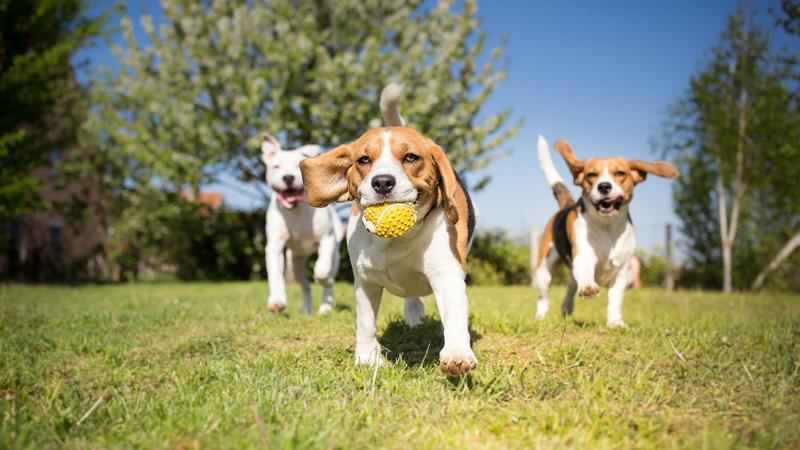}\\
        \includegraphics[height=80px]{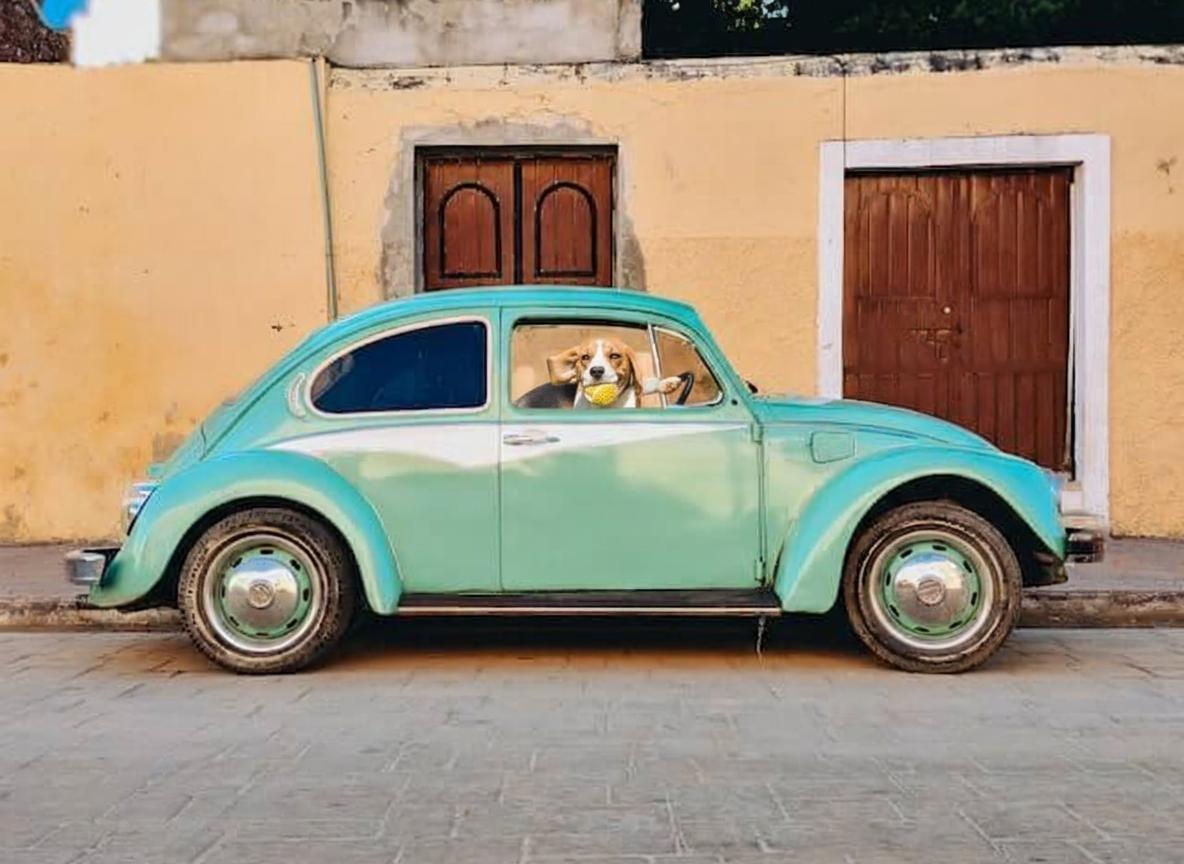}
        \caption{Image Inpainting}
    \end{subfigure}
    \hfill
    \begin{subfigure}{0.24\textwidth}
        \begin{minipage}{\textwidth}
        \raggedright\small\question{Generate this Golden Gate Bridge view in the style of the Van Gogh painting.}\vspace{0.5em}
        \end{minipage}
        \centering
        \includegraphics[height=80px]{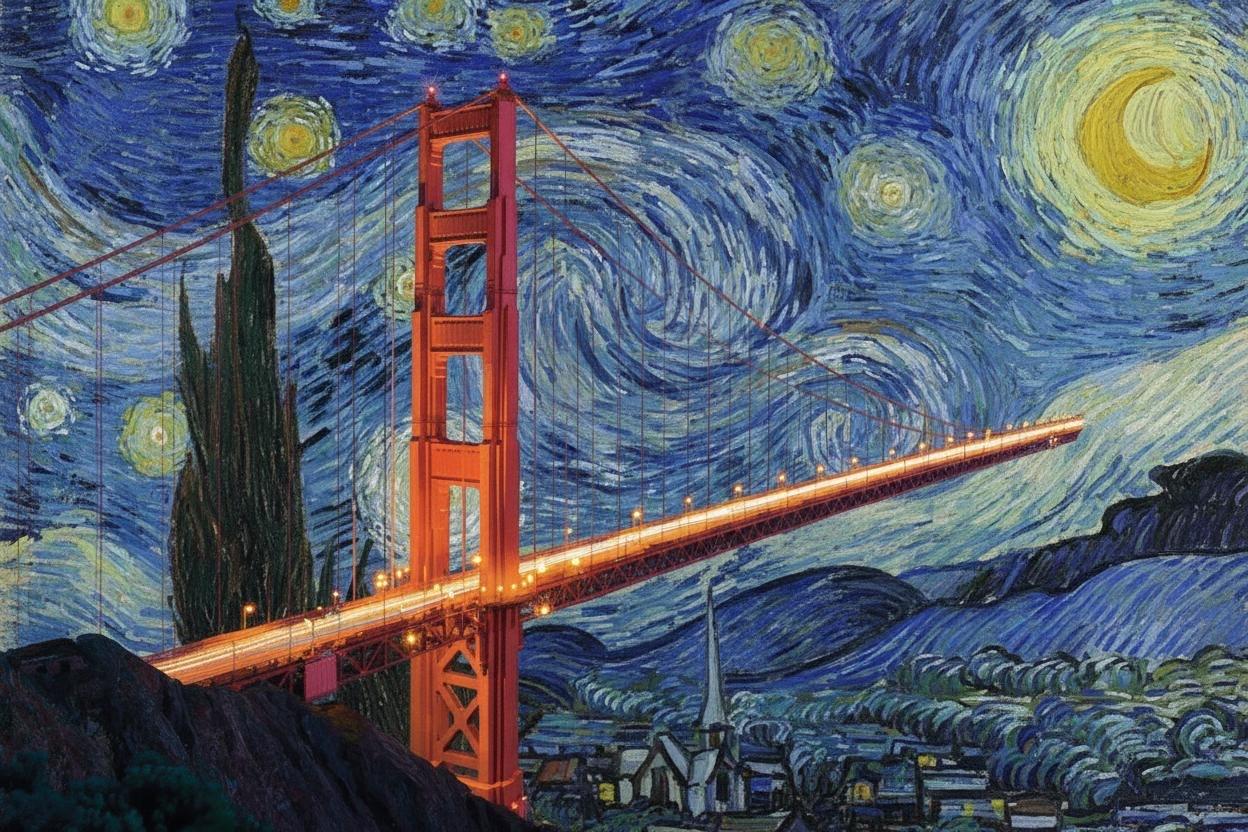}
        \caption{Style Transfer}
    \end{subfigure}

    \caption{Image generation and editing capabilities including text-to-image synthesis, image transformation, inpainting, and style transfer.}
    \label{fig:generation_examples}
\end{figure}

\subsubsection{Image OCR}
Robust in-the-wild optical character recognition (OCR) capable of extracting text, bounding boxes and word-level confidence scores from challenging visual contexts, including license plates, street signs, logos, product labels, and other unstructured text sources. \textbf{Orion} handles diverse fonts, orientations, and lighting conditions.
\begin{figure}[H]
\centering
\begin{subfigure}[t]{0.45\textwidth}
    \begin{minipage}{\textwidth}    
    \raggedright\small\question{Read the text in the image.}\vspace{0.5em}
    \end{minipage}
    \centering
    \includegraphics[width=\textwidth]{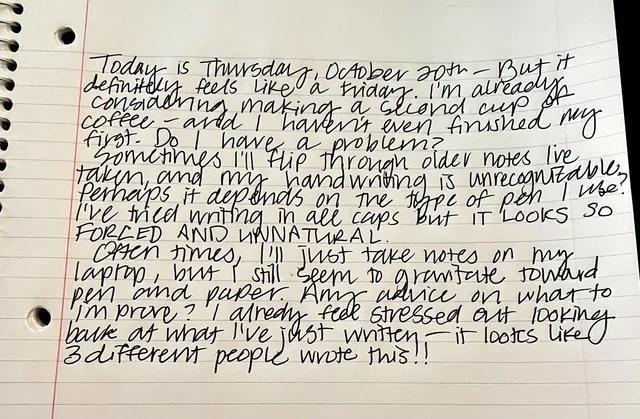}
    \caption{Handwritten Text Extraction\vspace{0.5em}\\
    \begin{minipage}{\textwidth}
    \raggedright\small\answer{Today is Thursday, October 20th – But it definitely feels like a Friday. I'm already considering making a second cup of coffee – and I haven't even finished my first. Do I have a problem? Sometimes I'll flip through older notes I've taken, and my handwriting is unrecognizable. Perhaps it depends on the type of pen I use. I've tried writing in all caps but IT LOOKS SO FORCED AND UNNATURAL. Often times, I'll just take notes on my laptop, but I still seem to gravitate toward pen and paper. Any advice on what to improve? I already feel stressed out looking back at what I've just written – it looks like 3 different people wrote this!!}
    \end{minipage}}
\end{subfigure}
\hfill
    \begin{subfigure}[t]{0.45\textwidth}
    \begin{minipage}{\textwidth}
    \raggedright\small\question{Read the text in the image.}\vspace{0.5em}
    \end{minipage}
    \centering
    \includegraphics[width=\textwidth]{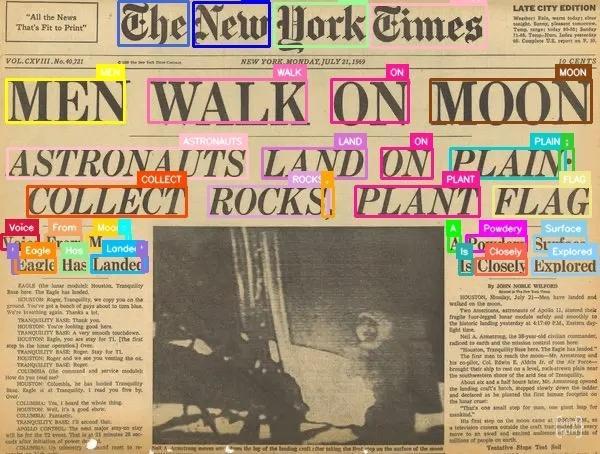}
    \caption{News Article Text Extraction\vspace{0.5em}\\
    \begin{minipage}{\textwidth}
    \raggedright\small\answer{All the News
    That's Fit to Print
    The New York Times
    LATE CITY EDITION
    Werdert Fait, want today'y aleer
    0-30
    35-46. Tengo-Hum, Lodre yesterday
    4A Complete V.I. rport en 7. 30.
    VOL.CXVIII.No.40,321
    NEW YORK, MONDAY, JULY 21, 1959
    10 CENTS
    MEN WALK ON MOON
    ASTRONAUTS LAND ON PLAIN;
    COLLECT ROCKS, PLANT FLAG.
    Voice From Moon:
    "Eagle Has Landed"
    \dots (truncated for brevity).
    }
    \end{minipage}}
\end{subfigure}

\caption{In-the-wild OCR examples showing robust text extraction from challenging visual contexts with diverse fonts and layouts.}
\label{fig:ocr_example}
\end{figure}

\subsubsection{UI Parsing}
Automatically parse and extract user interface elements from screenshots or application images, including buttons, text fields, links, icons, and other interactive components. \textbf{Orion} supports accessibility workflows, automated testing, and interface analysis. UI bounding boxes are returned in normalized xywh format.

\begin{figure}[H]
    \centering
    \begin{subfigure}{0.45\textwidth}
        \begin{minipage}{\textwidth}
        \raggedright\small\question{Detect and visualize all the UI elements in this screenshot.}\vspace{0.5em}
        \end{minipage}
        \centering
        \includegraphics[width=\textwidth]{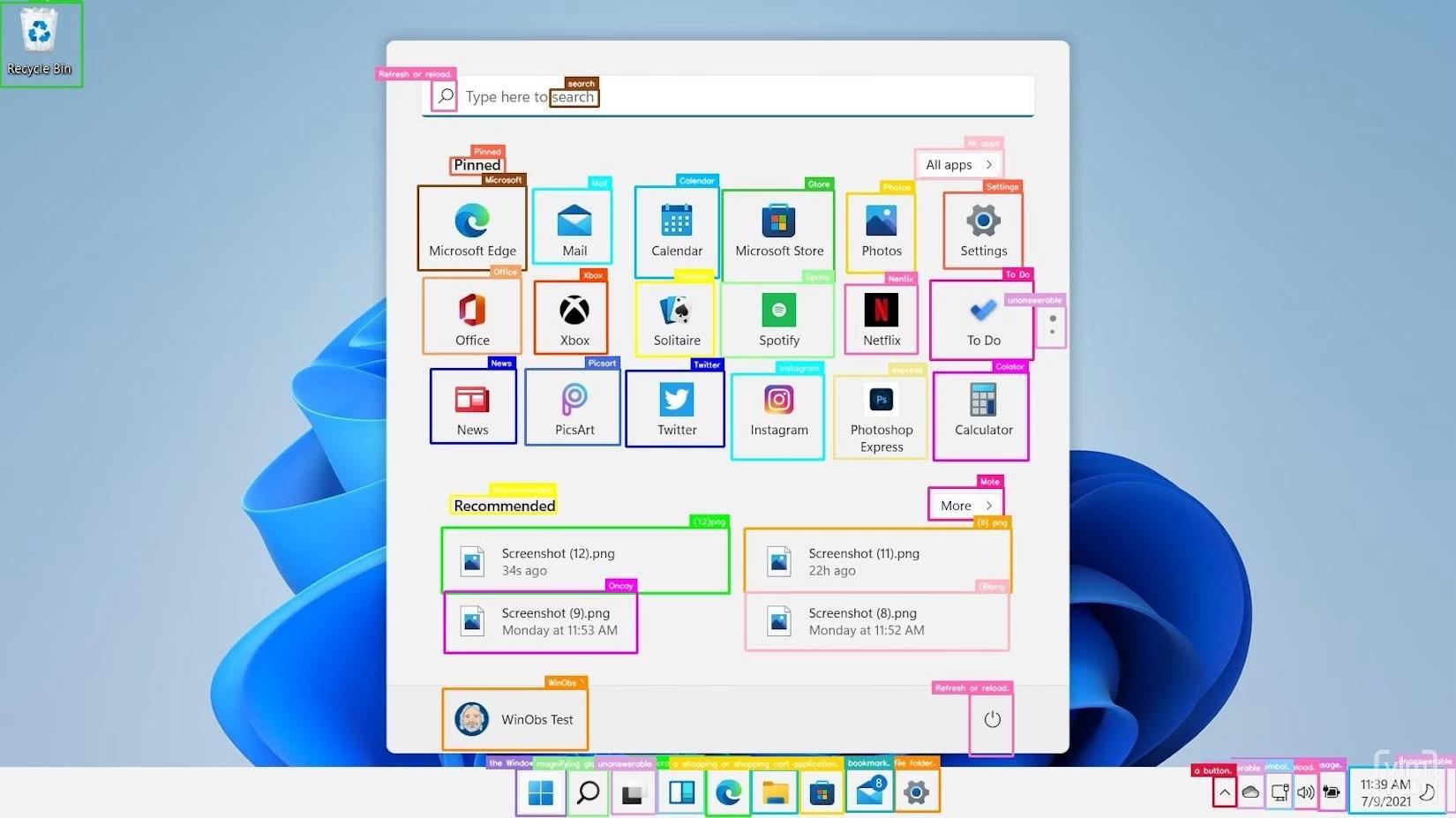}
        \caption{UI Element Detection and Parsing}
    \end{subfigure}
    \hfill
    \begin{subfigure}{0.45\textwidth}
        \begin{minipage}{\textwidth}
        \raggedright\small\question{Detect all the main UI elements excluding the sidebar and visualize the boxes.}\vspace{0.5em}
        \end{minipage}
        \centering
        \includegraphics[width=\textwidth]{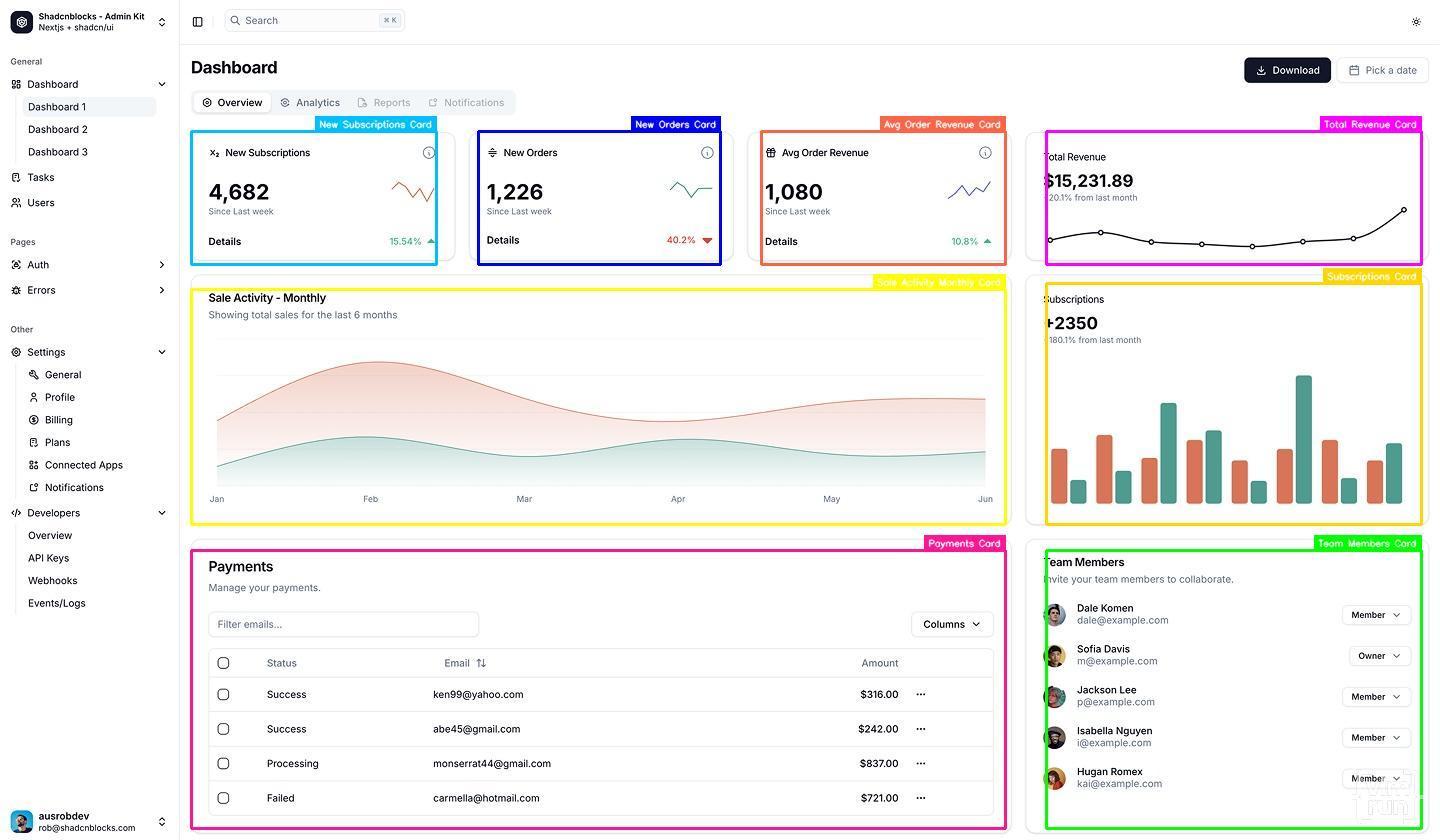}
        \caption{Card-based Interface}
    \end{subfigure}
    
    \caption{UI parsing examples showing automatic extraction of interface elements including buttons, text fields, and interactive components.}
    \label{fig:ui_parsing_example}
\end{figure}

\subsubsection{Image Tools}
Extend and compose visual workflows with pre-processing tools such as cropping, rotation, blurring, super-resolution, and other transformations. Users will soon be able to integrate custom image processing modules to extend the system's analytical and generative capabilities.

\begin{figure}[H]
    \centering
    \begin{subfigure}{0.24\textwidth}
        \centering
        \includegraphics[width=\textwidth]{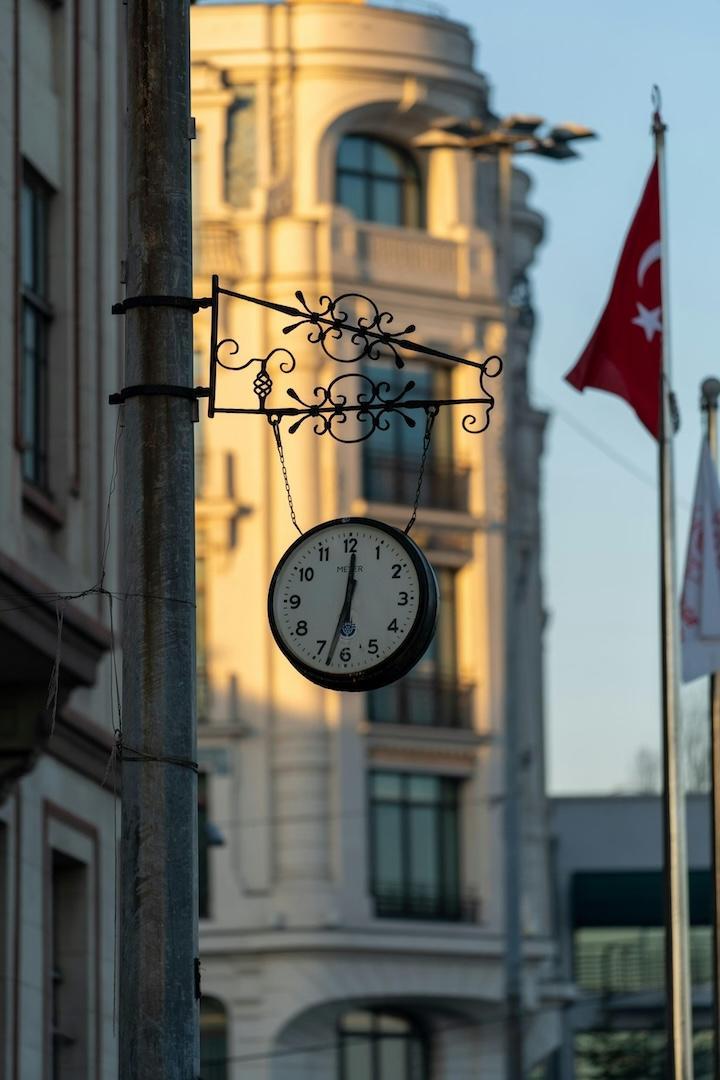}
        \caption{Original Image}
    \end{subfigure}
    \hfill
    \begin{subfigure}{0.24\textwidth}
        \centering
        \includegraphics[width=\textwidth]{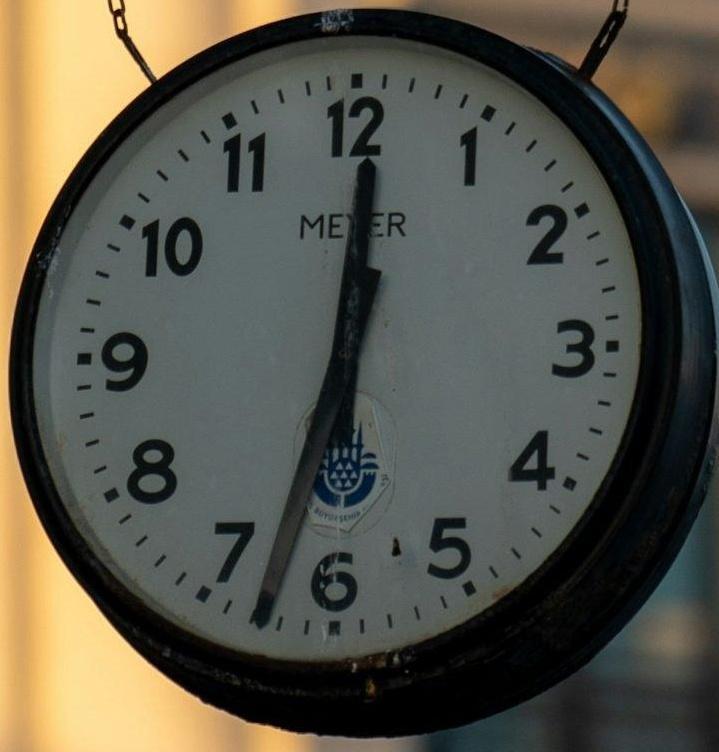}
        \caption{Cropped Image}
    \end{subfigure}
    \hfill
    \begin{subfigure}{0.24\textwidth}
        \centering
        \includegraphics[width=\textwidth]{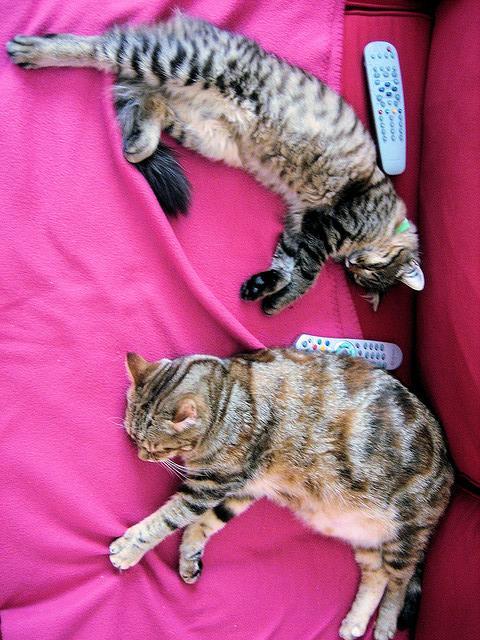}
        \caption{Rotated Image}
    \end{subfigure}
    \hfill
    \begin{subfigure}{0.24\textwidth}
        \centering
        \includegraphics[width=\textwidth]{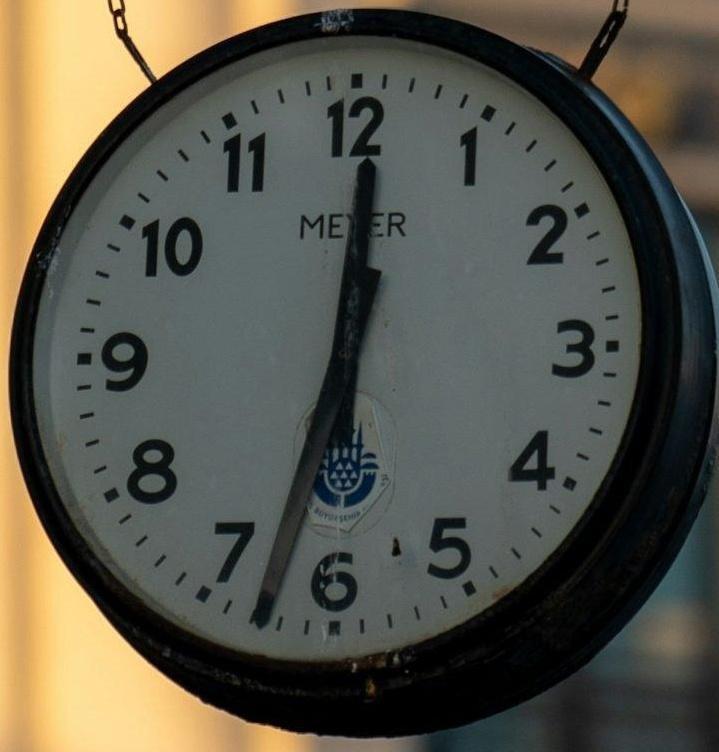}
        \caption{Crop + VQA}
    \end{subfigure}

    \vspace{1em}
    \begin{minipage}{0.9\textwidth}
    \small\question{(a) Crop the clock in this image (b) Rotate the image 90 degrees counterclockwise.}\\
    \answer{(b) I have cropped to the specific region. (c) I have rotated the image by 90 degrees counter-clockwise.}\\
    \\
    \small\question{(a) Crop into the clock in the image and extract the time shown.}\\
    \answer{(d) I have cropped the clock region from the image and extracted the time shown, providing both the cropped image and the extracted time value.}
    \end{minipage}

    \caption{Image transformation tools including cropping, rotation, enhancement, and time extraction.}
    \label{fig:image_tools_examples}
\end{figure}

\noindent
These capabilities enable \textbf{Orion} to serve as a versatile visual intelligence platform, supporting wide range of visual AI applications.

\newpage
\subsection{Document Understanding, Reasoning and Tool-Use}
\noindent
\textbf{\textbf{Orion}} provides extensive document understanding capabilities, supporting a variety of complex structure-aware tasks and extractive workflows. The system is designed to handle both scanned and digital documents, enabling robust analysis across diverse document types and formats. Our document understanding capabilities can be organized into the following functional categories:

\subsubsection{Document Layout Understanding}
Robustly analyze complex document layouts, extracting structural hierarchies including paragraphs, headers, tables, footnotes, and other semantic elements from both scanned and digital sources. The system preserves reading order and spatial relationships to maintain document coherence. Detection bounding boxes are returned in normalized xywh format.

\begin{figure}[H]
    \centering
    \begin{subfigure}{0.24\textwidth}
        \centering
        \includegraphics[width=\textwidth]{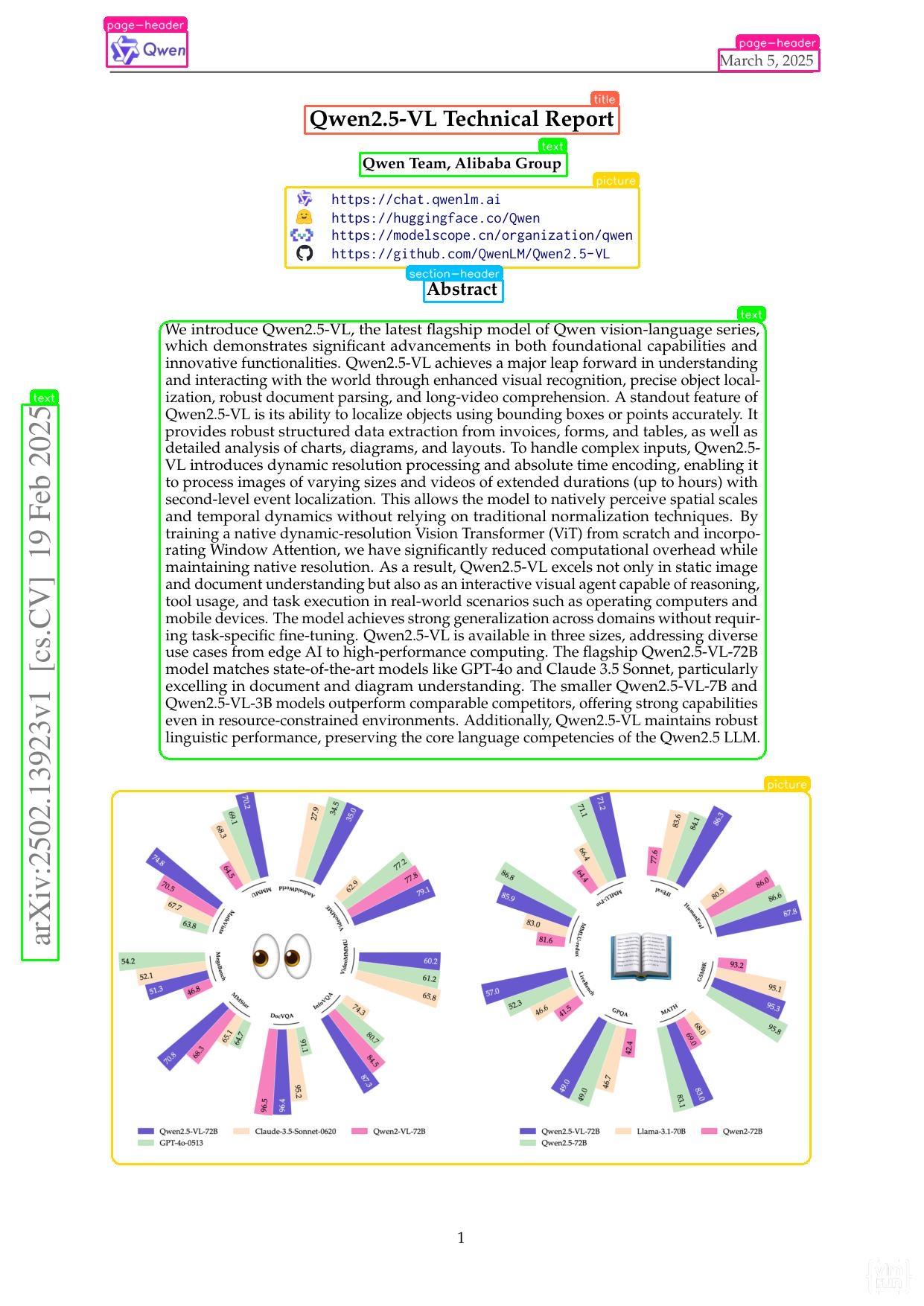}
        \caption{Layout Example 1}
    \end{subfigure}
    \hfill
    \begin{subfigure}{0.24\textwidth}
        \centering
        \includegraphics[width=\textwidth]{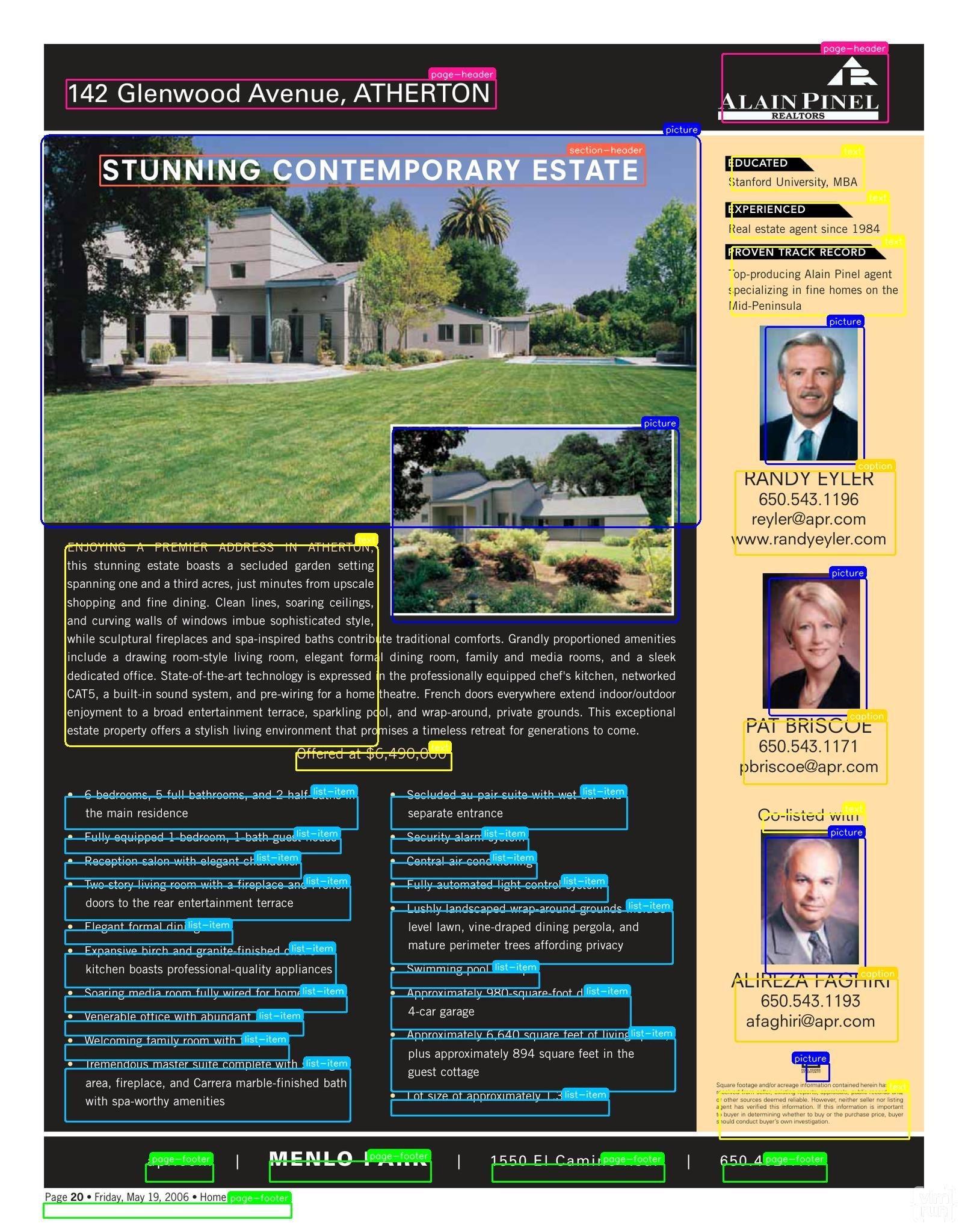}
        \caption{Layout Example 2}
    \end{subfigure}
    \hfill
    \begin{subfigure}{0.24\textwidth}
        \centering
        \includegraphics[width=\textwidth]{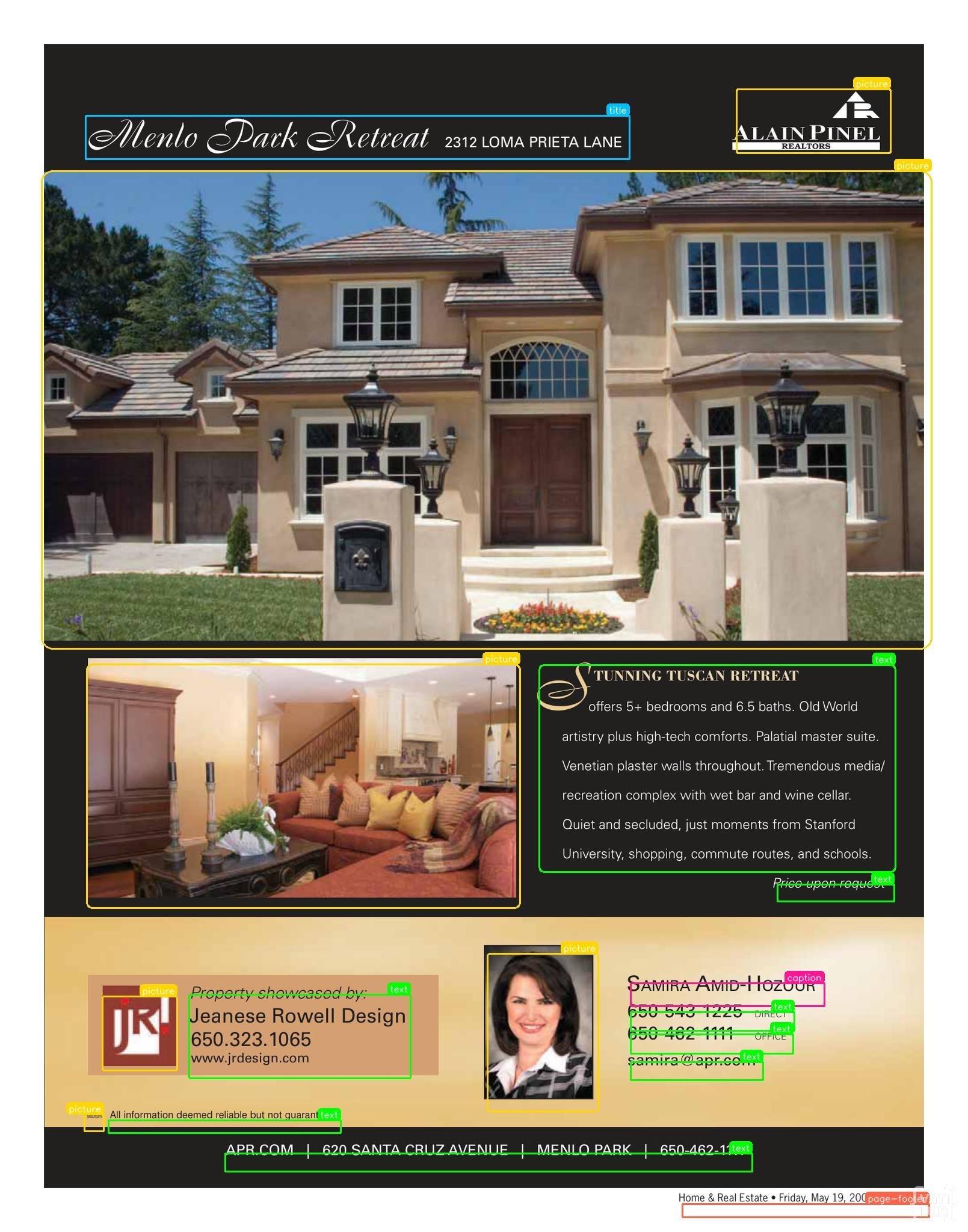}
        \caption{Layout Example 3}
    \end{subfigure}
    \hfill
    \begin{subfigure}{0.24\textwidth}
        \centering
        \includegraphics[width=\textwidth]{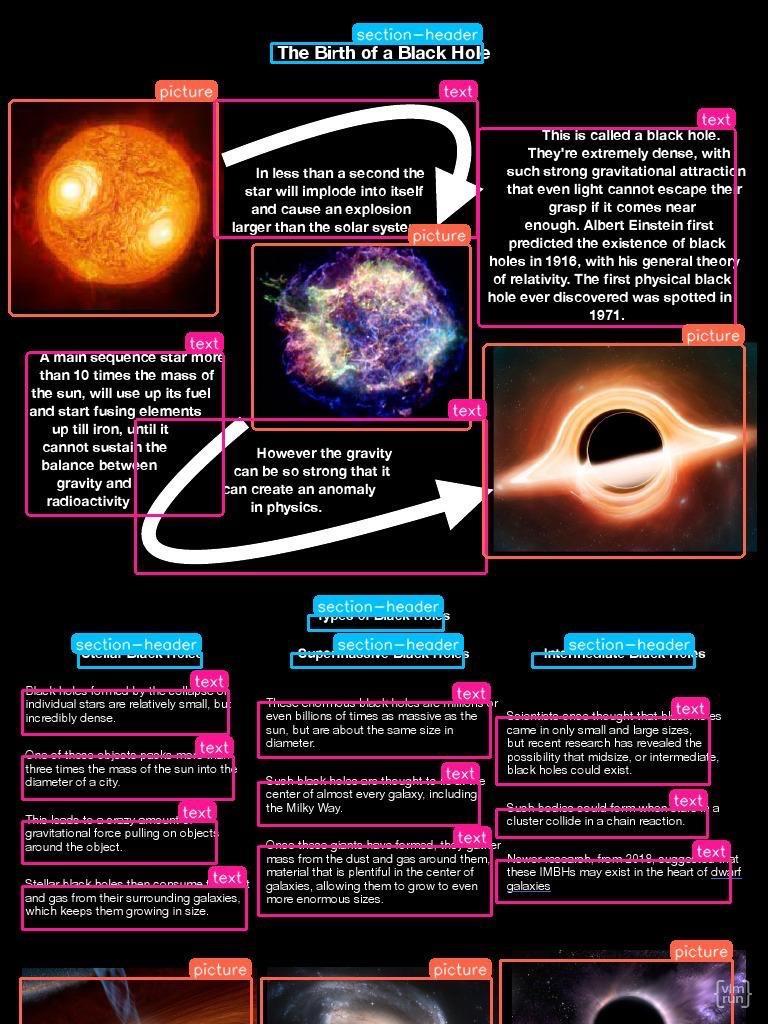}
        \caption{Layout Example 4}
    \end{subfigure}

    \vspace{1em}
    \begin{minipage}{0.9\textwidth}
    \small\question{Analyze the document layout and extract its structural elements.}\\
    \small\answer{I have analyzed the document layout and extracted the structural hierarchy, identifying headers, paragraphs, tables, footnotes, and other semantic elements while preserving the reading order and spatial relationships.}\\
    \\
    \small\question{Get the layout of the document and visualize all the text.}\\
    \small\answer{I have extracted the complete document layout and visualized all text elements with their precise locations, showing the spatial organization of content including text blocks, titles, and formatting structures.}
    \end{minipage}

    \caption{Document layout analysis showing detection of structural elements including headers, paragraphs, tables, and figures with preserved reading order.}
    \label{fig:layout_examples}
\end{figure}

\subsubsection{Document OCR}
Perform high-accuracy optical character recognition with specialized support for structured forms, tables, and handwritten text. The system handles both printed and handwritten content, enabling robust text extraction across diverse document types and quality levels.
\begin{itemize}
    \item \textbf{Form OCR}: Localize form fields, tables, checkboxes, signatures, and visual anchors in documents---enabling workflows such as form field extraction and field-value alignment, including multi-page analysis and cross-page field tracking.
\end{itemize}

\begin{figure}[H]
    \centering
    \begin{subfigure}{0.4\textwidth}
        \begin{minipage}{\textwidth}
        \raggedright\small\question{Localize form fields and align them with their values.}\vspace{0.5em}
        \end{minipage}
        \centering
        \includegraphics[height=240px]{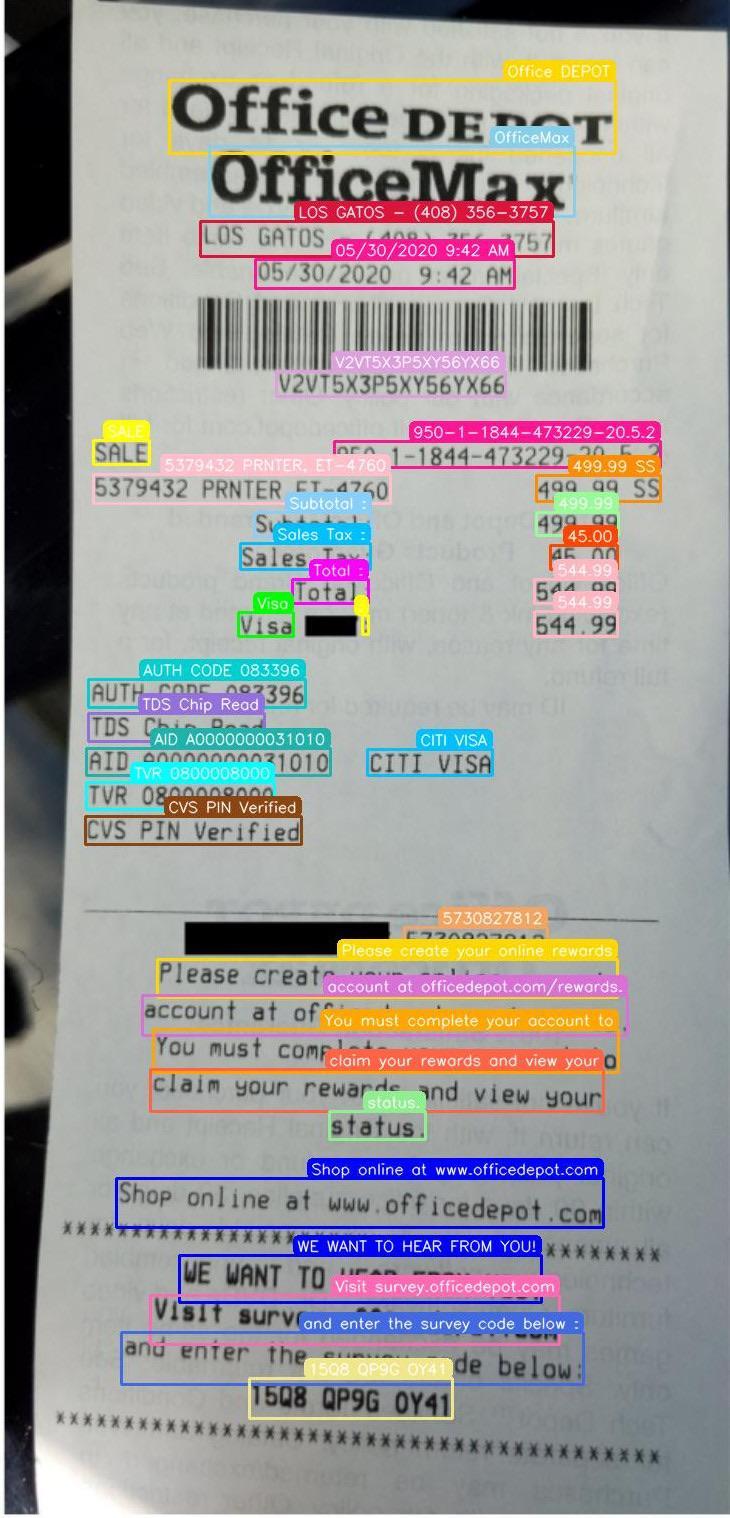}
        \caption{Form Extraction}
    \end{subfigure}
    \hfill
    \begin{subfigure}{0.55\textwidth}
        \begin{minipage}{\textwidth}
        \raggedright\small\question{Parse this document and extract all the handwritten fields. Detect the locations of each of the fields and visualize them.}\vspace{0.5em}
        \end{minipage}
        \centering
        \includegraphics[height=240px]{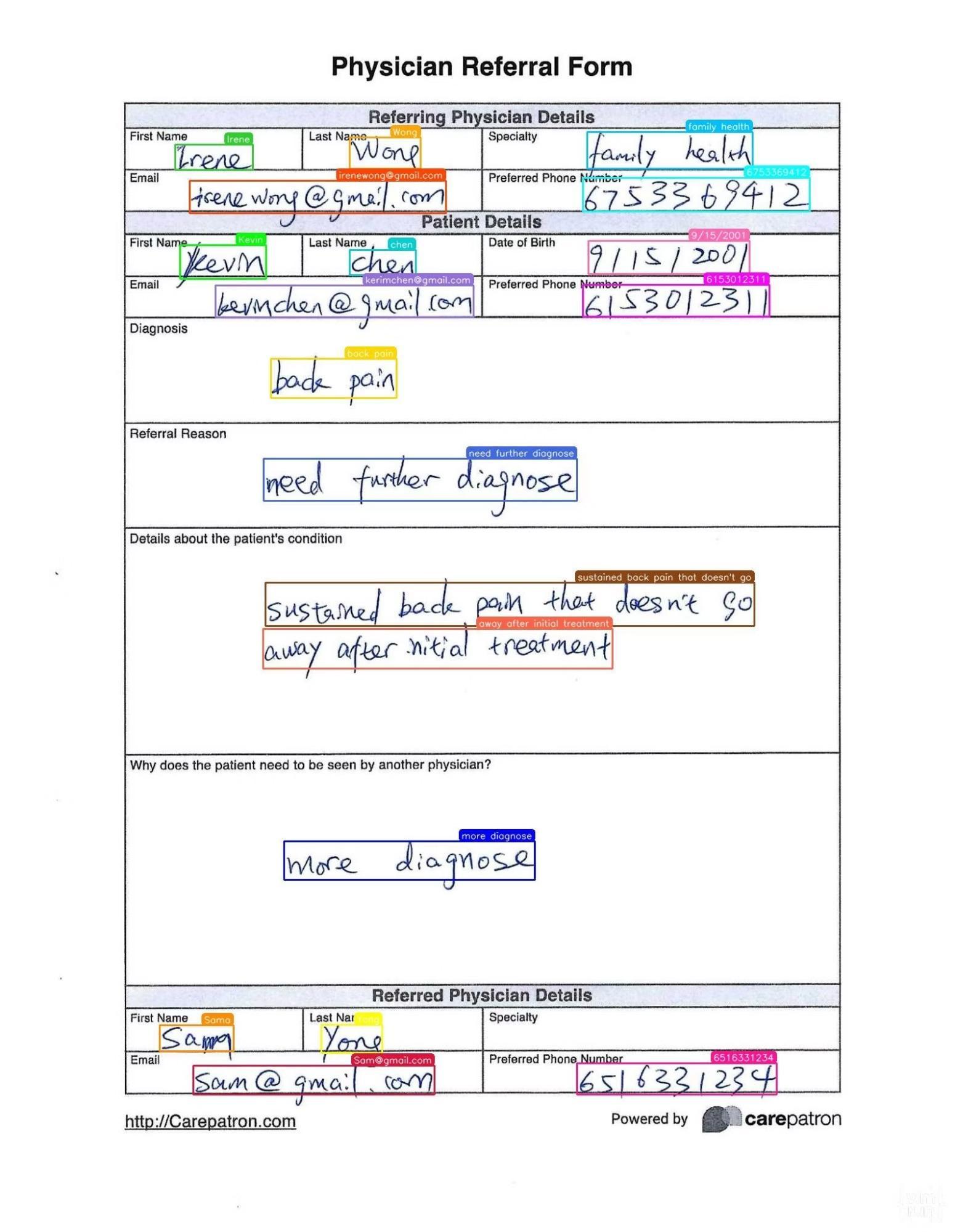}
        \caption{Handwritten Form Extraction}
    \end{subfigure}
    
    \caption{Visual grounding examples showing form field localization and field-value alignment with bounding box coordinates.}
    \label{fig:visual_grounding_example}
\end{figure}

\subsubsection{Document Tools}
Provide specialized modules for large-scale document processing, enabling efficient manipulation and analysis of extensive document collections. This includes intelligent pagination, in-place editing, and content redaction capabilities.
\begin{itemize}
    \item \textbf{Pagination}: Intelligently index and retrieve only relevant pages of a document, enabling efficient agentic processing of large documents (100+ pages) without loading entire files into memory.
    \item \textbf{Editing / Redaction}: Edit or redact the document in-place, enabling the agent to modify the document as it is being processed.
\end{itemize}

\noindent
These capabilities enable \textbf{orion} to serve as a comprehensive document intelligence platform, supporting use cases ranging from automated form processing and compliance workflows to large-scale document review and knowledge extraction.

\newpage
\subsection{Video Understanding, Reasoning and Tool-Use} \label{sec:video-understanding}

\noindent
\textbf{\textbf{Orion}} provides comprehensive capabilities for video analysis, generation, and tool-augmented workflows. The system supports a diverse range of video modalities and tasks, organized into the following functional categories:

\subsubsection{Video Captioning}
Automatically generate detailed descriptions and scene-level captions for entire videos and video segments. Supports multimodal prompts, per-frame analysis, and temporal alignment of captions with video timestamps, enabling fine-grained narrative generation.

\begin{figure}[H]
    \centering
    \begin{subfigure}{0.6\textwidth}
        \centering
        \includegraphics[width=\textwidth]{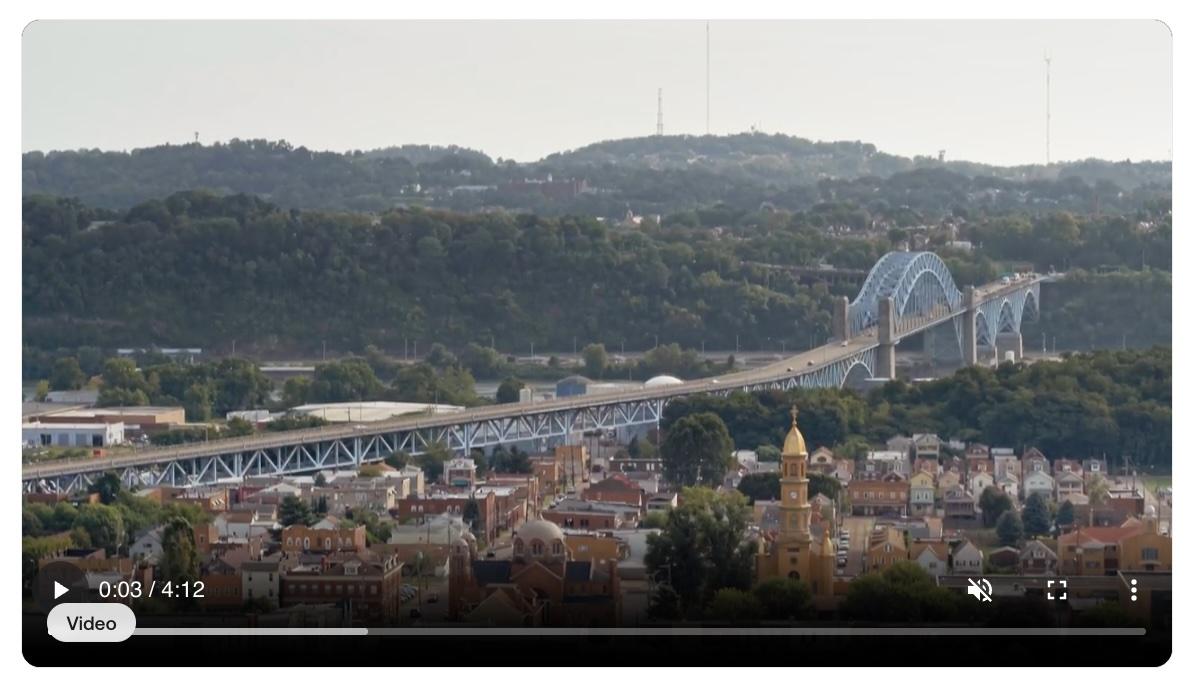}
        \caption{Video Input [\href{https://storage.googleapis.com/vlm-data-public-prod/hub/examples/video.agent/bakery.mp4}{link}]}
    \end{subfigure}
    \begin{center}
    \begin{minipage}{0.8\textwidth}
    \small\question{Parse the video and return a markdown summary with key scenes and timestamps.}\\
    {\begin{codebox}\color{vlmblue}
    \begin{minted}[breaklines, bgcolor=lightgray, tabsize=2]{markdown}
    
    Here is a markdown summary of the key scenes and timestamps:
    00:00 - 00:06: Establishing shot of a city with a large blue bridge and a town in the foreground.
    00:06 - 00:13: Scott Baker is interviewed in a modern bakery setting, discussing his family's long-standing connection to the community.
    00:10 - 00:15: Archival black and white footage shows a bustling bakery counter with cakes and pastries.
    00:14 - 00:16: A vintage black and white photo of a Jenny Lee Baked Goods delivery truck with two men.
    00:16 - 00:19: Exterior view of a "Jenny Lee Bakery" storefront.
    00:19 - 00:31: Donna is interviewed in a bakery setting, sharing her experiences.
    00:32 - 00:34: A vintage black and white photo of three men wearing chef hats.
    00:35 - 00:44: Bernie is interviewed in a bakery setting, reflecting on his career.
    00:39 - 00:44: A vintage black and white photo of children dressed in chef attire.
    00:45 - 00:48: Two men are seen reviewing documents in front of shelves stocked with bread.
    00:49 - 00:58: A train passes by under a large blue arched bridge.
    01:01 - 01:10: Donna recounts a difficult period for the business.
    ... (truncated for brevity).
    
    \end{minted}
    \end{codebox}
    }
    \end{minipage}
    \end{center}
    \caption{Video captioning with temporal grounding. See \href{https://chat.vlm.run/chat/4e6e2cd0-32c4-4c36-9234-a1c584f3e3ff}{full chat history}.}
    \label{fig:video_sampling_examples}
\end{figure}

\subsubsection{Temporal Grounding}
Localize and identify key moments or events within a video based on natural language instructions, providing precise start and end timestamps for any described action, object, or scene. This capability is particularly useful for tasks such as highlight extraction and video summarization. Temporal grounding is provided in the HH:MM:SS format and can also be used for sampling and trimming clips using the tools mentioned below. 

\begin{figure}[H]
    \centering
    \begin{subfigure}{0.45\textwidth}
        \begin{minipage}{\textwidth}
        \raggedright\small\question{Get the best highlight showing the most fireworks in this video.}\vspace{0.5em}
        \end{minipage}
        \centering
        \includegraphics[width=\textwidth]{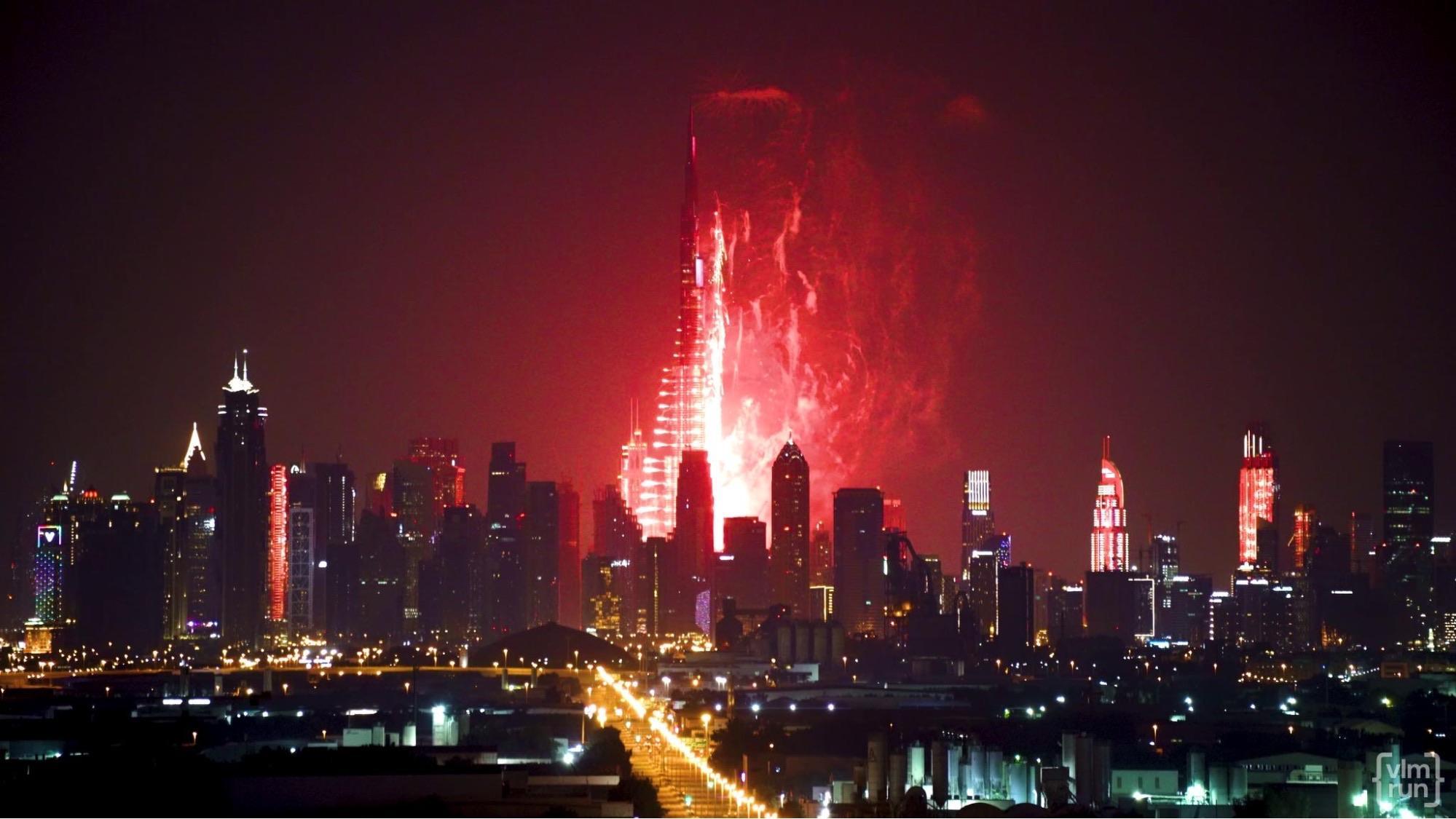}
        \caption{Fireworks highlight extraction\vspace{1em}\\
        \begin{minipage}{\textwidth}
        \small\answer{I have identified the most spectacular highlight of the fireworks display, featuring a breathtaking finale where the entire skyline is engulfed in a vibrant red glow, with precise timestamps (00:23 to 00:28).}
        \end{minipage}}
    \end{subfigure}
    \hfill
    \begin{subfigure}{0.45\textwidth}
        \begin{minipage}{\textwidth}
            \small\question{Step 1: Detect and highlight all people present in the image. Step 2: Apply deoldification techniques to restore true-to-life color and detail. Return a high-quality, artifact-free video clearly demonstrating the deoldified scene, focusing on preserving facial features and overall realism.}\vspace{0.5em}
        \end{minipage}
        \centering
        \includegraphics[width=\textwidth]{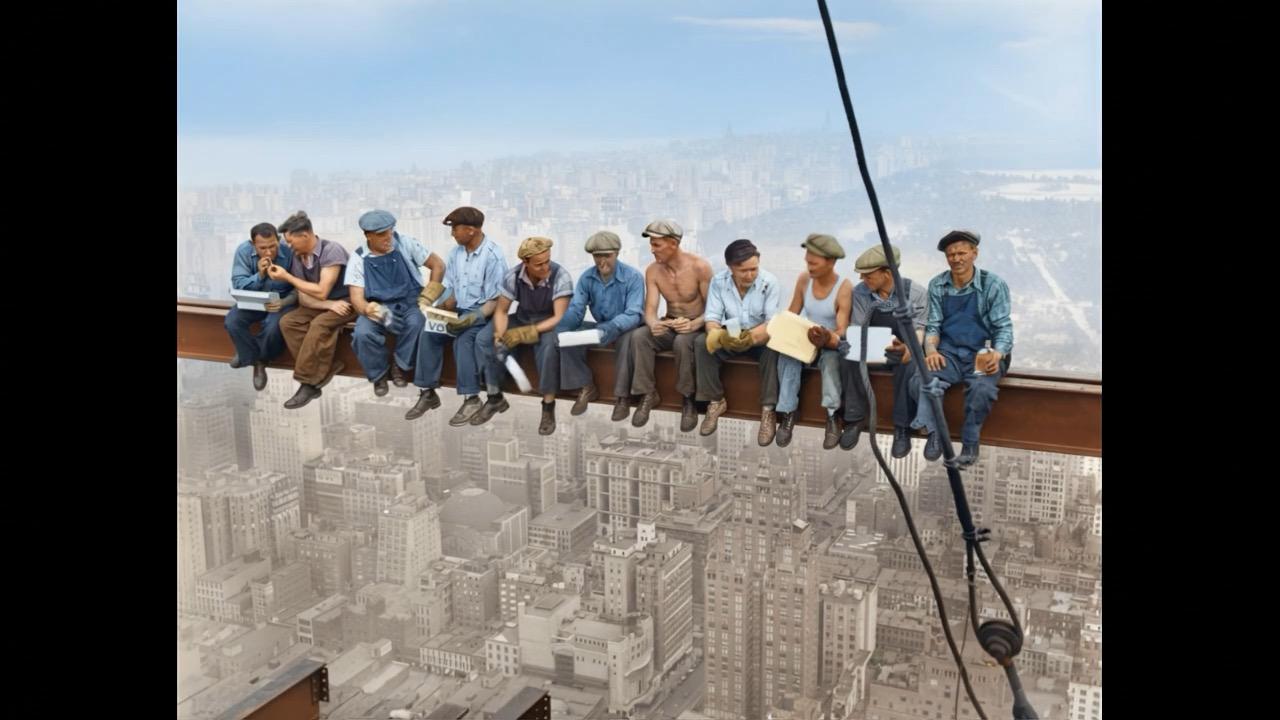}
        \caption{Video deoldification\vspace{1em}\\
        \begin{minipage}{\textwidth}
        \small\answer{I have detected and highlighted all people in the image, then applied deoldification techniques to restore true-to-life color and detail, producing a high-quality video demonstrating the enhanced scene with preserved facial features and overall realism.}
        \end{minipage}}
    \end{subfigure}

    \caption{Temporal grounding examples showing highlight extraction and video enhancement.}
    \label{fig:temporal_grounding_examples}
\end{figure}

\subsubsection{Video Generation}
Create short video sequences from text prompts or reference images, supporting open-domain video synthesis, style transfer, and dynamic scene animation.
    
\subsubsection{Video Tools}

Extensible toolkit for video editing and analysis, including frame sampling, highlight extraction, time-based trimming, and scene segmentation. Users can integrate custom video processing modules to extend the system's analytical and generative capabilities.
\begin{itemize}
    \item \textit{Frame Sampling}: Extract frames at regular intervals or specific timestamps for tasks such as thumbnail generation and visual analysis. See Figure~\ref{fig:video_sampling_examples}.
    \item \textit{Highlight Extraction}: Automatically identify and extract the most salient moments or key actions in a video, returning their start and end times along with segment URLs. See Figure~\ref{fig:video_sampling_examples}.
    \item \textit{Time-Based Trimming}: Precisely trims videos to specified time segments, supporting millisecond-level accuracy.
\end{itemize}
\begin{figure}[H]
    \centering
    \begin{subfigure}{0.24\textwidth}
        \centering
        \includegraphics[width=\textwidth]{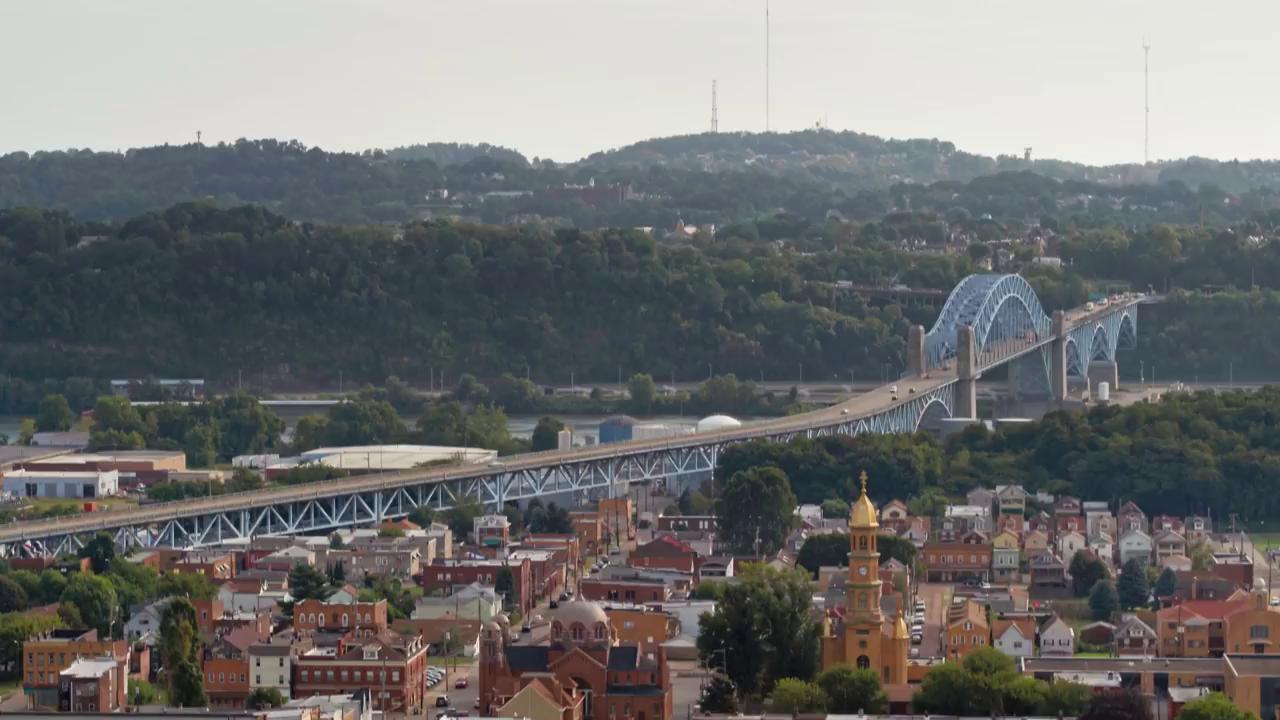}
        \caption{Frame 1}
    \end{subfigure}
    \hfill
    \begin{subfigure}{0.24\textwidth}
        \centering
        \includegraphics[width=\textwidth]{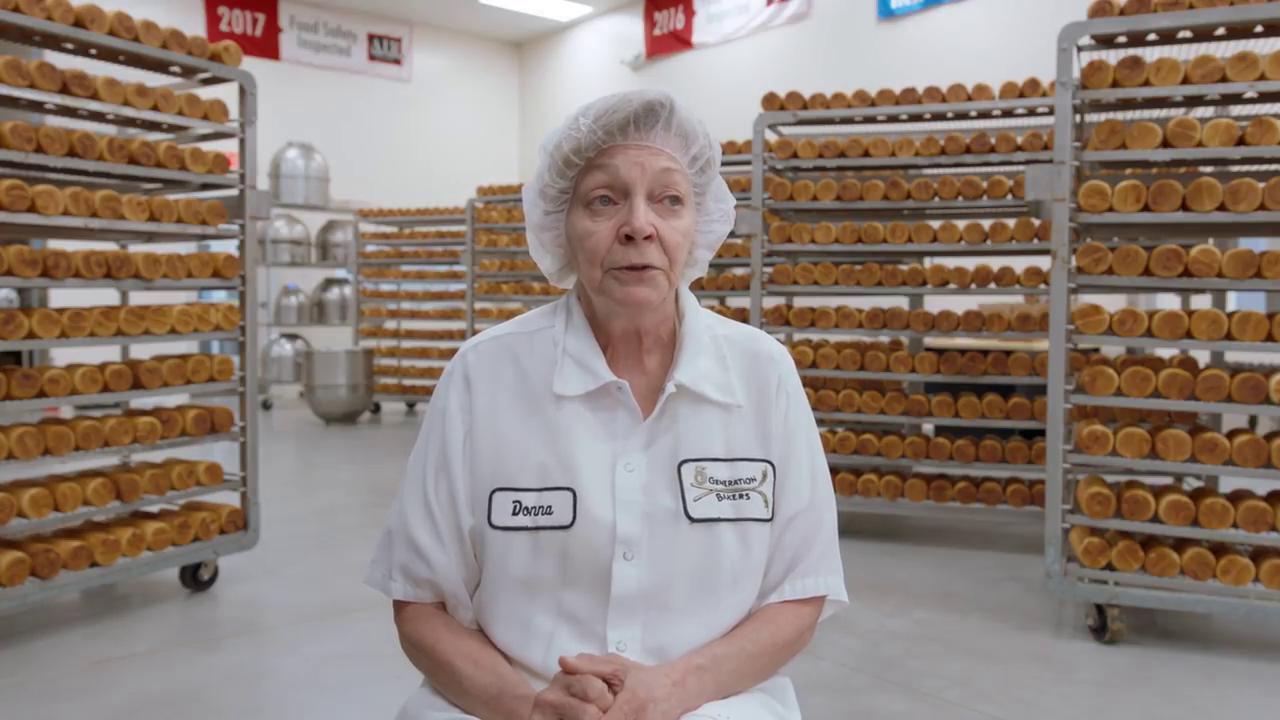}
        \caption{Frame 2}
    \end{subfigure}
    \hfill
    \begin{subfigure}{0.24\textwidth}
        \centering
        \includegraphics[width=\textwidth]{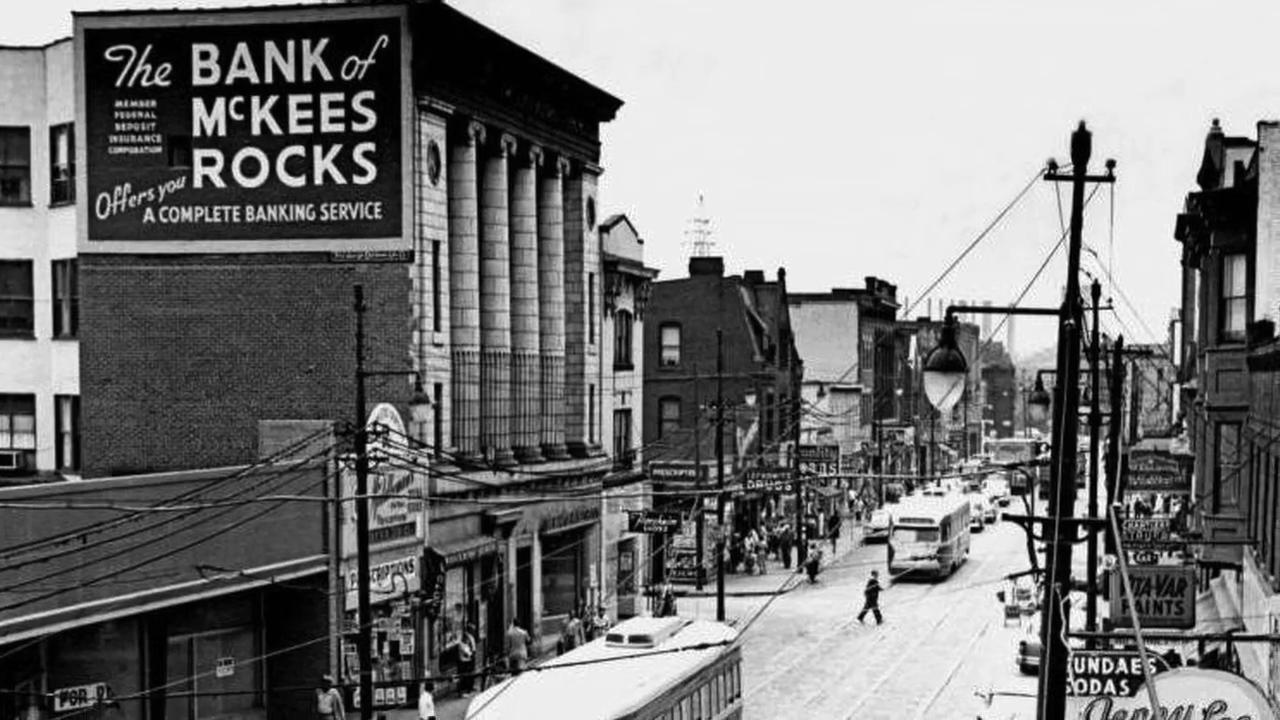}
        \caption{Frame 3}
    \end{subfigure}
    \hfill
    \begin{subfigure}{0.24\textwidth}
        \centering
        \includegraphics[width=\textwidth]{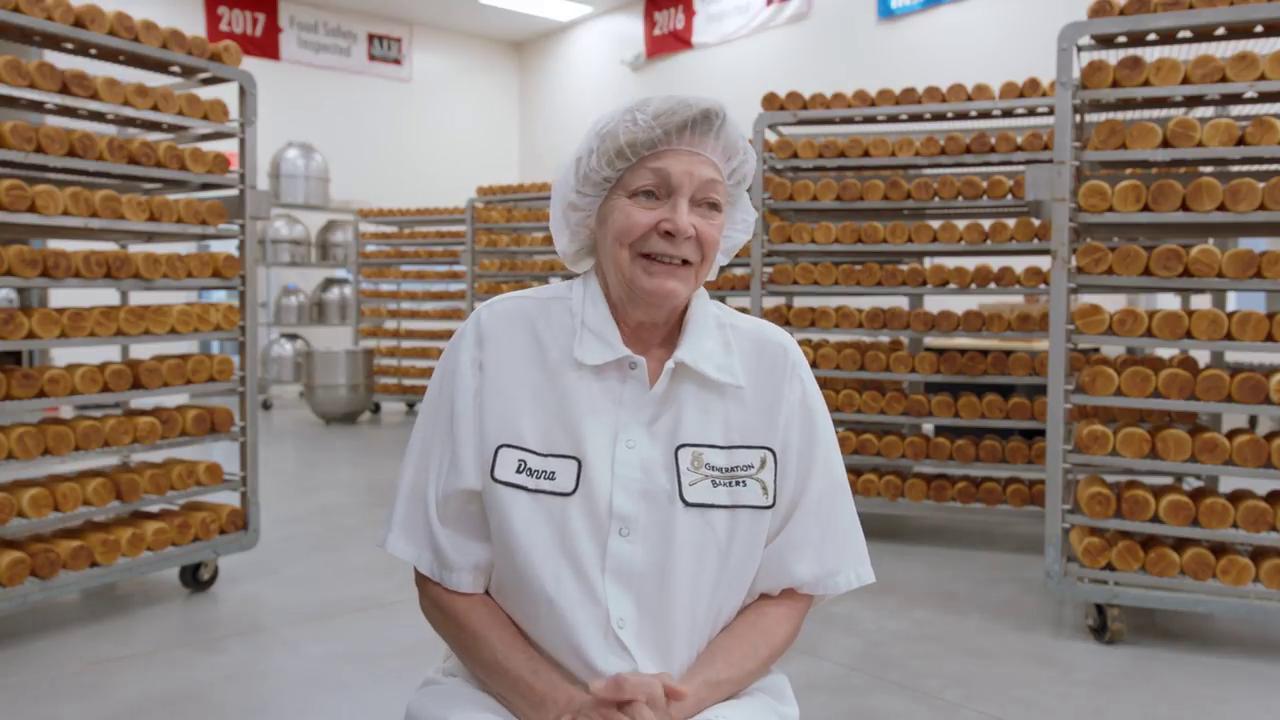}
        \caption{Frame 4}
    \end{subfigure}

    \vspace{1em}
    \begin{minipage}{0.9\textwidth}
    \small\question{Analyze the video content and extract key frames for captioning.}\\
    \small\answer{I have sampled key frames from the video and can generate detailed descriptions for each frame, enabling temporal analysis and scene understanding. The frame sampling provides a comprehensive view of the video content for captioning and analysis.}
    \end{minipage}

    \caption{Video frame sampling examples showing temporal analysis and keyframe extraction for video captioning and scene understanding.}
    \label{fig:video_sampling_examples}
\end{figure}

\noindent
All video workflows can be composed into multi-step pipelines, enabling rich, agentic parsing, annotation, and content creation for diverse use-cases across domains such as media, healthcare, and enterprise document review.

\subsection{Structured Outputs} \label{sec:structured-outputs}

\noindent
A core capability of \textbf{\textbf{Orion}} is robust, type-safe structured outputs for every step of agentic reasoning and execution. Unlike conventional LLM systems that produce free-form text, \textbf{\textbf{Orion}} emits outputs in strict, schema-validated formats—enabling precise downstream processing, reliable tool chaining, and seamless integration into applications and APIs.

\begin{itemize}[leftmargin=*]
    \item \textbf{Schema-driven Outputs} – Agents and tools are driven by explicit schemas, defining the required output types (e.g., arrays, key-value objects, tagged regions, time segments, entities). This ensures each step can be programmatically consumed and reliably orchestrated.
    \item \textbf{Multi-modal Structured Data} – Multimedia outputs, returned in the form of pre-signed URLs, can include mixed-modality values (e.g., image crops, video clips, or generated documents) in a single response, enabling unified downstream processing and interleaving of modalities across agent workflows.
    \item \textbf{API Ready and OpenAI-Compatible} – All outputs conform to Pydantic-compatible JSON output formats, supporting direct integration into developer pipelines, evaluation systems, or end-user applications with minimal custom handling.
\end{itemize}

\noindent
This structured output paradigm unlocks agentic planning, iterative tool invocation, reliable evaluation, and advanced workflow composition within \textbf{\textbf{Orion}}—transforming visual reasoning from opaque black-box outputs to transparent and verifiable computation.

\newpage
\section{Quantitative Evaluation}
\label{sec:results_quantitative}

We evaluate \textbf{Orion} across image, document, and video inputs, comparing its understanding, reasoning, and execution capabilities, against all frontier vision-language models. The benchmarks measure performance on diverse tasks including (but not restricted to) visual question answering, accurate text/object/face detection in images, document/video parsing and summarization, and more. 

\subsection{Benchmark Performance}

Our evaluation spans established multi-modal benchmarks including MMMU \cite{yue2024mmmu}, MMBench \cite{liu2024mmbench}, DocVQA \cite{docvqa2020}, and other comprehensive evaluation suites \cite{liu2023visual}. In the coming weeks, we will be updating these sections with additional benchmarks and updated numbers as our agent evolves. 

\begin{table}[h]
    \centering
    \begingroup
    \renewcommand{\arraystretch}{1.25}
    \setlength{\tabcolsep}{0.9em}
    \setlength{\arrayrulewidth}{0.5pt}
    \resizebox{\textwidth}{!}{%
    \colorbox{tablebg}{%
        \begin{tabular}{@{}p{3.6cm} p{4cm} *{4}{c}@{}}
        \toprule[1pt]
        \textbf{Capability} &
        \textbf{Benchmark} &
        \textcolor{vlmblue}{\textbf{Orion}} &
        \textcolor{geminicolor}{\textbf{Gemini 2.5 Flash}} &
        \textcolor{chatgptcolor}{\textbf{GPT-5 Mini}} &
        \textcolor{claudecolor}{\textbf{Claude Opus-4.1}} \\
        \midrule[0.5pt]
        \multirow{3}{*}{STEM \& Puzzle} &
        MMMU\textsubscript{val} & \textcolor{vlmblue}{\textbf{72.9}} & 72.7 & 67.9 & -- \\
        \cmidrule(l){2-6}
        &
        MMMU\_Pro & 58.2 & \textbf{60.7} & 53.7 & 60.7 \\
        \midrule[0.5pt]
        \multirow{3}{*}{General VQA} &
        MMBench EN (dev) & \textcolor{vlmblue}{\textbf{86.3}} & 82.4 & 78.5 & 84.1 \\
        \cmidrule(l){2-6}
        &
        RealWorldQA & 69.2 & \textbf{70.6} & 73.3 & 68.5 \\
        \cmidrule(l){2-6}
        &
        MMStar & 68.4 & \textbf{71.3} & 61.3 & 71.0 \\
        \midrule[0.5pt]
        \multirow{2}{*}{Hallucination} &
        HallusionBench & \textcolor{vlmblue}{\textbf{69.7}} & 53.6 & 55.9 & 55.1 \\
        \cmidrule(l){2-6}
        &
        MM\_MT\_Bench & \textcolor{vlmblue}{\textbf{8.2}} & 7.1 & 7.4 & 7.9 \\
        \midrule[0.5pt]
        \multirow{3}{*}{Docs \& Charts} &
        AI2D\textsubscript{TEST} & \textcolor{vlmblue}{\textbf{85.2}} & 84.8 & 82.9 & 84.4 \\
        \cmidrule(l){2-6}
        &
        MMLongBench-Doc & 33.3 & \textbf{38.3} & 42.4 & 48.1 \\
        \cmidrule(l){2-6}
        &
        OCRBench & 798 & \textbf{813} & 807 & 750 \\
        \midrule[0.5pt]
        \multirow{2}{*}{Multi-image Reasoning} &
        BLINK & \textcolor{vlmblue}{\textbf{64.2}} & 62.0 & 56.7 & 62.9 \\
        \cmidrule(l){2-6}
        &
        MUIREBENCH & \textcolor{vlmblue}{\textbf{67.5}} & 67.0 & 57.5 & -- \\
        \bottomrule[1pt]
      \end{tabular}%
    }%
    }
    \endgroup
    \caption{Comparison of \textbf{\textcolor{vlmblue}{Orion}} against leading VLMs on selected benchmarks. Bold indicates highest score per benchmark.}
    \label{tab:results_orion_fast_only}
\end{table}

\subsection{Human Evaluation Methodology}

To comprehensively assess \textbf{Orion}'s performance across diverse visual tasks, we conducted a qualitative human evaluation comparing outputs from \textbf{Orion}, ChatGPT (GPT-5), Gemini 2.5 Pro, and Claude Sonnet 4.5 across 46 diverse visual tasks. This evaluation methodology was designed to provide rigorous, unbiased assessment of model outputs across a wide spectrum of visual AI capabilities, from object detection and segmentation to complex multi-step visual reasoning, document analysis, and image generation.

\textbf{Evaluation Design.} To ensure objective assessment, we implement a double-blind evaluation protocol where all model outputs are presented to evaluators in randomized order, labeled only as Model A, B, C, or D. Each task is evaluated by at least three independent evaluators who undergo calibration training to establish consistent scoring standards.

\textbf{Composite Overall Score.} We develop a composite overall score (0--10) that aggregates multiple dimensions of output quality. The score is calculated as a weighted combination of four components: 
\begin{itemize}
\item \textit{Helpfulness} (30\%), measuring whether the model addresses all the requested requirements; 
\item \textit{Correctness} (35\%), assessing factual correctness; 
\item \textit{Presentation} (20\%, for visual outputs), evaluating presentation and artifact-free rendering; and
\item \textit{Instruction Following} (15\%), determining whether the output meets specified constraints and requirements.
\end{itemize}

\textbf{Evaluation Process.} Ten independent evaluators assess 46 diverse visual tasks spanning object detection, multi-step reasoning, medical image analysis, document processing, and image generation. For each task, evaluators receive the input and prompt, along with outputs from all four models in randomized order, and score each output independently across all four dimensions.

\begin{figure}[H]
    \centering
    \includegraphics[width=0.95\textwidth]{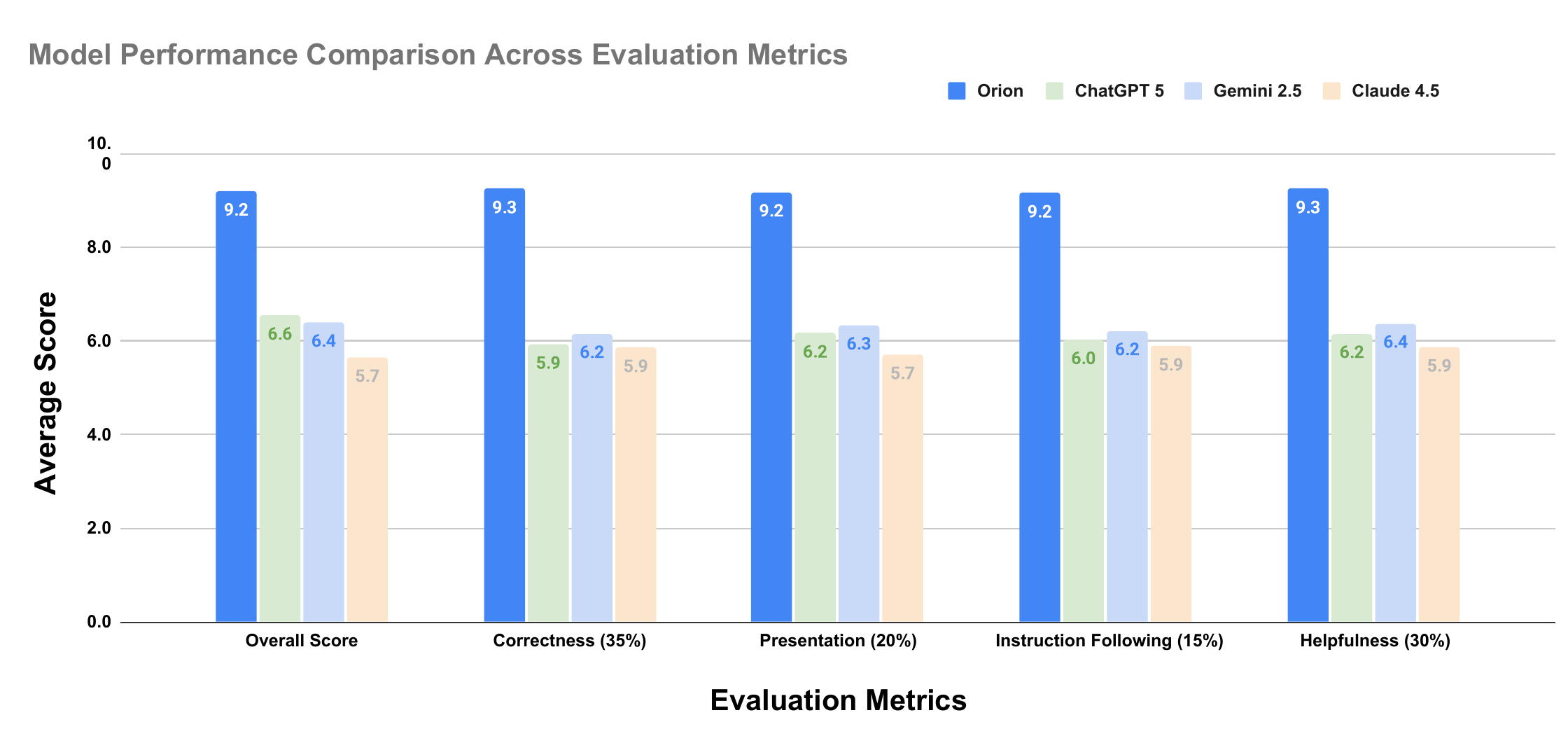}
    \caption{Comprehensive evaluation results showing \textbf{Orion}'s superior performance across 46 diverse visual tasks compared to frontier Vision-Language Models. The evaluation demonstrates \textbf{Orion}'s consistent advantage in complex visual reasoning, specialized skill execution, and multi-step workflow completion.}
    \label{fig:orion_evaluation}
\end{figure}

\textbf{Results and Analysis.} The human evaluation results, visualized in Figure~\ref{fig:orion_evaluation}, demonstrate \textbf{Orion}'s superior performance across the majority of evaluated tasks. The double-blind design and multi-dimensional scoring ensure these results reflect genuine performance differences, establishing \textbf{Orion} as the leading system for production-grade visual tasks requiring precision, multi-step reasoning, and specialized tool integration.

\subsection{Performance Analysis and Comparison}

As shown in Table~\ref{tab:results_orion_fast_only}, \textbf{Orion} demonstrates superior performance across multiple vision capabilities when compared to frontier vision-language models. The comprehensive benchmark results reveal that \textbf{Orion}'s agentic architecture, which orchestrates specialized computer vision tools, enables it to achieve more accurate and reliable performance across diverse visual understanding tasks compared to monolithic models. For an interactive version of the results showcased in the table~\ref{tab:results_orion_fast_only}, see our showdown page at \href{https://chat.vlm.run/showdown}{https://chat.vlm.run/showdown}.

\textbf{Reduced Hallucination and Enhanced Accuracy} \textbf{Orion} exhibits significantly lower hallucination rates compared to other frontier models, as evidenced by its state-of-the-art performance on hallucination detection benchmarks. The system's ability to leverage precise visual analysis tools—including object detection, segmentation, and spatial reasoning—enables it to ground its responses in actual visual content rather than generating hallucinated details. This tool-augmented approach provides built-in verification mechanisms that help the system distinguish between real visual information and spurious patterns, resulting in more trustworthy and accurate outputs across all vision tasks.

\textbf{Superior Multi-Capability Vision Performance.} \textbf{Orion} demonstrates enhanced accuracy across multiple vision capabilities compared to other frontier models. The system's specialized tool suite enables precise execution of diverse visual tasks including object detection and localization, semantic and instance segmentation, optical character recognition, spatial reasoning, and multi-image understanding. Unlike monolithic models that rely solely on learned representations, \textbf{Orion} can invoke specialized tools to perform exact visual computations—such as precise bounding box detection, pixel-level segmentation masks, and character-level OCR—resulting in more accurate results across these fundamental vision capabilities.

\textbf{Comprehensive Visual Understanding.} The benchmark results demonstrate that \textbf{Orion} achieves competitive or superior performance across a wide spectrum of visual understanding tasks, from general visual question answering to specialized document analysis, chart interpretation, and multi-image reasoning. The system's ability to seamlessly integrate multiple vision tools within a single agentic workflow enables it to handle complex visual scenarios that require combining multiple capabilities—such as detecting objects, extracting text, analyzing spatial relationships, and reasoning about visual content—more accurately than models that attempt to perform all these tasks through a single neural architecture.

\textbf{Production-Ready Reliability.} The consistent performance improvements demonstrated across benchmarks indicate that \textbf{Orion}'s tool-augmented approach provides more reliable and accurate vision capabilities suitable for production deployment. The system's reduced hallucination rates and enhanced accuracy across multiple vision tasks make it particularly well-suited for real-world applications where precision and trustworthiness are critical. By combining neural reasoning with symbolic visual tools, \textbf{Orion} achieves a balance between the flexibility of learned models and the precision of specialized computer vision algorithms, resulting in superior overall performance on image understanding and analysis tasks.

\subsection{Future Work}

While \texttt{Orion} demonstrates significant capabilities in multi-modal reasoning and tool use, several directions remain for future exploration:

\textbf{Enhanced Benchmarking and Evaluation.} Current multi-modal benchmarks predominantly focus on single-turn interactions and relatively simple tasks. This limitation stems largely from the difficulty of evaluating complex, multi-step outputs that require visual verification. We are developing a comprehensive suite of benchmarks specifically designed to assess performance on intricate, multi-turn tasks where outputs must be visually validated against ground truth.

\textbf{Improved Tool Integration.} Expanding the repertoire of available tools and refining the tool selection mechanism could further enhance the system's capabilities. This includes integrating domain-specific tools for specialized applications and developing more sophisticated strategies for tool composition.

\textbf{Efficiency Optimizations.} While the current system achieves strong performance, there remain opportunities to optimize computational efficiency, particularly in reducing the number of inference calls required for complex tasks and improving the speed of tool execution.

\textbf{Robustness and Error Recovery.} Enhancing the system's ability to detect and recover from errors, particularly in long-running multi-step tasks, represents an important area for improvement. This includes developing better mechanisms for self-correction and validation of intermediate outputs.

\begingroup
\setlength{\tabcolsep}{0.9em}
\renewcommand{\arraystretch}{1.6}
\setlength{\LTcapwidth}{\textwidth}
\setlength{\arrayrulewidth}{0.5pt}
\begin{longtable}{@{}>{\centering\arraybackslash}p{3cm} >{\centering\arraybackslash}p{3cm} >{\centering\arraybackslash}p{3cm} >{\centering\arraybackslash}p{3cm} >{\centering\arraybackslash}p{3cm}@{}}
\caption{Comparison of \textcolor{vlmblue}{\textbf{Orion}} with frontier vision-language models across diverse tasks. See our interactive showdown at \href{https://chat.vlm.run/showdown}{https://chat.vlm.run/showdown}.}
\label{fig:vlm_comparison} \\
\toprule[1pt]
\textbf{Input} & \textcolor{vlmblue}{\textbf{Orion}} & \textcolor{chatgptcolor}{\textbf{GPT 5}} & \textcolor{geminicolor}{\textbf{Gemini 2.5}} & \textcolor{claudecolor}{\textbf{Claude 4.5}} \\
\midrule[0.5pt]
\endfirsthead
\caption[]{Comparison of \textcolor{vlmblue}{\textbf{Orion}} with frontier vision-language models across diverse tasks. See our interactive showdown at \href{https://chat.vlm.run/showdown}{https://chat.vlm.run/showdown}.\textit{(continued)}} \\
\toprule[1pt]
\textbf{Input} & \textcolor{vlmblue}{\textbf{Orion}} & \textcolor{chatgptcolor}{\textbf{GPT 5}} & \textcolor{geminicolor}{\textbf{Gemini 2.5}} & \textcolor{claudecolor}{\textbf{Claude 4.5}} \\
\midrule[0.5pt]
\endhead
\midrule[0.5pt]
\multicolumn{5}{r}{\footnotesize\textit{Continued on next page\ldots}} \\
\endfoot
\bottomrule[1pt]
\endlastfoot


    \parbox{3cm}{\centering \includegraphics[height=100px]{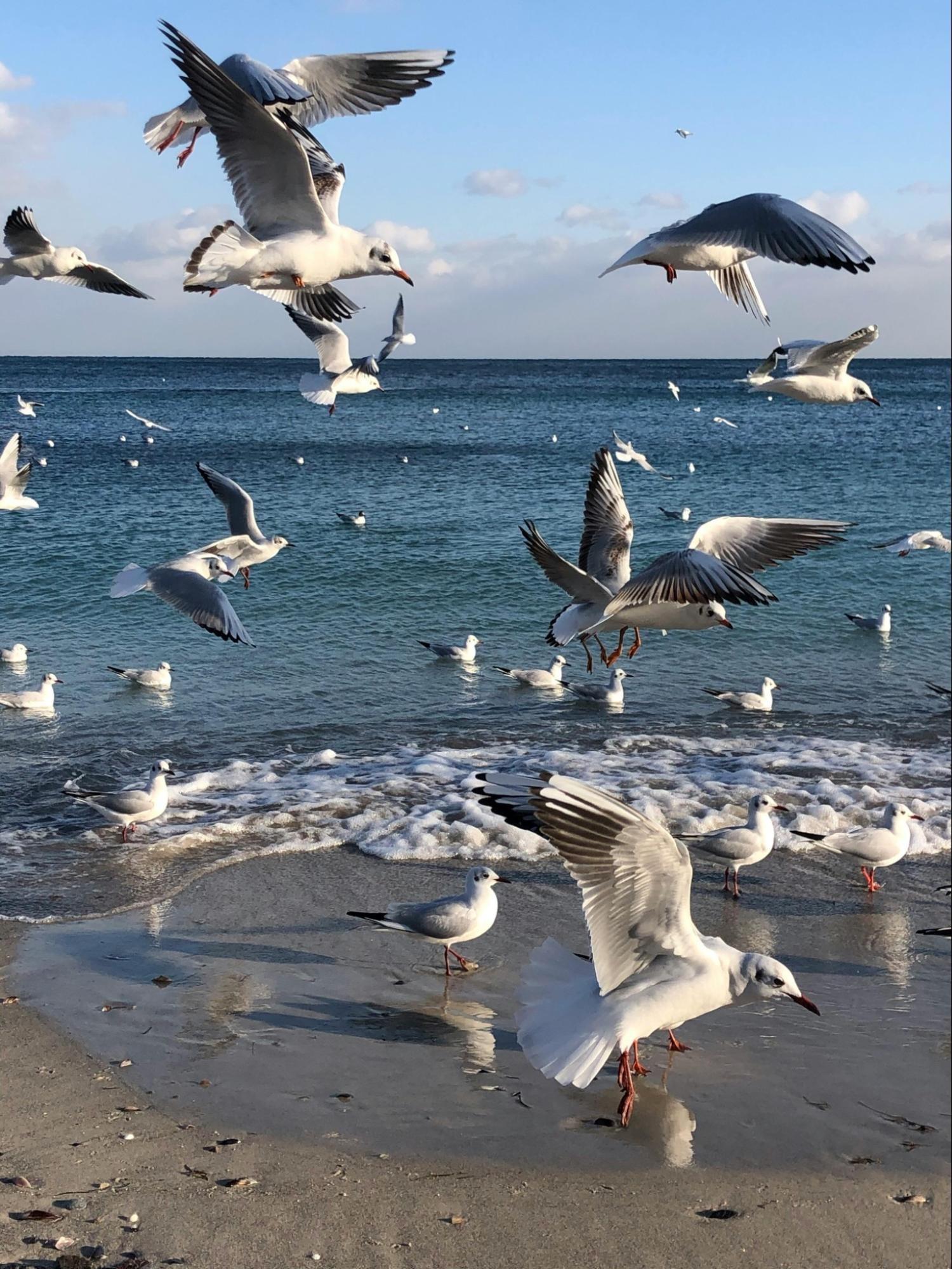}} & \parbox{3cm}{\centering \includegraphics[height=100px]{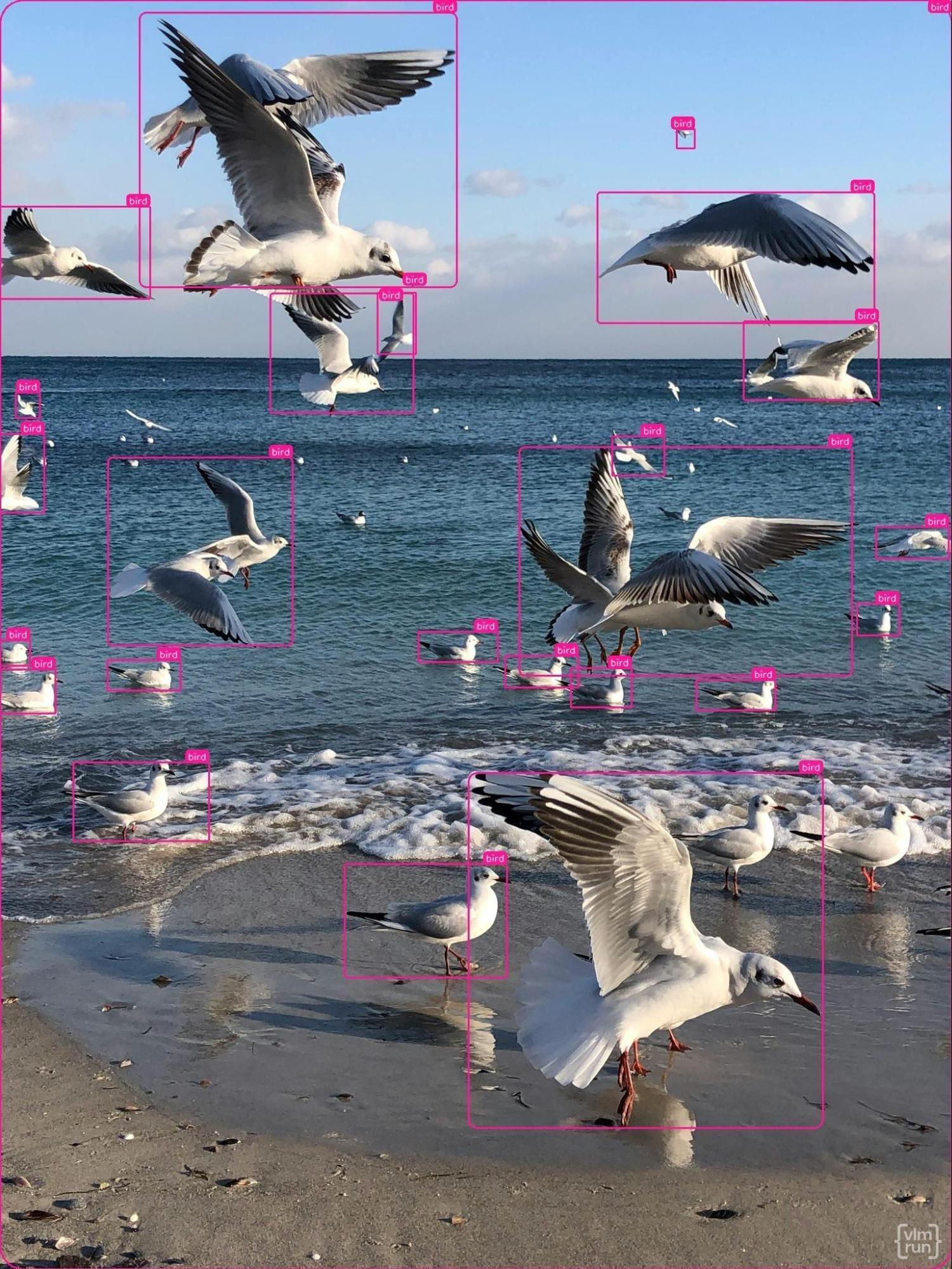}} &      \parbox{3cm}{\centering \includegraphics[height=100px]{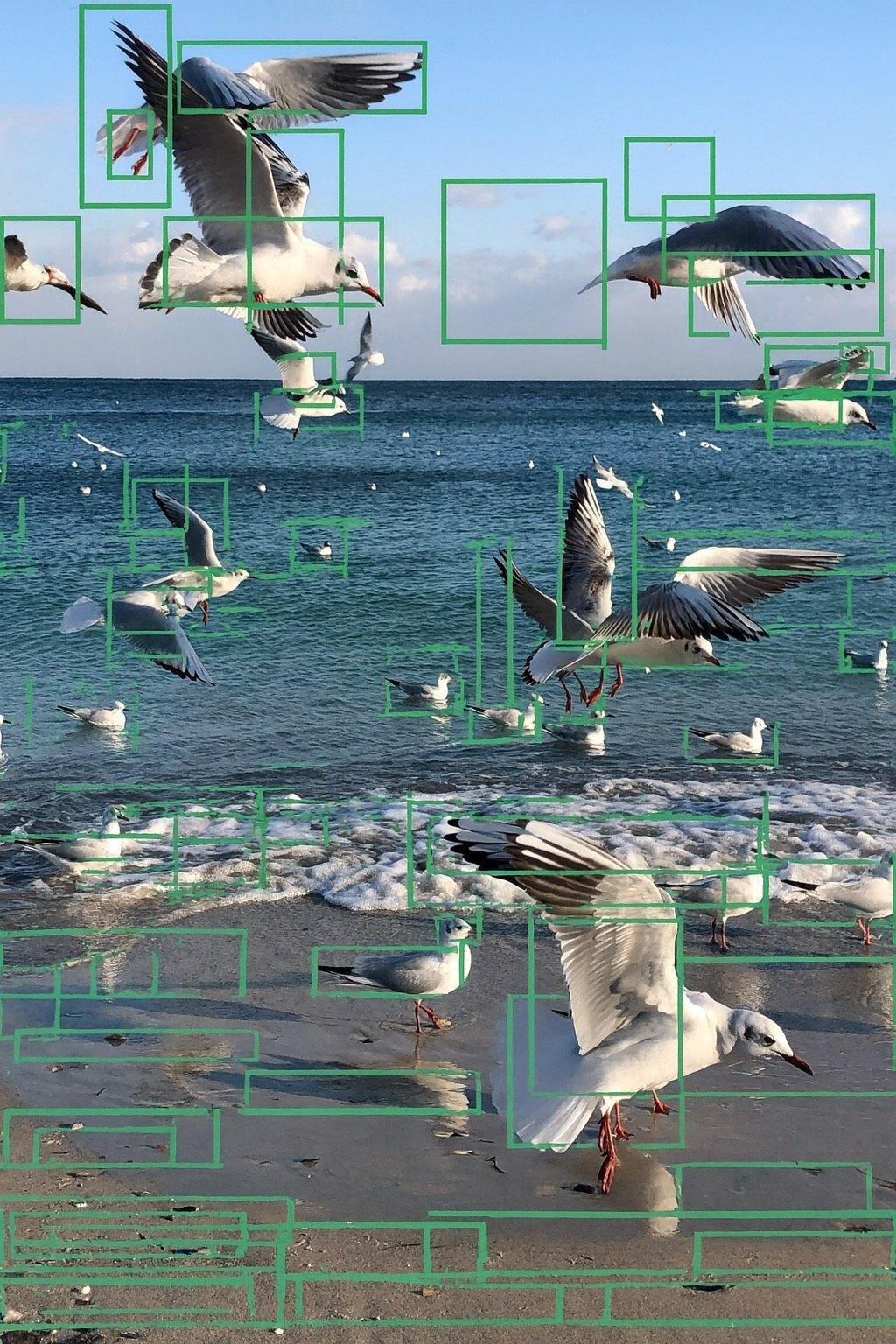}} &      \parbox{3cm}{\centering \includegraphics[height=100px]{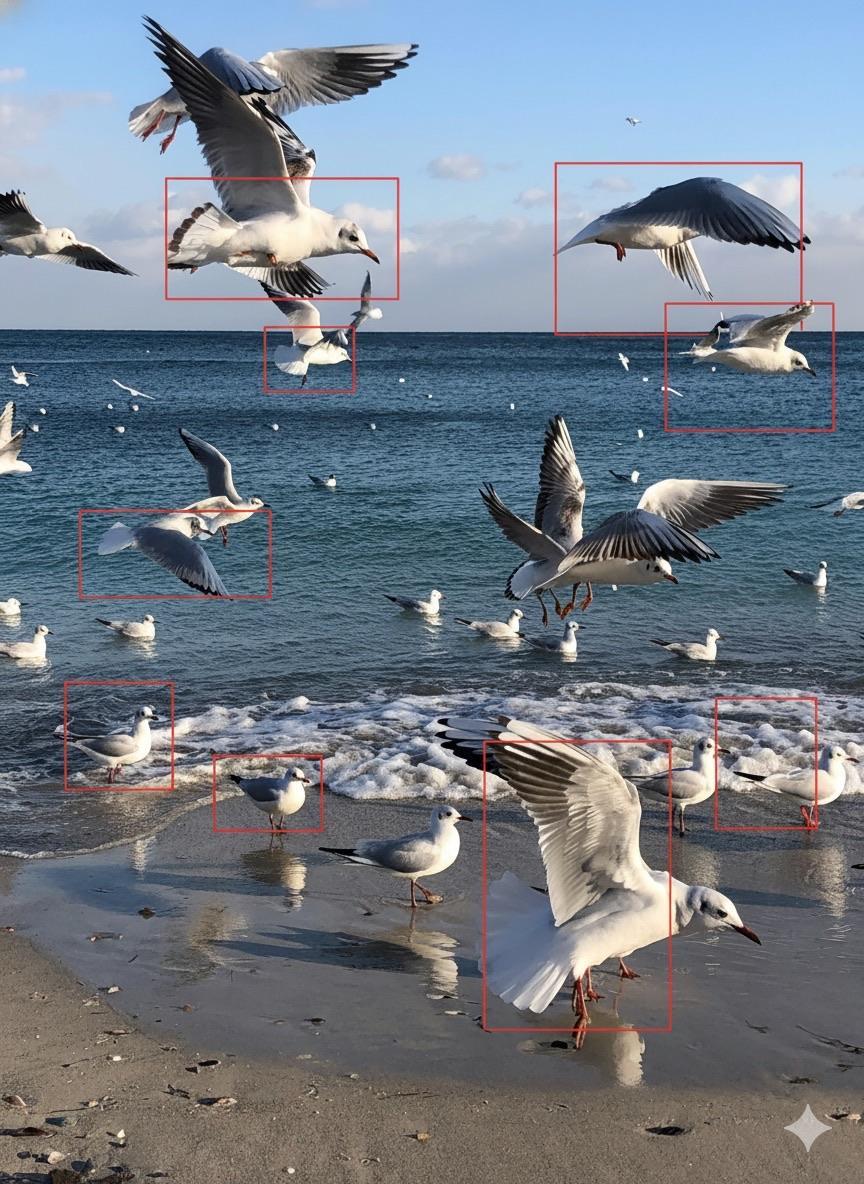}} &      \parbox{3cm}{\centering \includegraphics[height=100px]{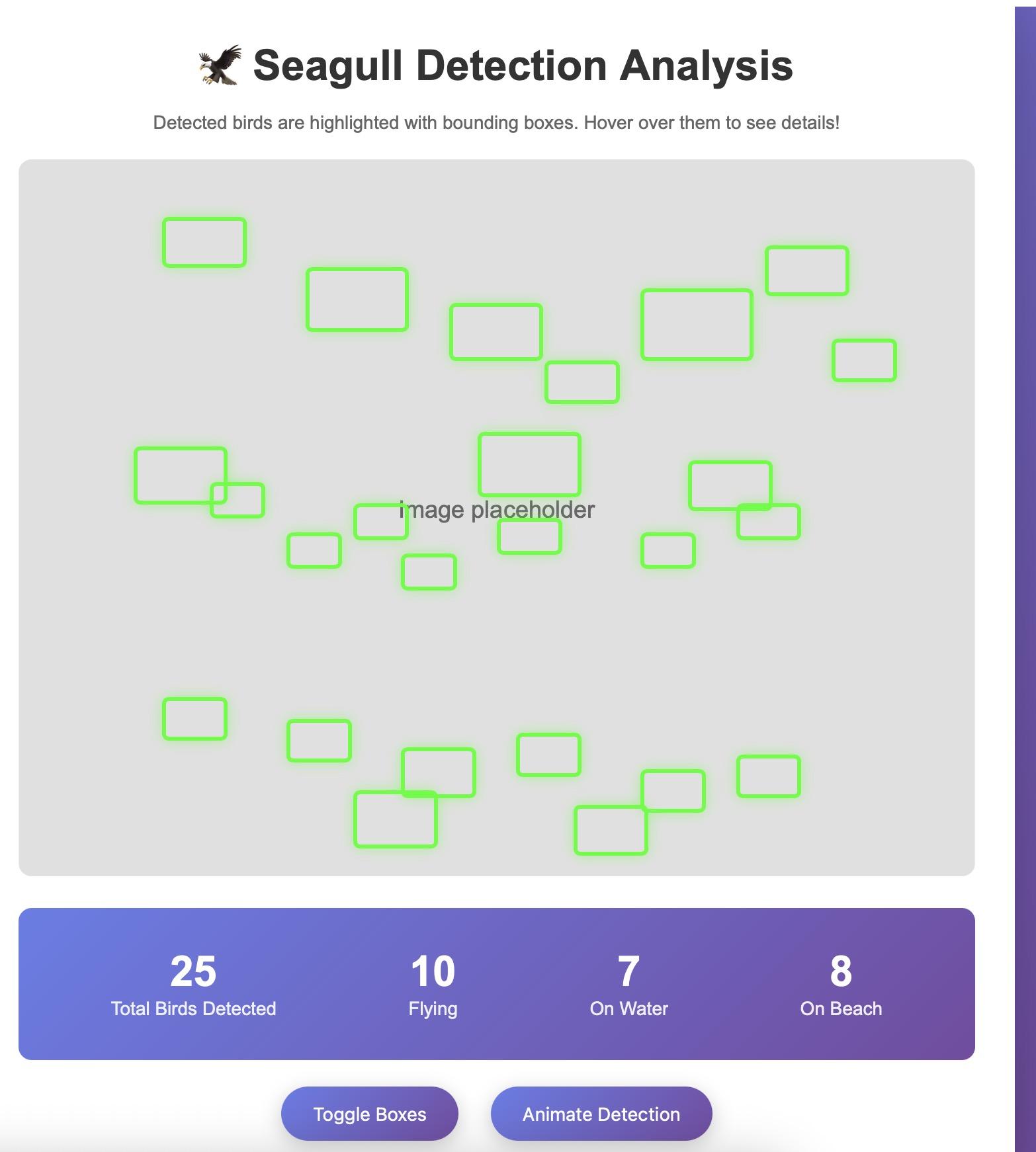}} \\    
    \multicolumn{5}{l}{\cellcolor{subtlegray!20}\parbox{\textwidth}{\scriptsize\textit{\textcolor{questiongray}{1. Detect all the birds in the image and visualize them.}}}}\\
    \arrayrulecolor{gray!30}\midrule[0.3pt]\arrayrulecolor{black}

    \parbox{3cm}{\centering \includegraphics[width=3cm]{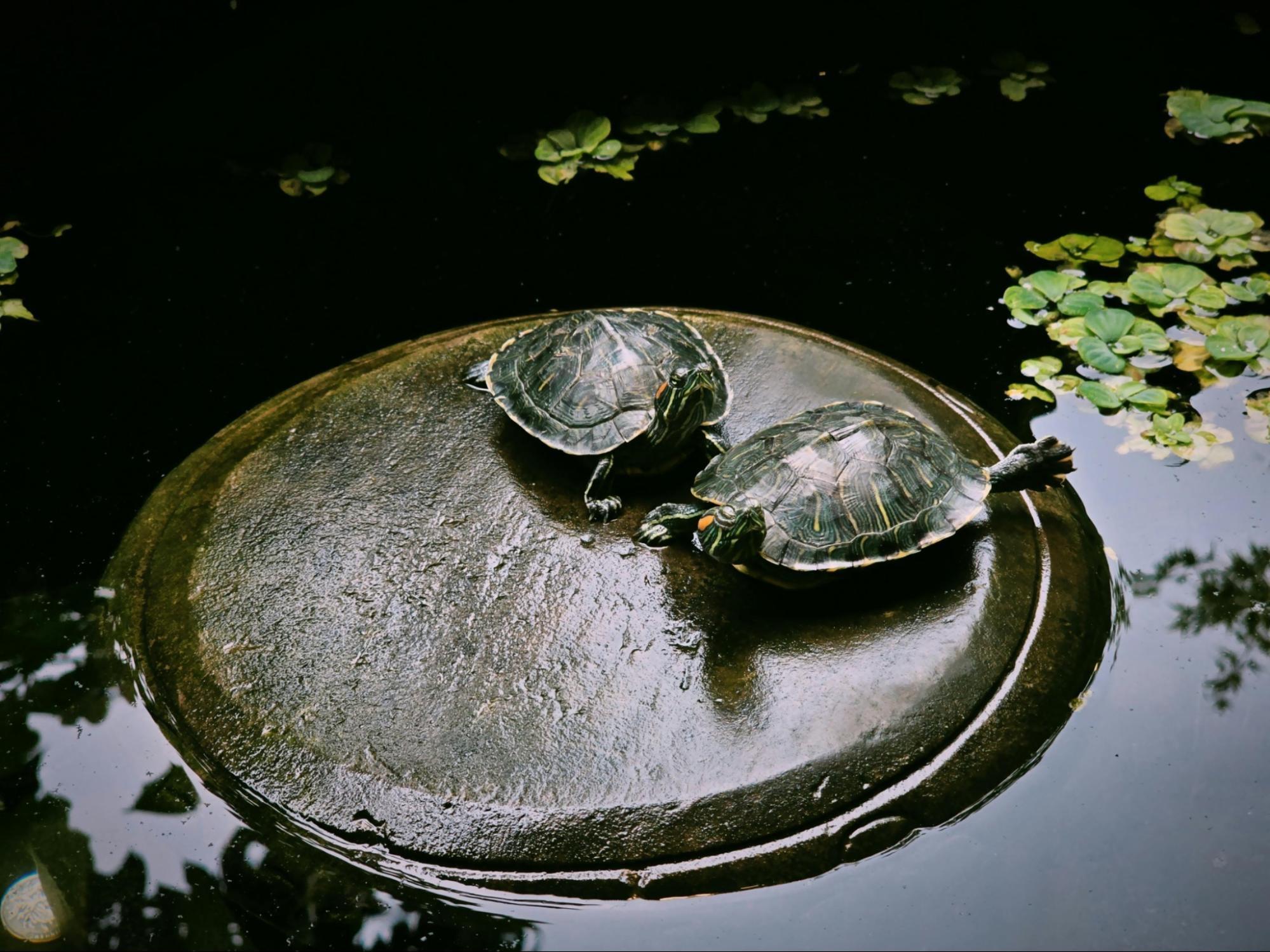}} & \parbox{3cm}{\centering \includegraphics[width=3cm]{3_results_sections/comparision_table/images/image163.jpg}} &      \parbox{3cm}{\centering \includegraphics[width=3cm]{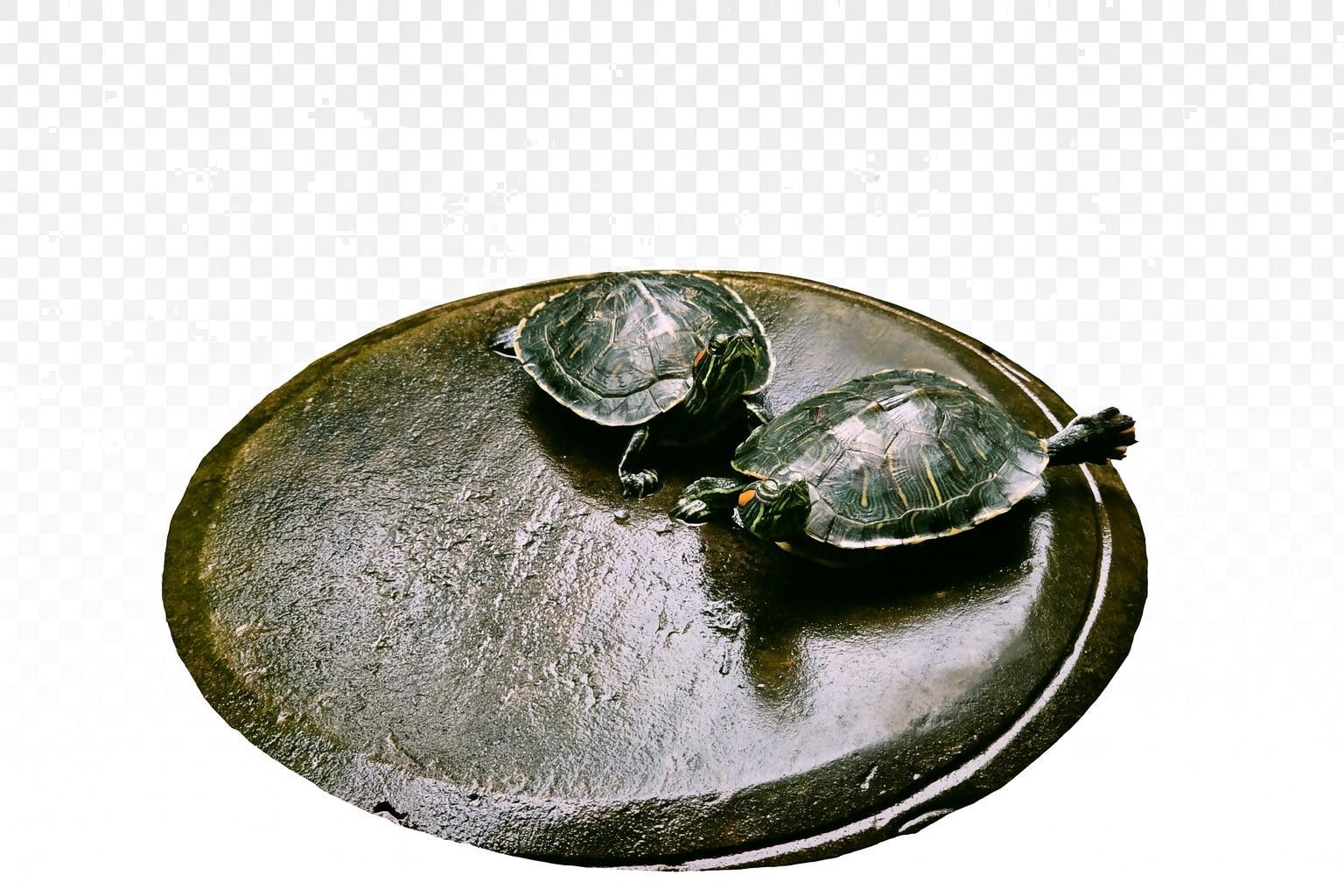}} &      \parbox{3cm}{\centering \includegraphics[width=3cm]{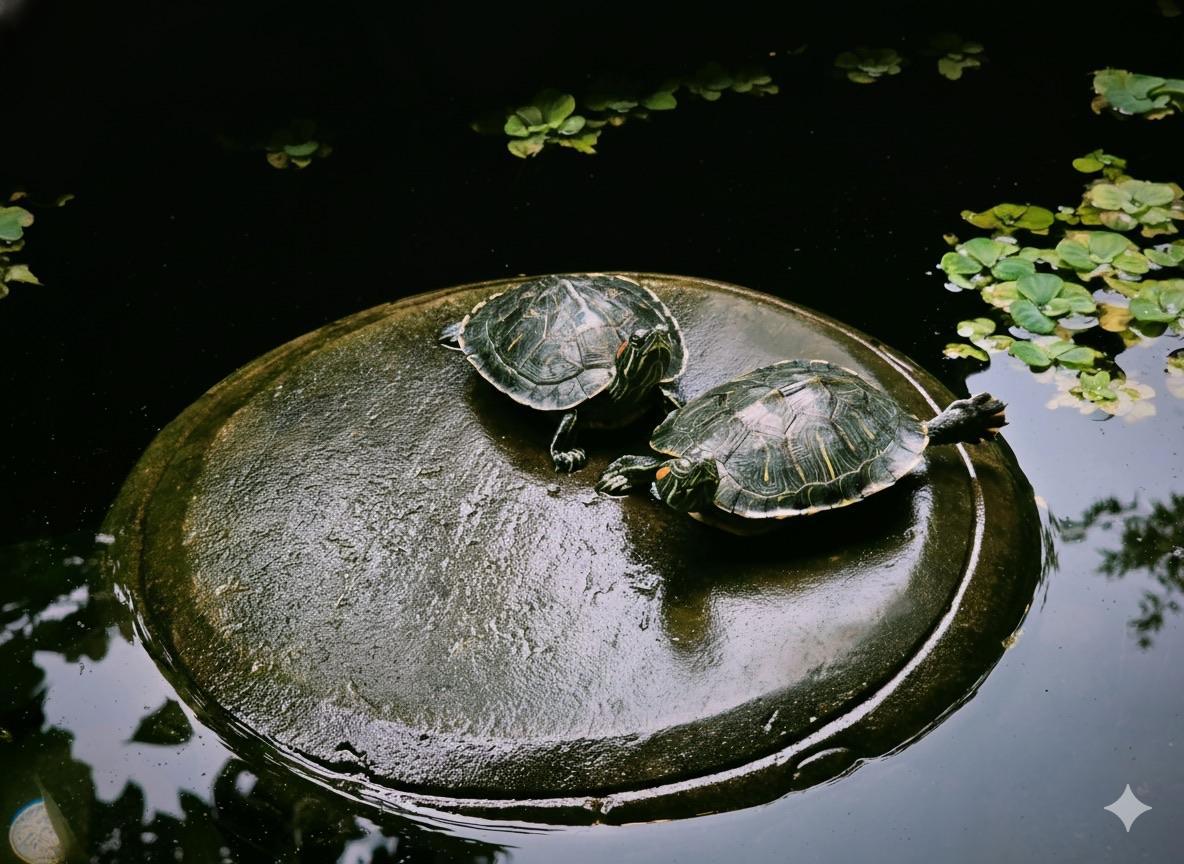}} &      \parbox{3cm}{\centering \includegraphics[width=3cm]{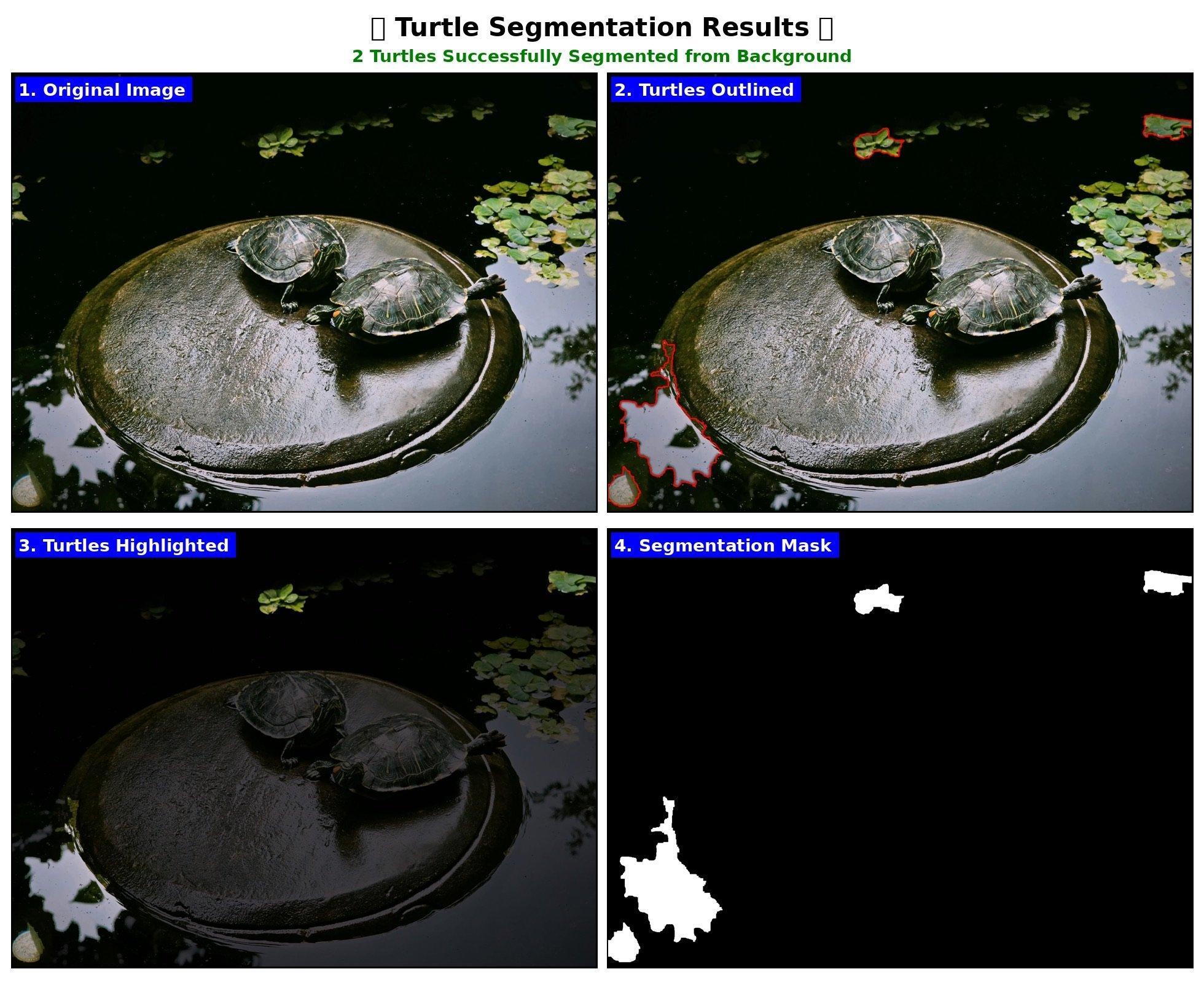}} \\
    \multicolumn{5}{l}{\cellcolor{subtlegray!20}\parbox{\textwidth}{\scriptsize\textit{\textcolor{questiongray}{2. Segment out the turtles in the image.}}}}\\
    \arrayrulecolor{gray!30}\midrule[0.3pt]\arrayrulecolor{black}

    \parbox{3cm}{\centering \includegraphics[width=3cm]{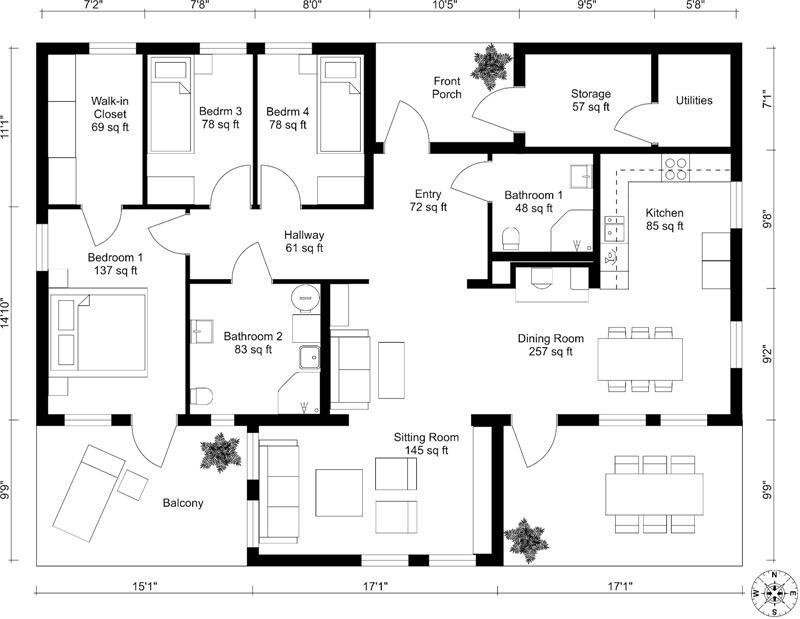}} & \parbox{3cm}{\centering \includegraphics[width=3cm]{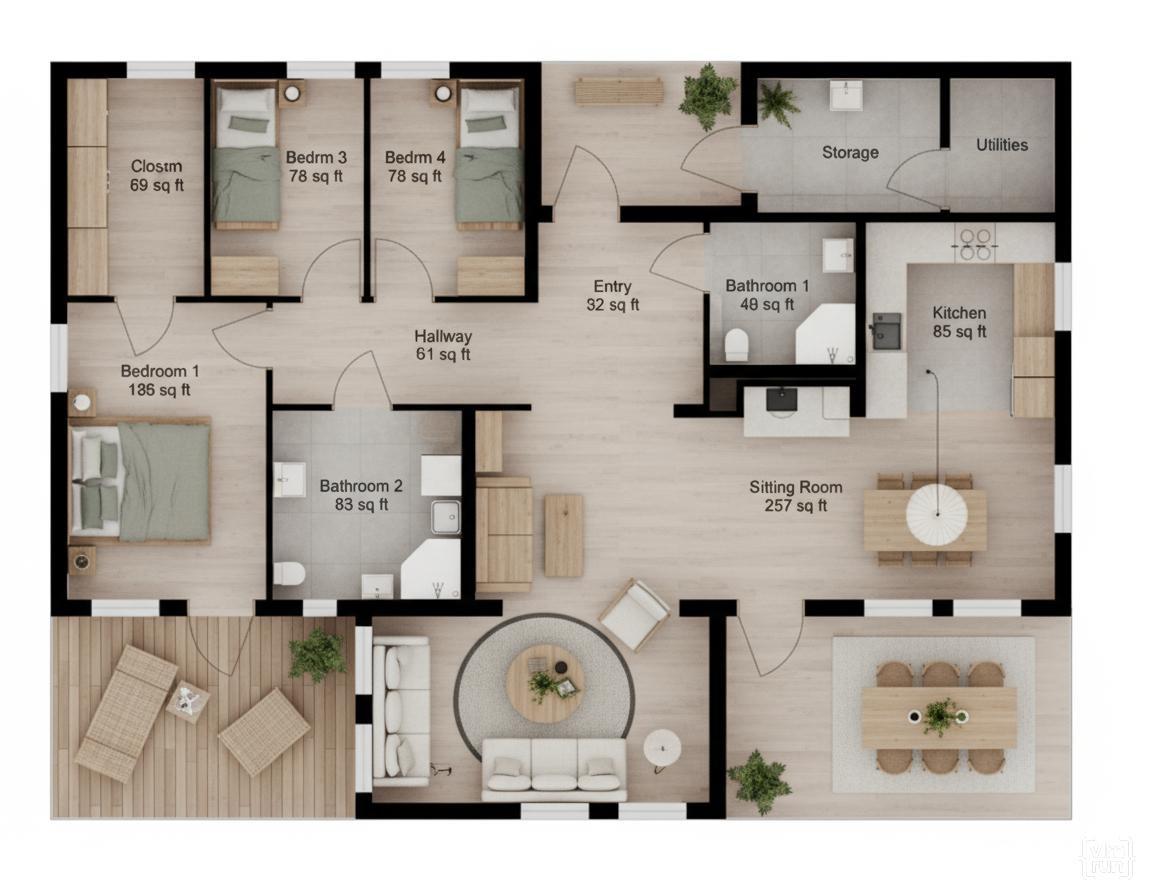}} &      \parbox{3cm}{\centering \includegraphics[width=3cm]{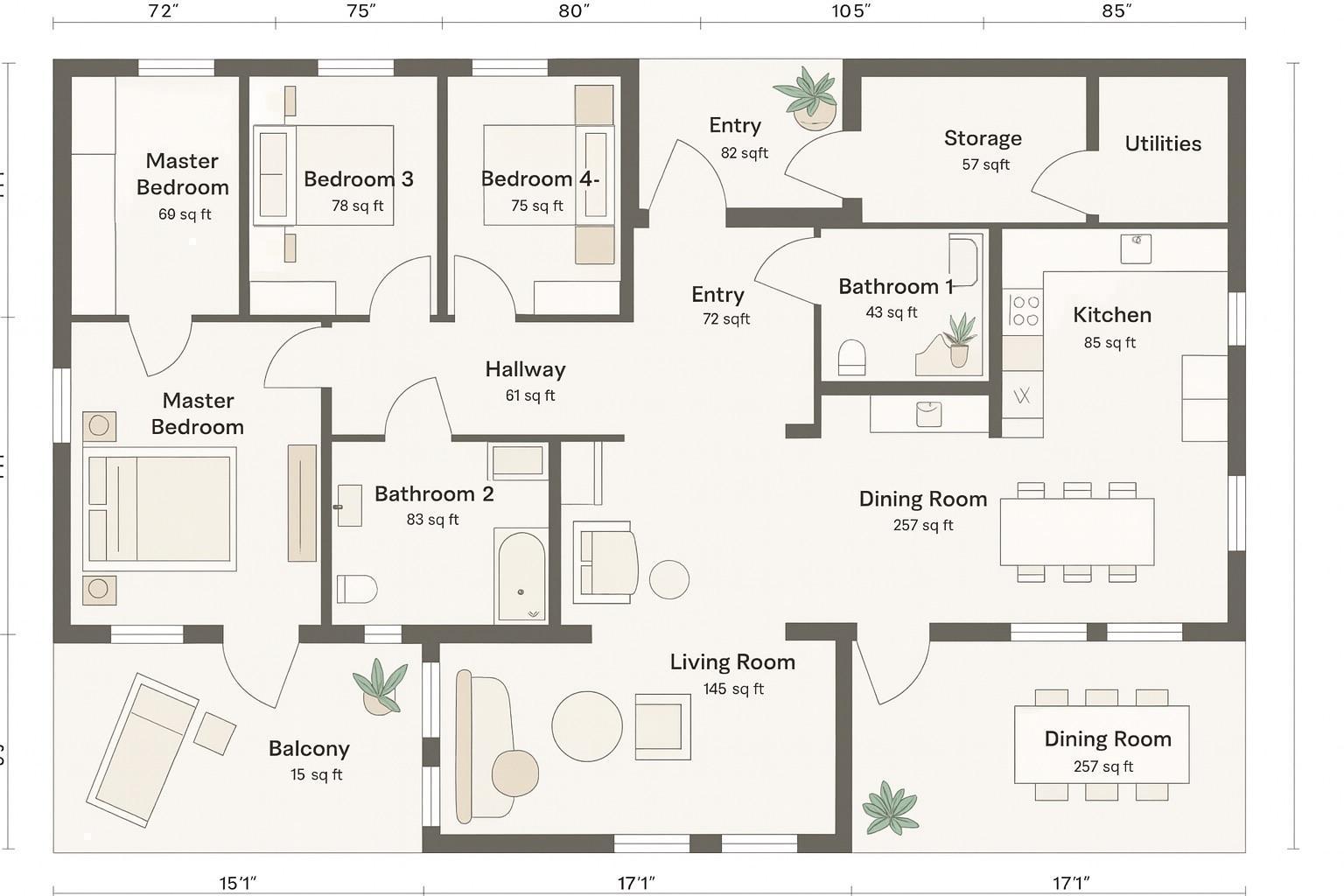}} &      \parbox{3cm}{\centering \includegraphics[width=3cm]{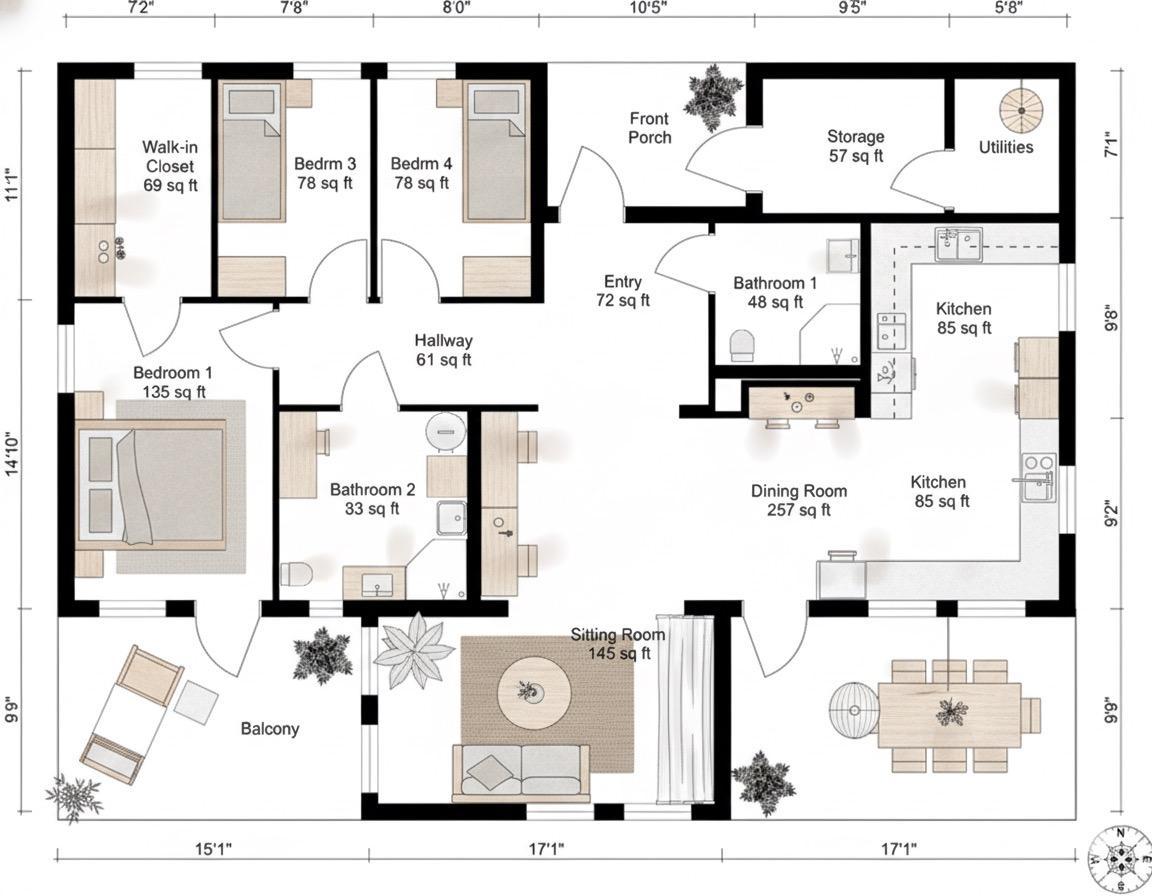}} &      \parbox{3cm}{\centering \includegraphics[width=3cm]{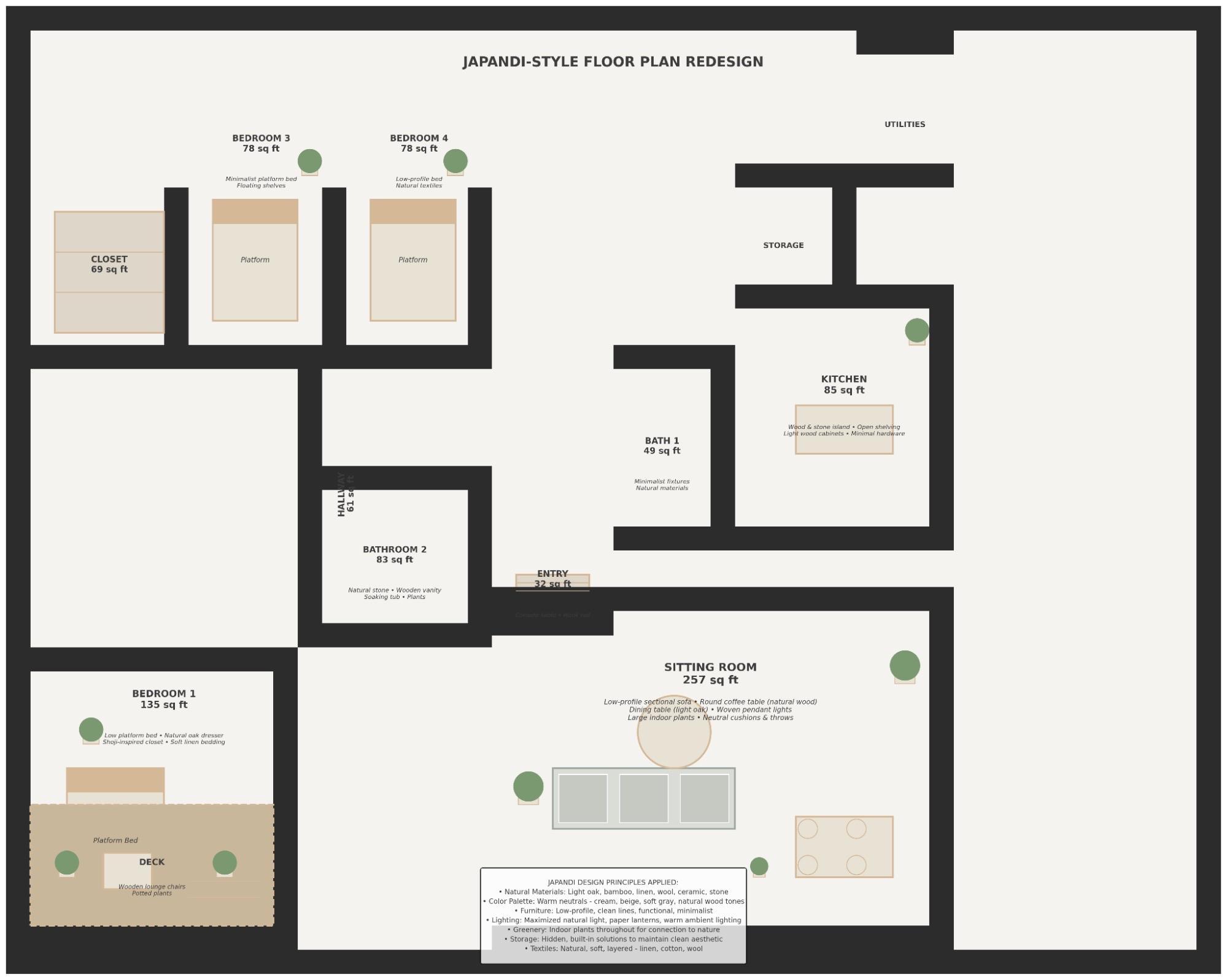}} \\
    \multicolumn{5}{l}{\cellcolor{subtlegray!20}\parbox{\textwidth}{\scriptsize\textit{\textcolor{questiongray}{3. Identify and list all major rooms in the floor plan, providing the square footage for each. Redesign the space using Japandi-style furnishings, emphasizing natural light, neutral color palettes, and minimalist forms. Return an updated floor plan with labeled rooms and suggested furniture layout.}}}}\\
    \arrayrulecolor{gray!30}\midrule[0.3pt]\arrayrulecolor{black}

    \parbox{3cm}{\centering \includegraphics[width=3cm]{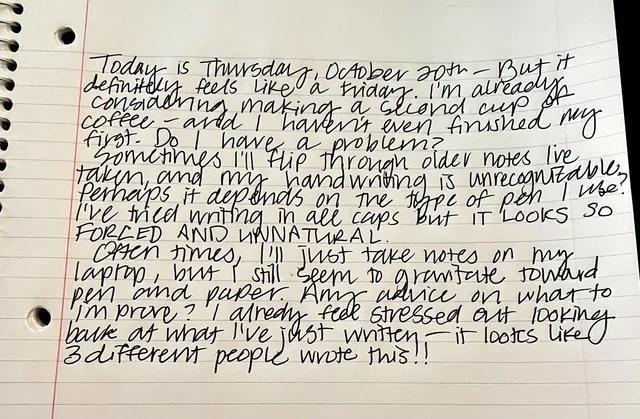}} & \parbox{3cm}{\centering \includegraphics[width=3cm]{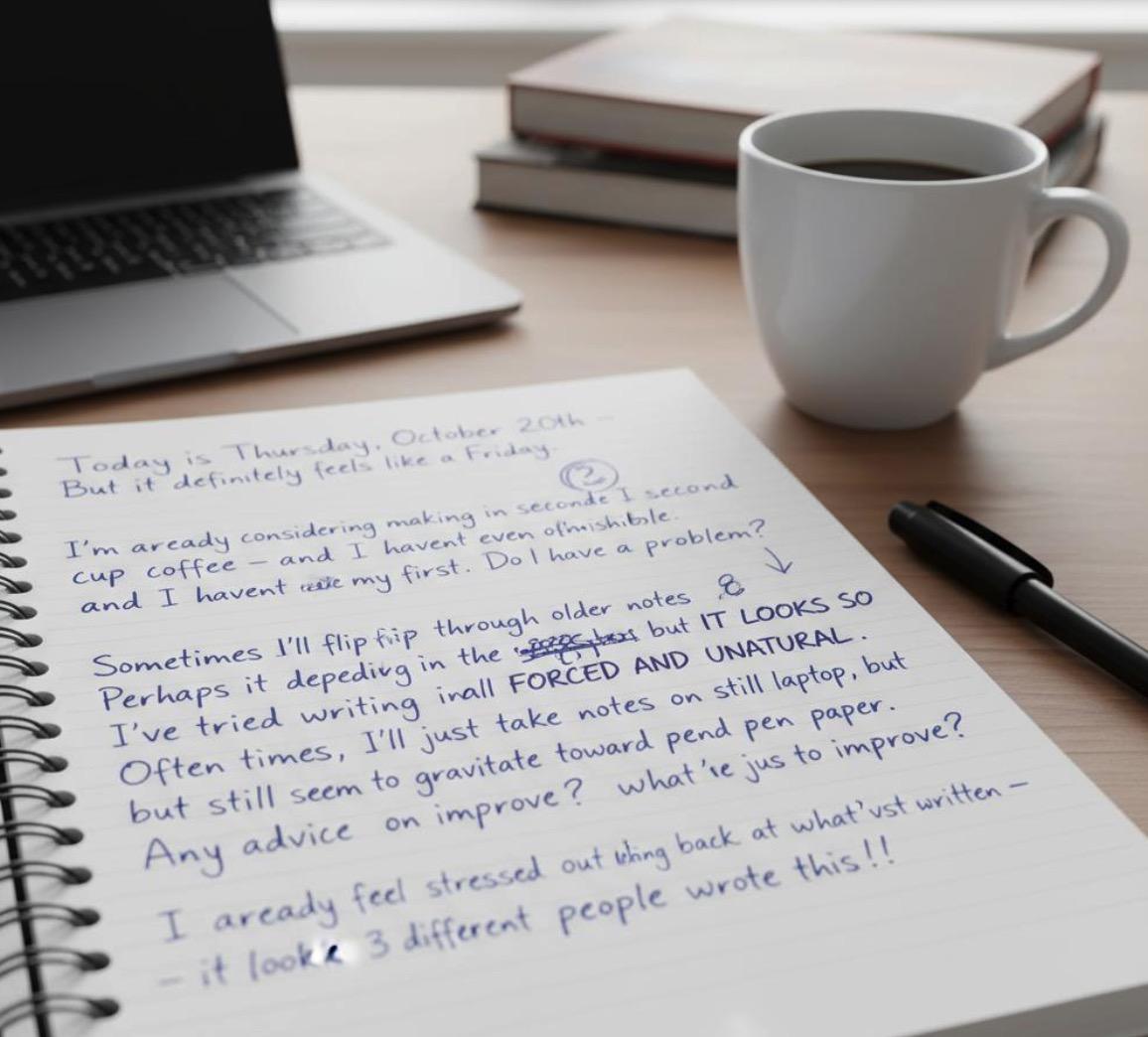}} &      \parbox{3cm}{\centering \includegraphics[width=3cm]{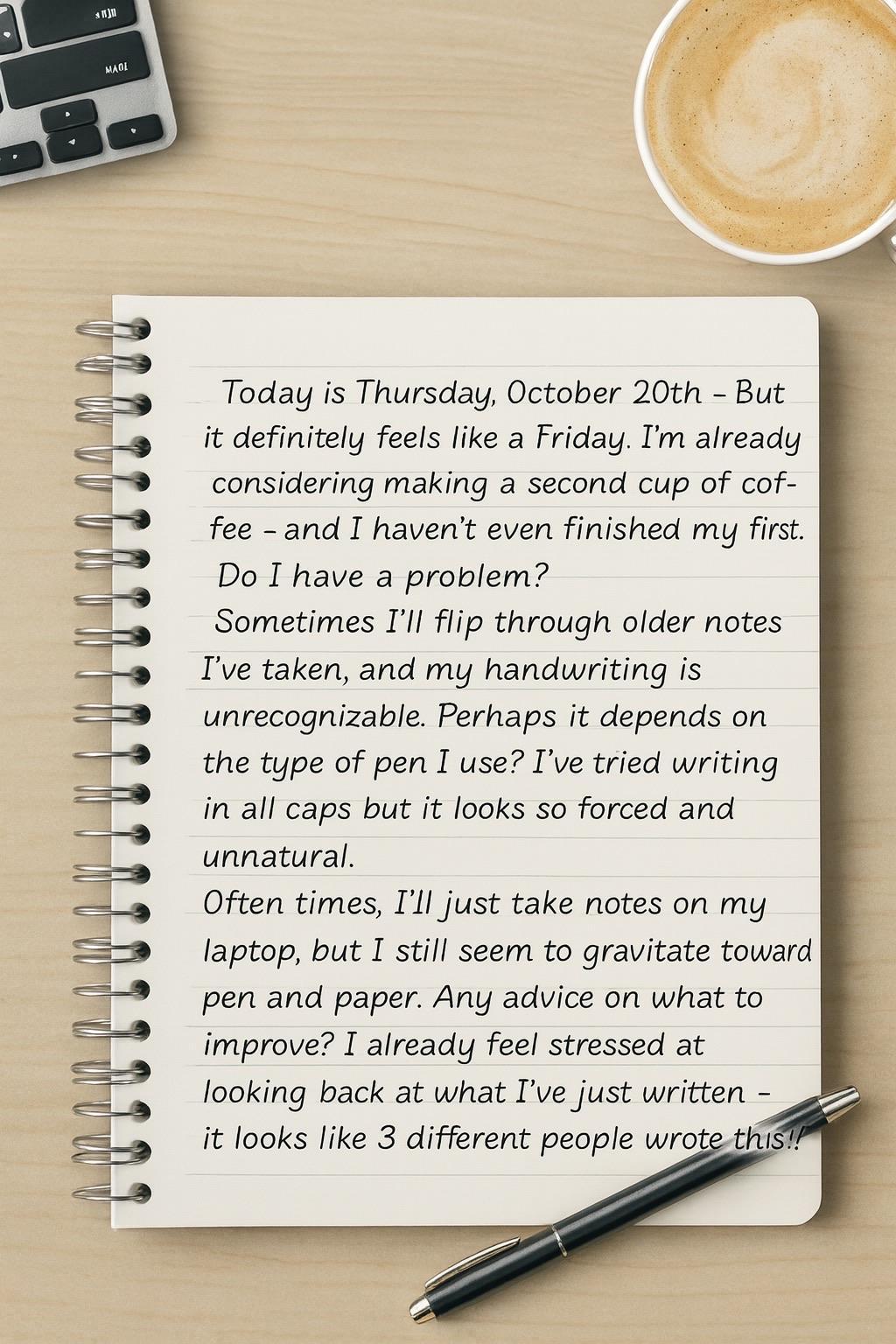}} &      \parbox{3cm}{\centering \includegraphics[width=3cm]{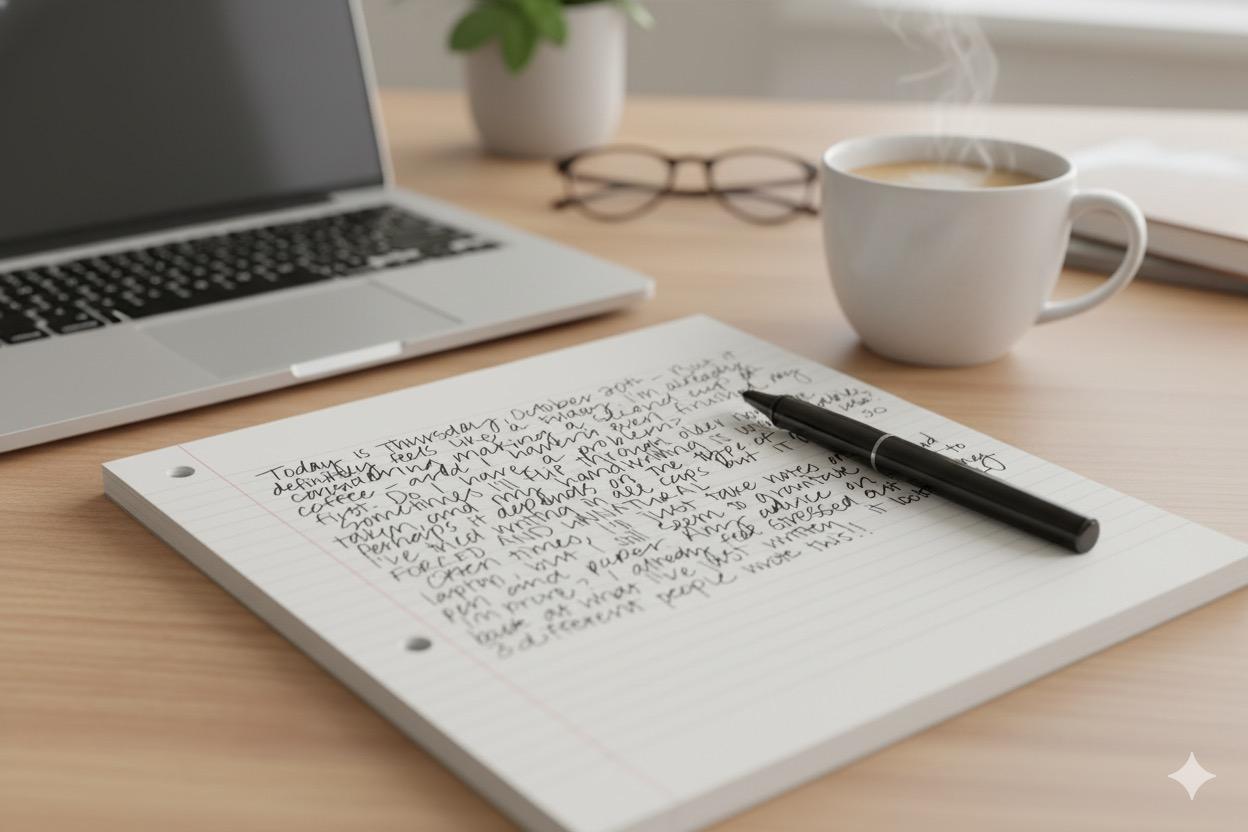}} &      \parbox{3cm}{\centering \includegraphics[width=3cm]{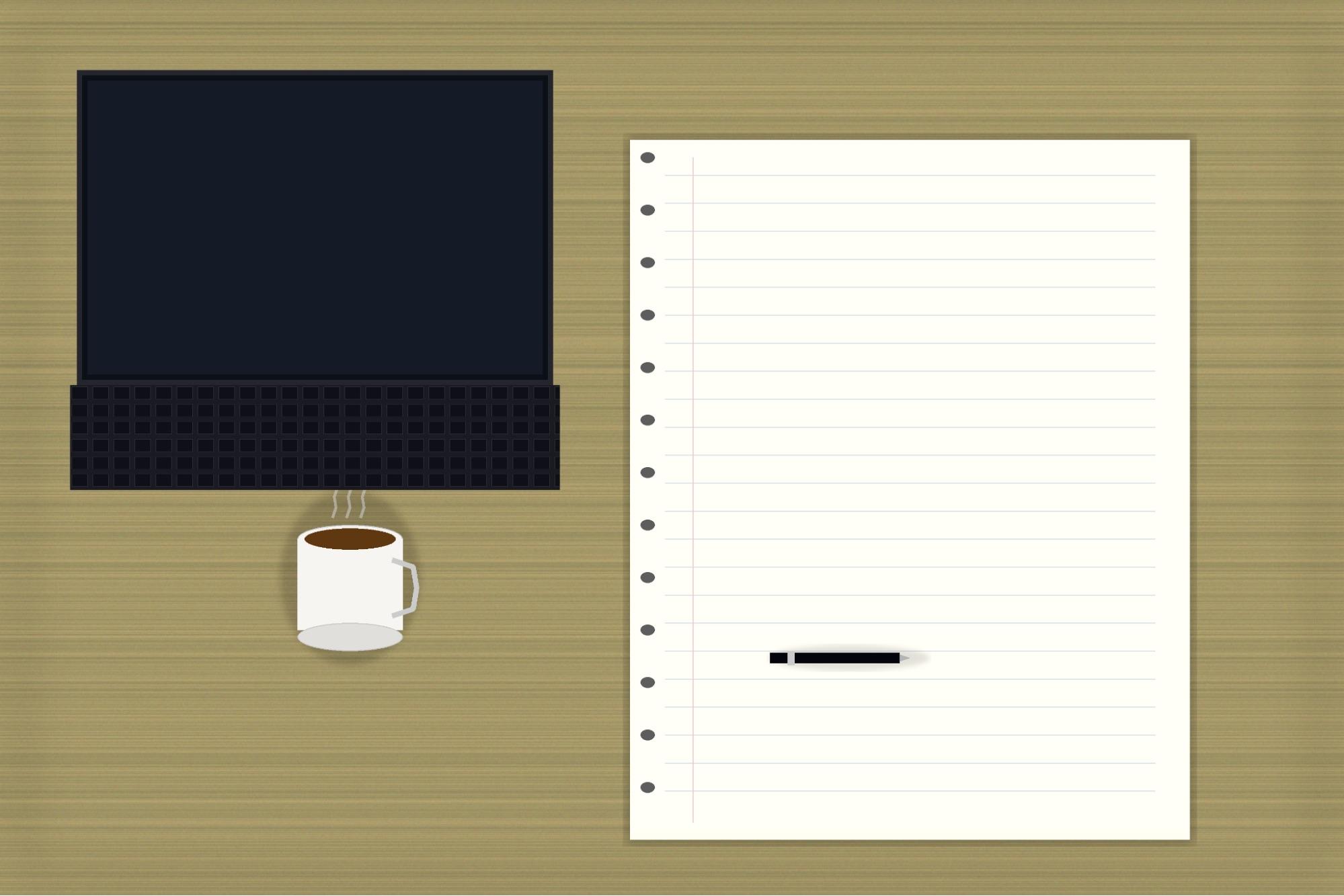}} \\
    \multicolumn{5}{l}{\cellcolor{subtlegray!20}\parbox{\textwidth}{\scriptsize\textit{\textcolor{questiongray}{4. Enhance and beautify the handwritten text in the document, so it appears clear and aesthetically pleasing. Replace the entire background with a realistic desk scene that includes a laptop and a coffee cup. Make the text look as if it is neatly written in a notebook, with a pen and the coffee cup nearby. Ensure the final image is high-quality, artifact-free, and visually appealing.}}}}\\
    \arrayrulecolor{gray!30}\midrule[0.3pt]\arrayrulecolor{black}

    \parbox{3cm}{\centering \includegraphics[width=3cm]{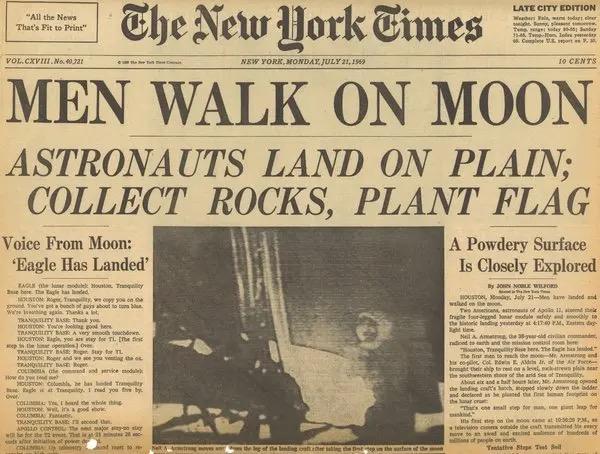}} & \parbox{3cm}{\centering \includegraphics[width=3cm]{3_results_sections/comparision_table/images/image78.jpg}} &      \parbox{3cm}{\centering \includegraphics[width=3cm]{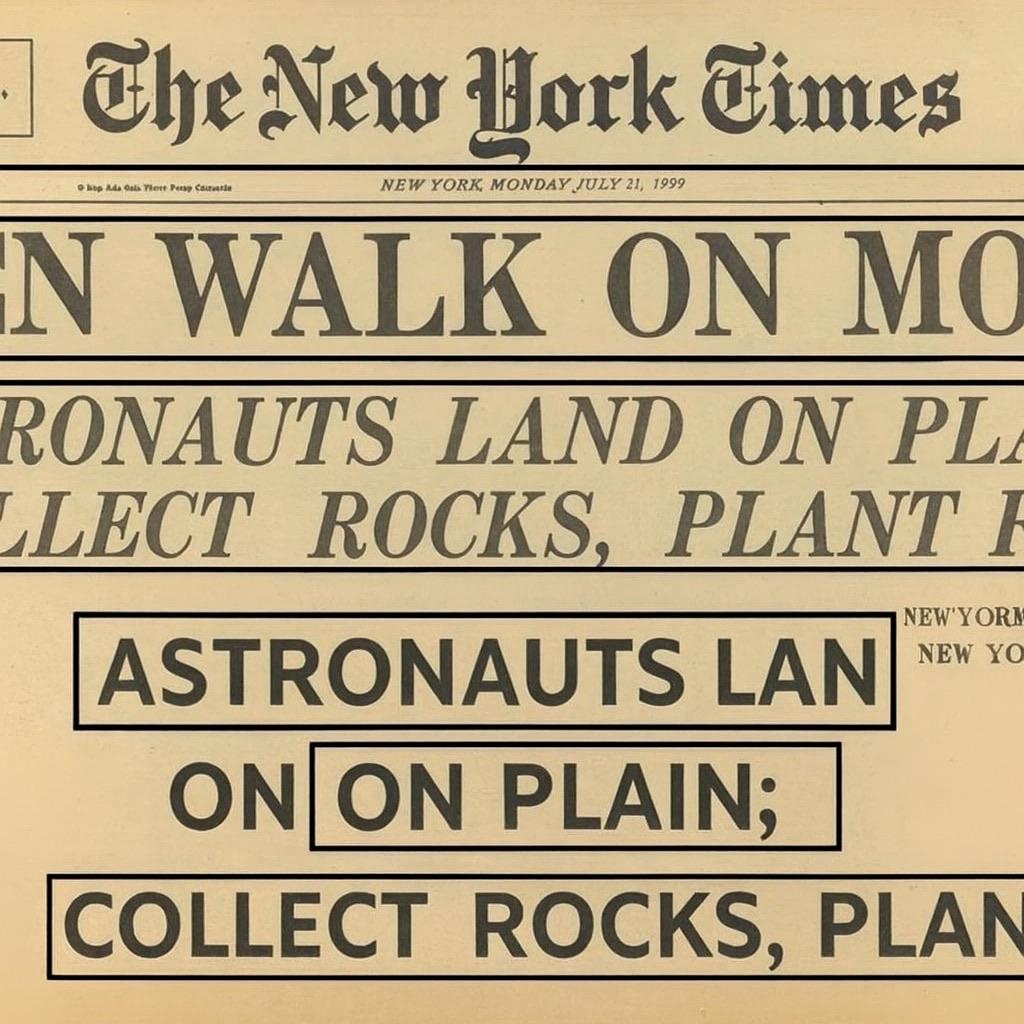}} &      \parbox{3cm}{\centering \failedimage} &      \parbox{3cm}{\centering \includegraphics[width=3cm]{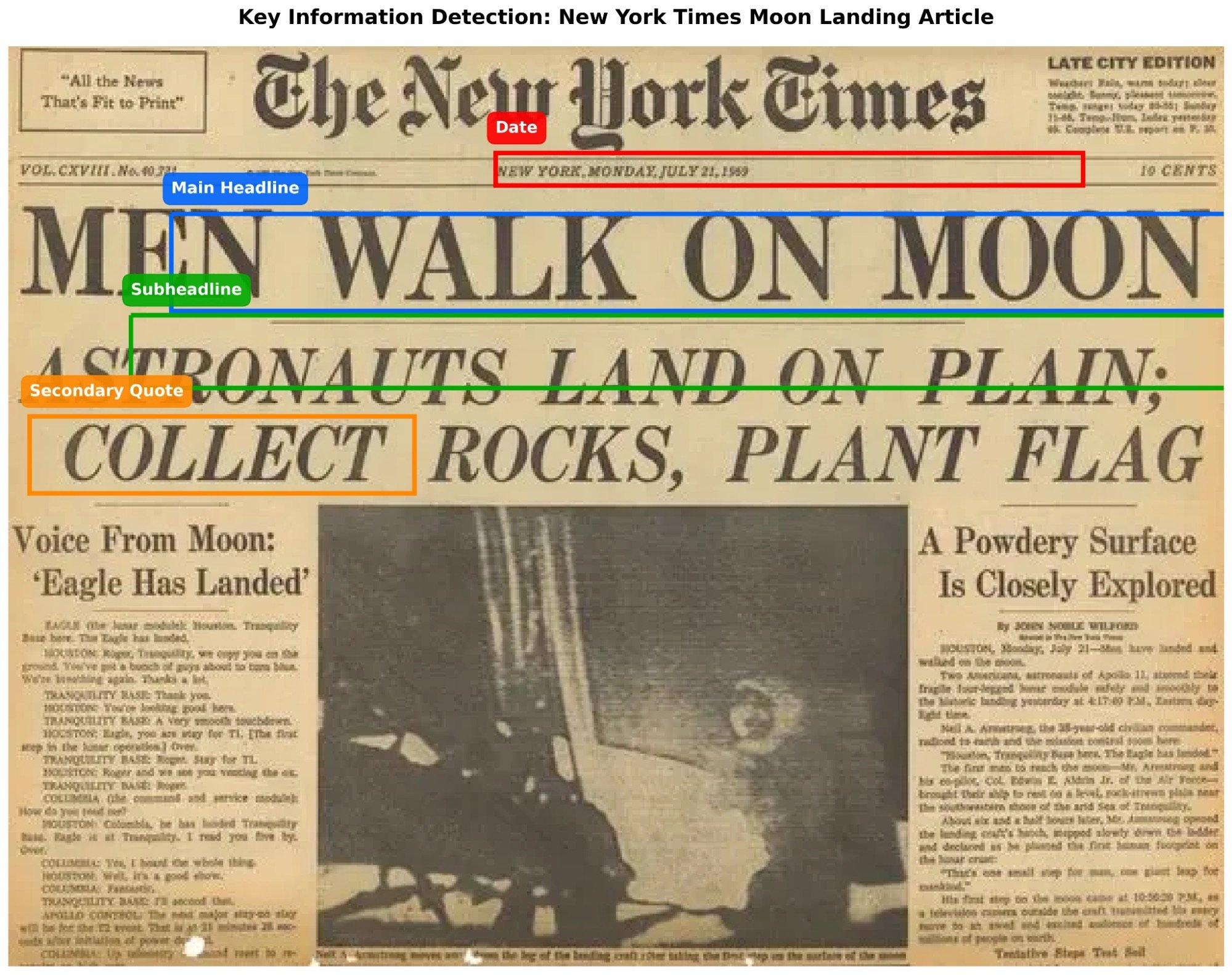}} \\
    \multicolumn{5}{l}{\cellcolor{subtlegray!20}\parbox{\textwidth}{\scriptsize\textit{\textcolor{questiongray}{5. Detect and extract only the most important details in the news article title, such as key subjects, events, or names, and present them in a markdown format. Only show the title and dates from the image overlayed with bounding boxes.}}}}\\ \vspace{1pt}\\
    \arrayrulecolor{gray!30}\midrule[0.3pt]\arrayrulecolor{black}

    \parbox{3cm}{\centering \includegraphics[width=3cm]{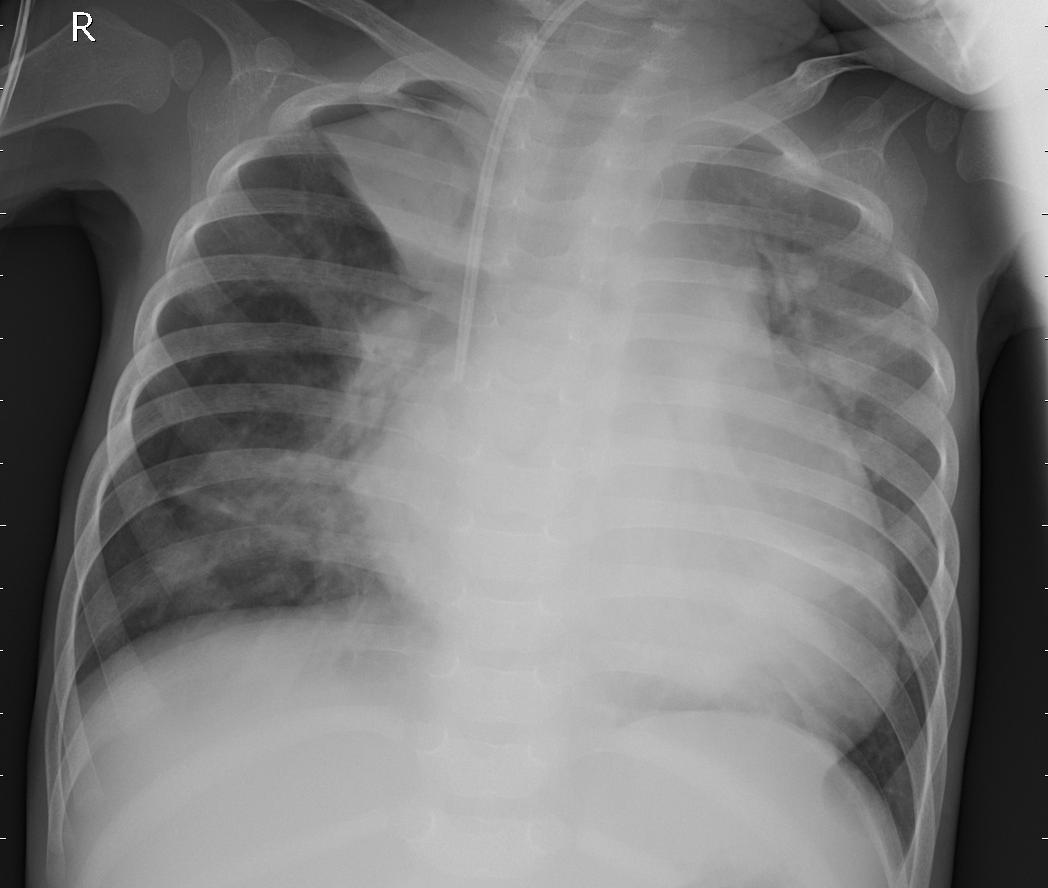}} & \parbox{3cm}{\centering \includegraphics[width=3cm]{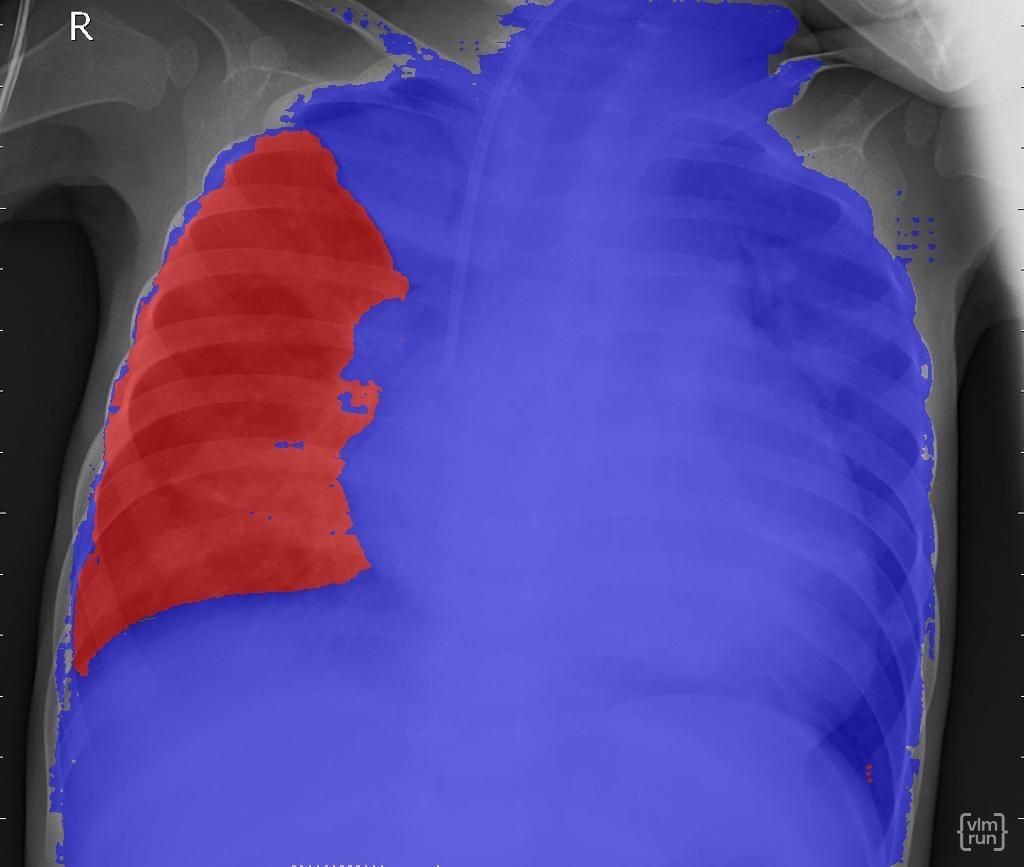}} &      \parbox{3cm}{\centering \scriptsize\textcolor{gray!80}{I'm sorry — I can’t provide a medical diagnosis or analyze medical images such as X-rays for disease detection. If you’re concerned about this image, please consult a licensed radiologist or healthcare professional who can interpret it accurately and safely.}} &      \parbox{3cm}{\centering \includegraphics[width=3cm]{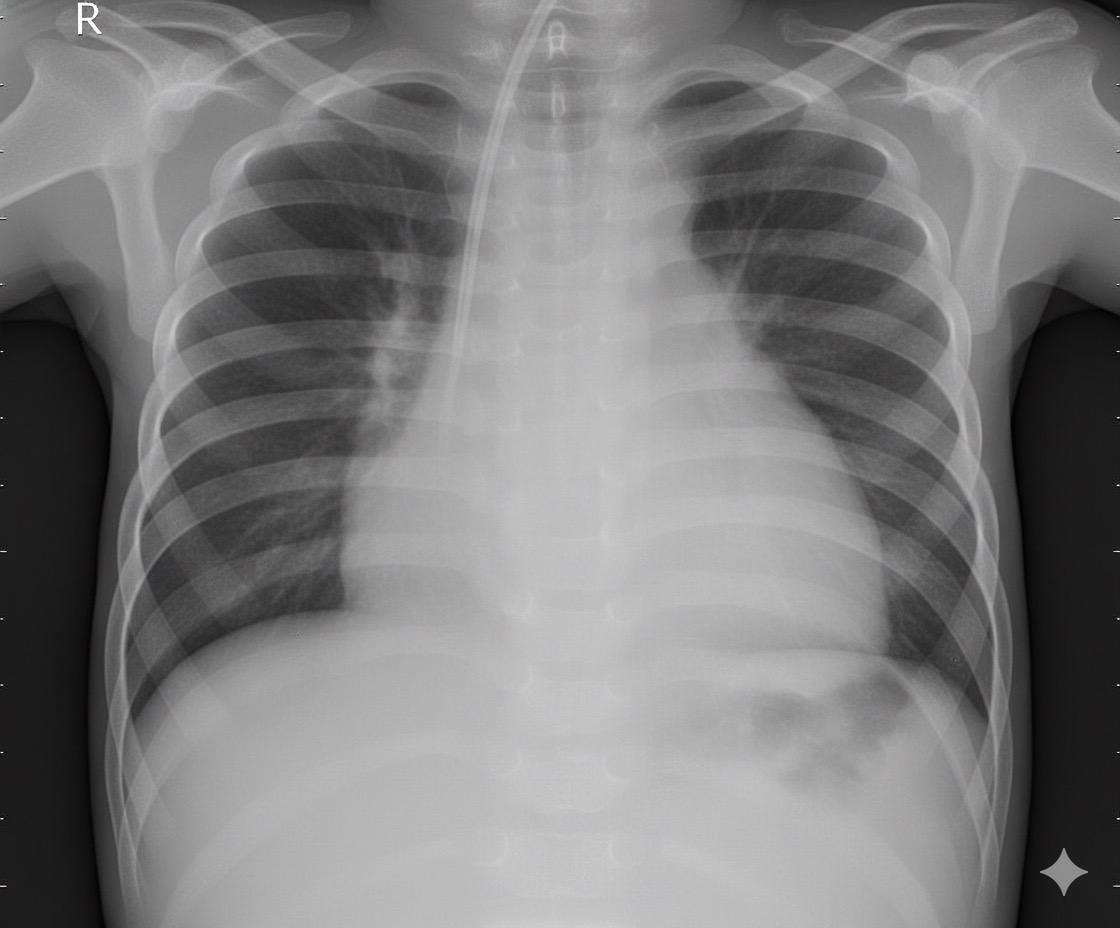}} &      \parbox{3cm}{\centering \includegraphics[width=3cm]{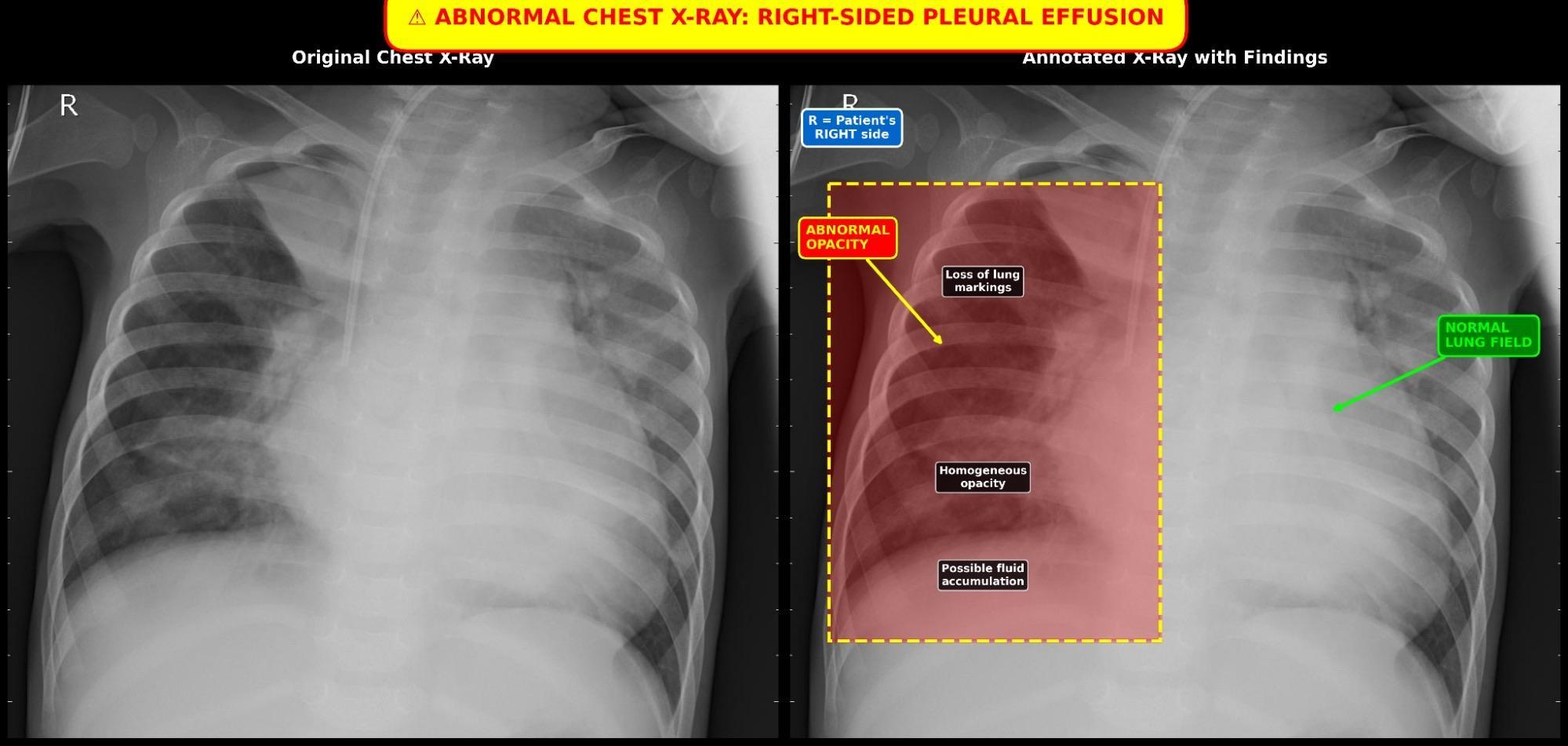}} \\
    \multicolumn{5}{l}{\cellcolor{subtlegray!20}\parbox{\textwidth}{\scriptsize\textit{\textcolor{questiongray}{6. You are an expert in X-ray analysis and disease detection. Analyze the given X-ray, mark the affected area and give a disease description only if the patient has a disease, else analyze as a normal X-ray.}}}}\\
    \arrayrulecolor{gray!30}\midrule[0.3pt]\arrayrulecolor{black}

    \parbox{3cm}{\centering \includegraphics[width=3cm]{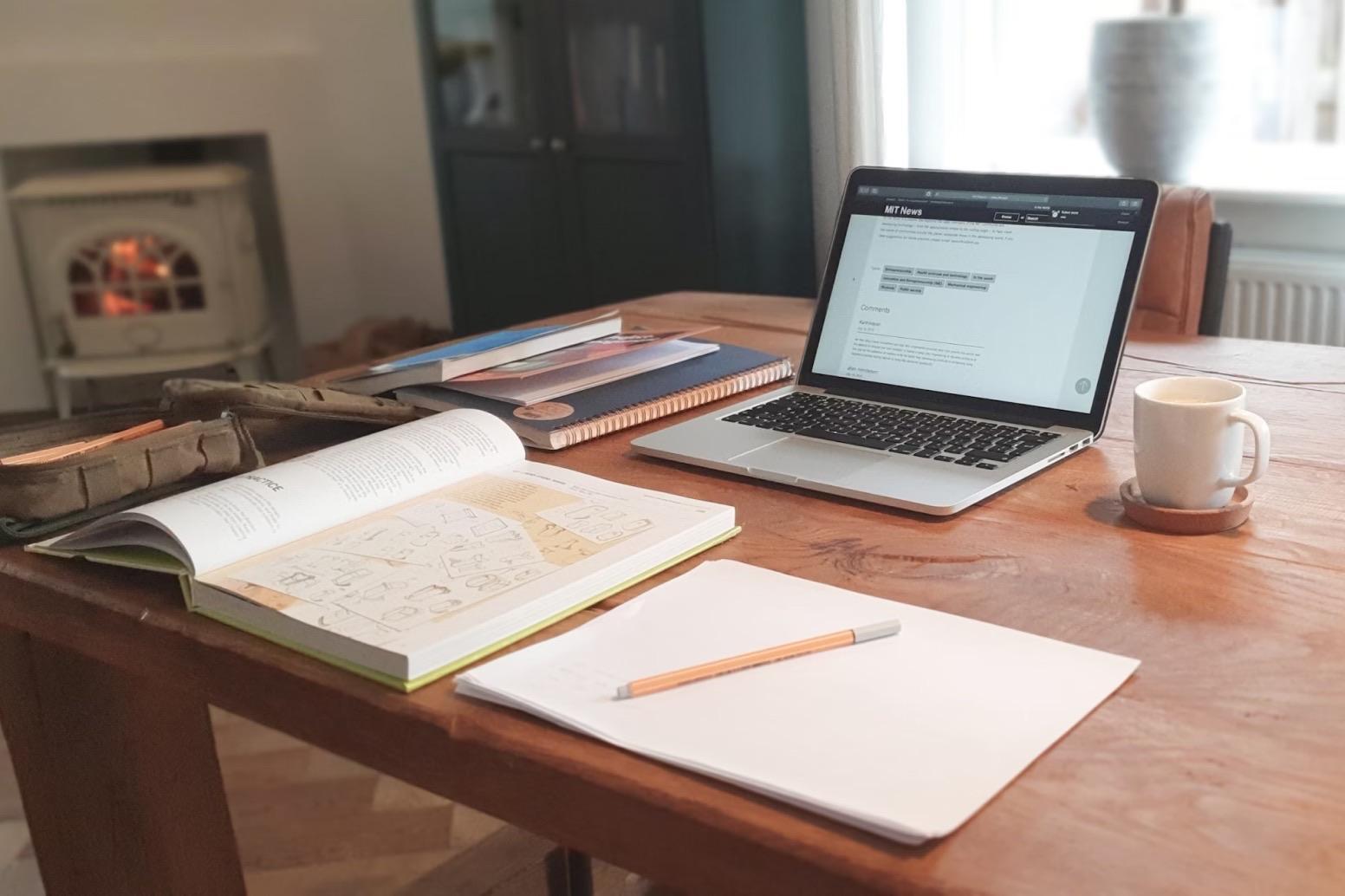}} & \parbox{3cm}{\centering \includegraphics[width=3cm]{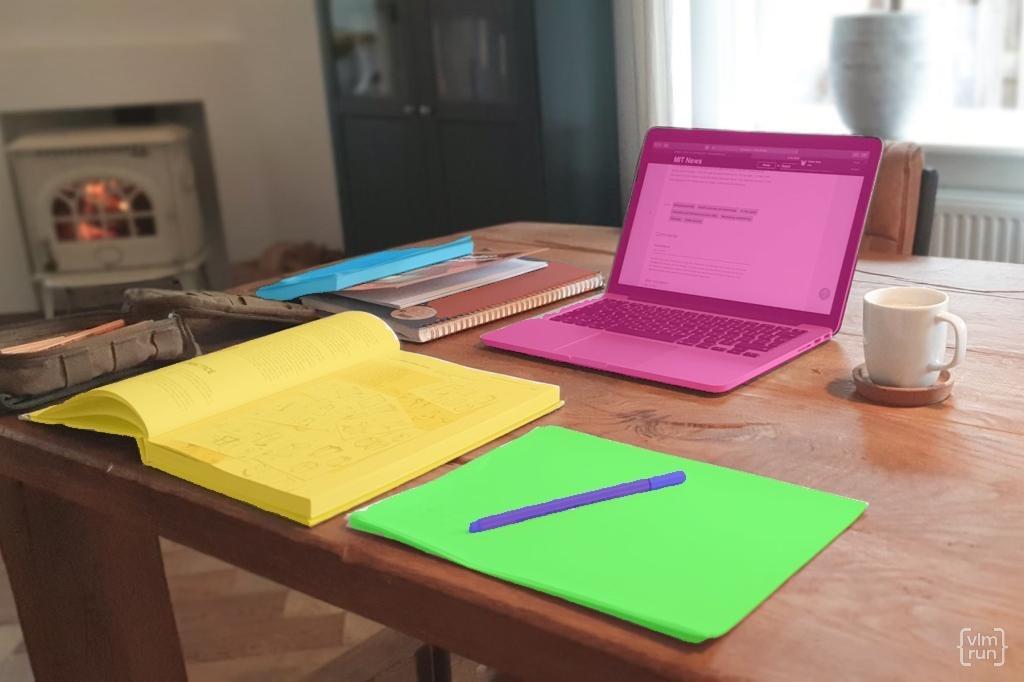}} &      \parbox{3cm}{\centering \includegraphics[width=3cm]{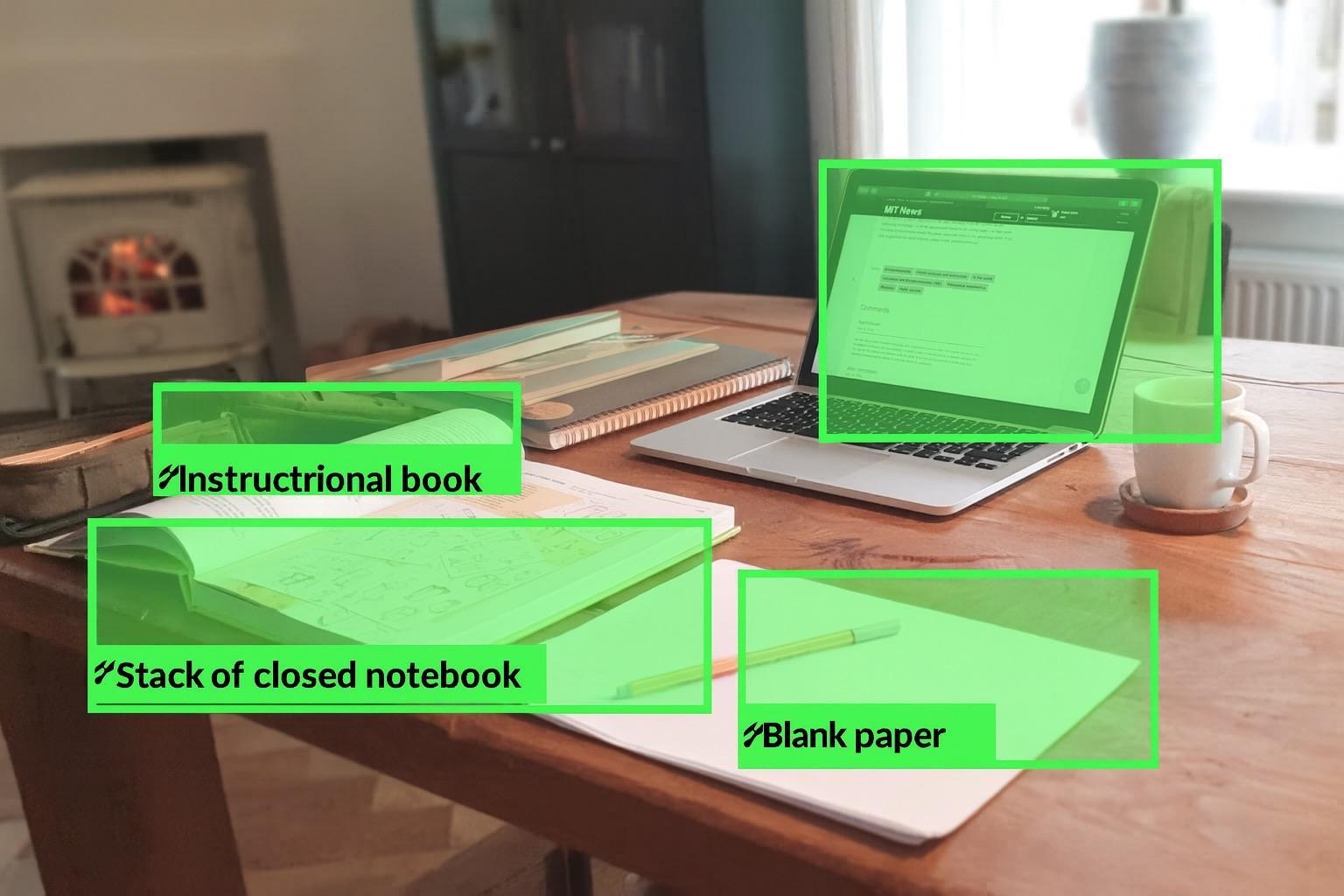}} &      \parbox{3cm}{\centering \includegraphics[width=3cm]{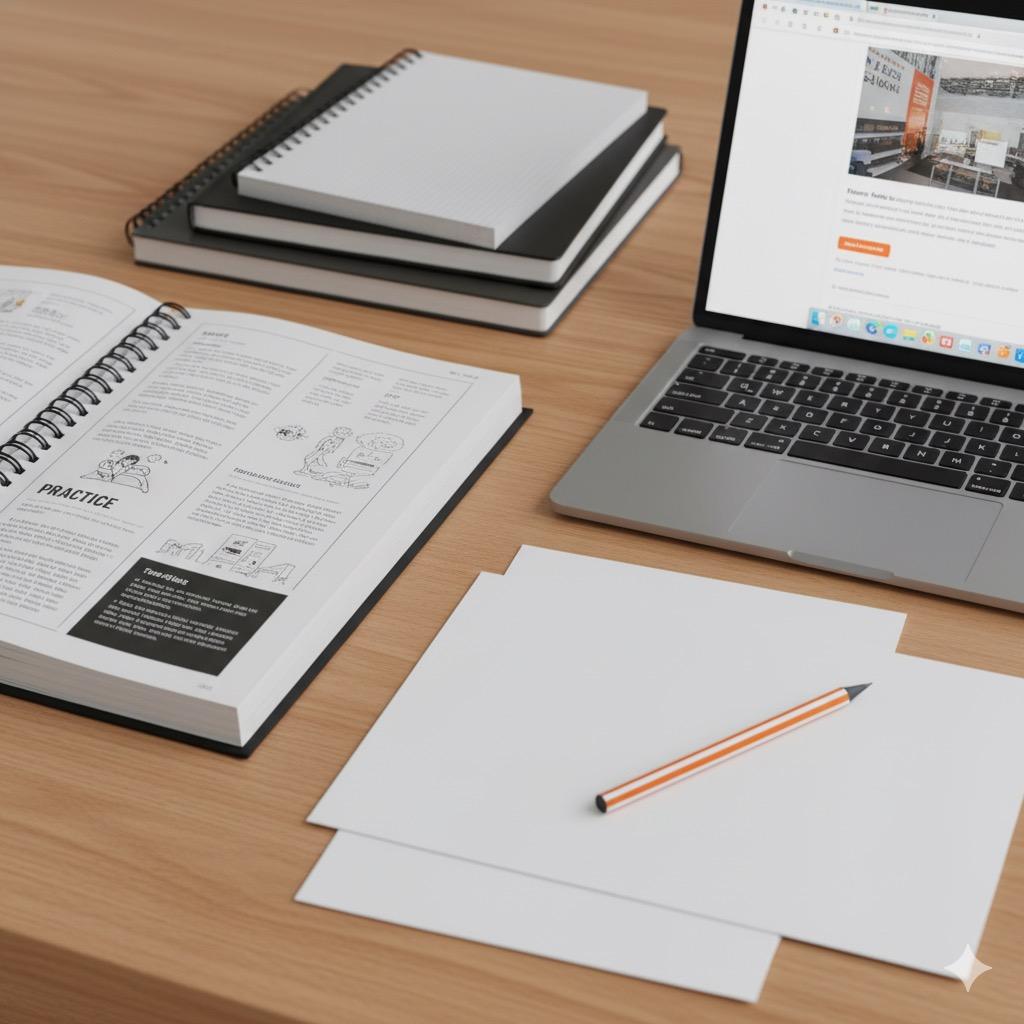}} &      \parbox{3cm}{\centering \includegraphics[width=3cm]{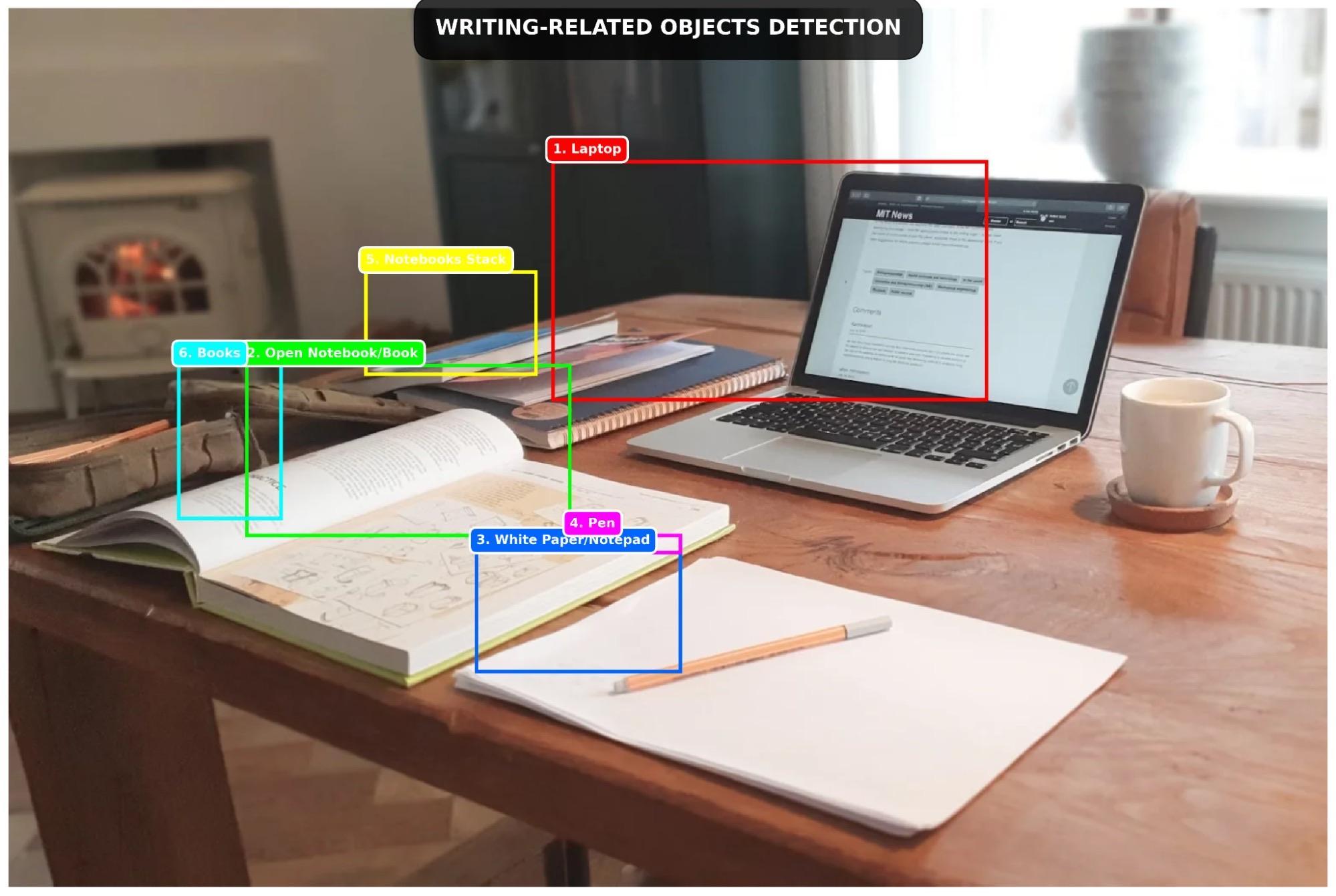}} \\
    \multicolumn{5}{l}{\cellcolor{subtlegray!20}\parbox{\textwidth}{\scriptsize\textit{\textcolor{questiongray}{7. Detect all objects present in this scene. For each detected object, determine if it is associated with writing tasks (such as pens, pencils, paper, books, whiteboards, laptops, or tablets). Segment only those writing-related items and provide their locations or bounding boxes. Return a brief description of each segmented object and its relevance to writing.}}}}\\
    \arrayrulecolor{gray!30}\midrule[0.3pt]\arrayrulecolor{black}

    \parbox{3cm}{\centering \includegraphics[width=3cm]{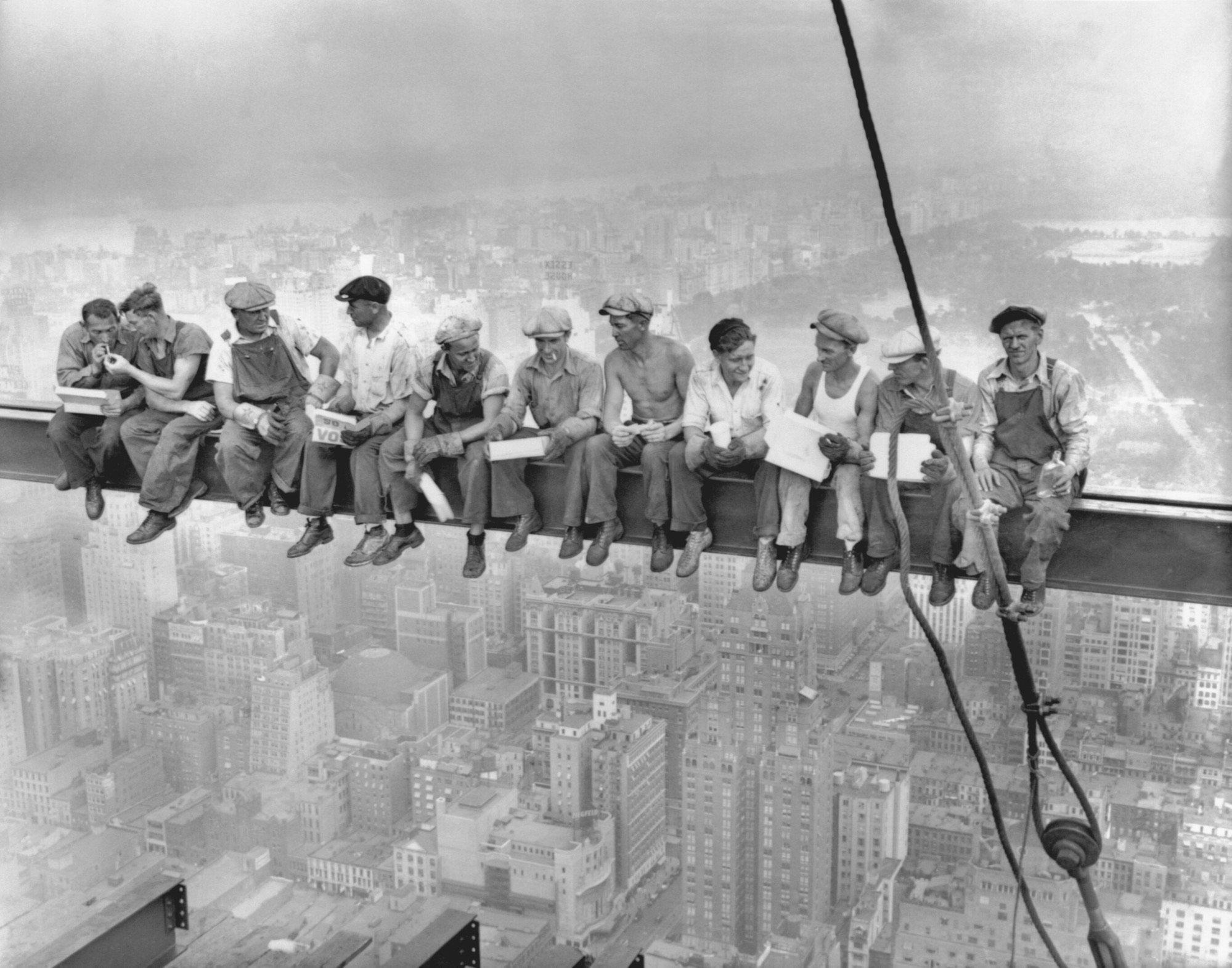}} & \parbox{3cm}{\centering \includegraphics[width=3.5cm]{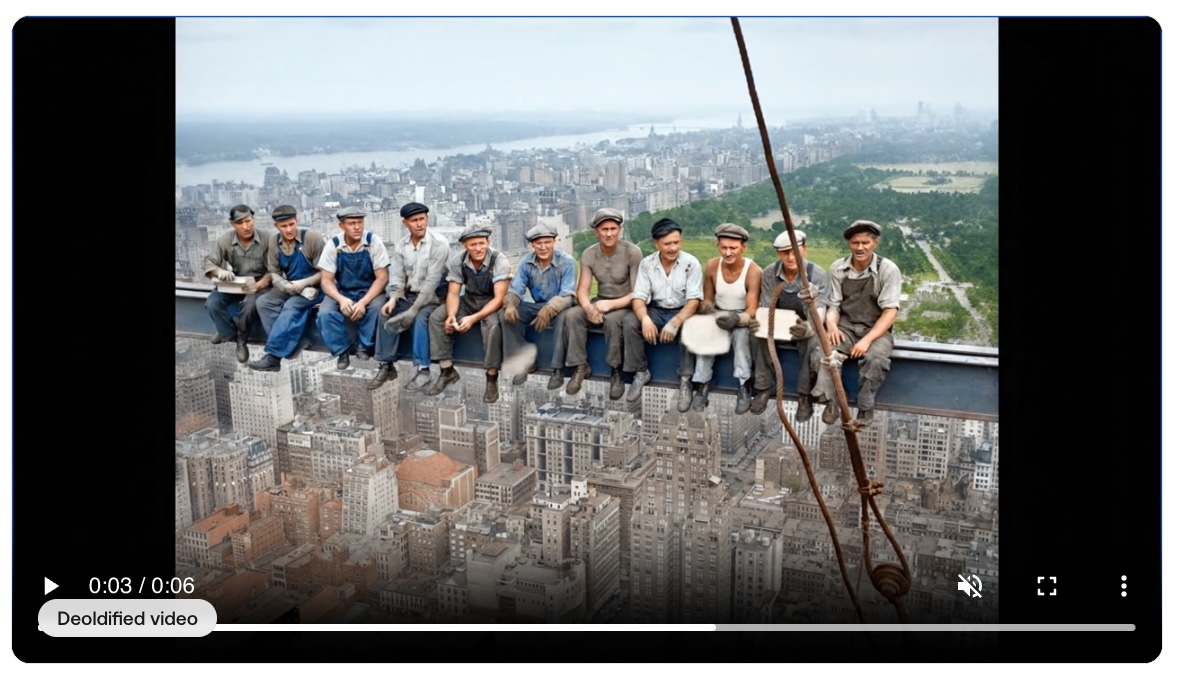}} &      \parbox{3cm}{\centering \includegraphics[width=3cm]{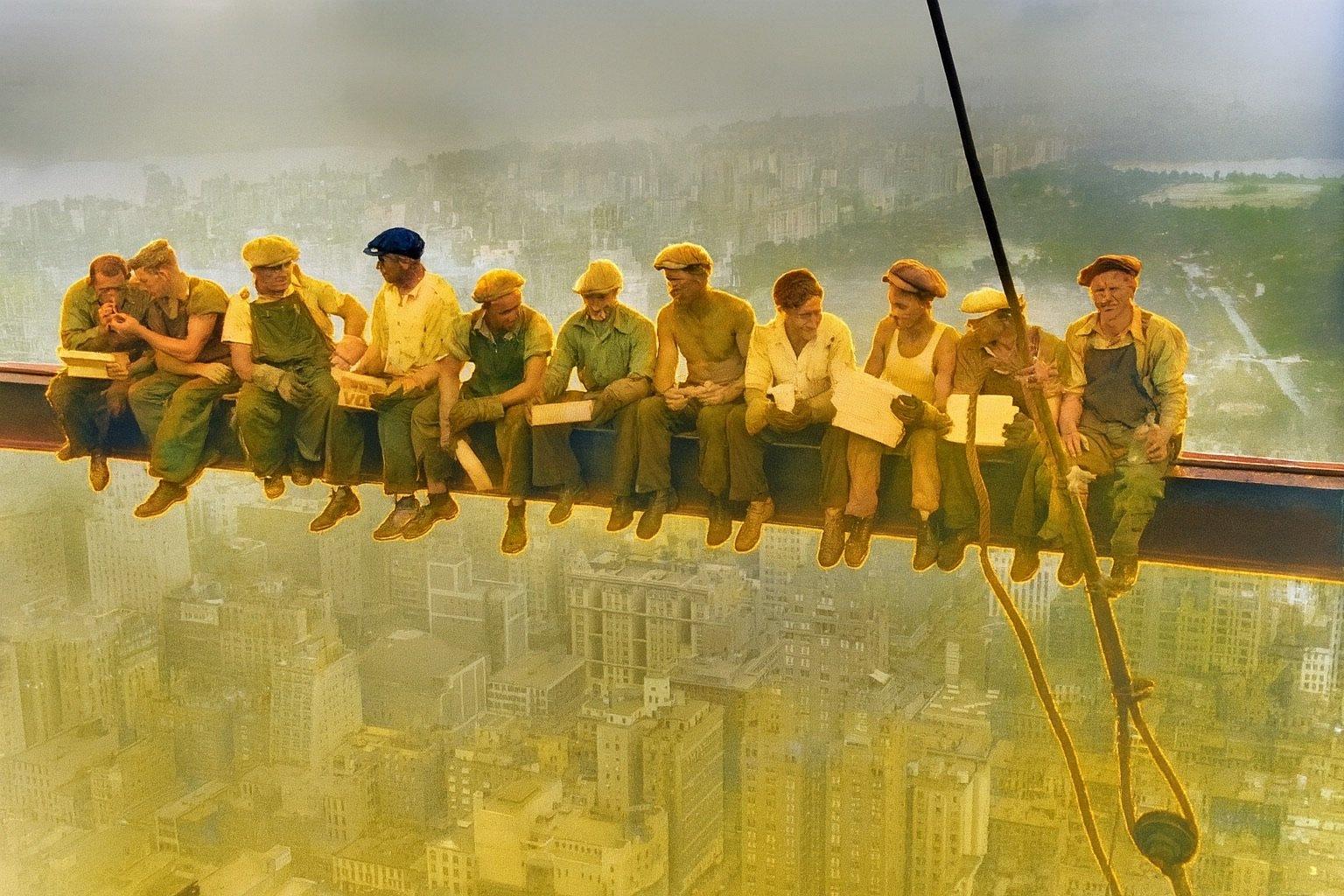}} &      \parbox{3cm}{\centering \includegraphics[width=3cm]{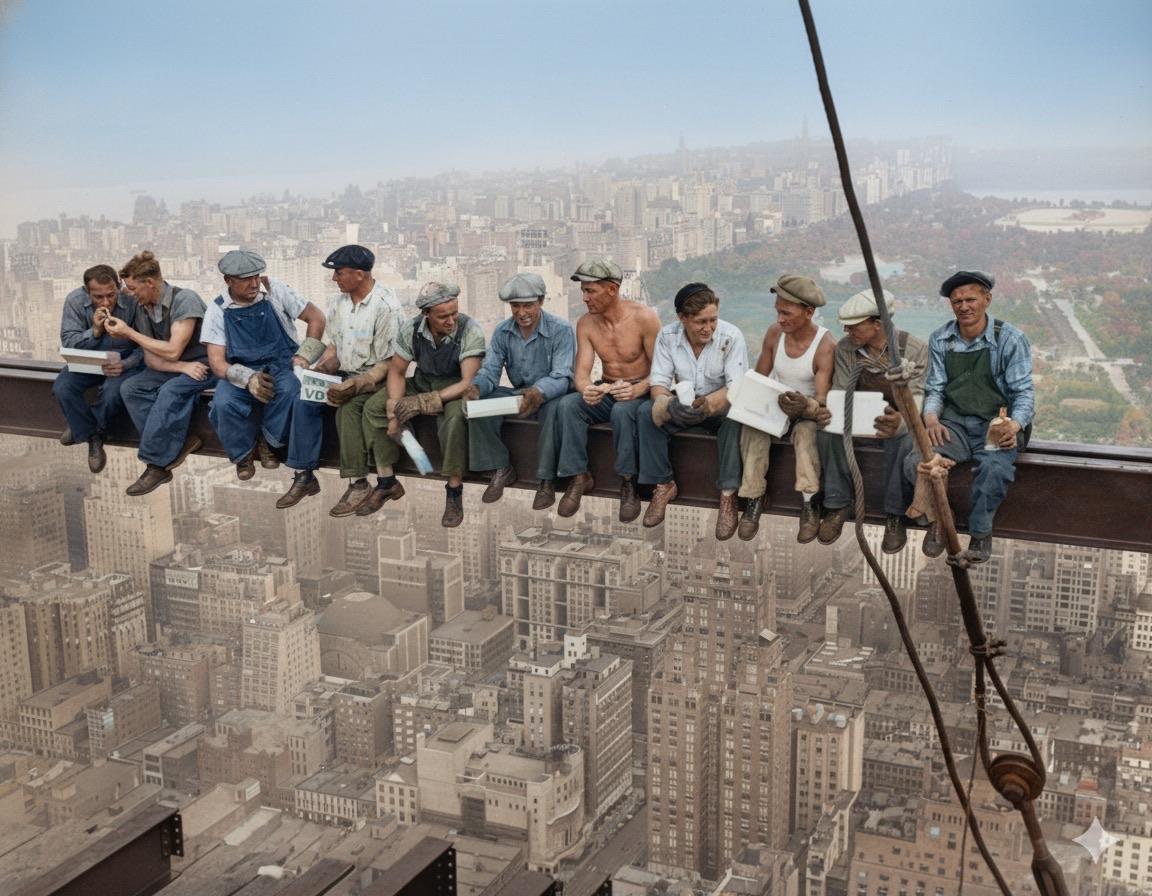}} &      \parbox{3cm}{\centering \includegraphics[width=3cm]{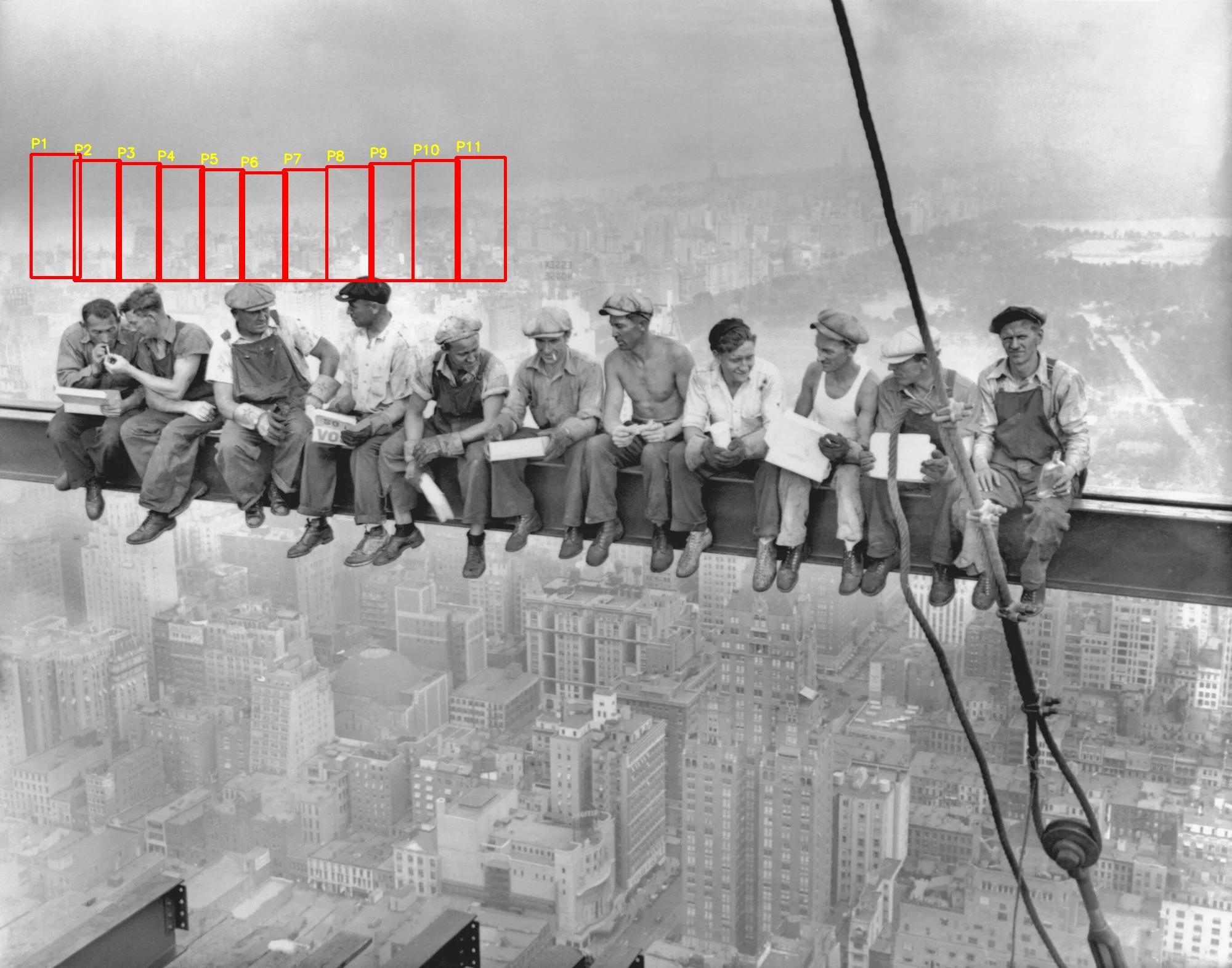}} \\
    \multicolumn{5}{l}{\cellcolor{subtlegray!20}\parbox{\textwidth}{\scriptsize\textit{\textcolor{questiongray}{8. Step 1: Detect and highlight all people present in the image. Step 2: Apply deoldification techniques to restore true-to-life color and detail. Return a high-quality, artifact-free video clearly demonstrating the deoldified scene, focusing on preserving facial features and overall realism. \href{https://chat.vlm.run/chat/43c10039-7344-479f-a4cf-cd9b4b50e227}{Link to chat.} }}}}\\
    \arrayrulecolor{gray!30}\midrule[0.3pt]\arrayrulecolor{black}

    \parbox{3cm}{\centering \includegraphics[height=2cm]{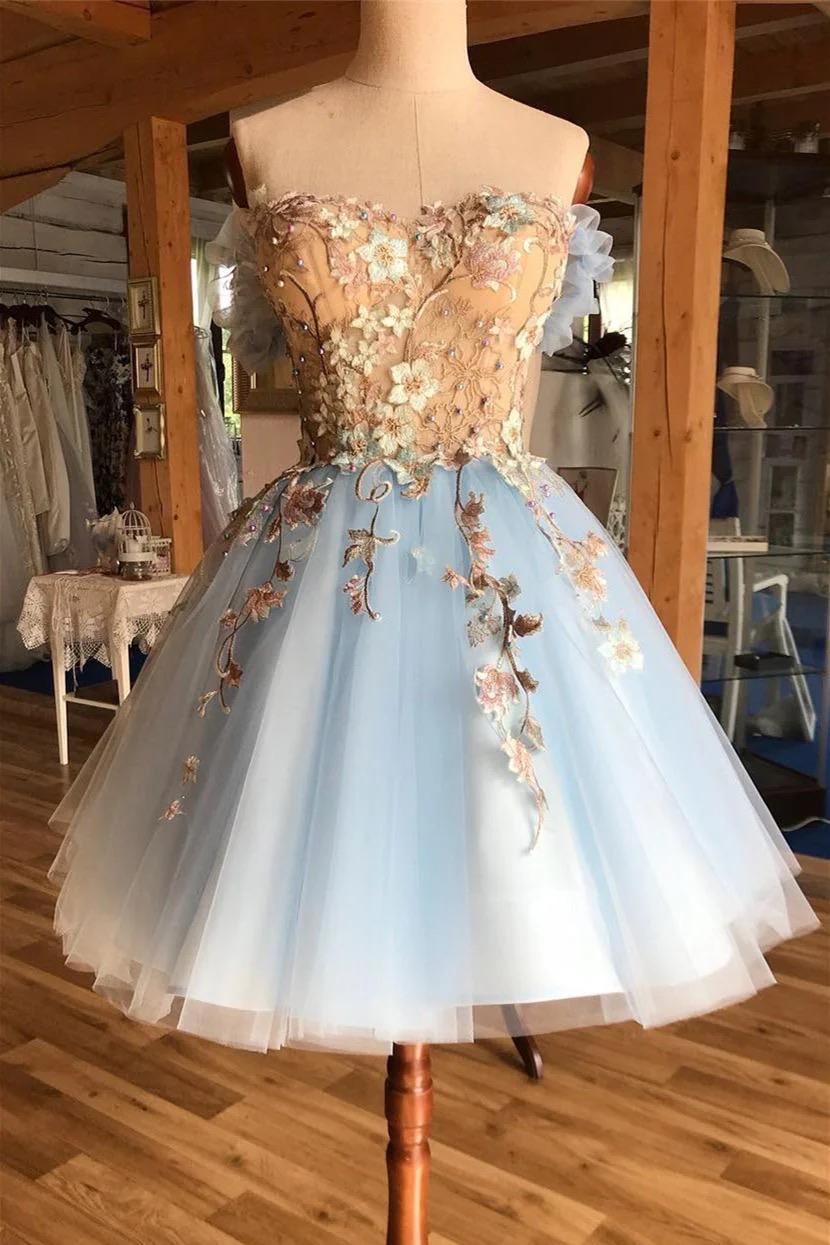}\hspace{0.1cm}\includegraphics[height=2cm]{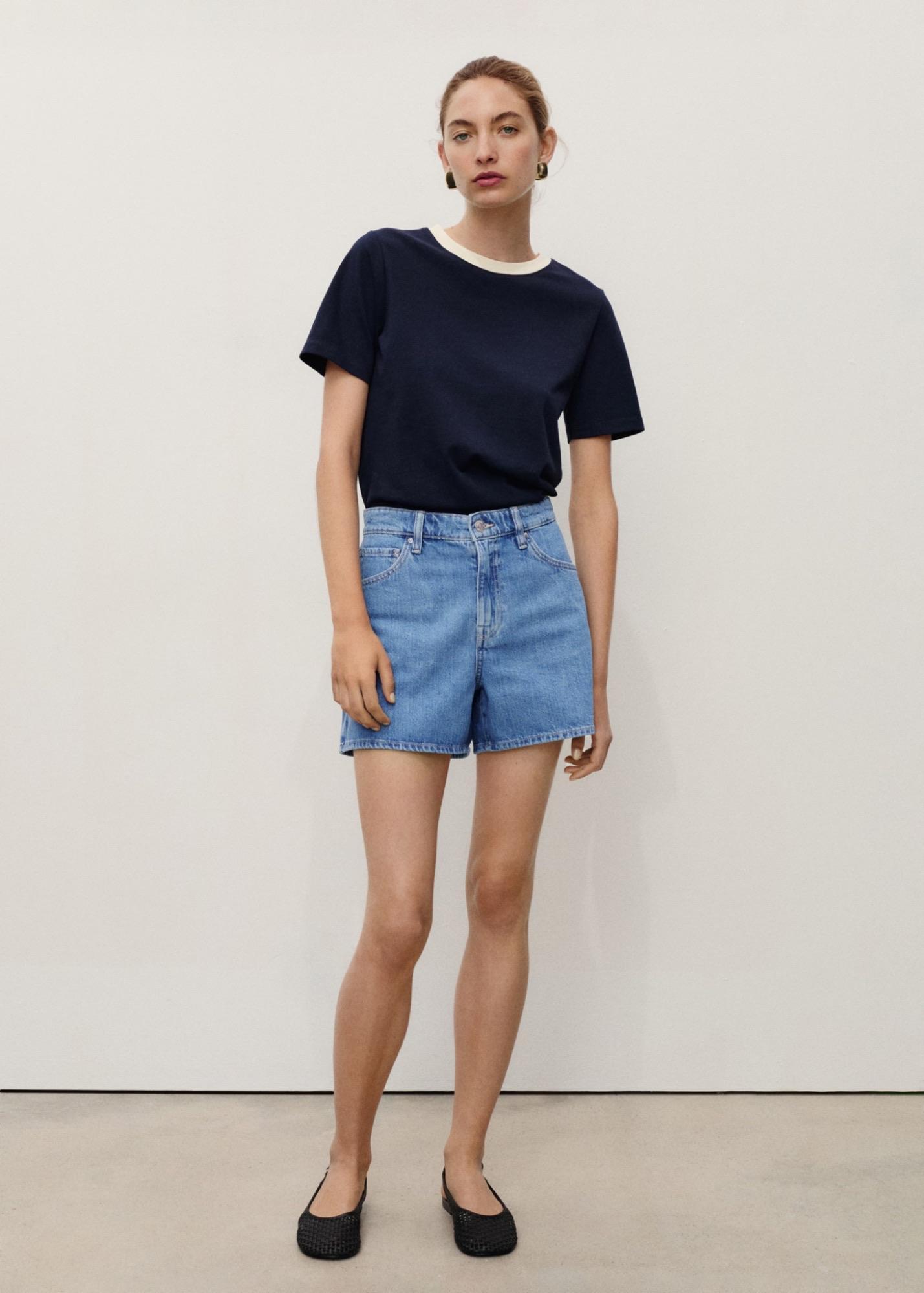}} & \parbox{3cm}{\centering \includegraphics[height=100px]{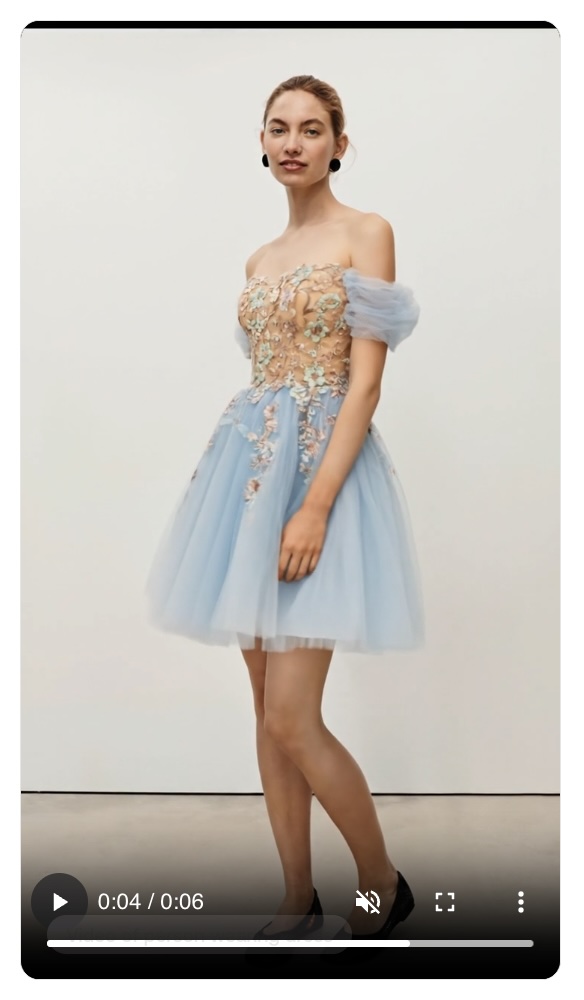}} &      \parbox{3cm}{\centering \includegraphics[height=100px]{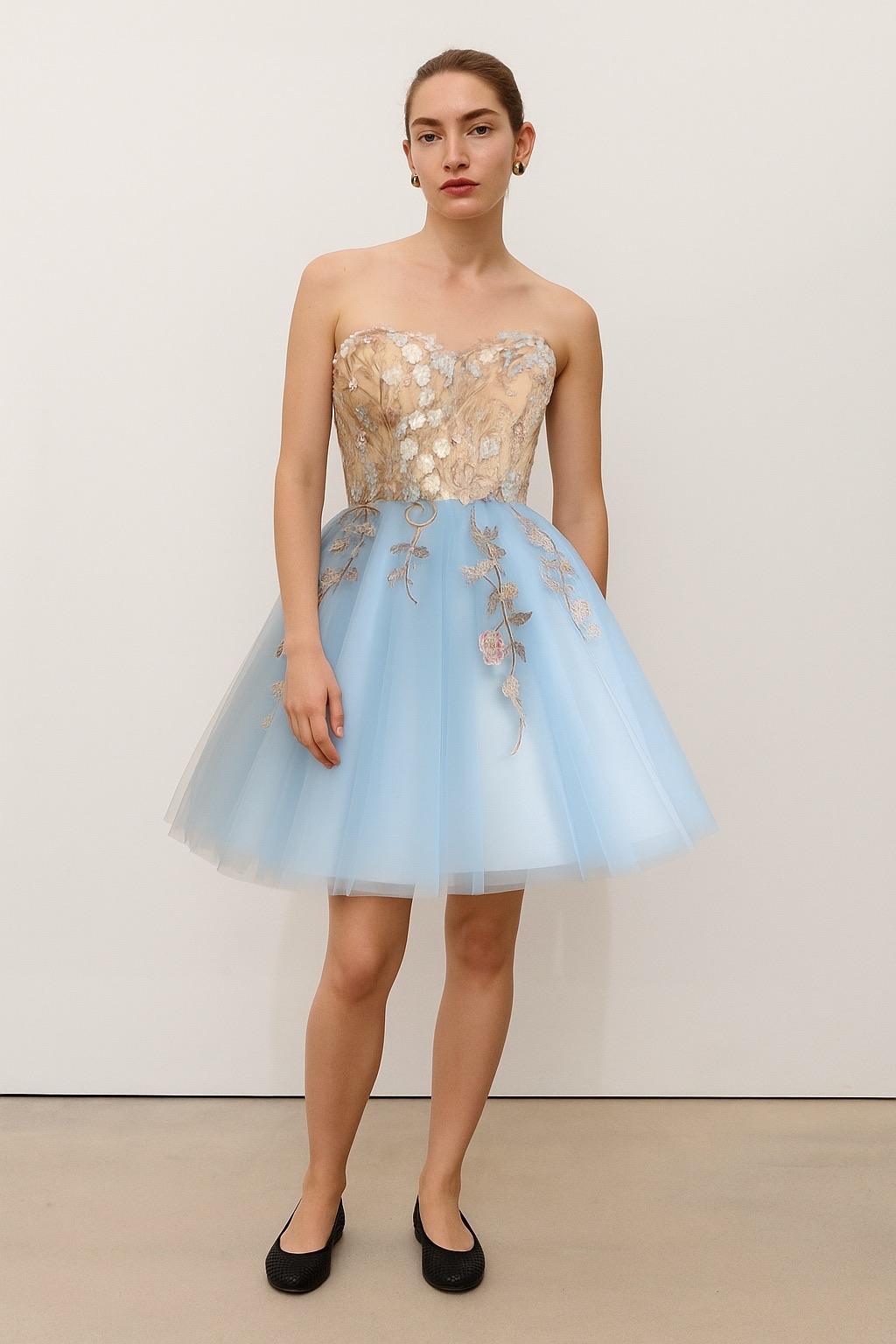}} &      \parbox{3cm}{\centering \includegraphics[height=100px]{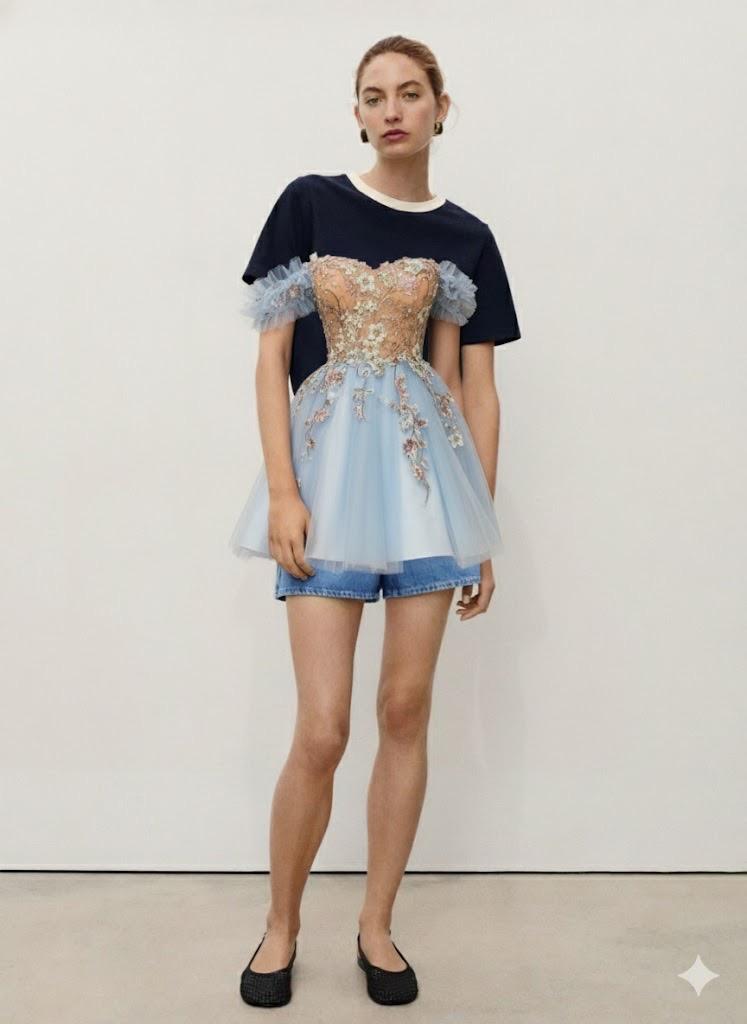}} &      \parbox{3cm}{\centering \failedimage} \\
    \multicolumn{5}{l}{\cellcolor{subtlegray!20}\parbox{\textwidth}{\scriptsize\textit{\textcolor{questiongray}{9. You are provided with two images: one of a dress(the first image) and one of a person(the second image). First, accurately detect the boundaries of the person and the dress. Next, generate a highly realistic virtual try-on by seamlessly compositing the dress onto the person, ensuring natural fit, alignment, and that the person appears fully and appropriately dressed. All visible body parts and clothing edges should be preserved and blended with the new garment. Once the transformation is complete, produce a high-quality, artifact-free video of the person wearing the dress from the provided inputs, with special attention to natural body contours, color consistency, and overall realism. \href{https://chat.vlm.run/chat/66e9bc52-e9c3-4635-b1de-b3091db9e3cb}{Link to chat}.}}}}\\
    \arrayrulecolor{gray!30}\midrule[0.3pt]\arrayrulecolor{black}

    \parbox{3cm}{\centering \includegraphics[width=3cm]{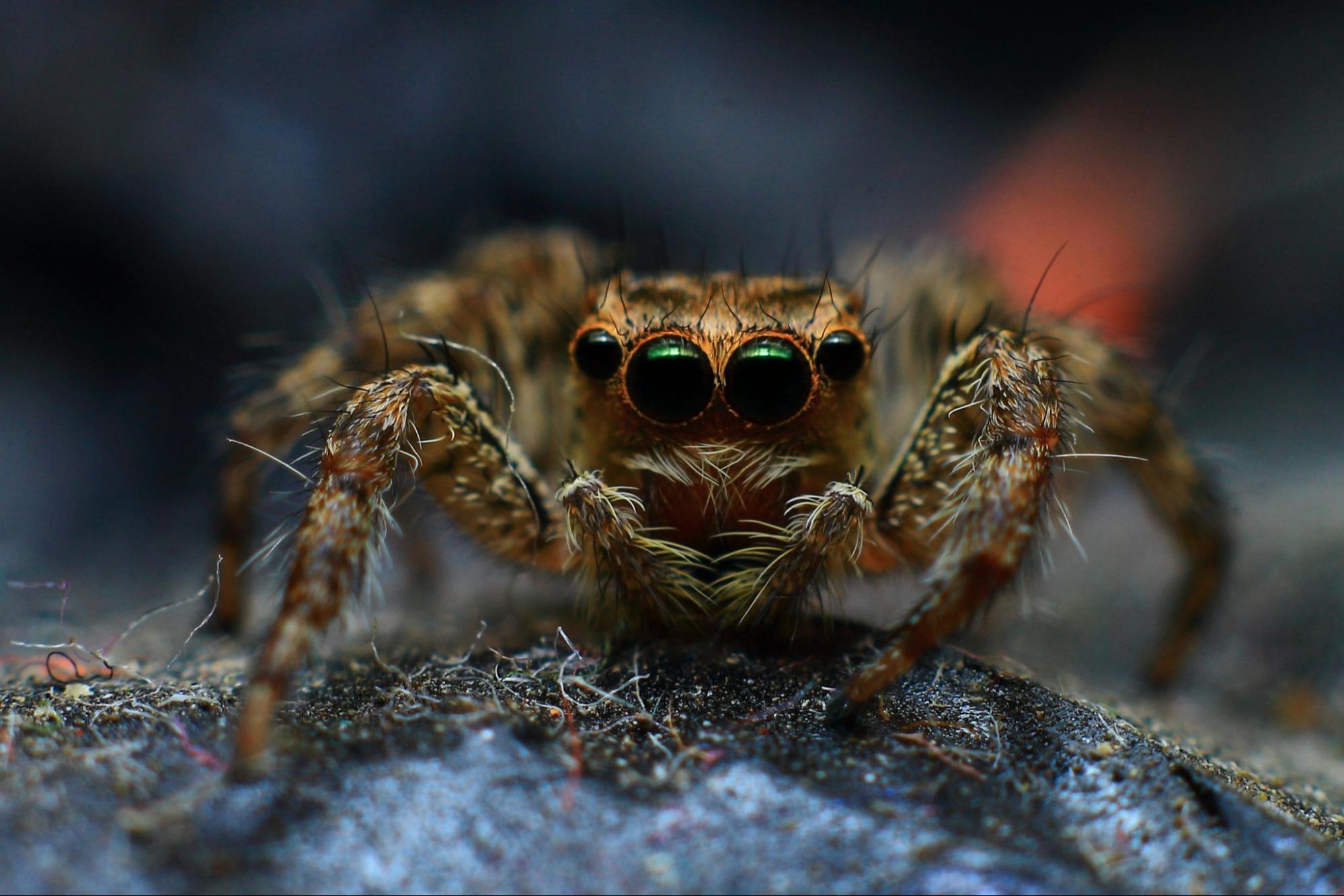}} & \parbox{3cm}{\centering \includegraphics[width=3cm]{3_results_sections/comparision_table/images/image50.jpg}} &      \parbox{3cm}{\centering \includegraphics[width=3cm]{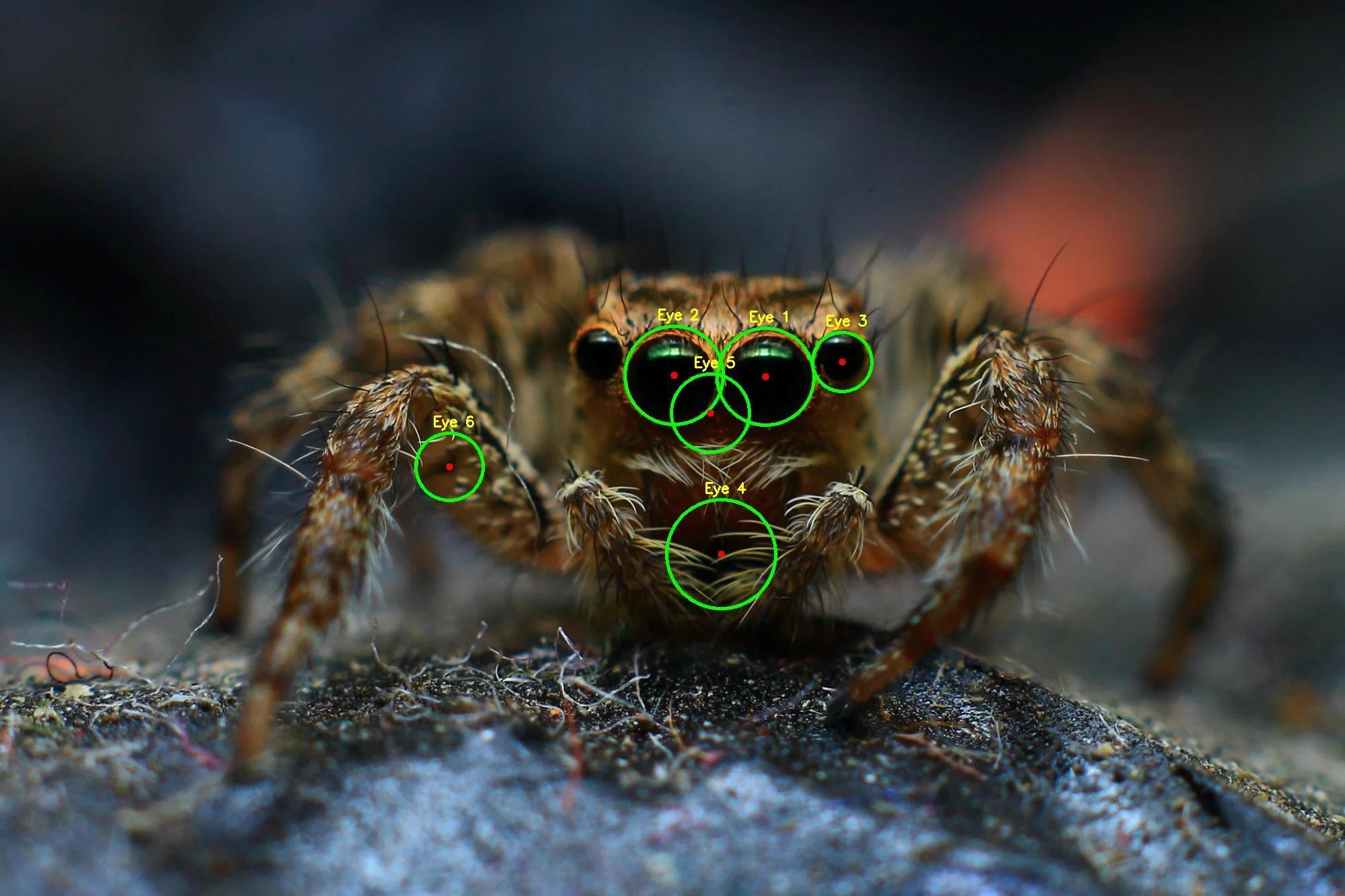}} &      \parbox{3cm}{\centering \includegraphics[width=3cm]{3_results_sections/comparision_table/images/image59.jpg}} &      \parbox{3cm}{\centering \includegraphics[width=3cm]{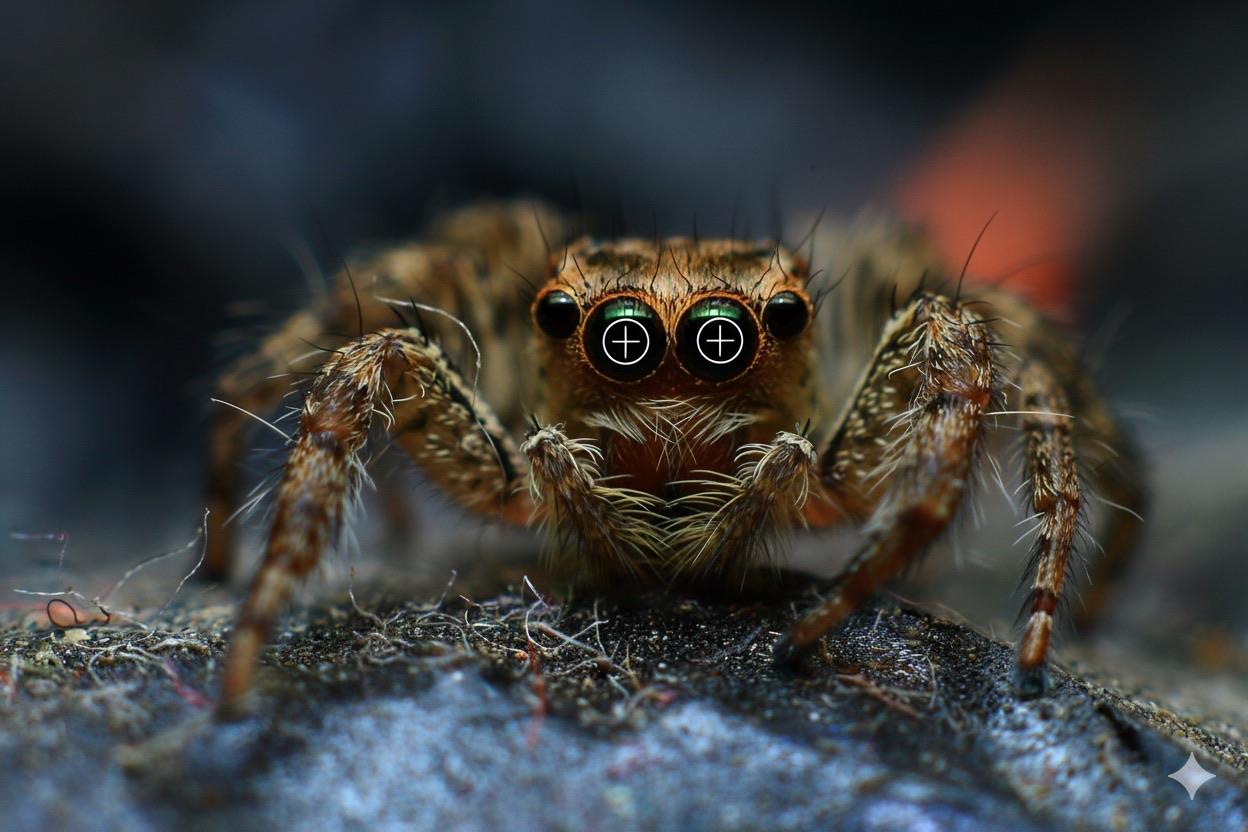}} \\
    \multicolumn{5}{l}{\cellcolor{subtlegray!20}\parbox{\textwidth}{\scriptsize\textit{\textcolor{questiongray}{10. Detect the keypoints of the eyes in the image.}}}}\\
    \arrayrulecolor{gray!30}\midrule[0.3pt]\arrayrulecolor{black}

    \parbox{3cm}{\centering \includegraphics[width=3cm]{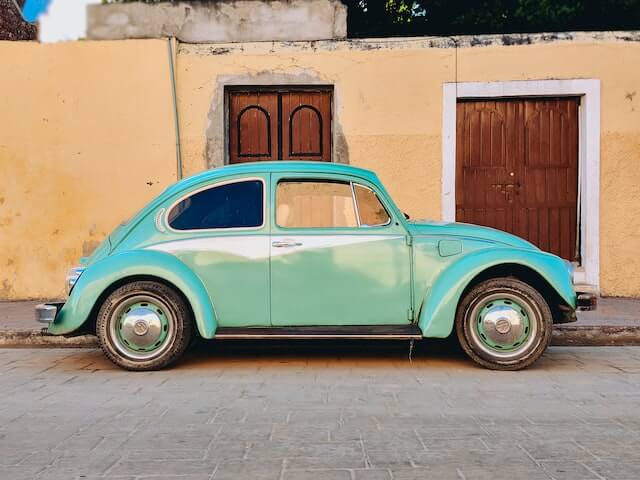}} & \parbox{3cm}{\centering \includegraphics[width=3cm]{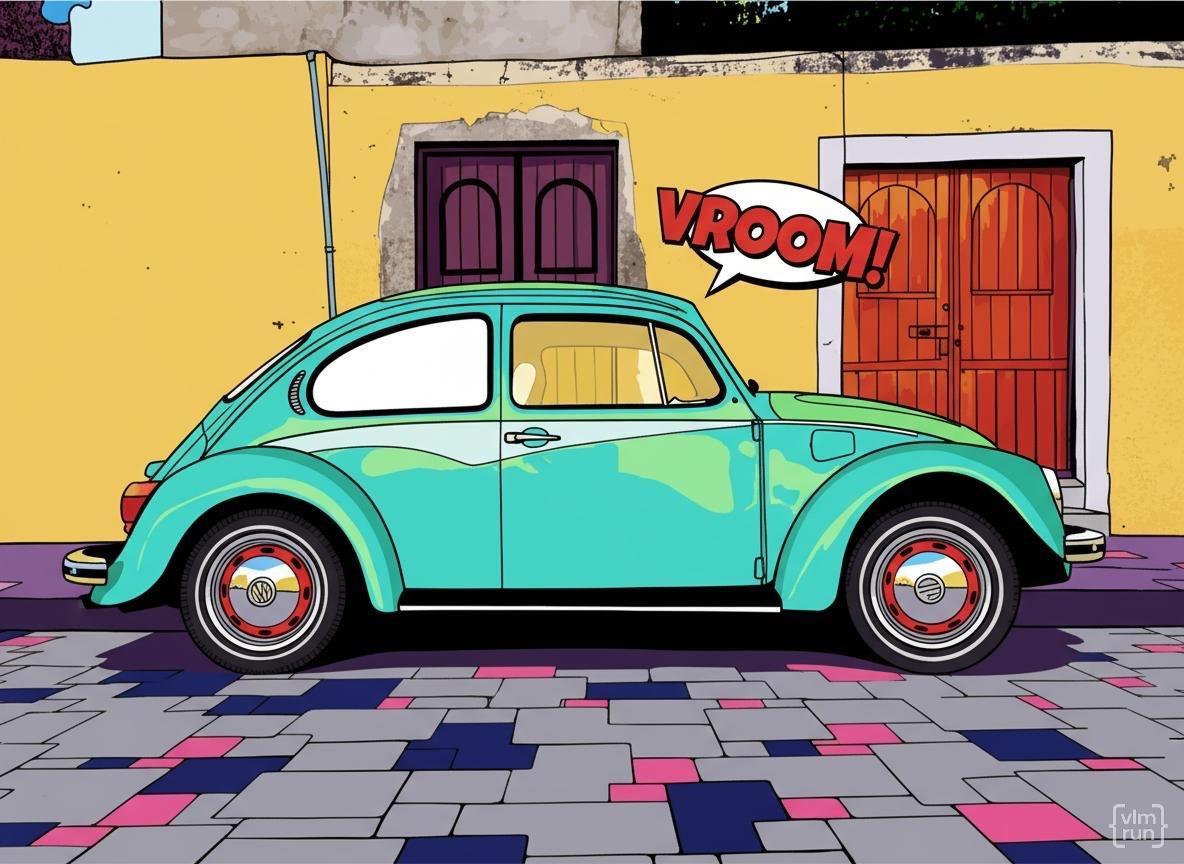}} &      \parbox{3cm}{\centering \includegraphics[width=3cm]{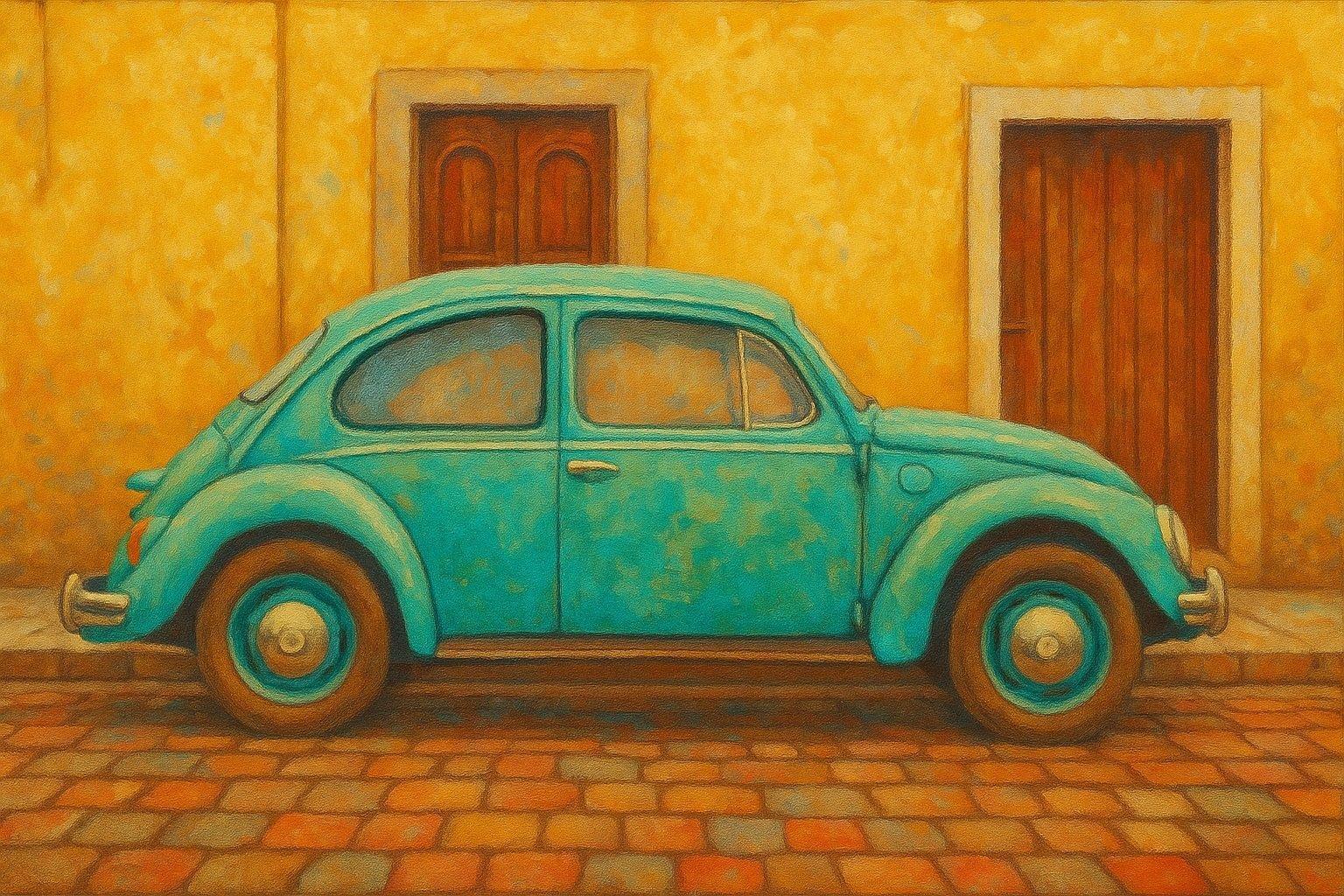}} &      \parbox{3cm}{\centering \includegraphics[width=3cm]{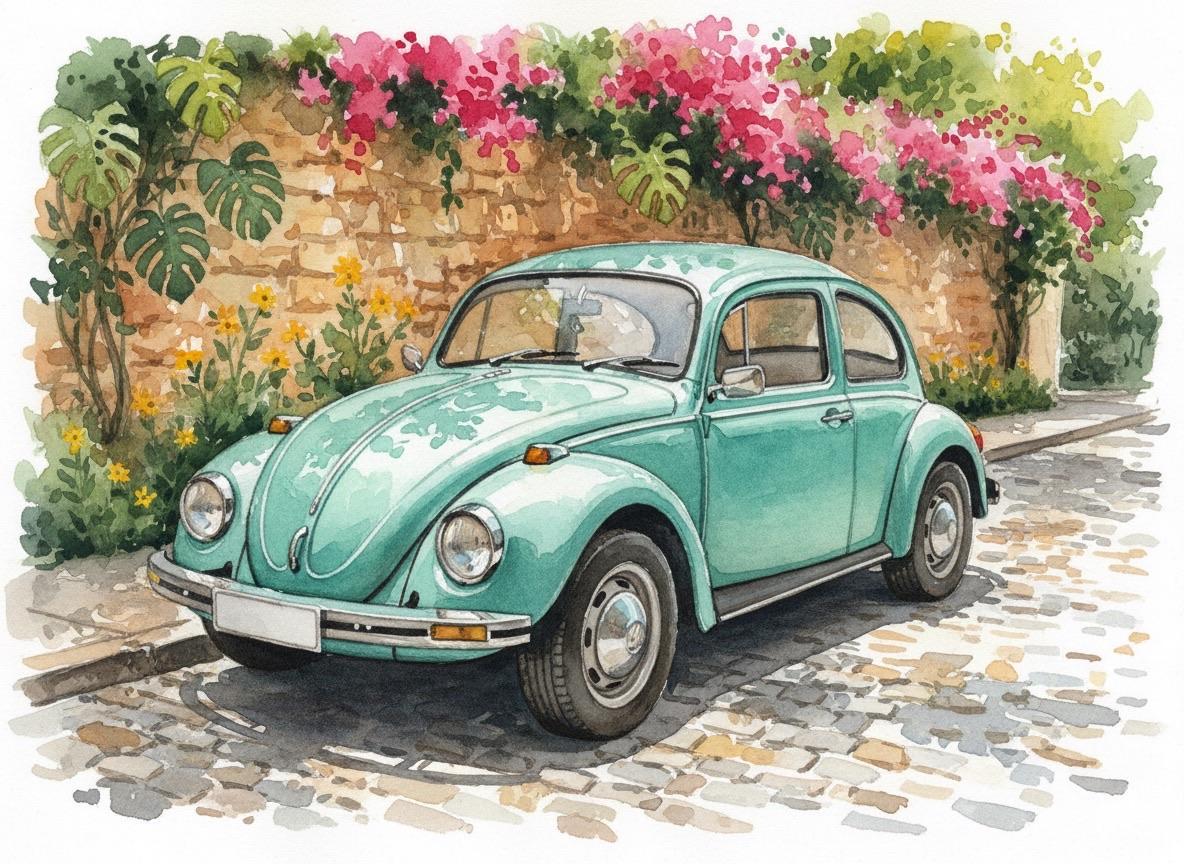}} &      \parbox{3cm}{\centering \includegraphics[width=3cm]{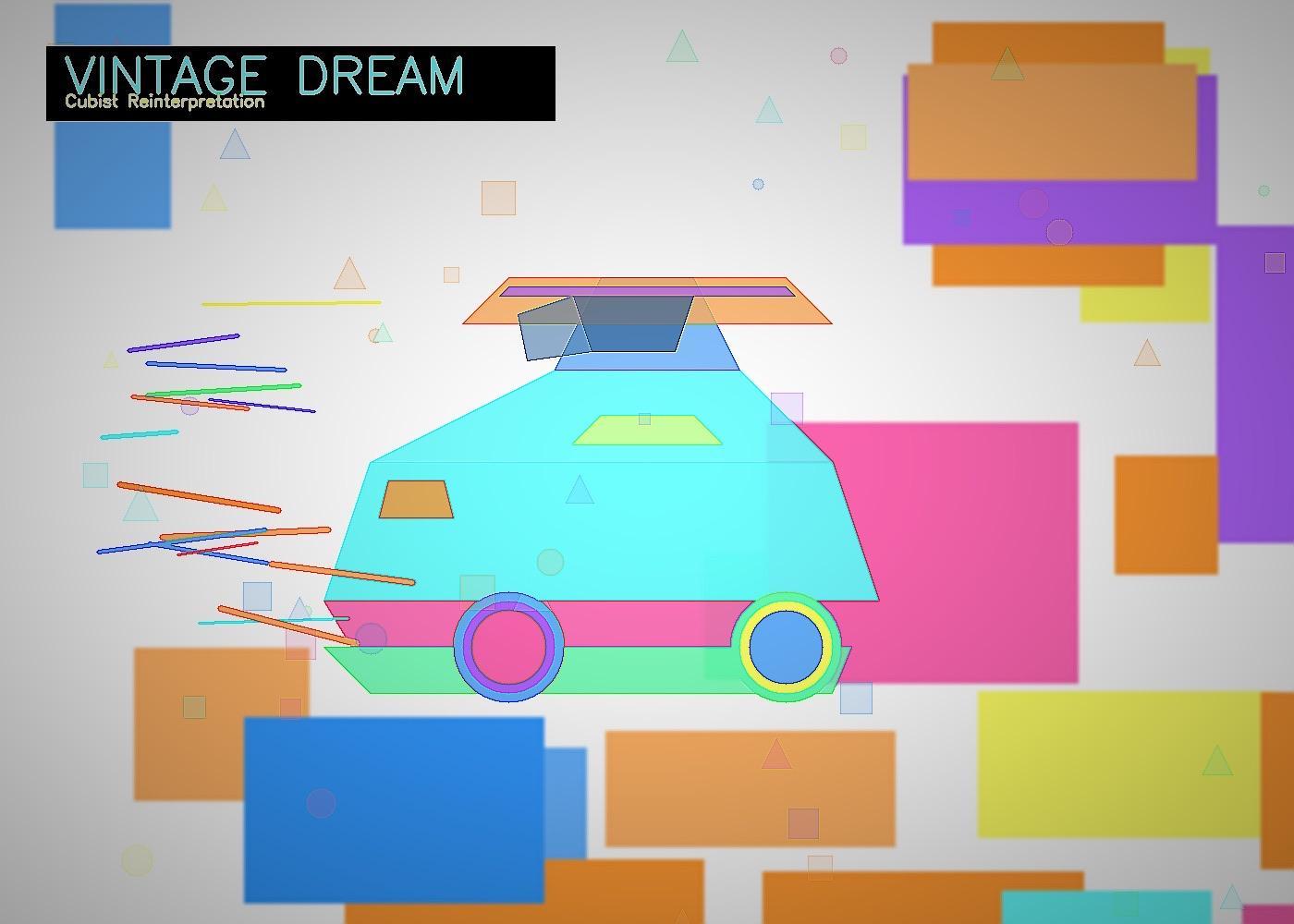}} \\
    \multicolumn{5}{l}{\cellcolor{subtlegray!20}\parbox{\textwidth}{\scriptsize\textit{\textcolor{questiongray}{11. Generate a new creative image inspired by this reference, but with a completely different artistic style and composition.}}}}\\
    \arrayrulecolor{gray!30}\midrule[0.3pt]\arrayrulecolor{black}

    \parbox{3cm}{\centering \includegraphics[width=3cm]{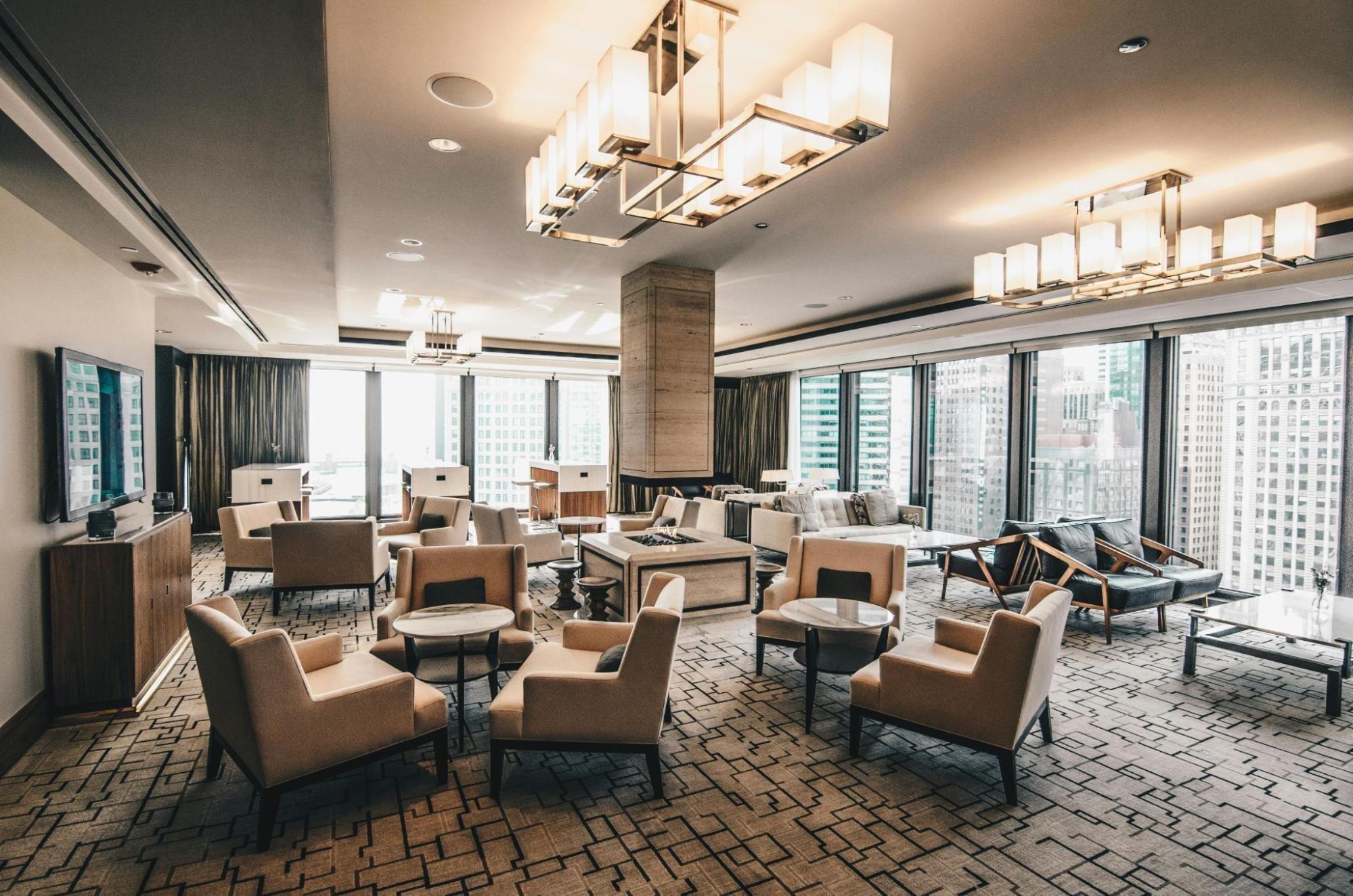}} & \parbox{3cm}{\centering \includegraphics[width=3cm]{3_results_sections/comparision_table/images/image115.jpg}} &      \parbox{3cm}{\centering \includegraphics[width=3cm]{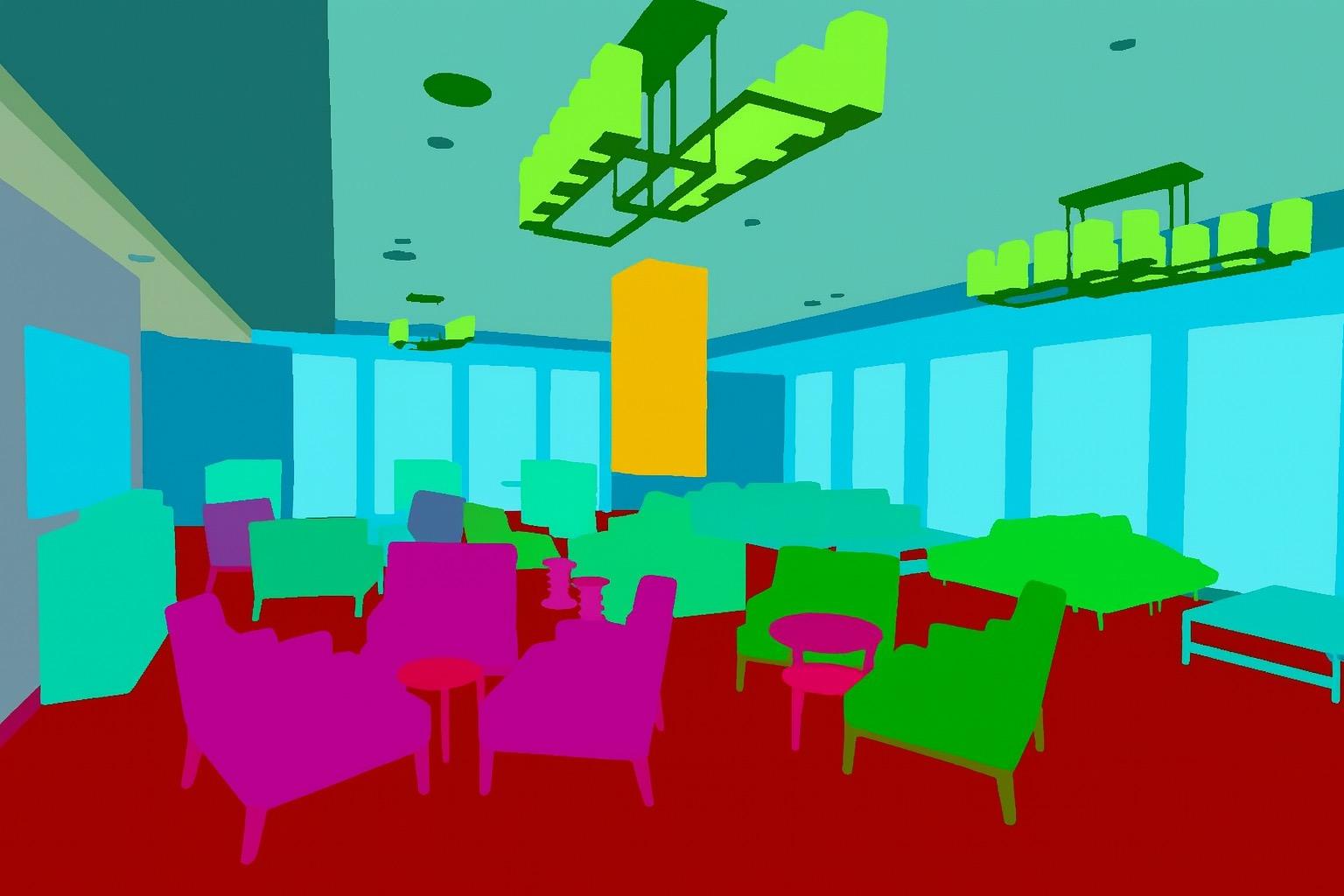}} &      \parbox{3cm}{\centering \includegraphics[width=3cm]{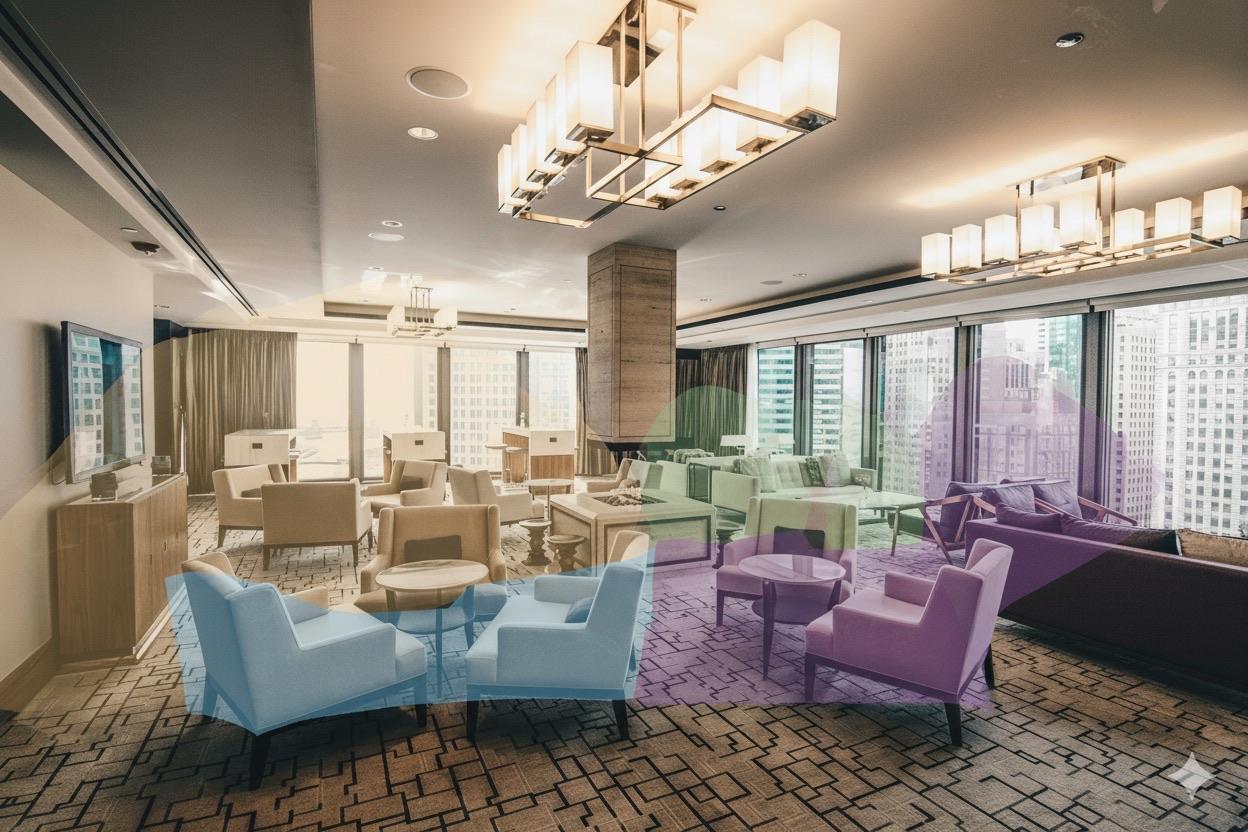}} &      \parbox{3cm}{\centering \includegraphics[width=3cm]{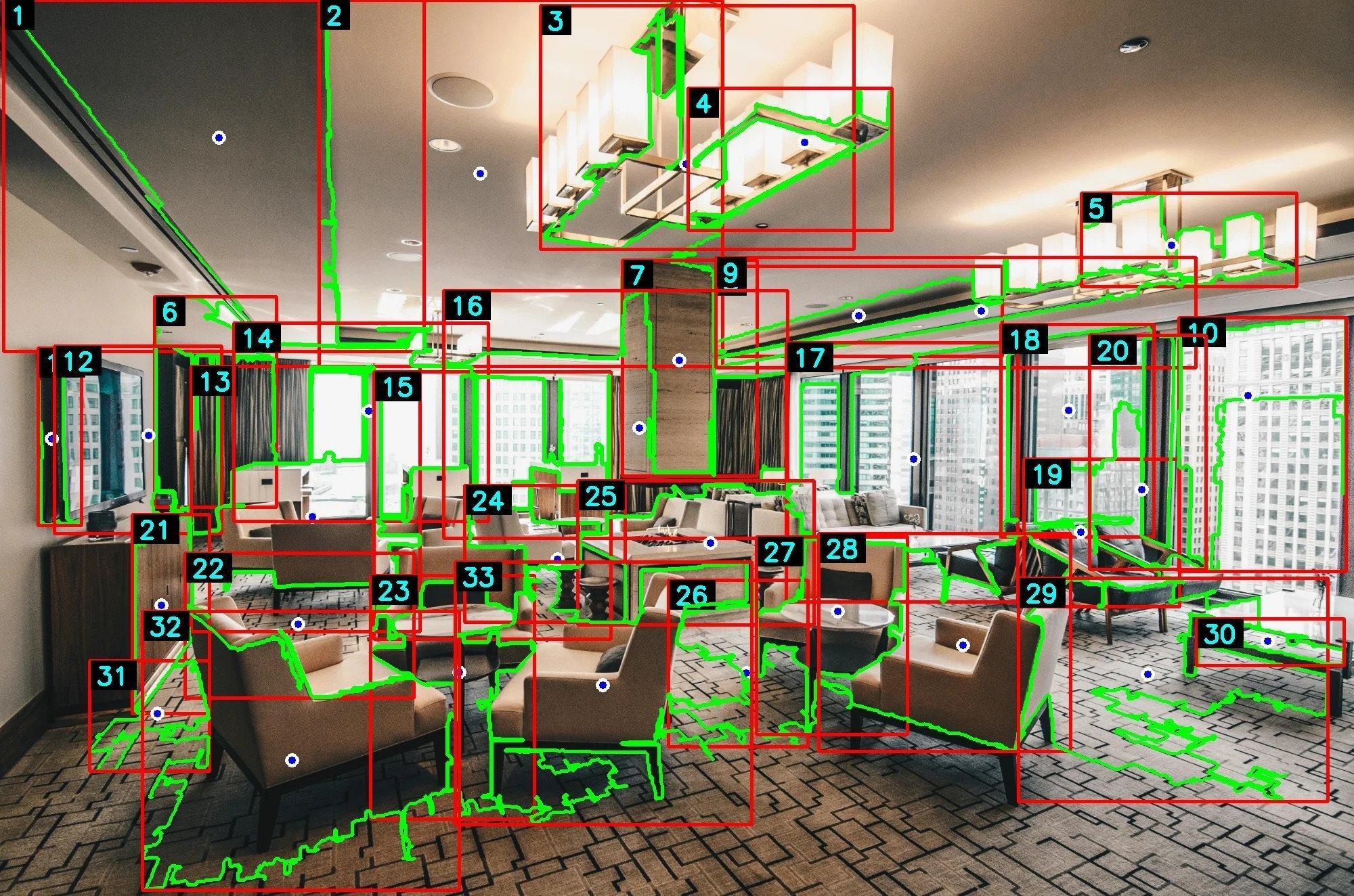}} \\
    \multicolumn{5}{l}{\cellcolor{subtlegray!20}\parbox{\textwidth}{\scriptsize\textit{\textcolor{questiongray}{12. Segment all the furniture in the living room.}}}}\\
    \arrayrulecolor{gray!30}\midrule[0.3pt]\arrayrulecolor{black}

    \parbox{3cm}{\centering \includegraphics[width=3cm]{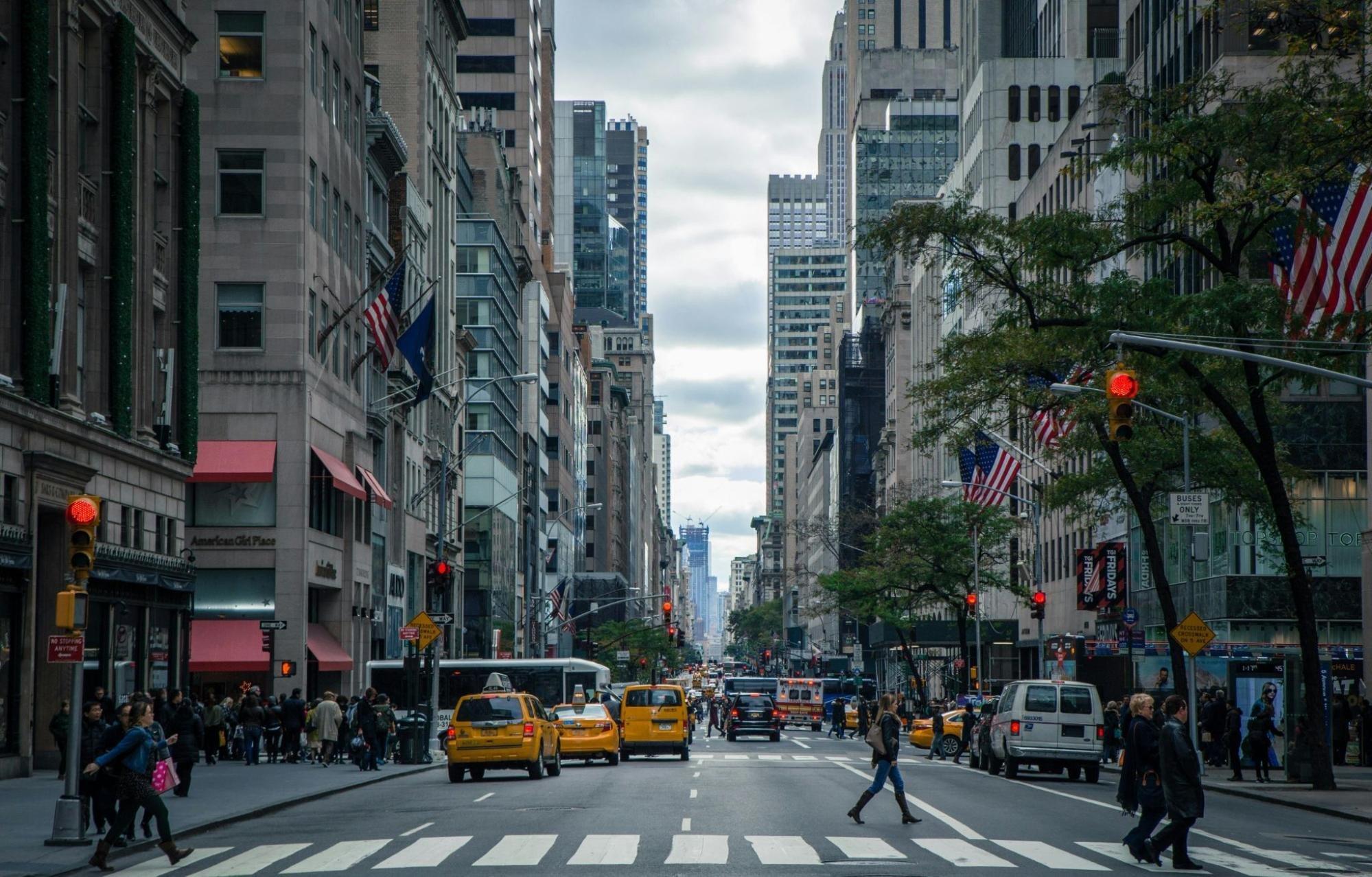}} & \parbox{3cm}{\centering \includegraphics[width=3cm]{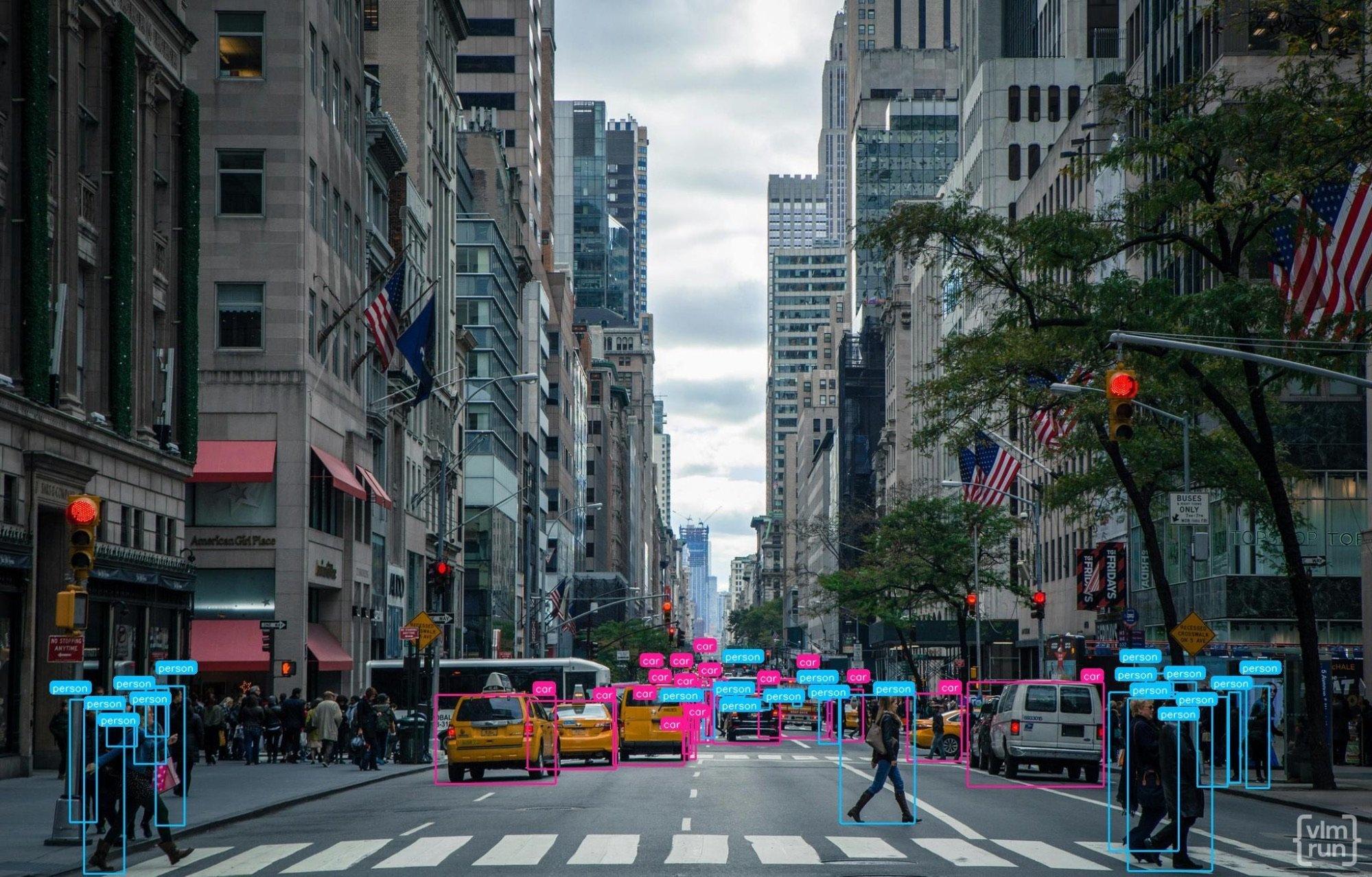}} &      \parbox{3cm}{\centering \includegraphics[width=3cm]{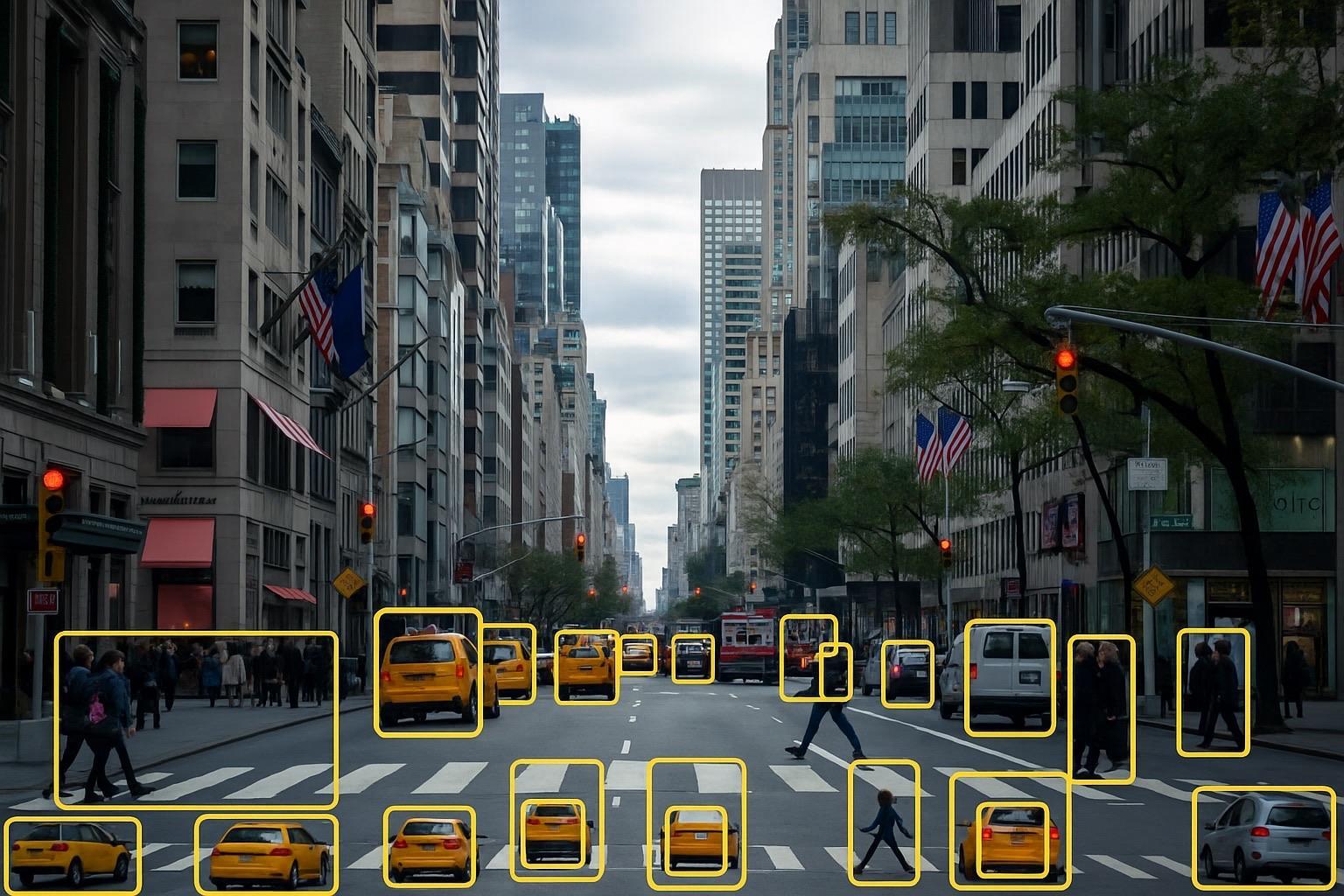}} &      \parbox{3cm}{\centering \includegraphics[width=3cm]{3_results_sections/comparision_table/images/image21.jpg}} &      \parbox{3cm}{\centering \includegraphics[width=3cm]{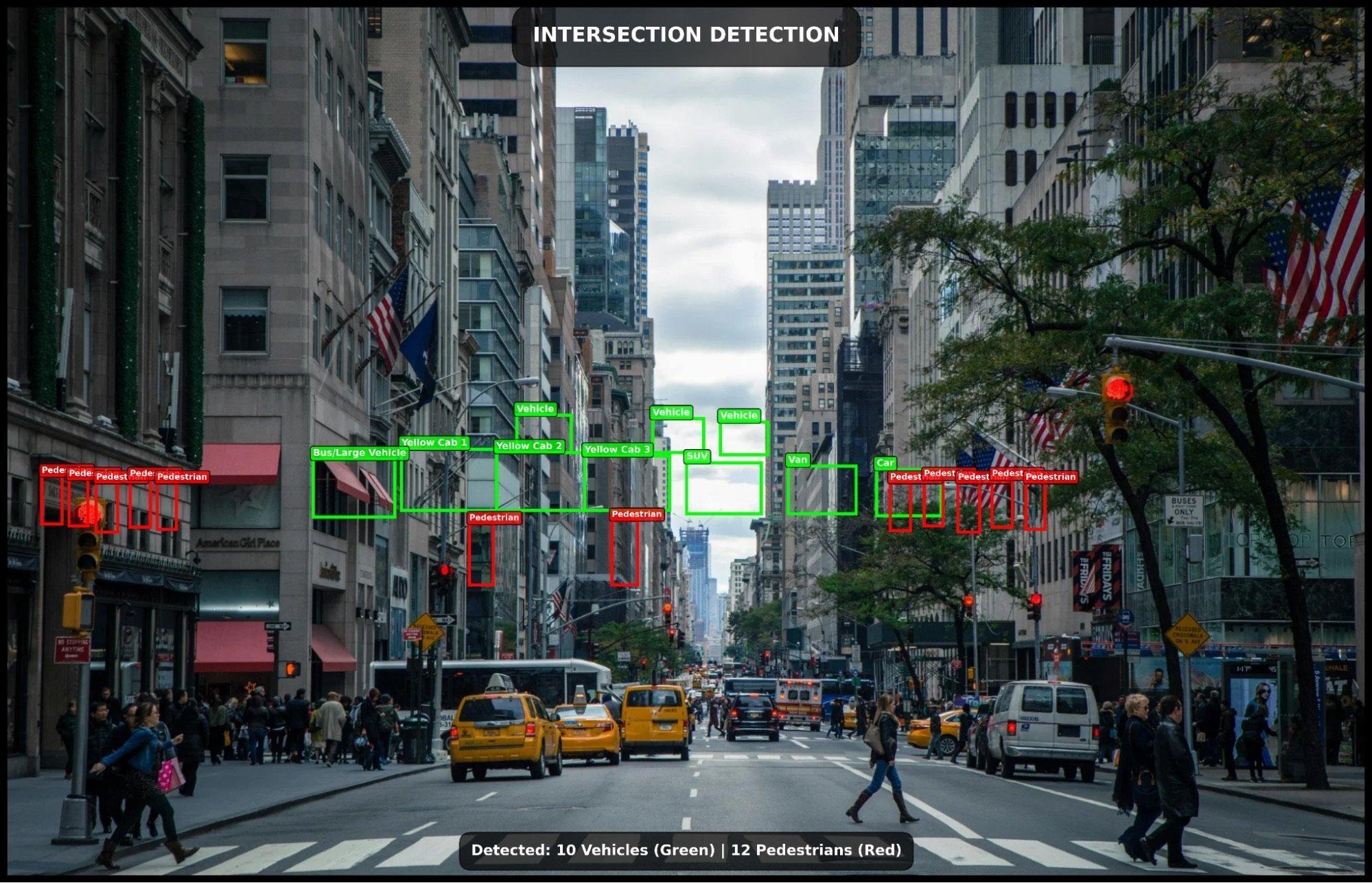}} \\
    \multicolumn{5}{l}{\cellcolor{subtlegray!20}\parbox{\textwidth}{\scriptsize\textit{\textcolor{questiongray}{13. Detect all the cars and pedestrians at the intersection.}}}}\\
    \arrayrulecolor{gray!30}\midrule[0.3pt]\arrayrulecolor{black}

    \parbox{3cm}{\centering \includegraphics[width=3cm]{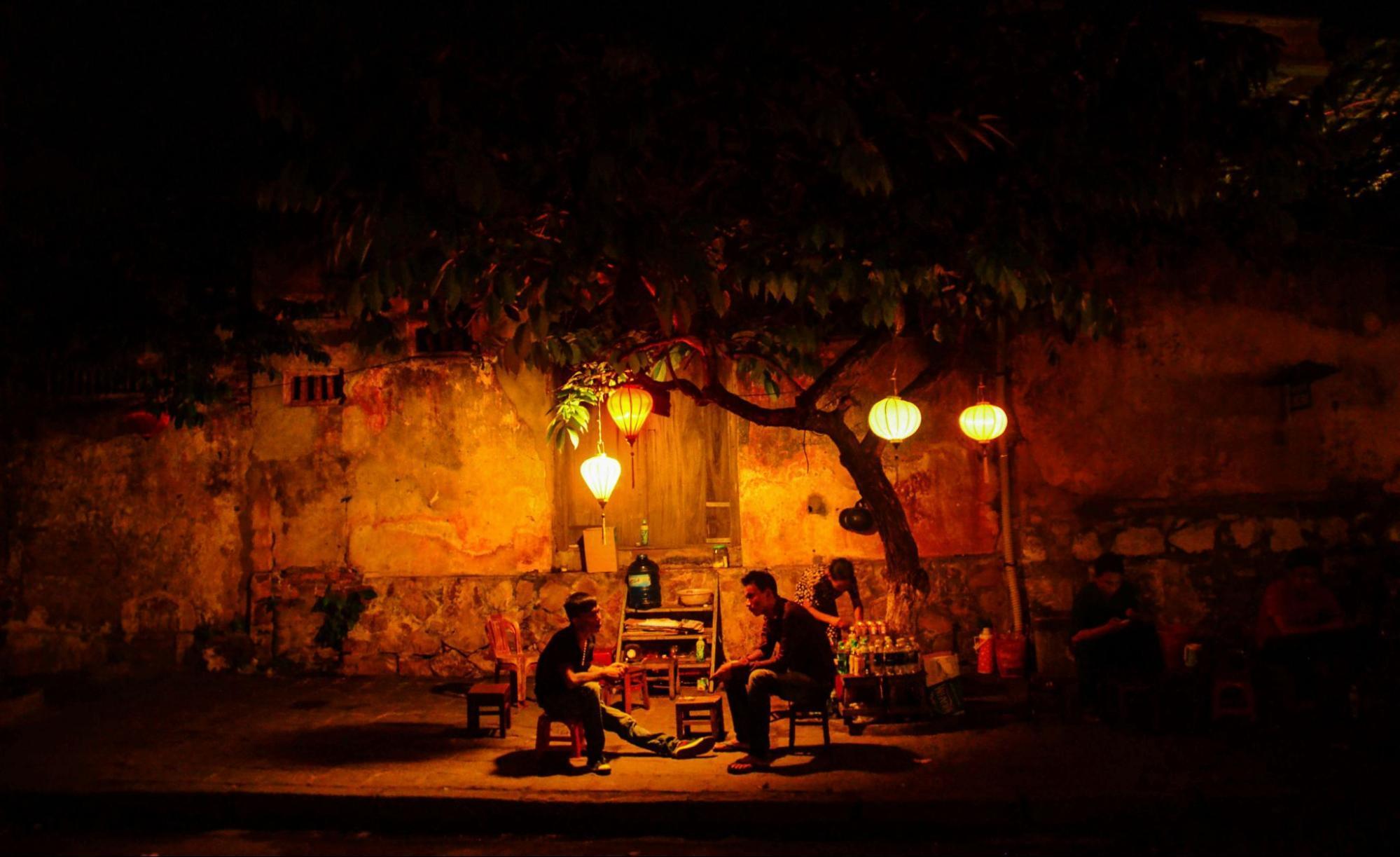}} & \parbox{3cm}{\centering \includegraphics[width=3cm]{3_results_sections/comparision_table/images/image25.jpg}} &      \parbox{3cm}{\centering \includegraphics[width=3cm]{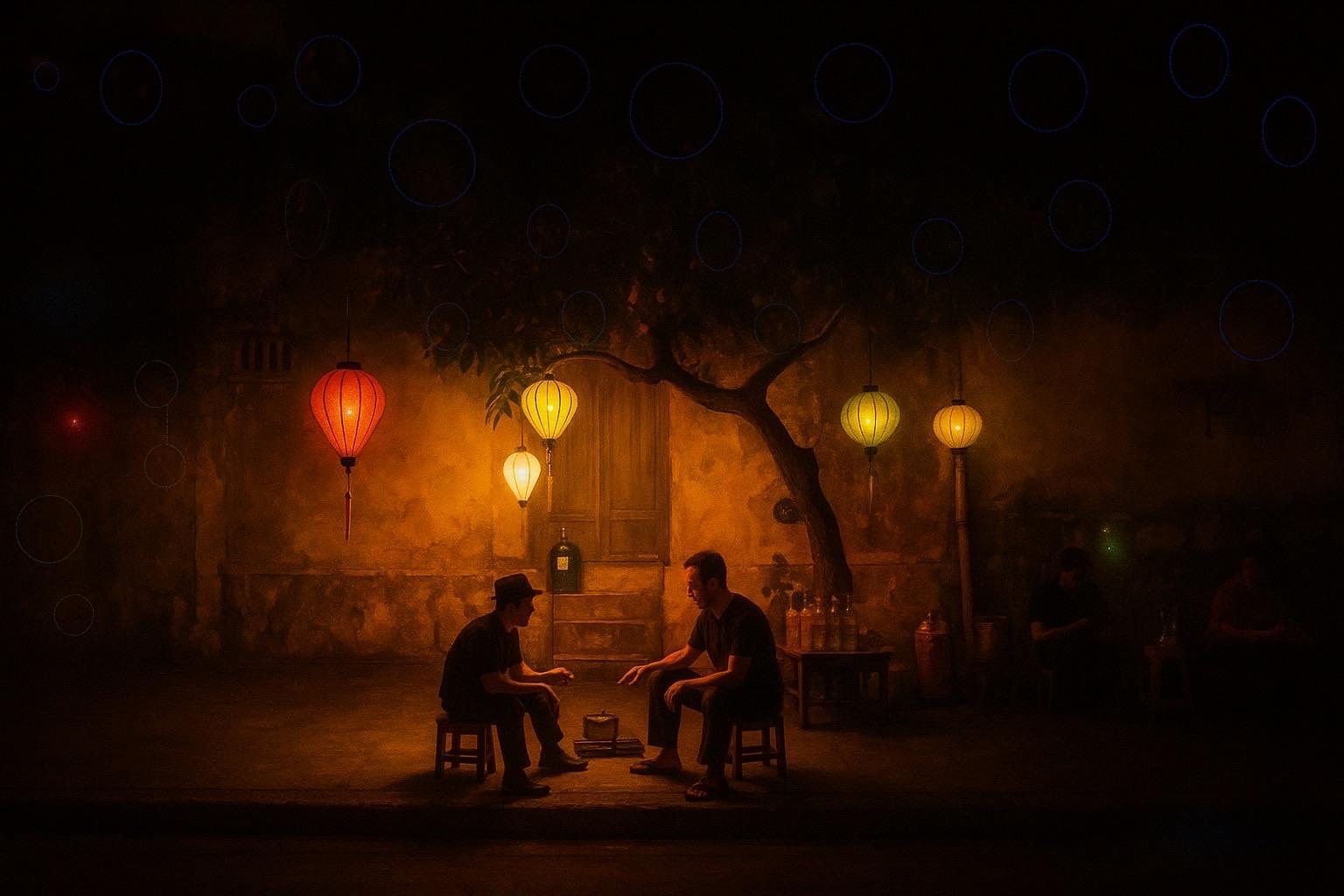}} &      \parbox{3cm}{\centering \includegraphics[width=3cm]{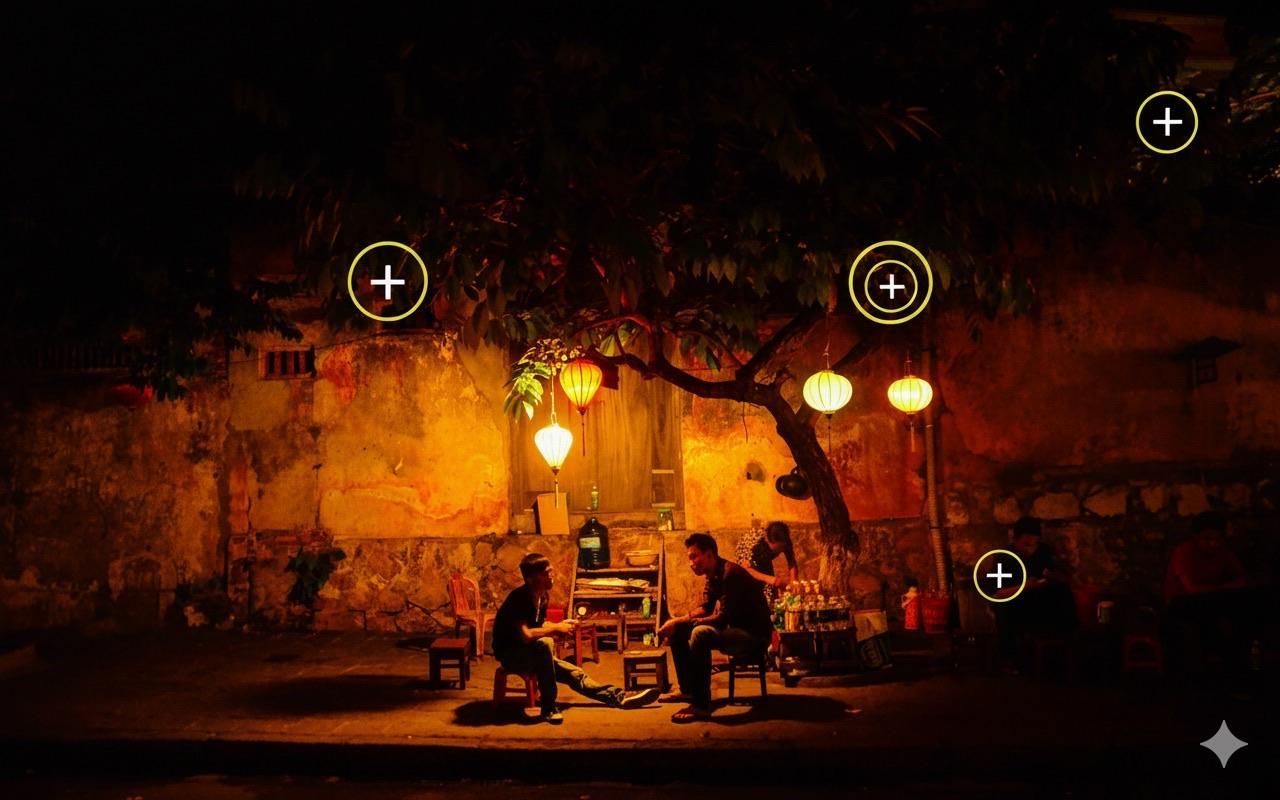}} &      \parbox{3cm}{\centering \includegraphics[width=3cm]{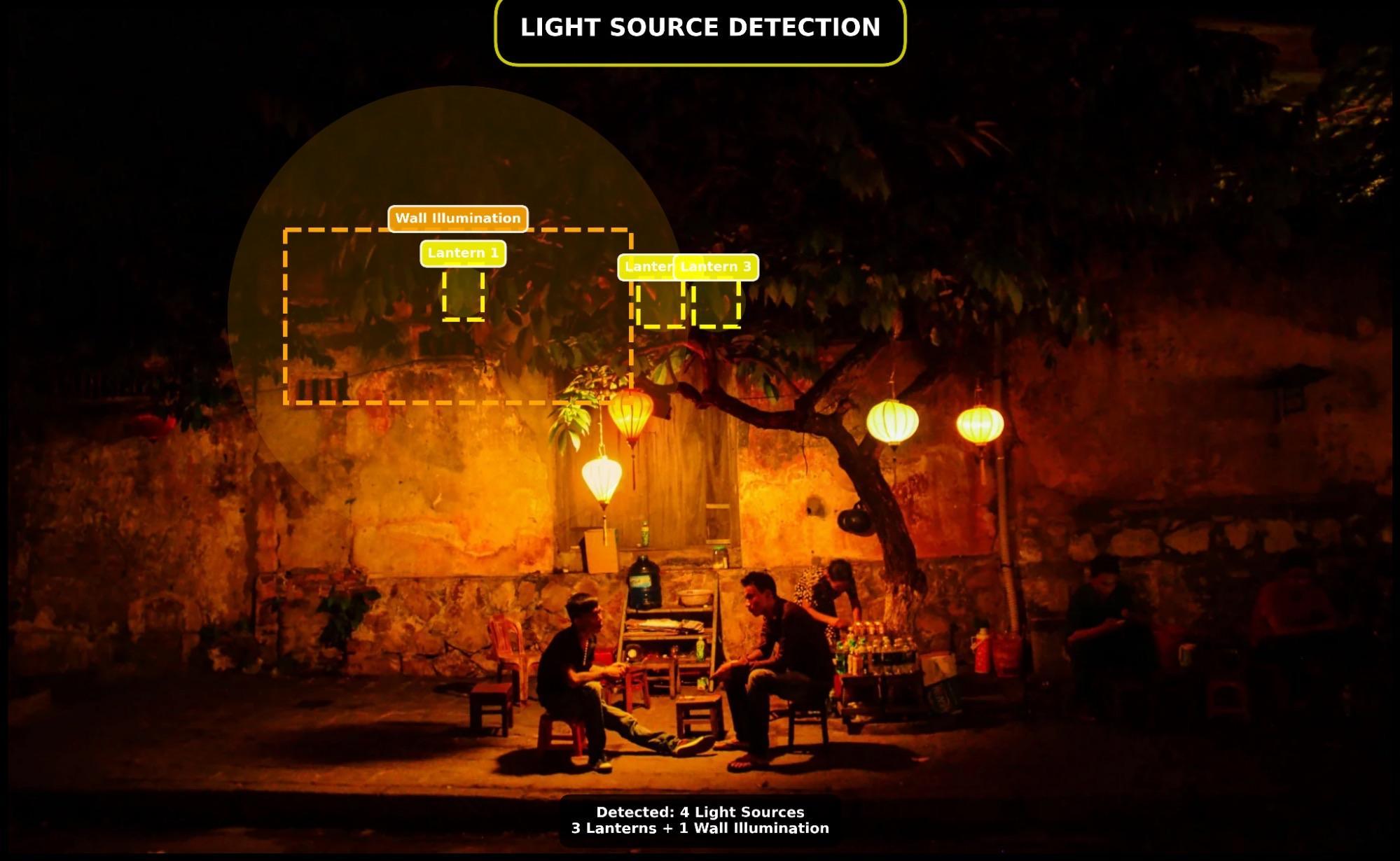}} \\
    \multicolumn{5}{l}{\cellcolor{subtlegray!20}\parbox{\textwidth}{\scriptsize\textit{\textcolor{questiongray}{14. Detect the light sources in the image.}}}}\\
    \arrayrulecolor{gray!30}\midrule[0.3pt]\arrayrulecolor{black}

    \parbox{3cm}{\centering \includegraphics[width=3cm]{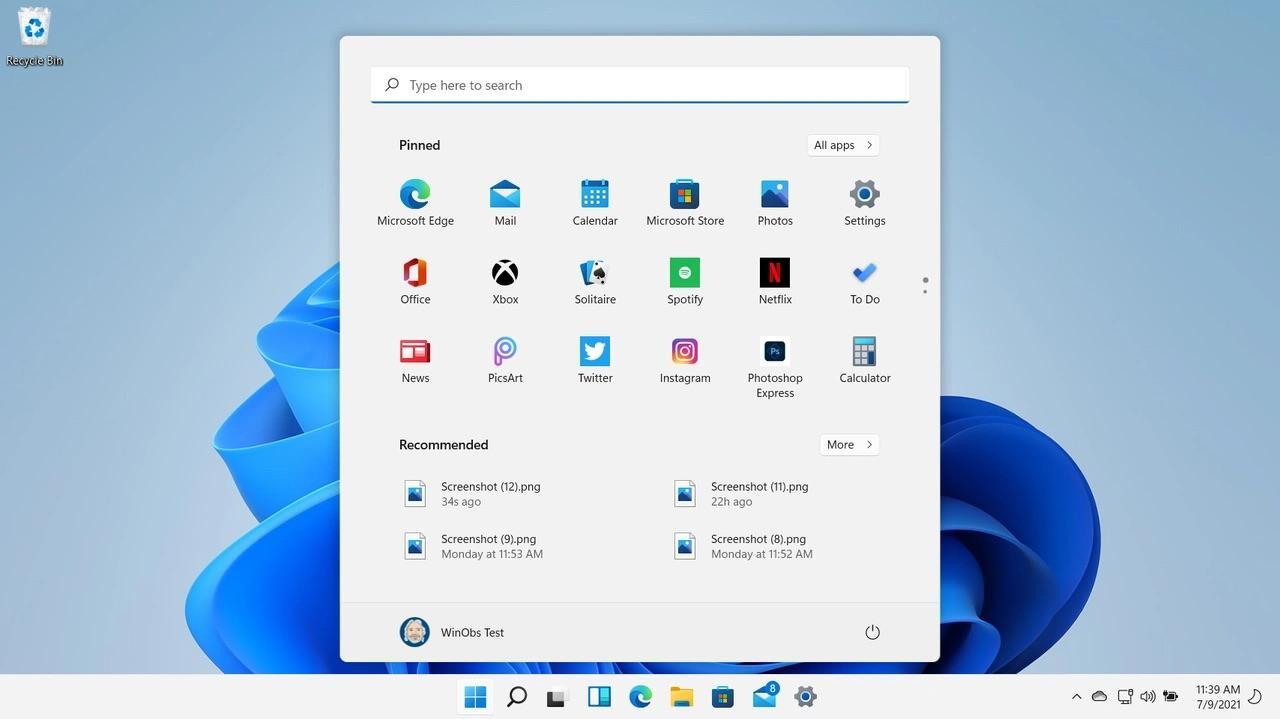}} & \parbox{3cm}{\centering \includegraphics[width=3cm]{3_results_sections/comparision_table/images/image132.jpg}} &      \parbox{3cm}{\centering \includegraphics[width=3cm]{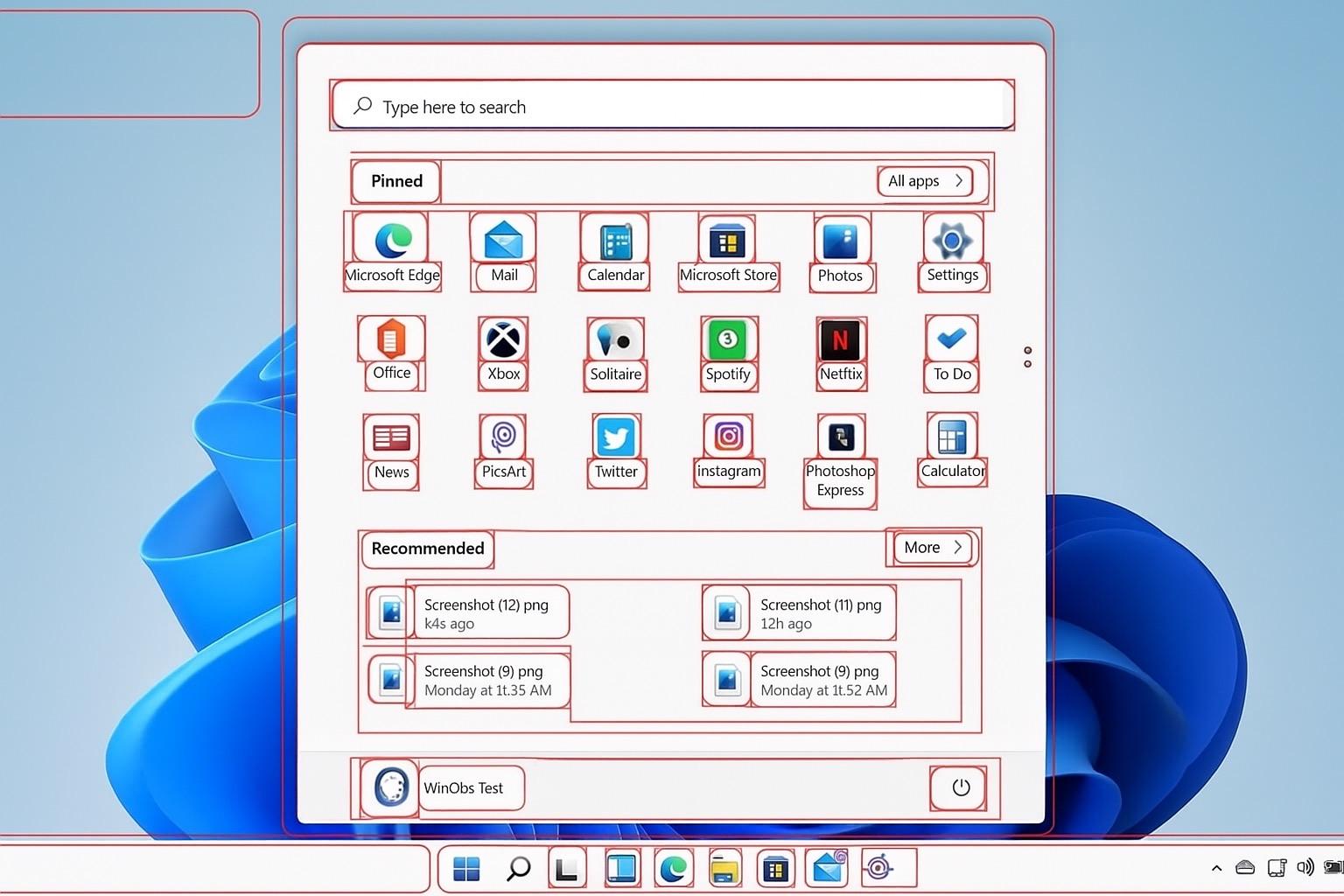}} &      \parbox{3cm}{\centering \includegraphics[width=3cm]{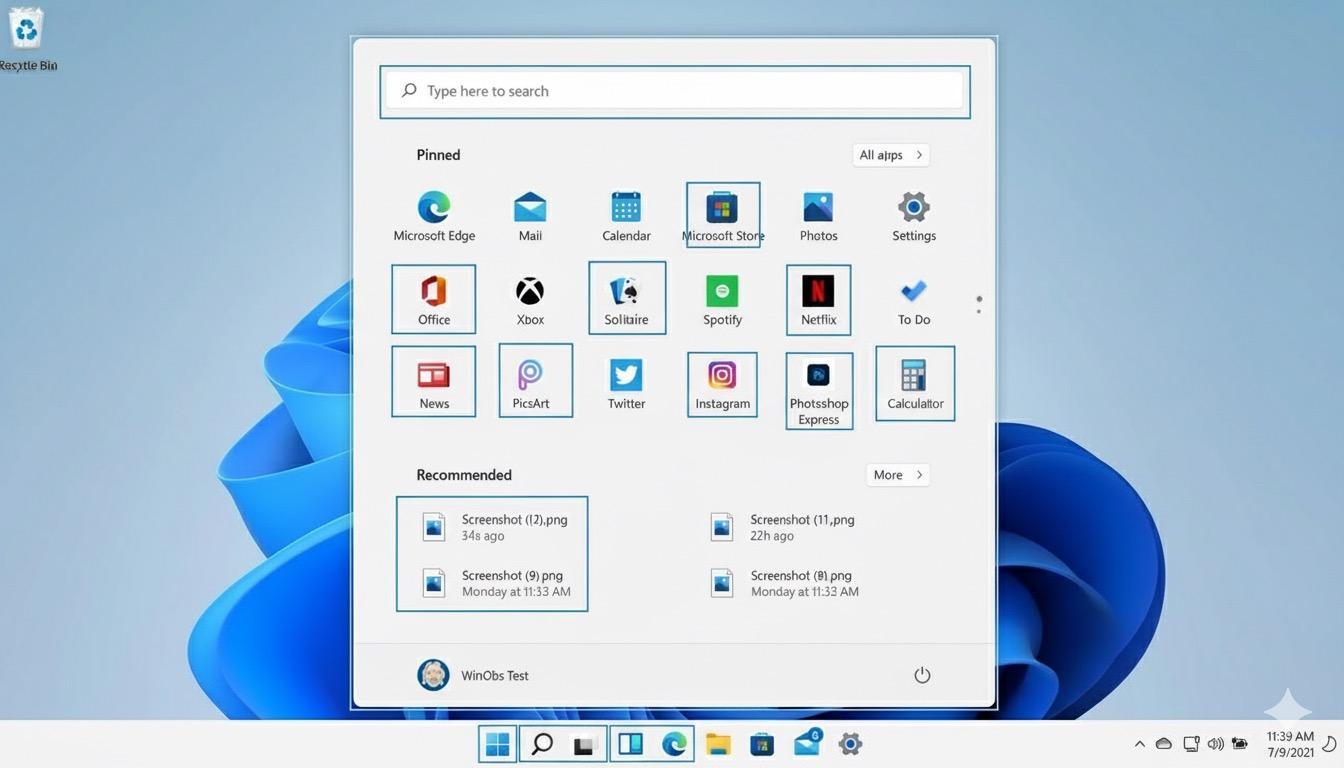}} &      \parbox{3cm}{\centering \includegraphics[width=3cm]{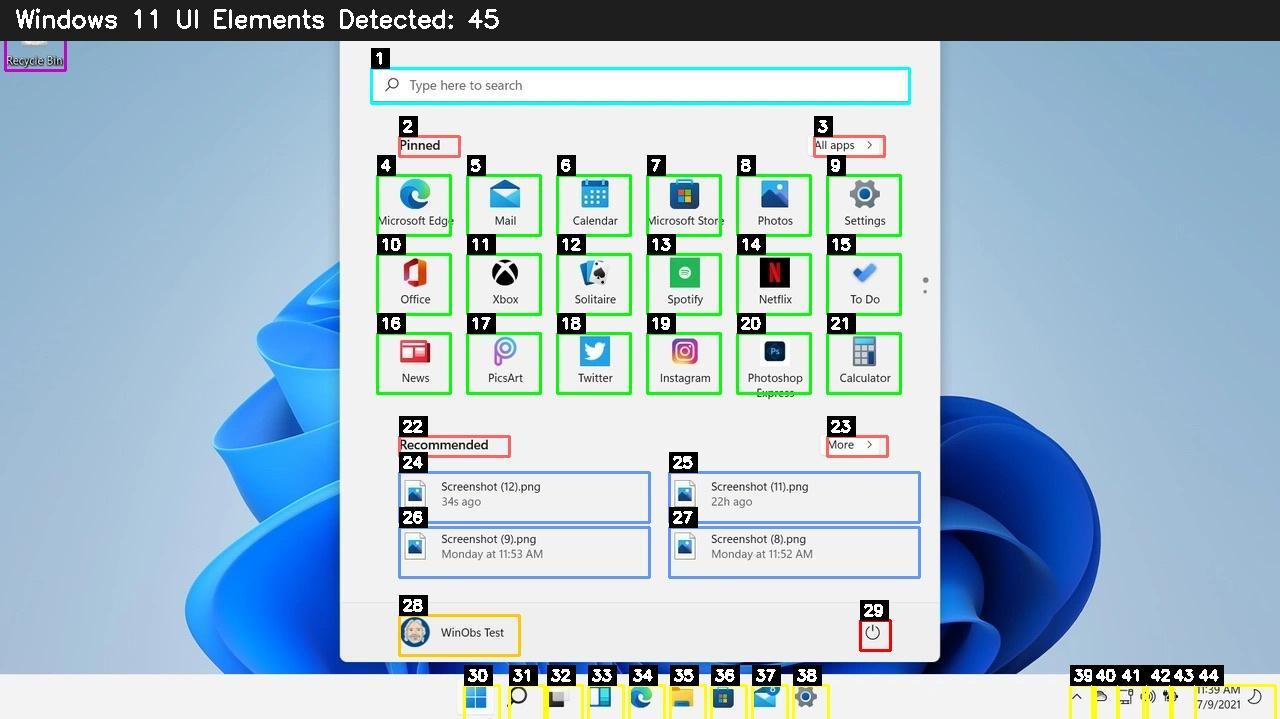}} \\
    \multicolumn{5}{l}{\cellcolor{subtlegray!20}\parbox{\textwidth}{\scriptsize\textit{\textcolor{questiongray}{15. Detect all the UI elements in this image and visualize the boxes.}}}}\\
    \arrayrulecolor{gray!30}\midrule[0.3pt]\arrayrulecolor{black}

    \parbox{3cm}{\centering \includegraphics[width=3cm]{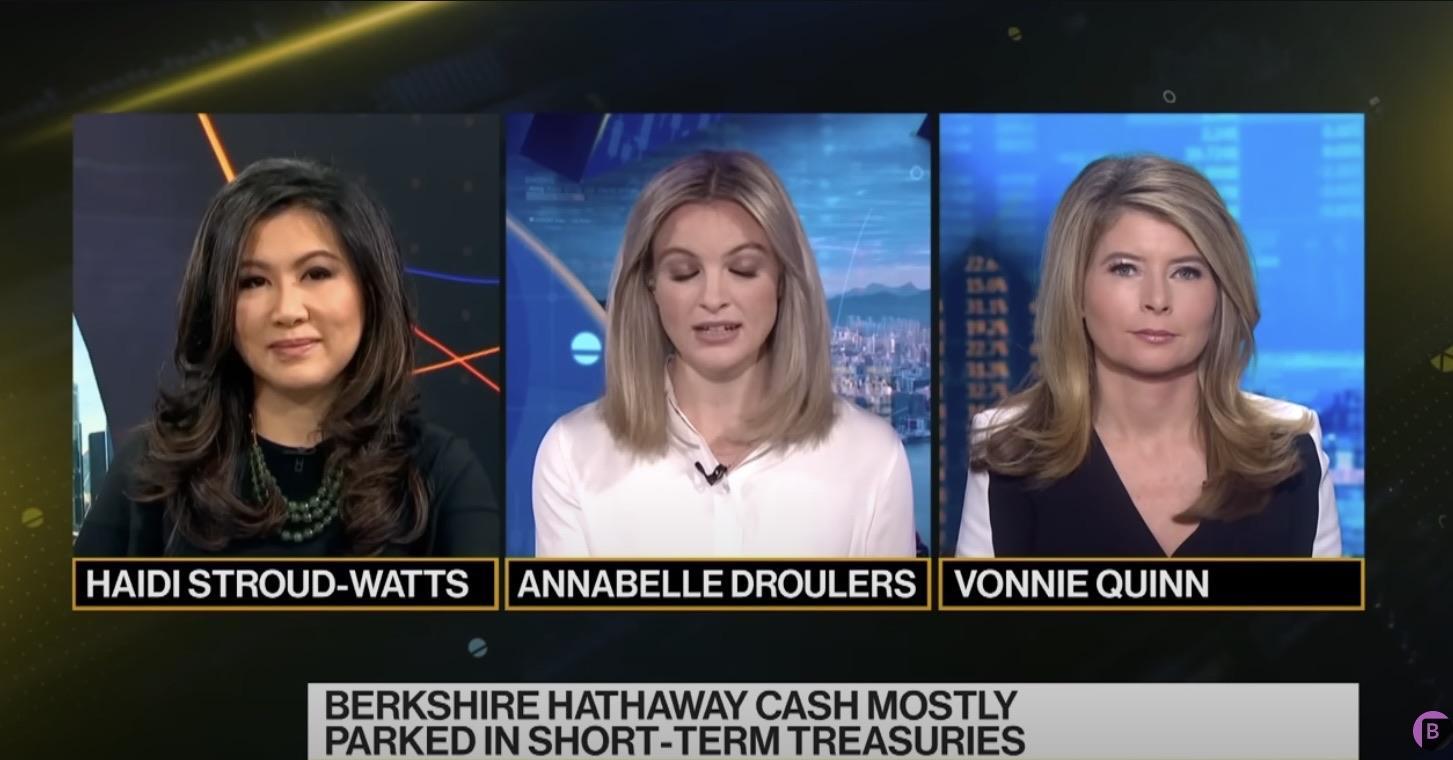}} & \parbox{3cm}{\centering \includegraphics[width=3cm]{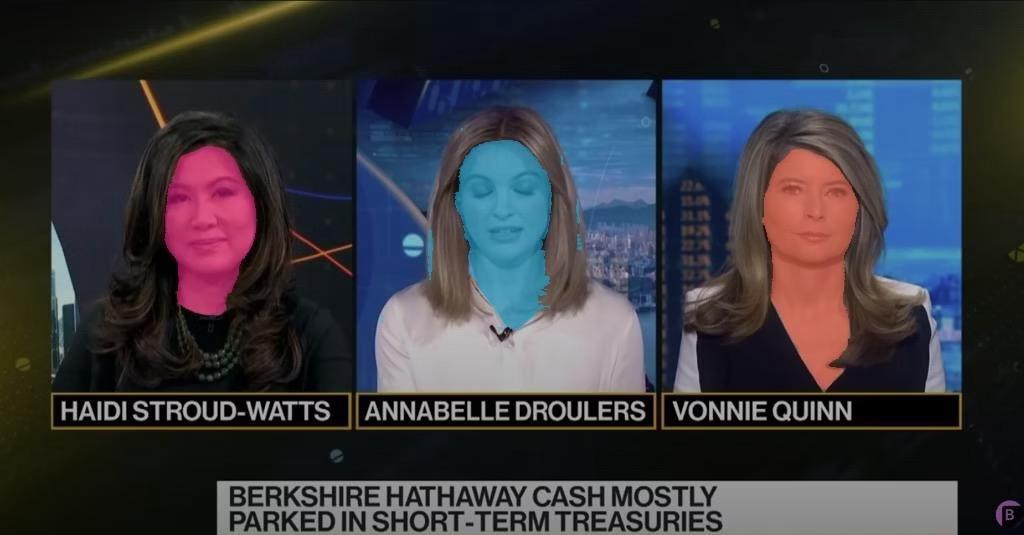}} &      \parbox{3cm}{\centering \includegraphics[width=3cm]{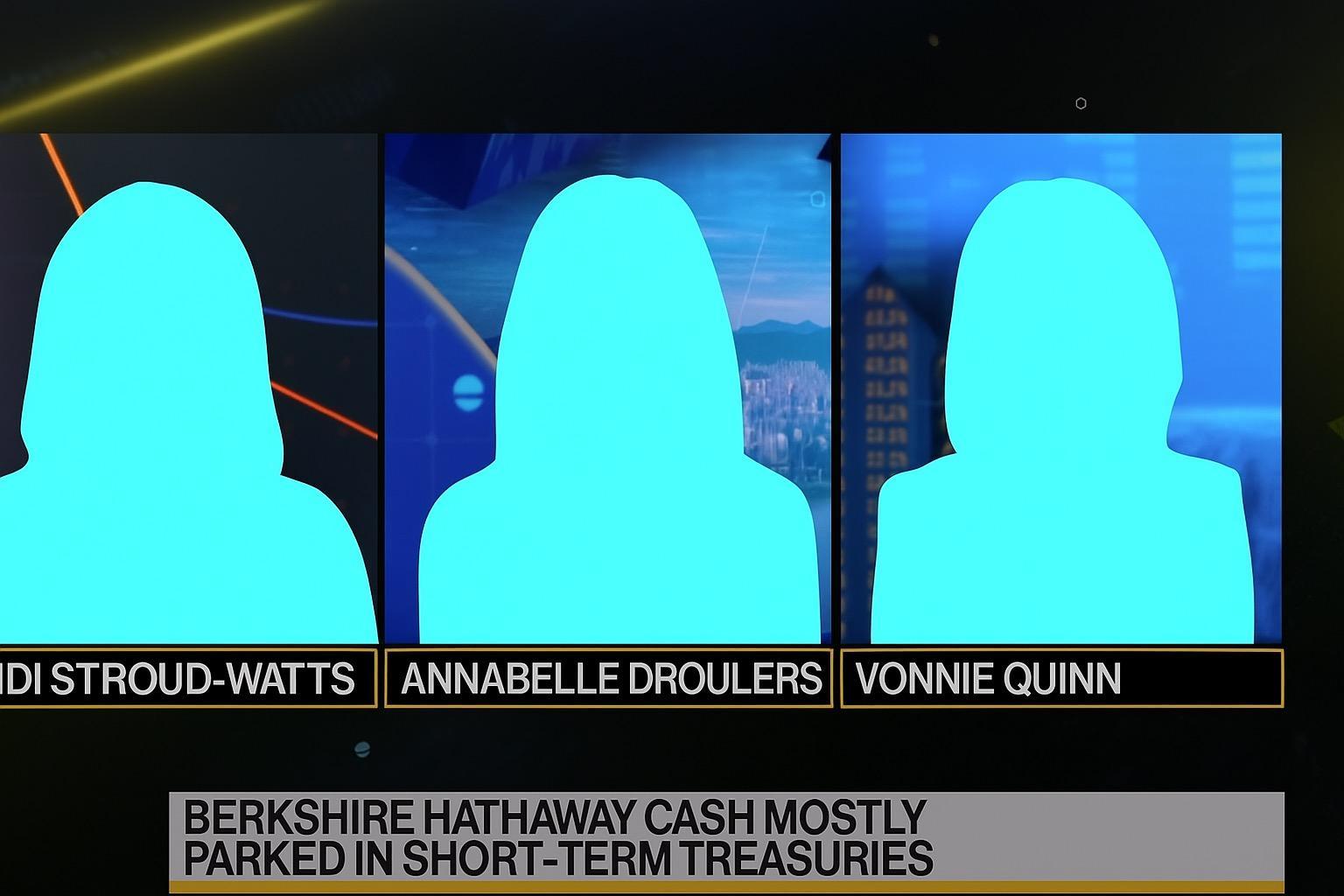}} &      \parbox{3cm}{\centering \includegraphics[width=3cm]{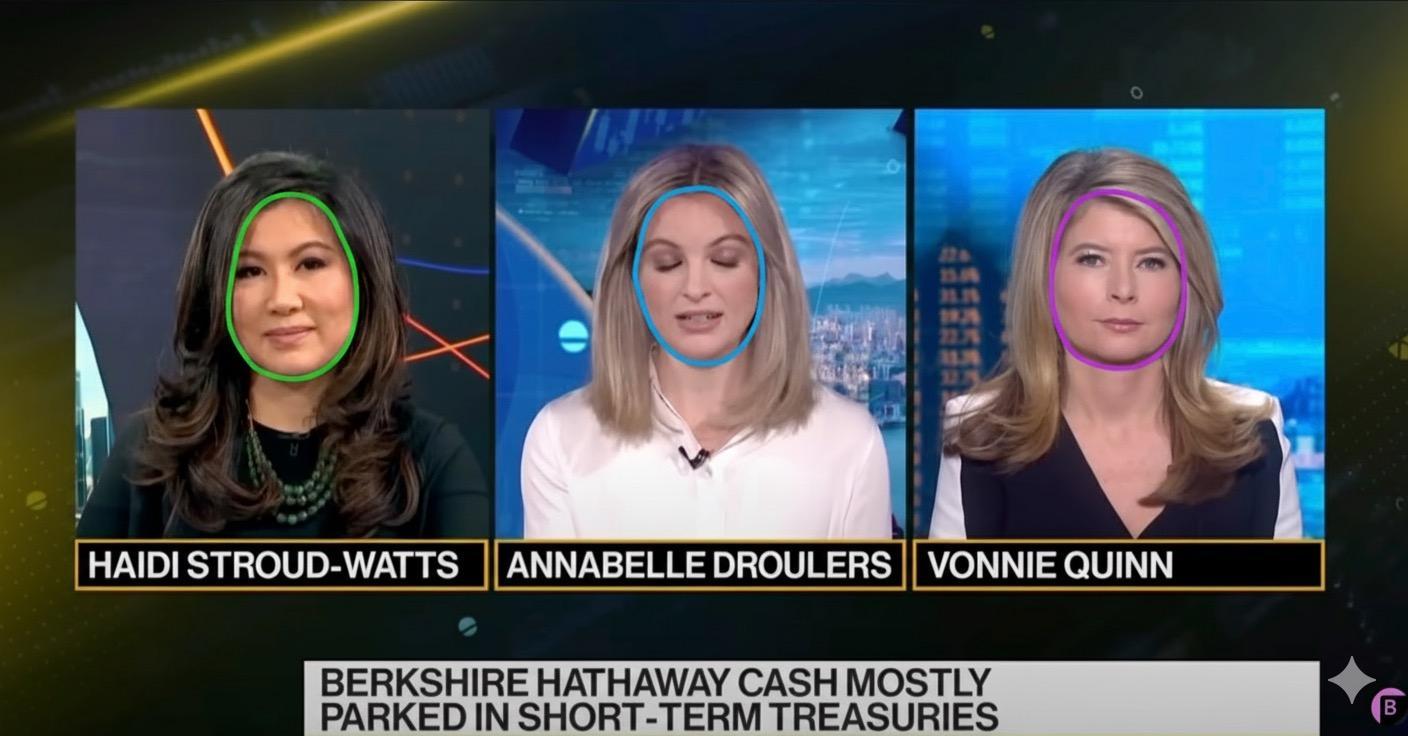}} &      \parbox{3cm}{\centering \includegraphics[width=3cm]{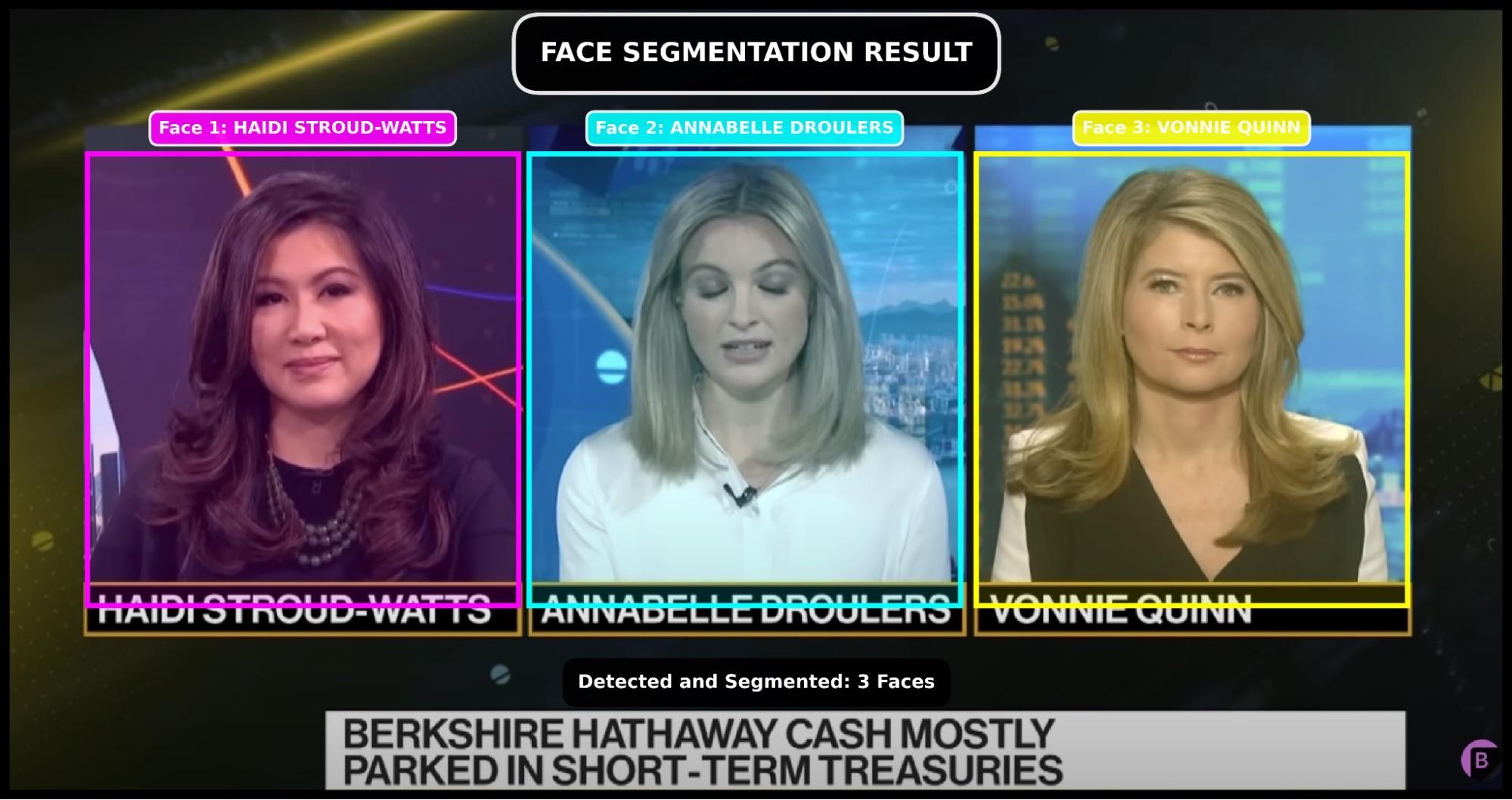}} \\
    \multicolumn{5}{l}{\cellcolor{subtlegray!20}\parbox{\textwidth}{\scriptsize\textit{\textcolor{questiongray}{16. Segment out the faces in this image and visulize the segments.}}}}\\
    \arrayrulecolor{gray!30}\midrule[0.3pt]\arrayrulecolor{black}

    \parbox{3cm}{\centering \includegraphics[width=3cm]{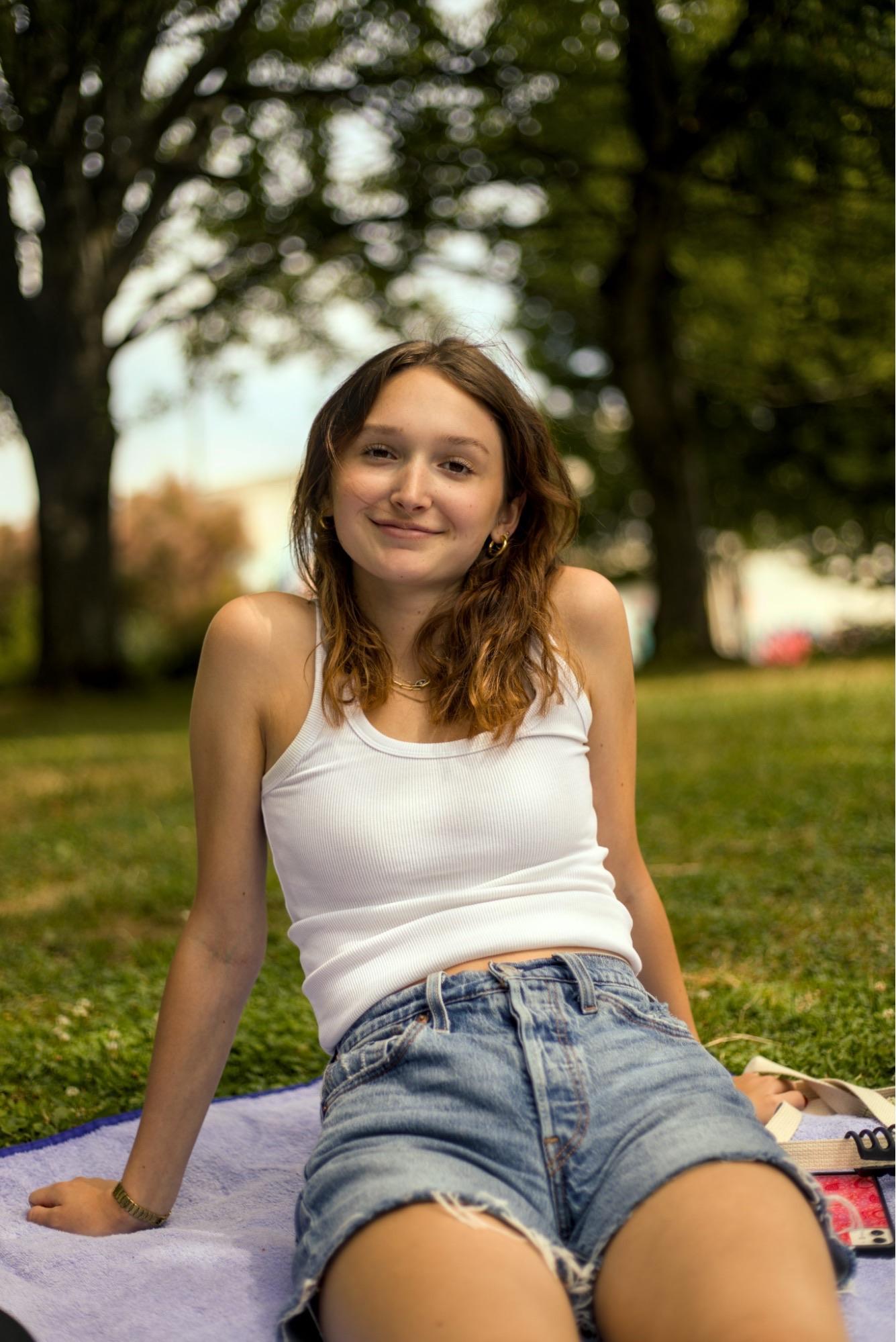}} & \parbox{3cm}{\centering \includegraphics[width=3cm]{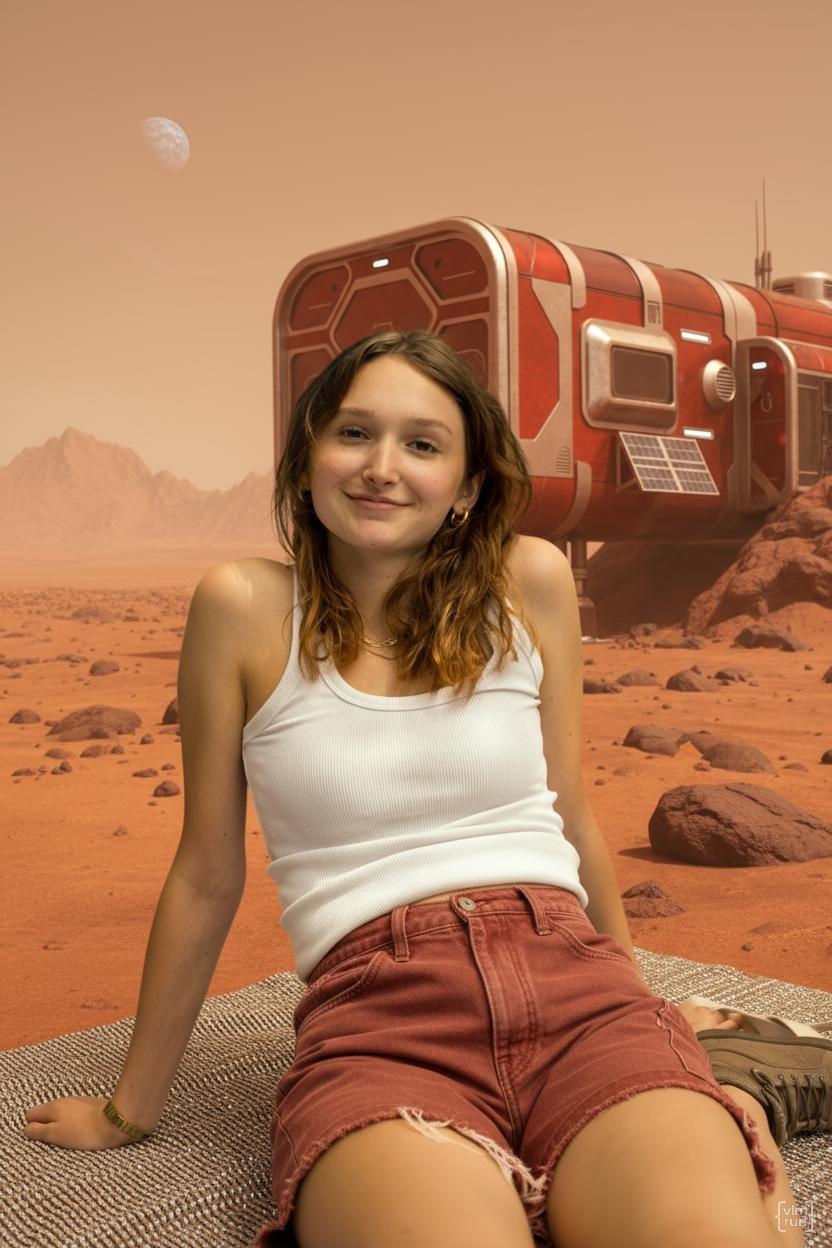}} &      \parbox{3cm}{\centering \includegraphics[width=3cm]{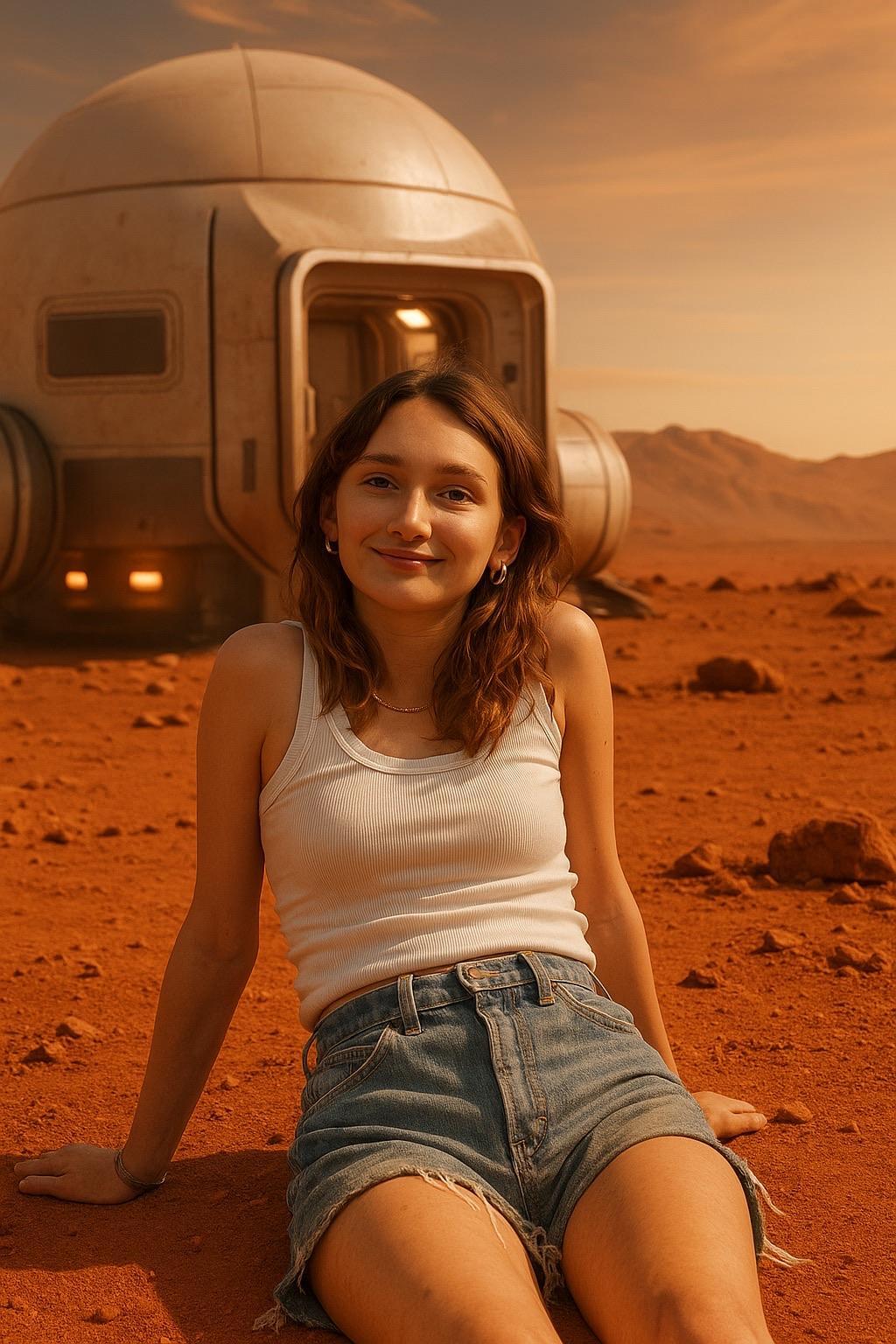}} &      \parbox{3cm}{\centering \includegraphics[width=3cm]{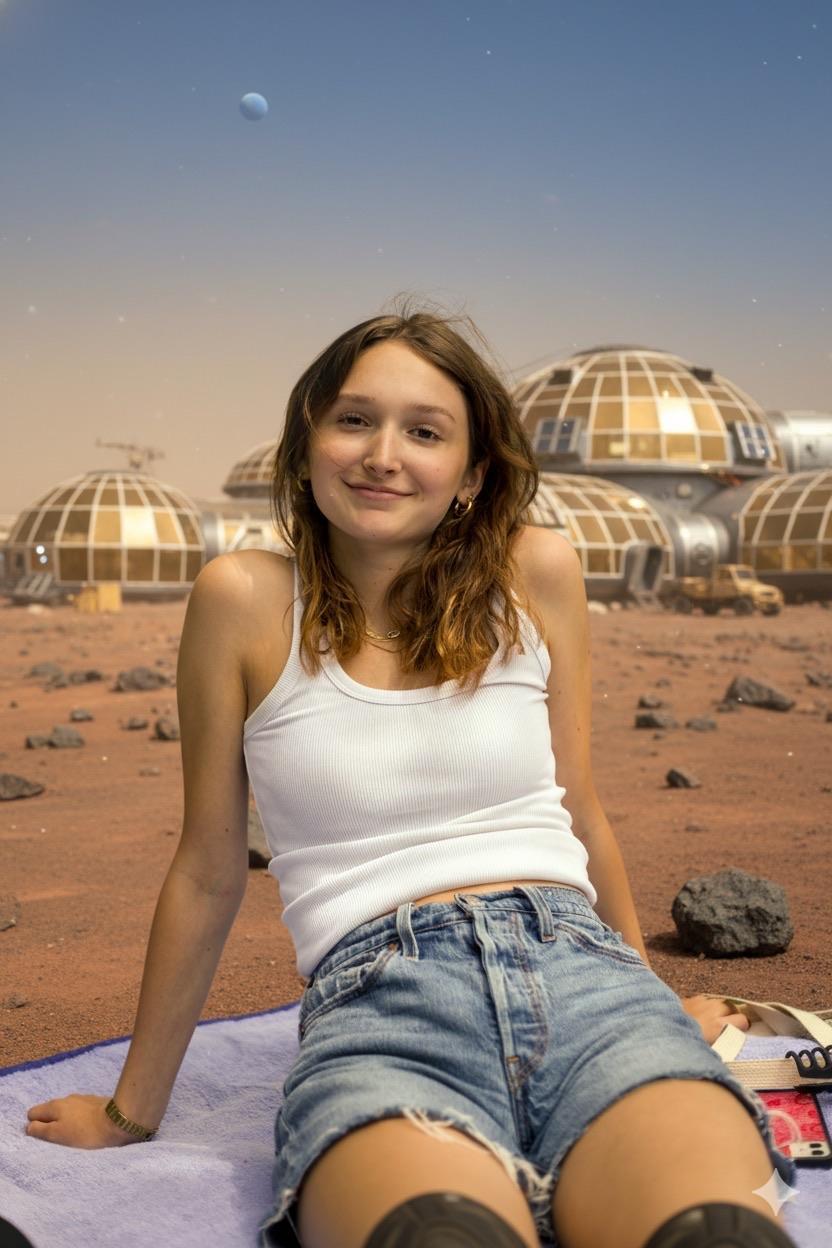}} &      \parbox{3cm}{\centering \includegraphics[width=3cm]{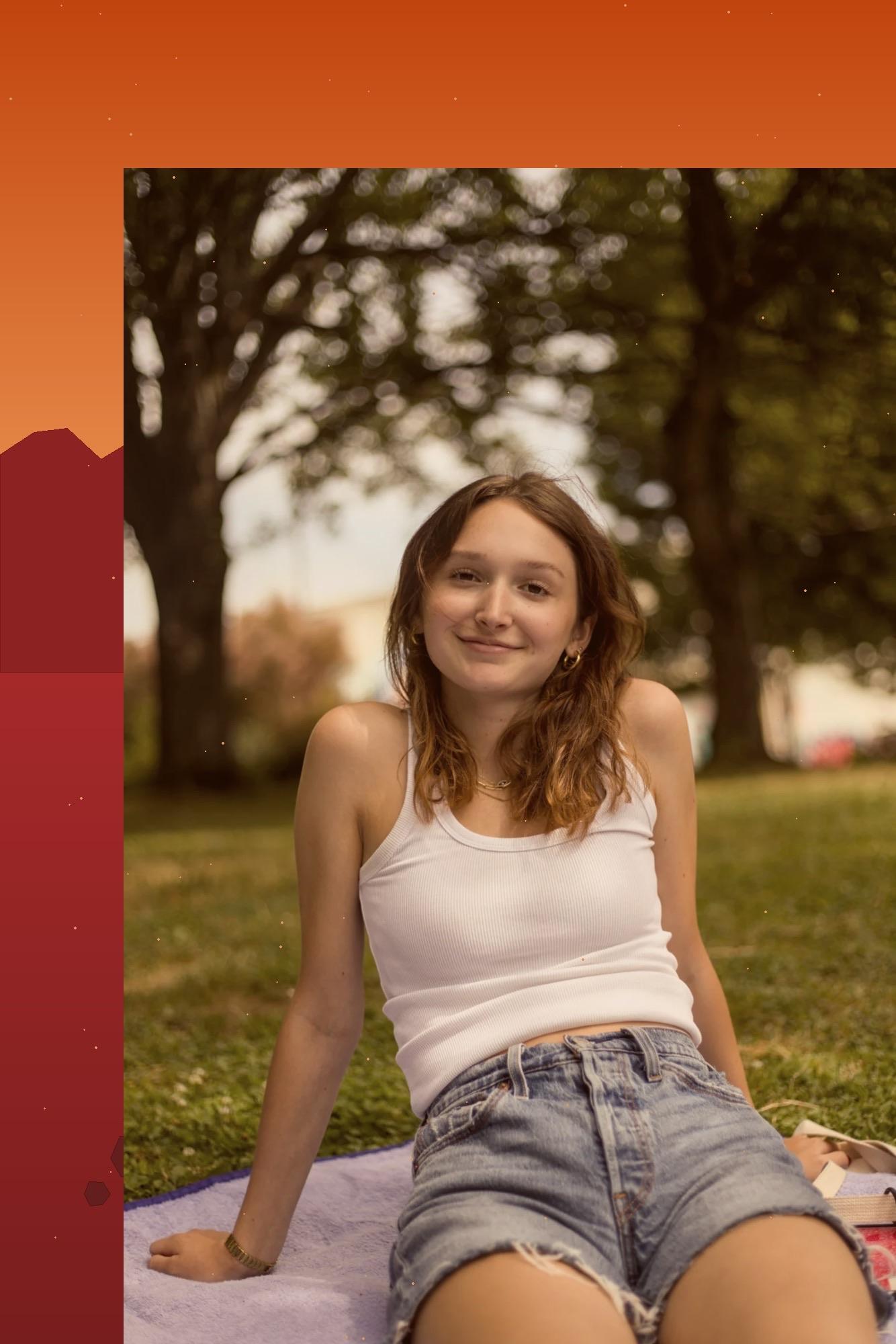}} \\
    \multicolumn{5}{l}{\cellcolor{subtlegray!20}\parbox{\textwidth}{\scriptsize\textit{\textcolor{questiongray}{17. Generate an image of this girl on the mars. I want her in front of a futuristic settlement house on mars as if she has been there for her whole lif enjoying the day. I want her outfit and pose to blend into the surrounding and look natural.}}}}\\
    \arrayrulecolor{gray!30}\midrule[0.3pt]\arrayrulecolor{black}

    \parbox{3cm}{\centering \includegraphics[width=3cm]{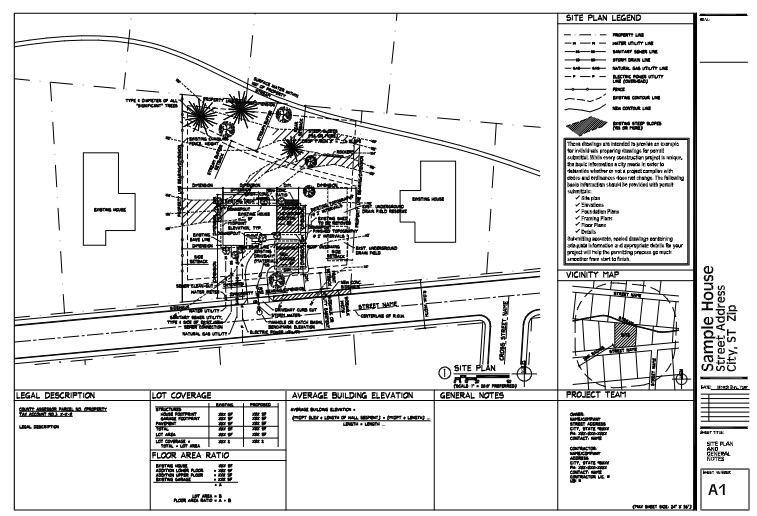}} & \parbox{3cm}{\centering \includegraphics[width=3cm]{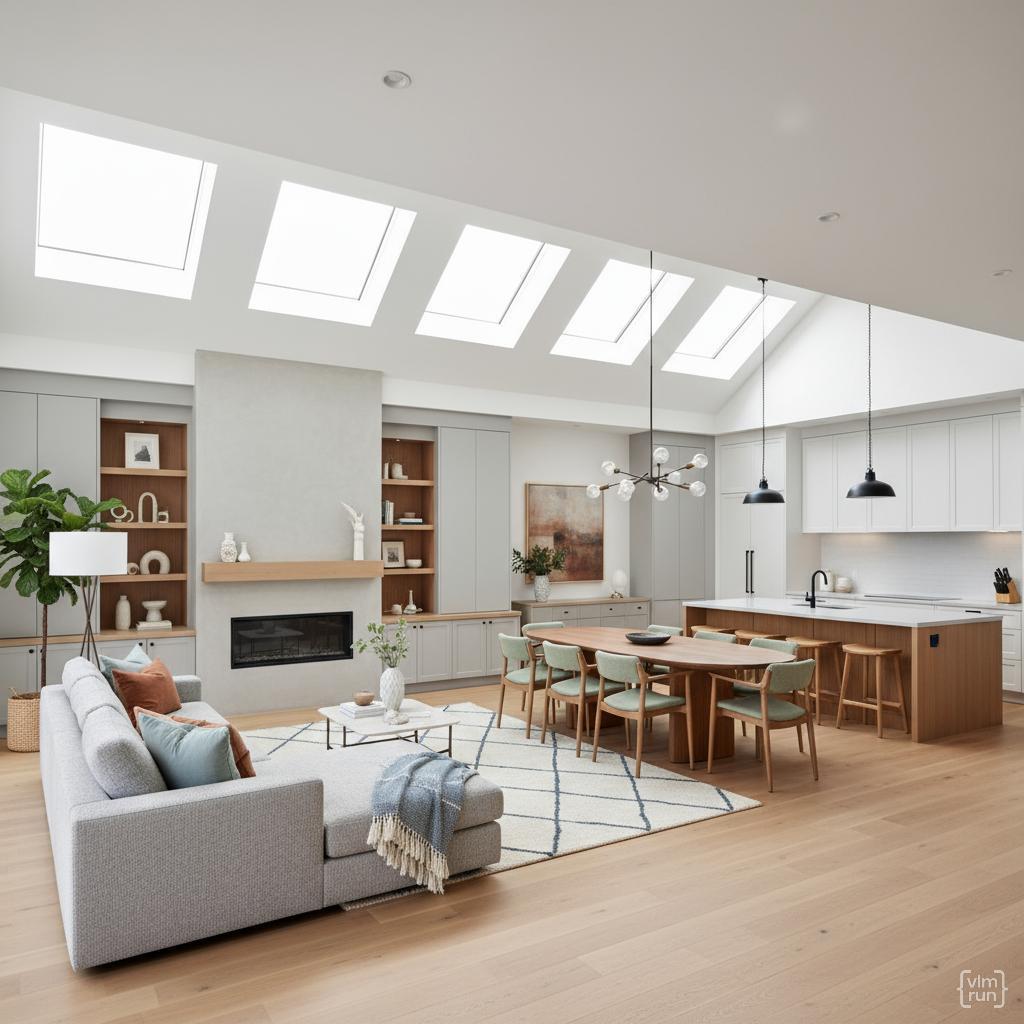}} &      \parbox{3cm}{\centering \includegraphics[width=3cm]{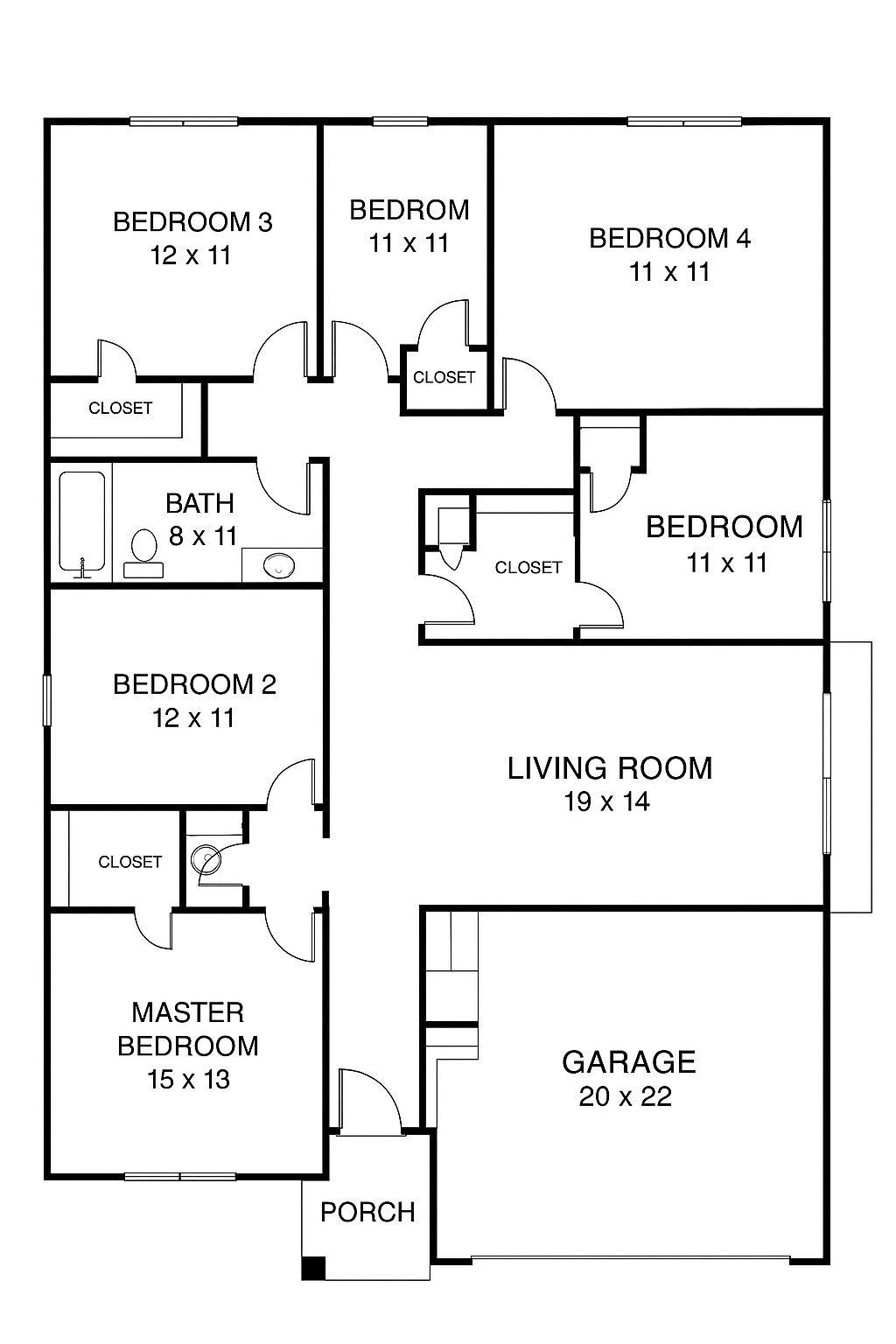}} &      \parbox{3cm}{\centering \includegraphics[width=3cm]{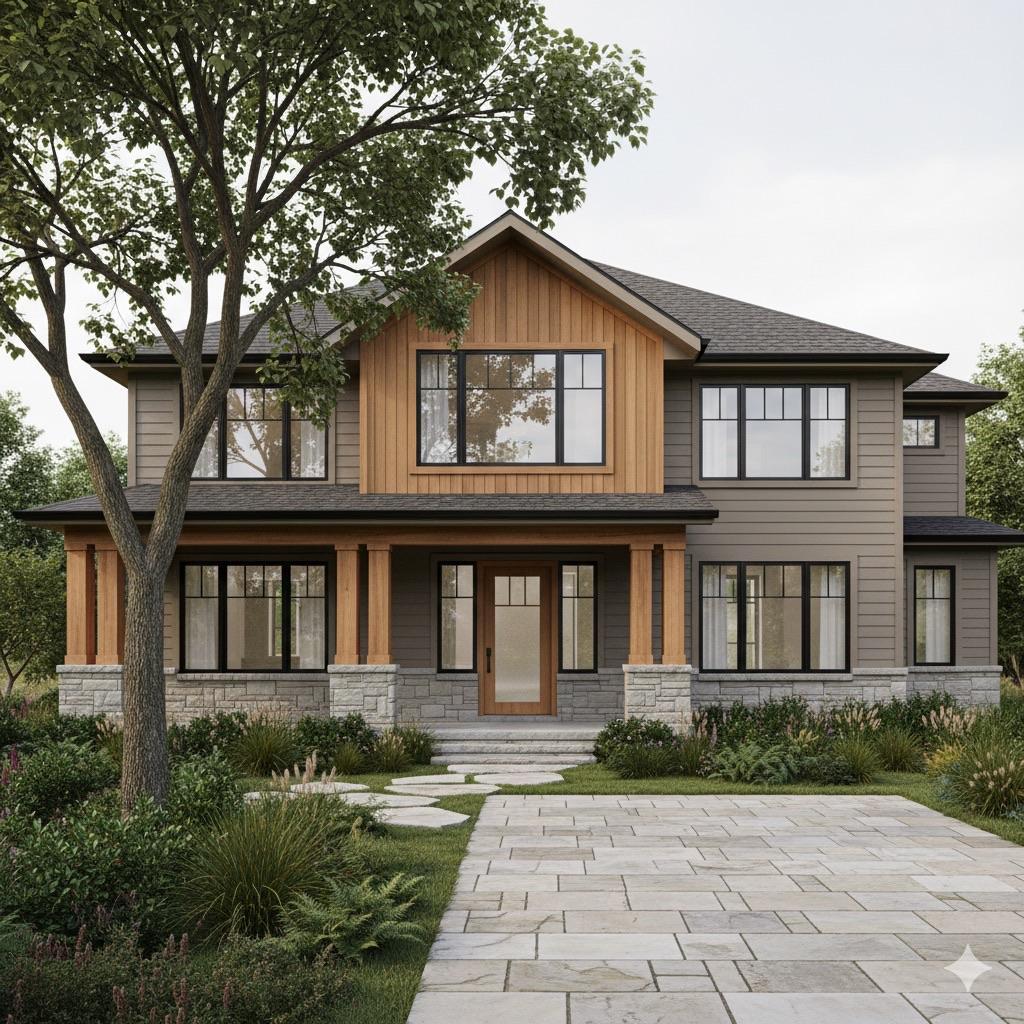}} &      \parbox{3cm}{\centering \includegraphics[width=3cm]{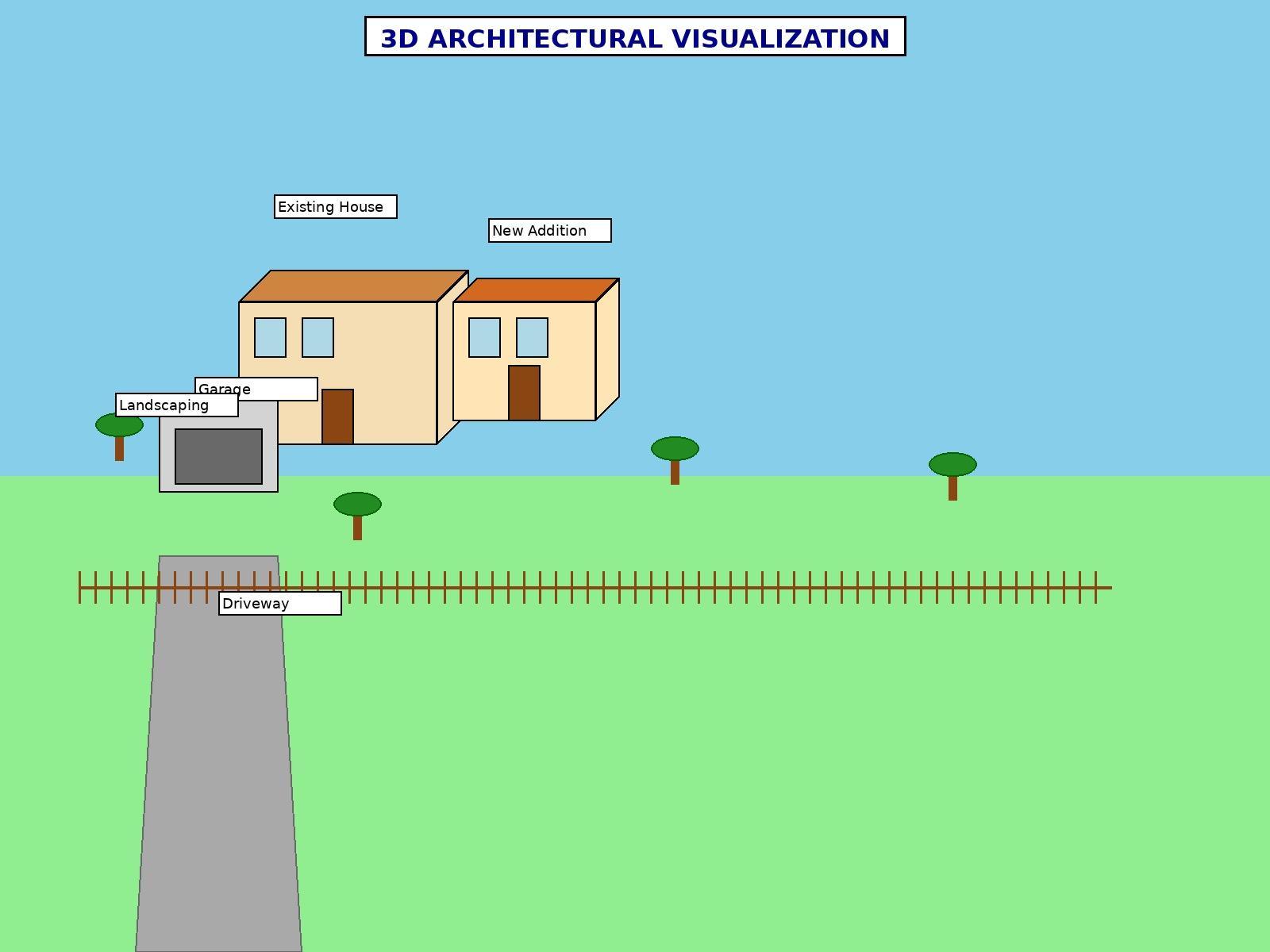}} \\
    \multicolumn{5}{l}{\cellcolor{subtlegray!20}\parbox{\textwidth}{\scriptsize\textit{\textcolor{questiongray}{18. Give a detailed explanation of the plan and generate detailed 3D architectural visualizations and interior design concepts based on the plan.}}}}\\
    \arrayrulecolor{gray!30}\midrule[0.3pt]\arrayrulecolor{black}

    \parbox{3cm}{\centering \includegraphics[width=3cm]{3_results_sections/comparision_table/images/image30.jpg}} & \parbox{3cm}{\centering \includegraphics[width=3cm]{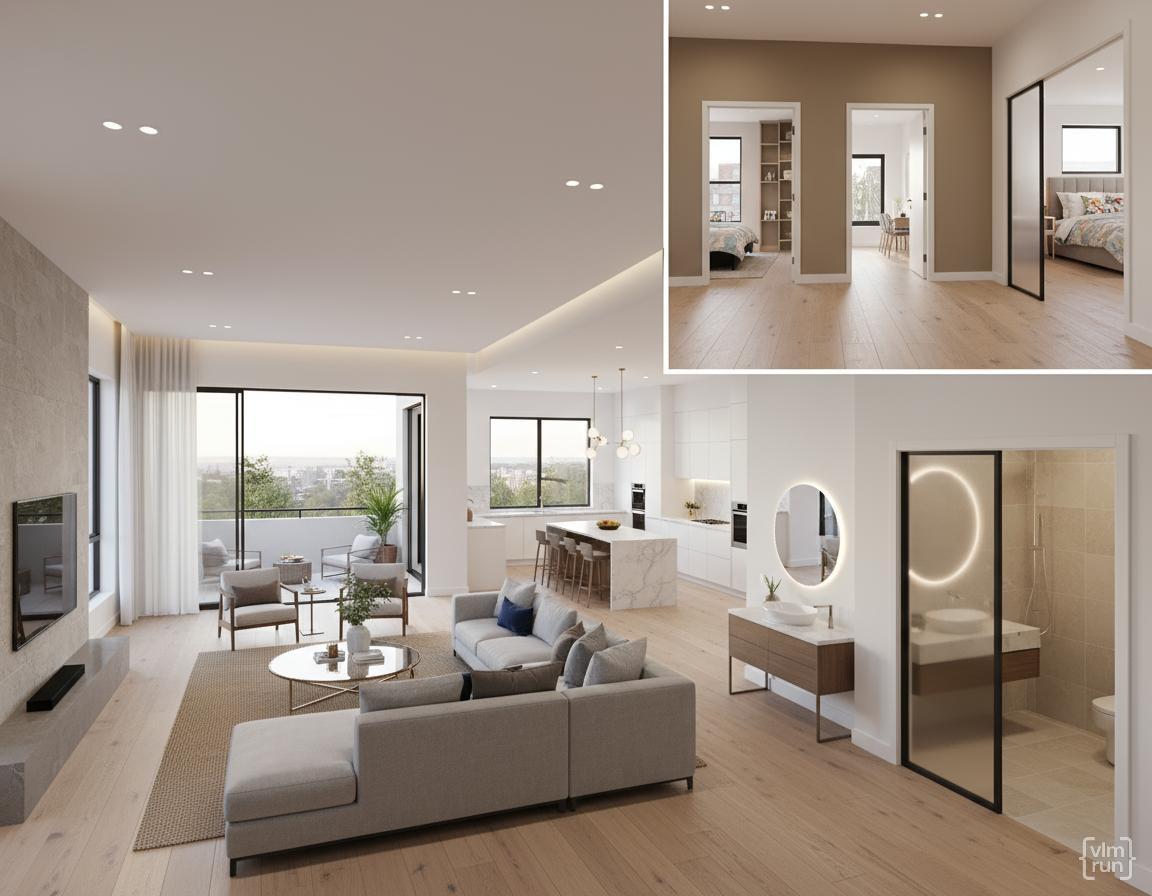}} &      \parbox{3cm}{\centering \includegraphics[width=3cm]{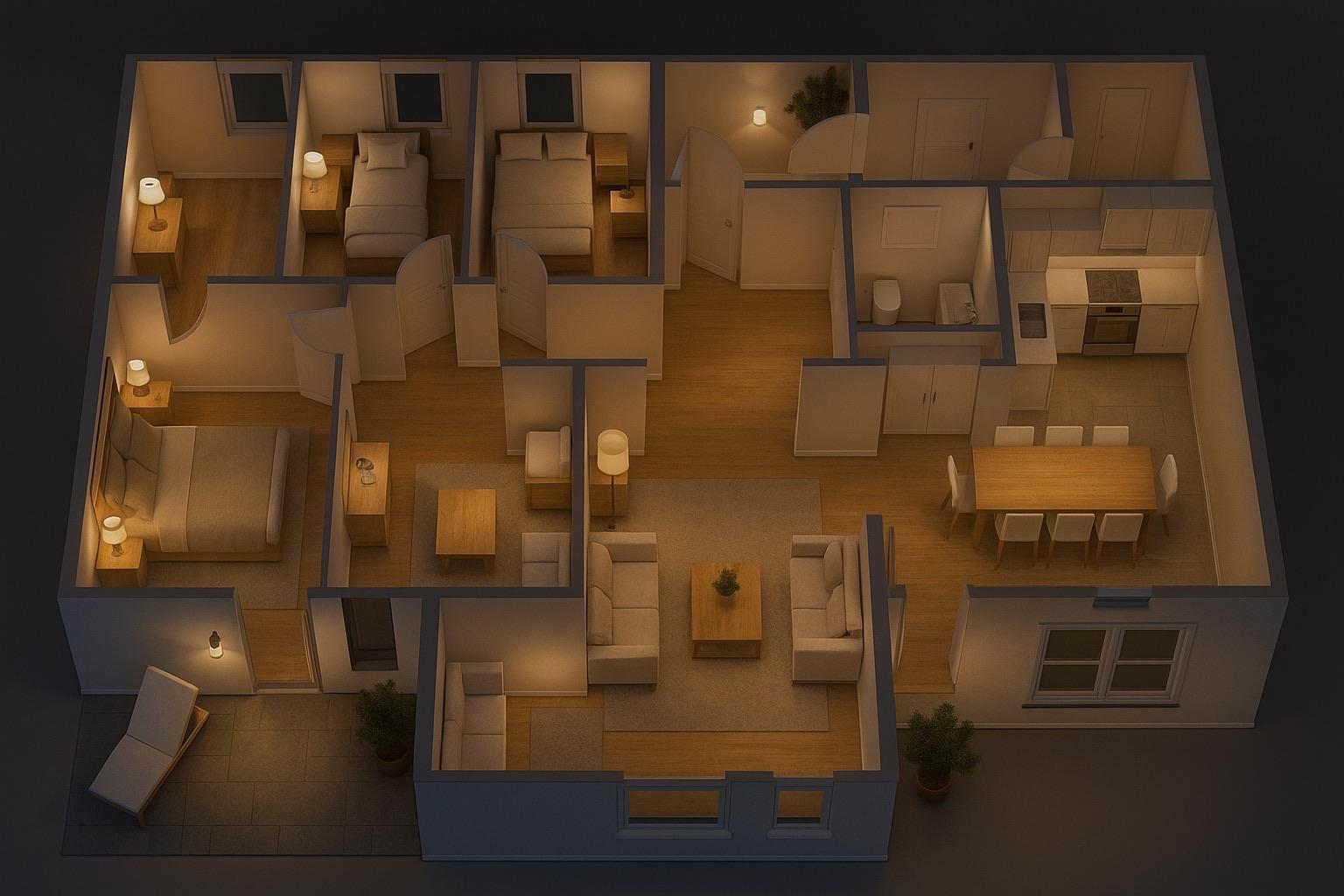}} &      \parbox{3cm}{\centering \includegraphics[width=3cm]{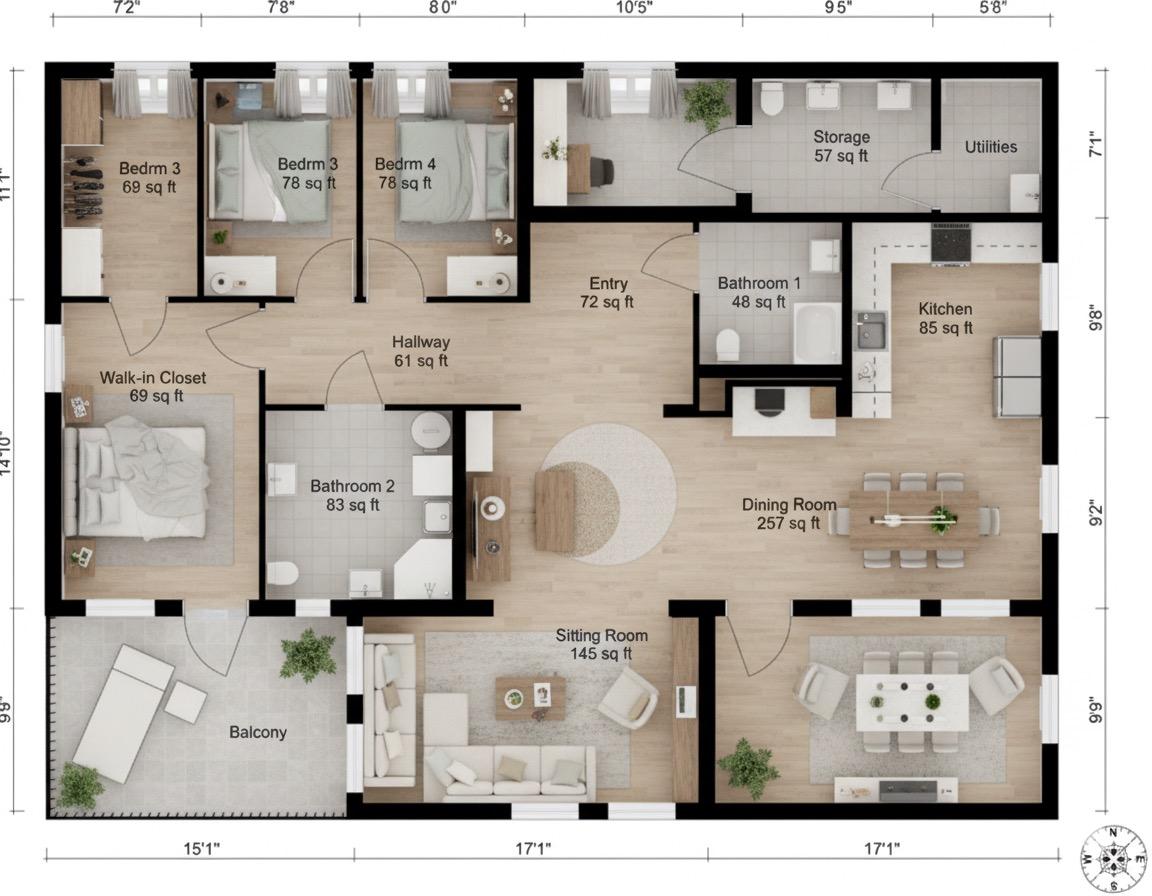}} &      \parbox{3cm}{\centering \includegraphics[width=3cm]{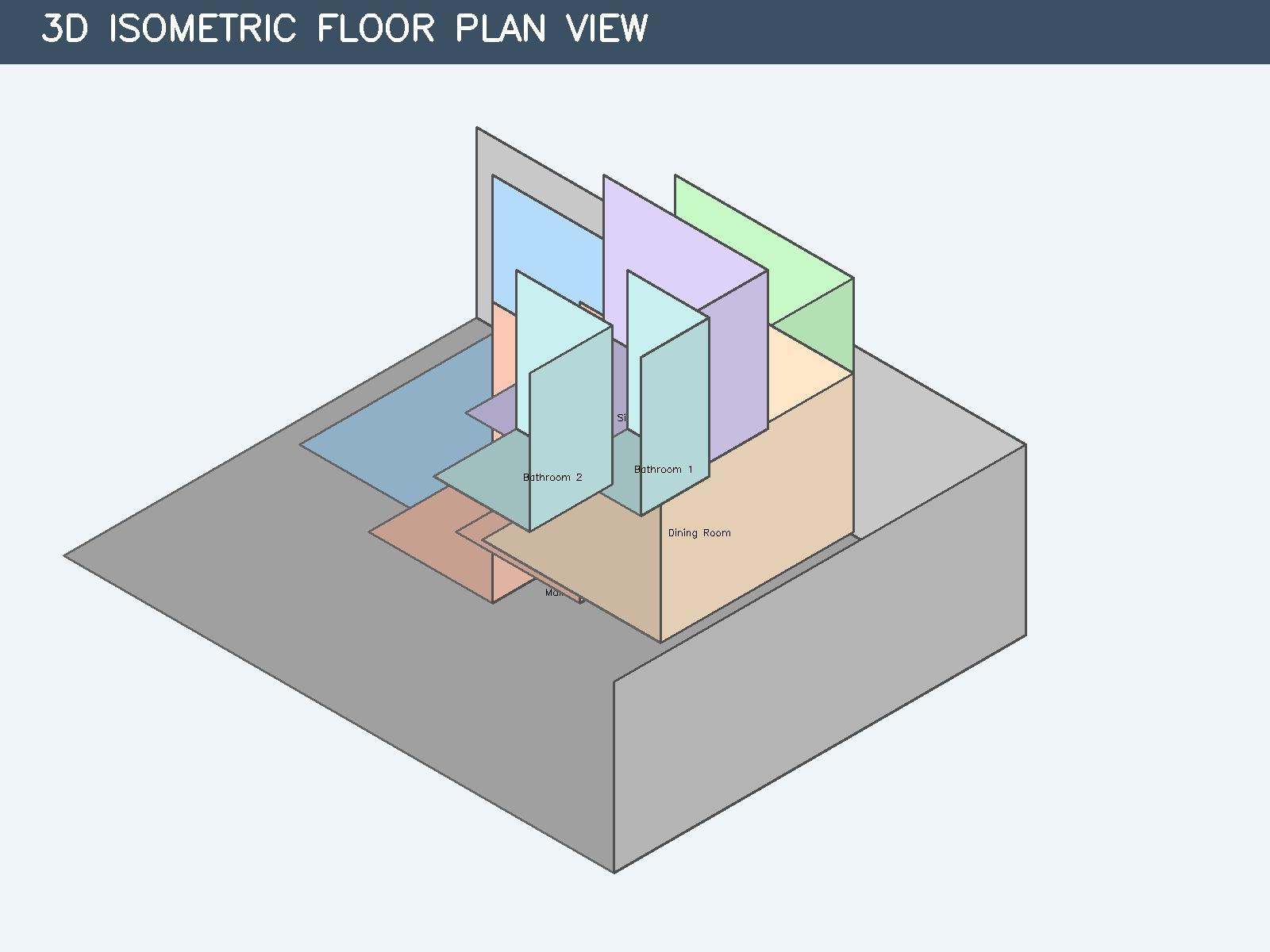}} \\
    \multicolumn{5}{l}{\cellcolor{subtlegray!20}\parbox{\textwidth}{\scriptsize\textit{\textcolor{questiongray}{19. Generate detailed 3D architectural visualizations and interior design concepts based on this floor plan, including furniture placement and lighting.}}}}\\
    \arrayrulecolor{gray!30}\midrule[0.3pt]\arrayrulecolor{black}

    \parbox{3cm}{\centering \includegraphics[width=3cm]{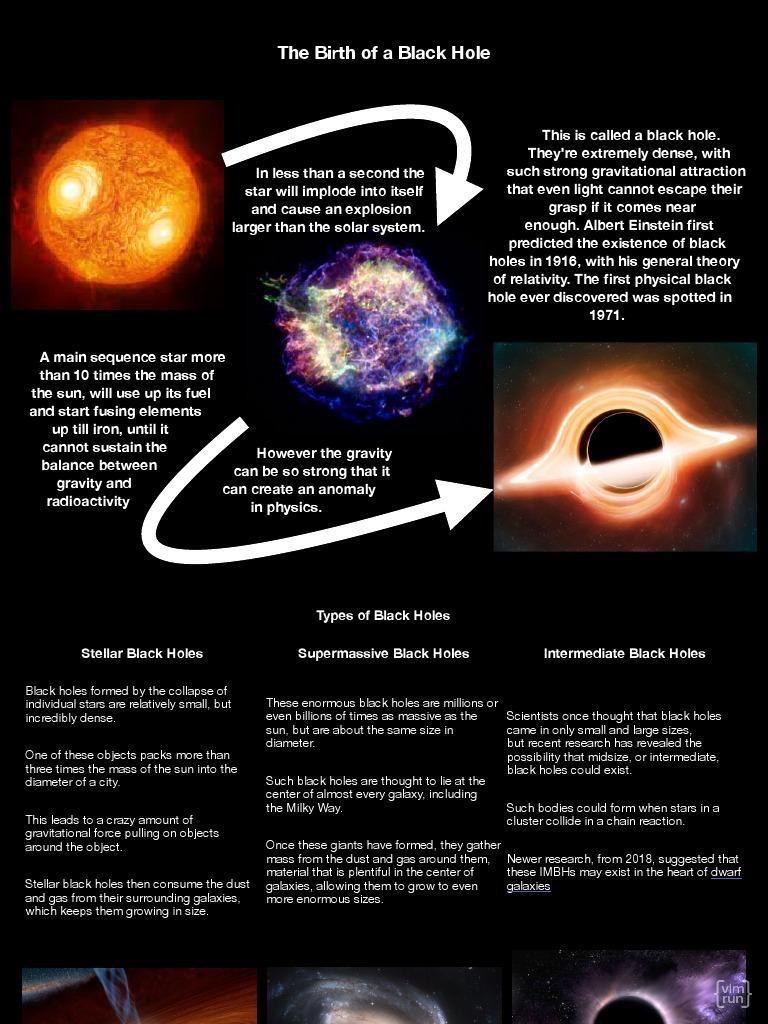}} & \parbox{3cm}{\centering \includegraphics[width=3cm]{3_results_sections/comparision_table/images/image156.jpg}} &      \parbox{3cm}{\centering \includegraphics[width=3cm]{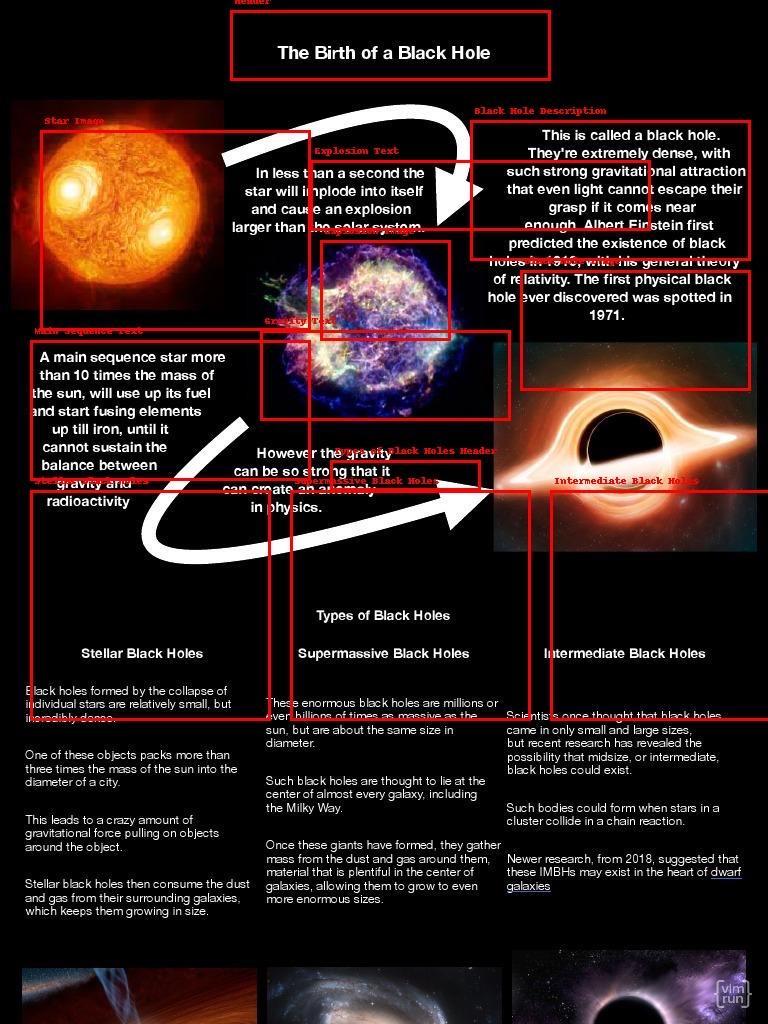}} &      \parbox{3cm}{\centering \failedimage} &      \parbox{3cm}{\centering \failedimage} \\
    \multicolumn{5}{l}{\cellcolor{subtlegray!20}\parbox{\textwidth}{\scriptsize\textit{\textcolor{questiongray}{20. Get the layout of the document and visualize all the text.}}}}\\
    \arrayrulecolor{gray!30}\midrule[0.3pt]\arrayrulecolor{black}

    \parbox{3cm}{\centering \includegraphics[width=3cm]{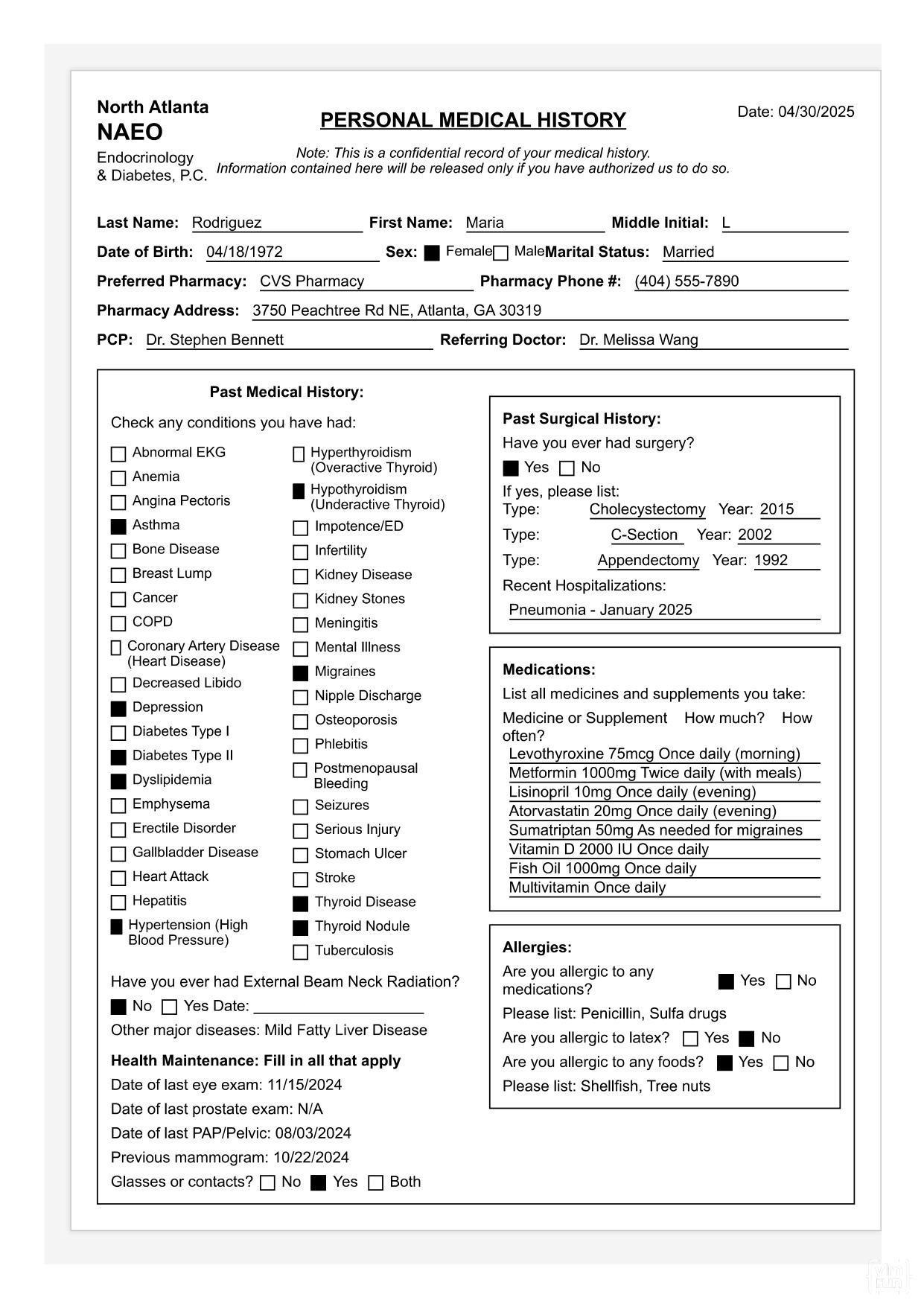}} & \parbox{3cm}{\centering \includegraphics[width=3cm]{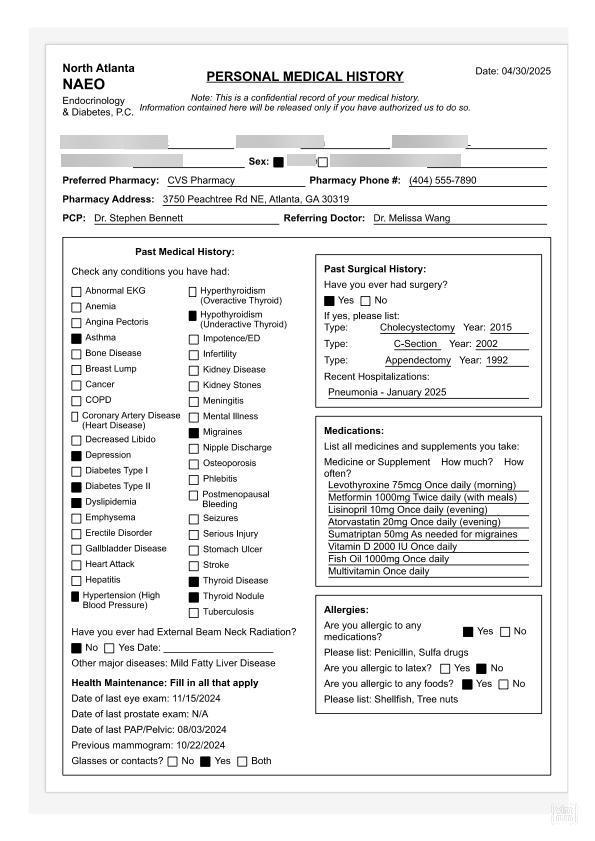}} &      \parbox{3cm}{\centering \includegraphics[width=3cm]{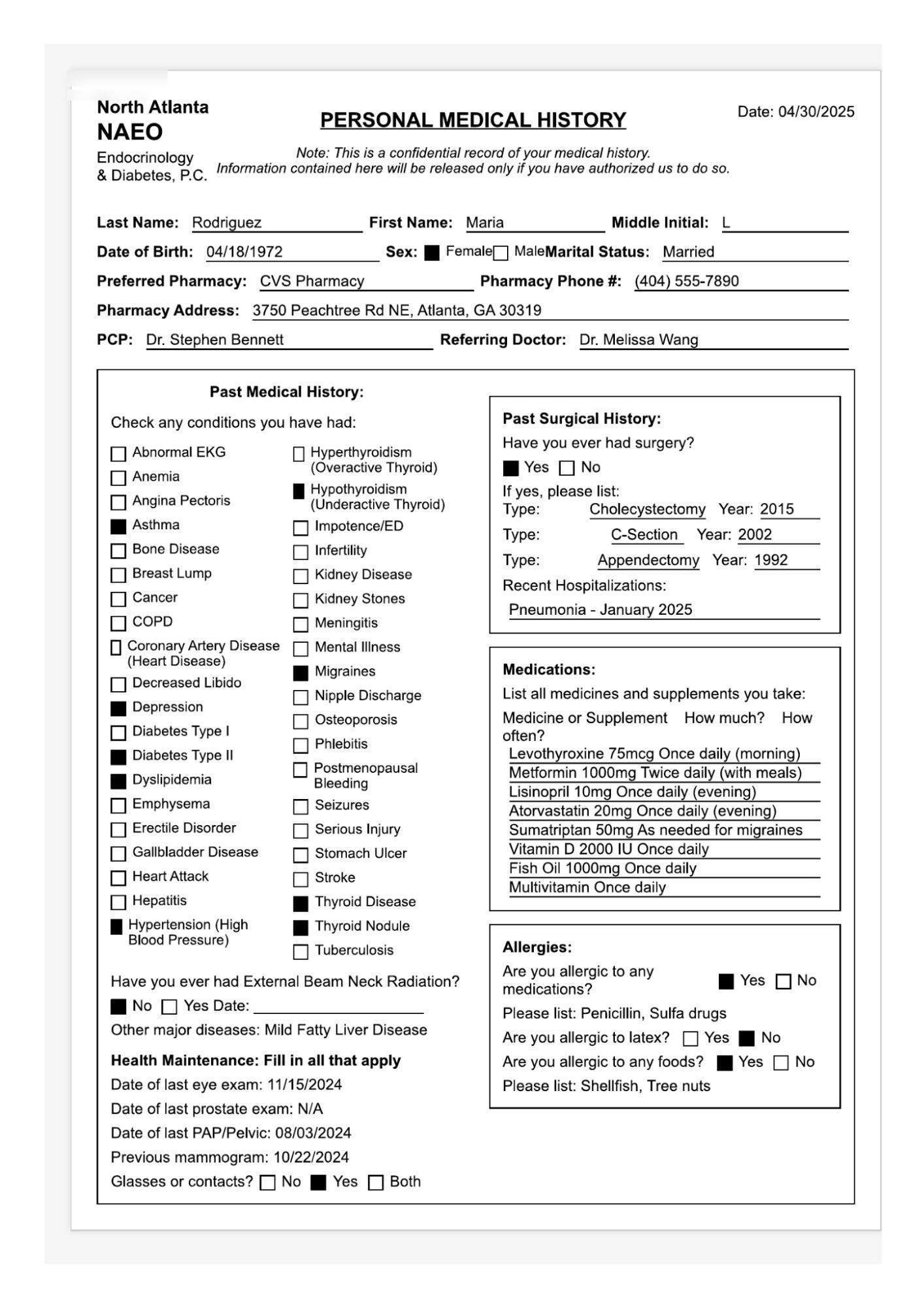}} &      \parbox{3cm}{\centering \includegraphics[width=3cm]{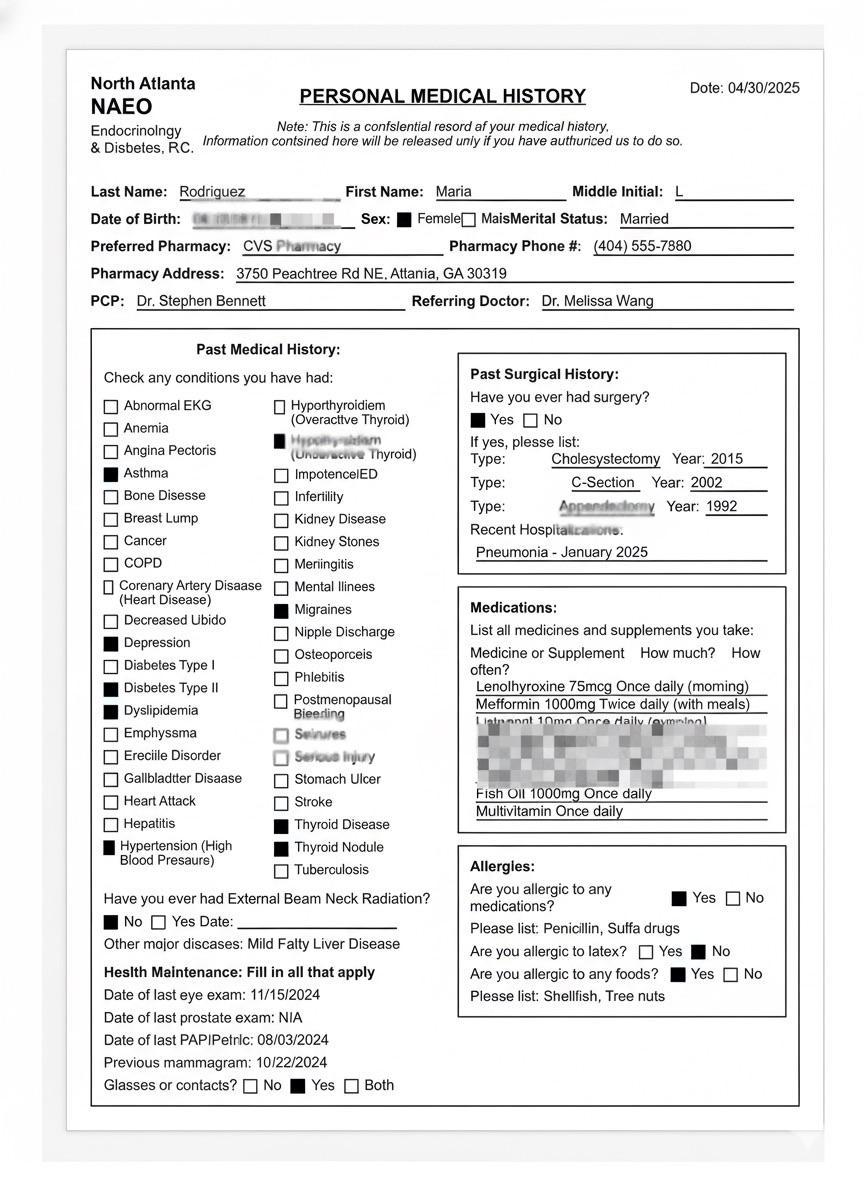}} &      \parbox{3cm}{\centering \includegraphics[width=3cm]{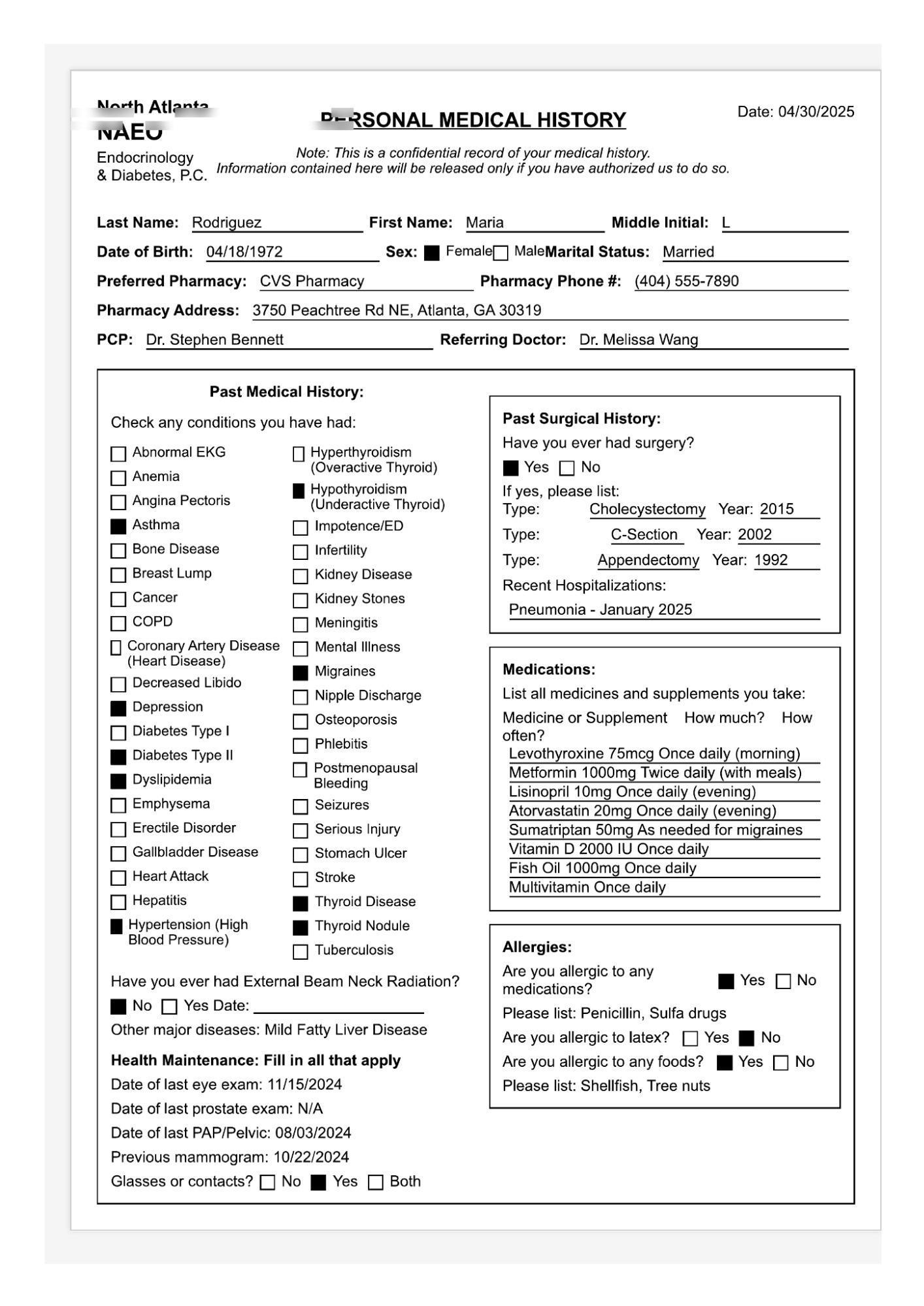}} \\
    \multicolumn{5}{l}{\cellcolor{subtlegray!20}\parbox{\textwidth}{\scriptsize\textit{\textcolor{questiongray}{21. Can you blur out the patient name, dob, sex and marriage status?}}}}\\
    \arrayrulecolor{gray!30}\midrule[0.3pt]\arrayrulecolor{black}

    \parbox{3cm}{\centering \includegraphics[width=3cm]{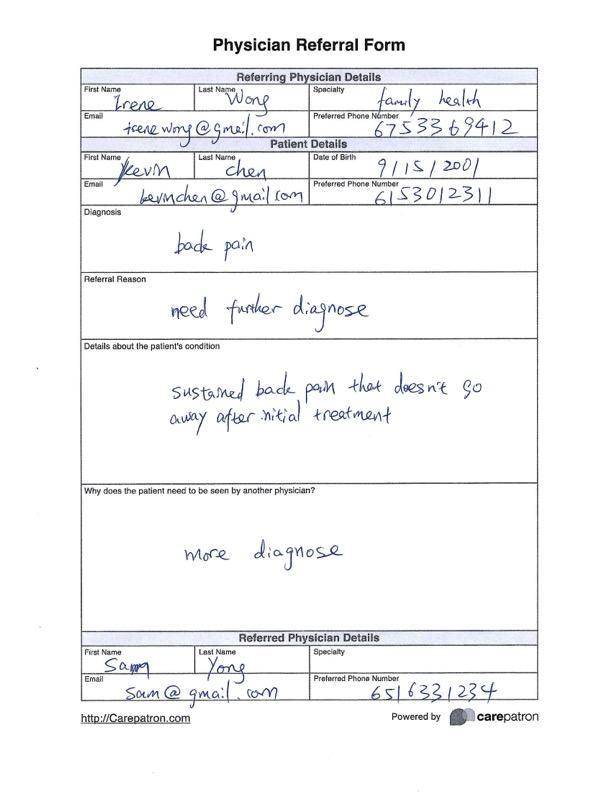}} & \parbox{3cm}{\centering \includegraphics[width=3cm]{3_results_sections/comparision_table/images/image99.jpg}} &      \parbox{3cm}{\centering \includegraphics[width=3cm]{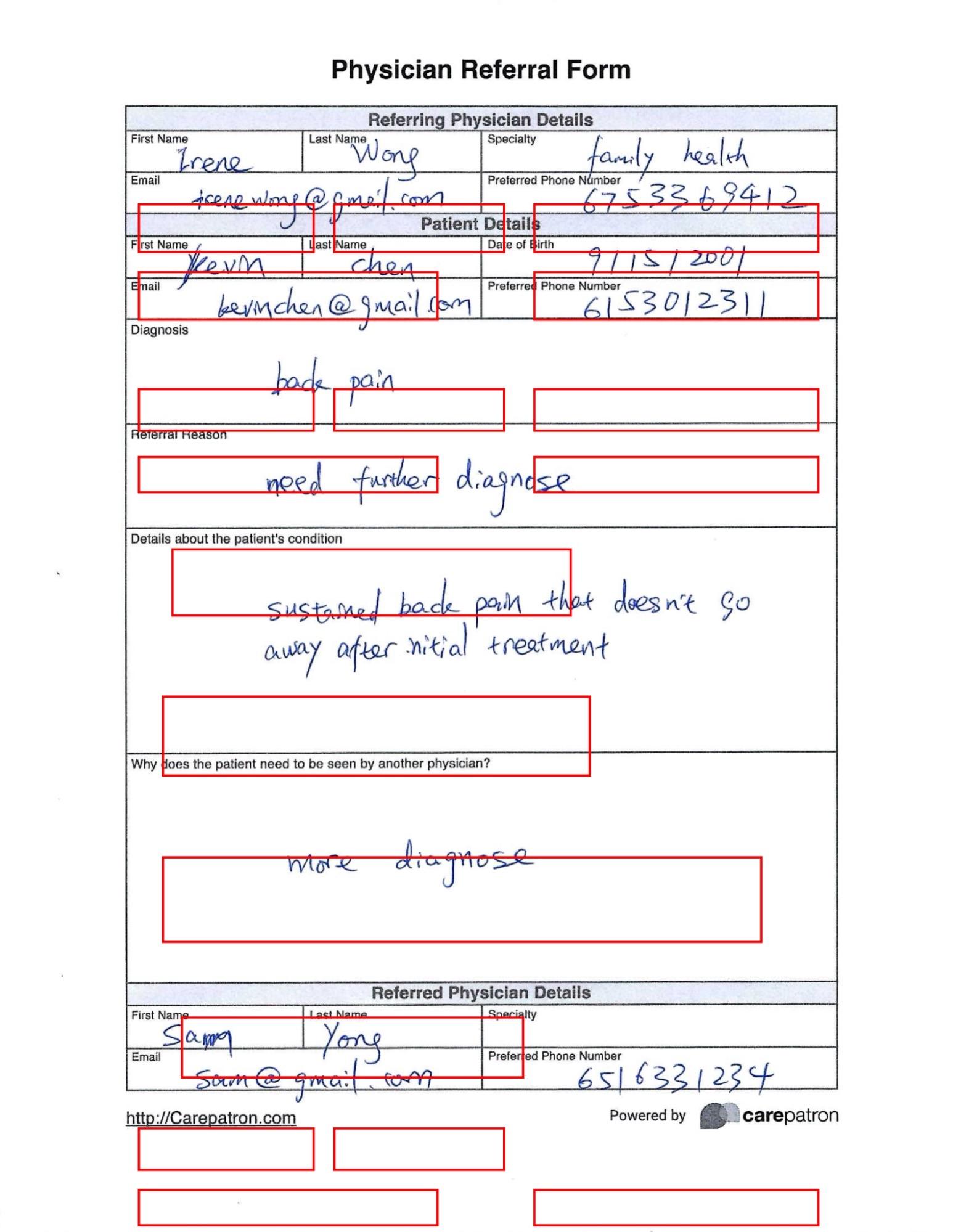}} &      \parbox{3cm}{\centering \includegraphics[width=3cm]{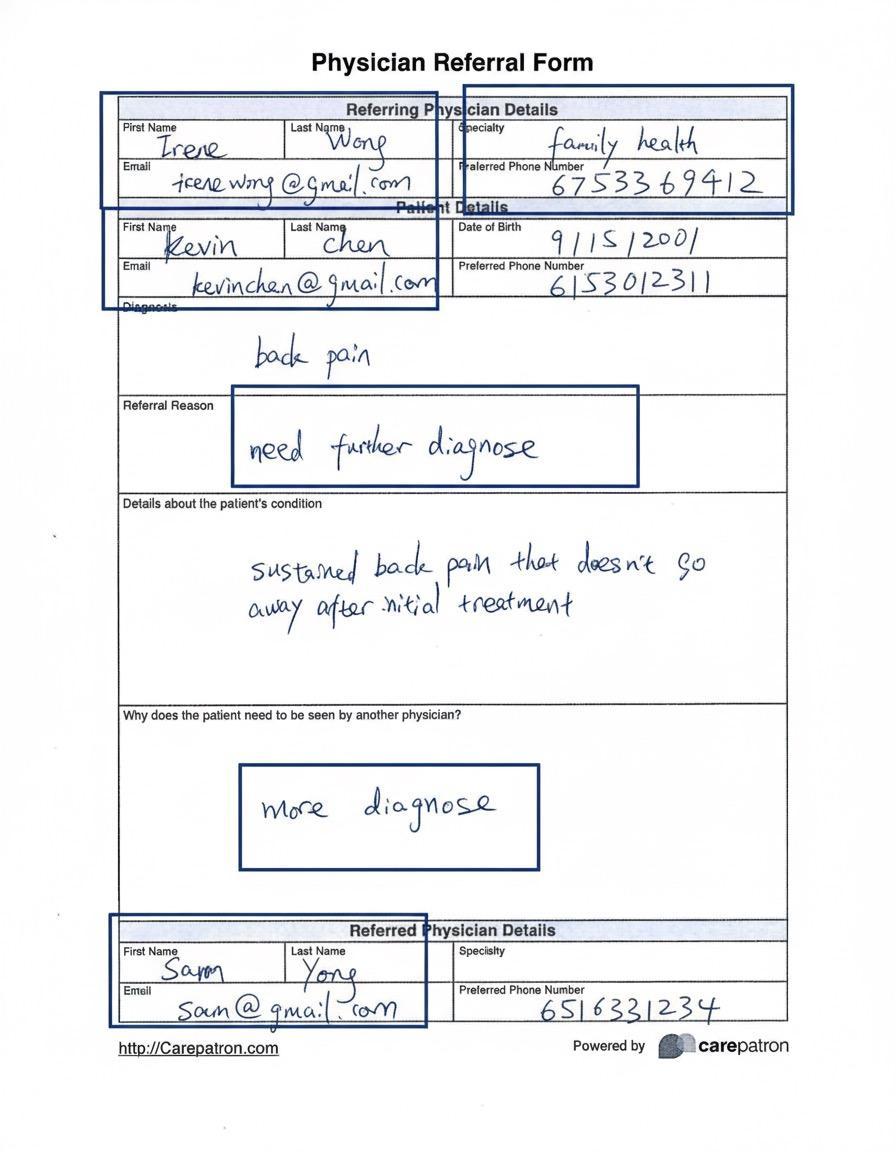}} &      \parbox{3cm}{\centering \includegraphics[width=3cm]{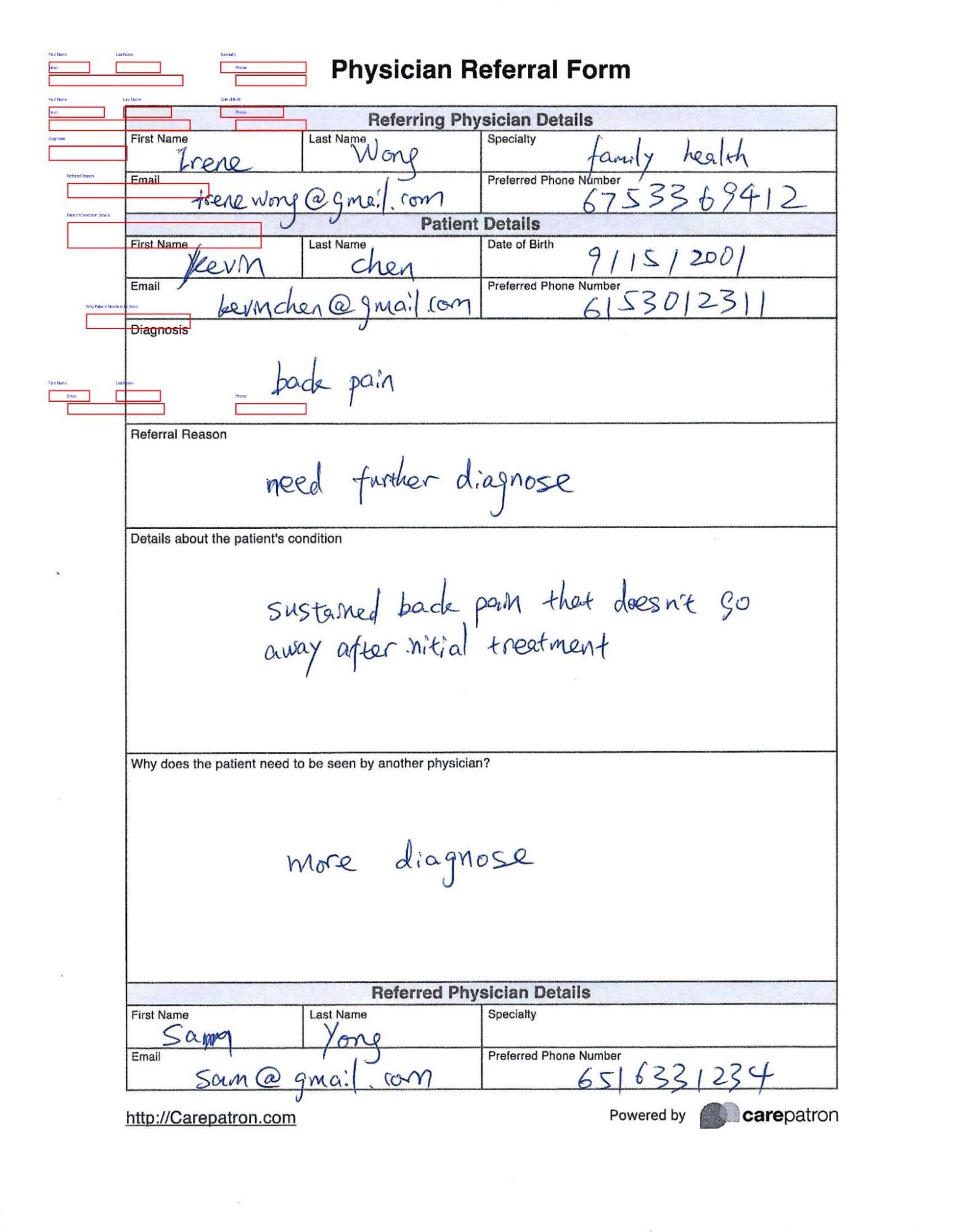}} \\
    \multicolumn{5}{l}{\cellcolor{subtlegray!20}\parbox{\textwidth}{\scriptsize\textit{\textcolor{questiongray}{22. Parse this document and extract all the handwritten fields. Detect the locations of each of the fields and visualize them.}}}}\\
    \arrayrulecolor{gray!30}\midrule[0.3pt]\arrayrulecolor{black}

    \parbox{3cm}{\centering \includegraphics[width=3cm]{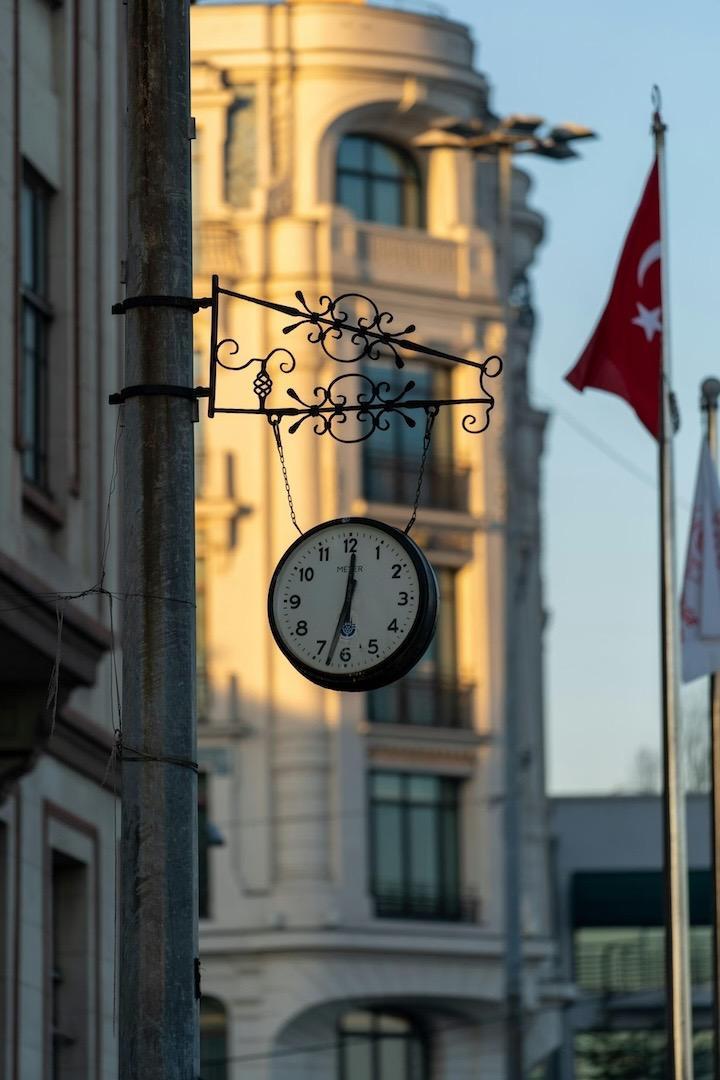}} & \parbox{3cm}{\centering \includegraphics[width=3cm]{3_results_sections/comparision_table/images/image77.jpg}} &      \parbox{3cm}{\centering \includegraphics[width=3cm]{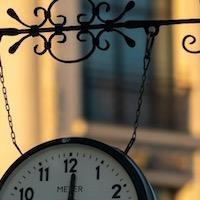}} &      \parbox{3cm}{\centering \includegraphics[width=3cm]{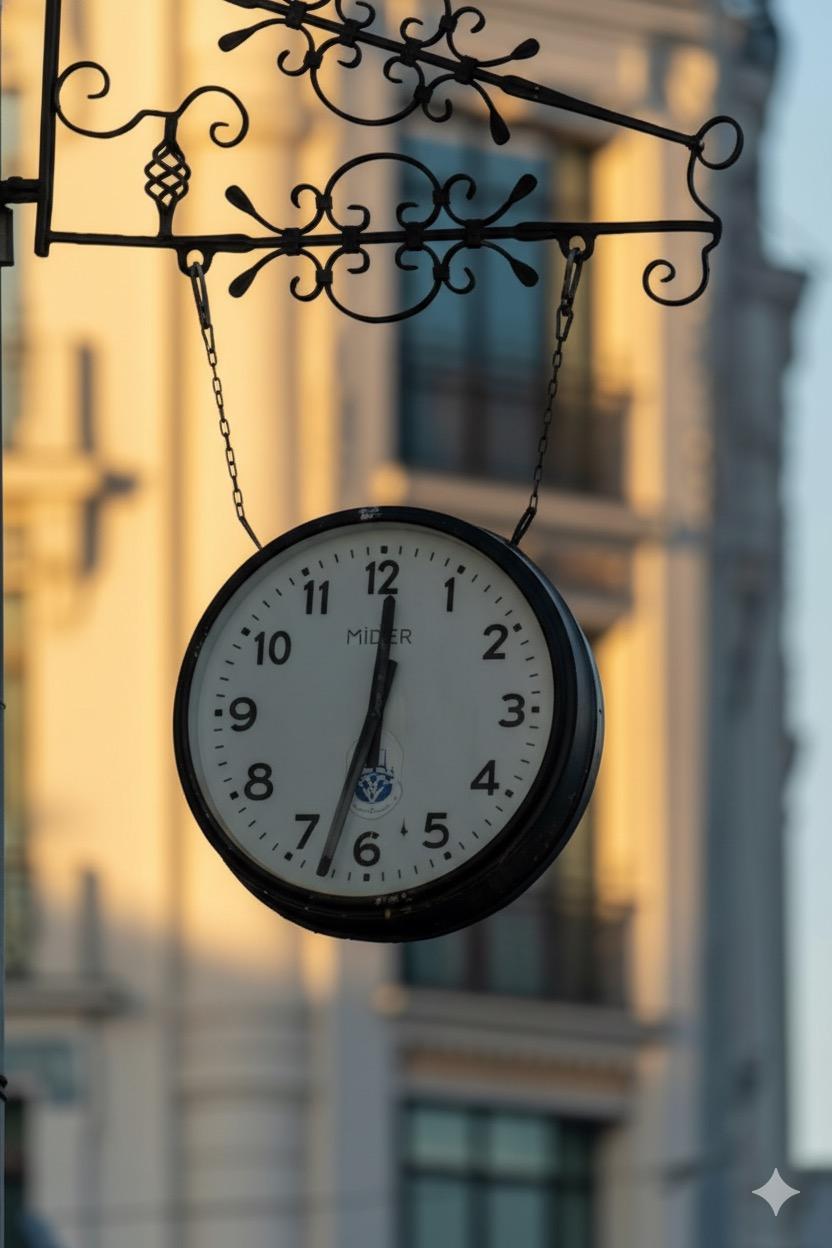}} &      \parbox{3cm}{\centering \failedimage} \\
    \multicolumn{5}{l}{\cellcolor{subtlegray!20}\parbox{\textwidth}{\scriptsize\textit{\textcolor{questiongray}{23. Crop into the clock in the image and extract the time shown.}}}}\\
    \arrayrulecolor{gray!30}\midrule[0.3pt]\arrayrulecolor{black}

    \parbox{3cm}{\centering \includegraphics[width=3cm]{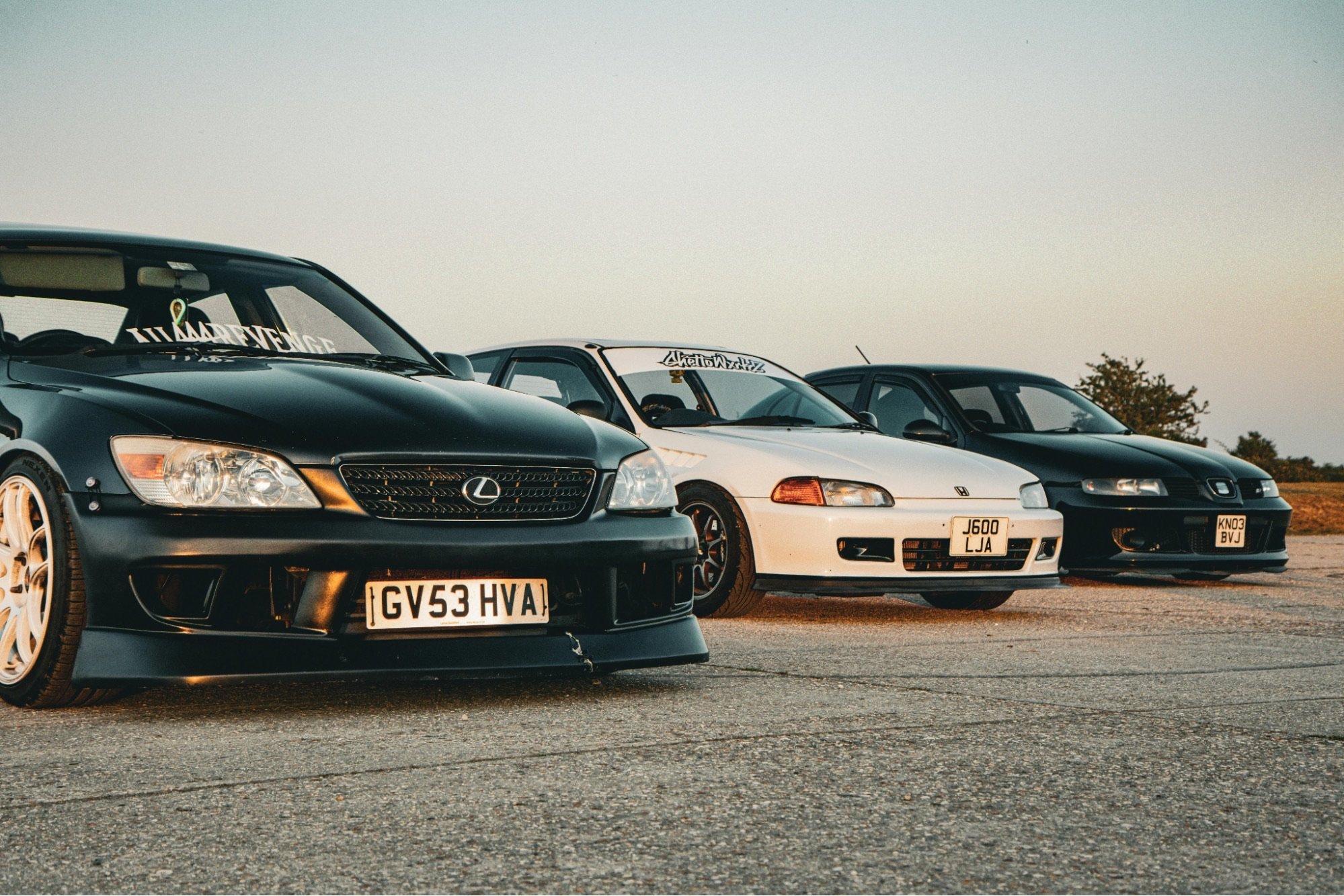}} & \parbox{3cm}{\centering \includegraphics[width=3cm]{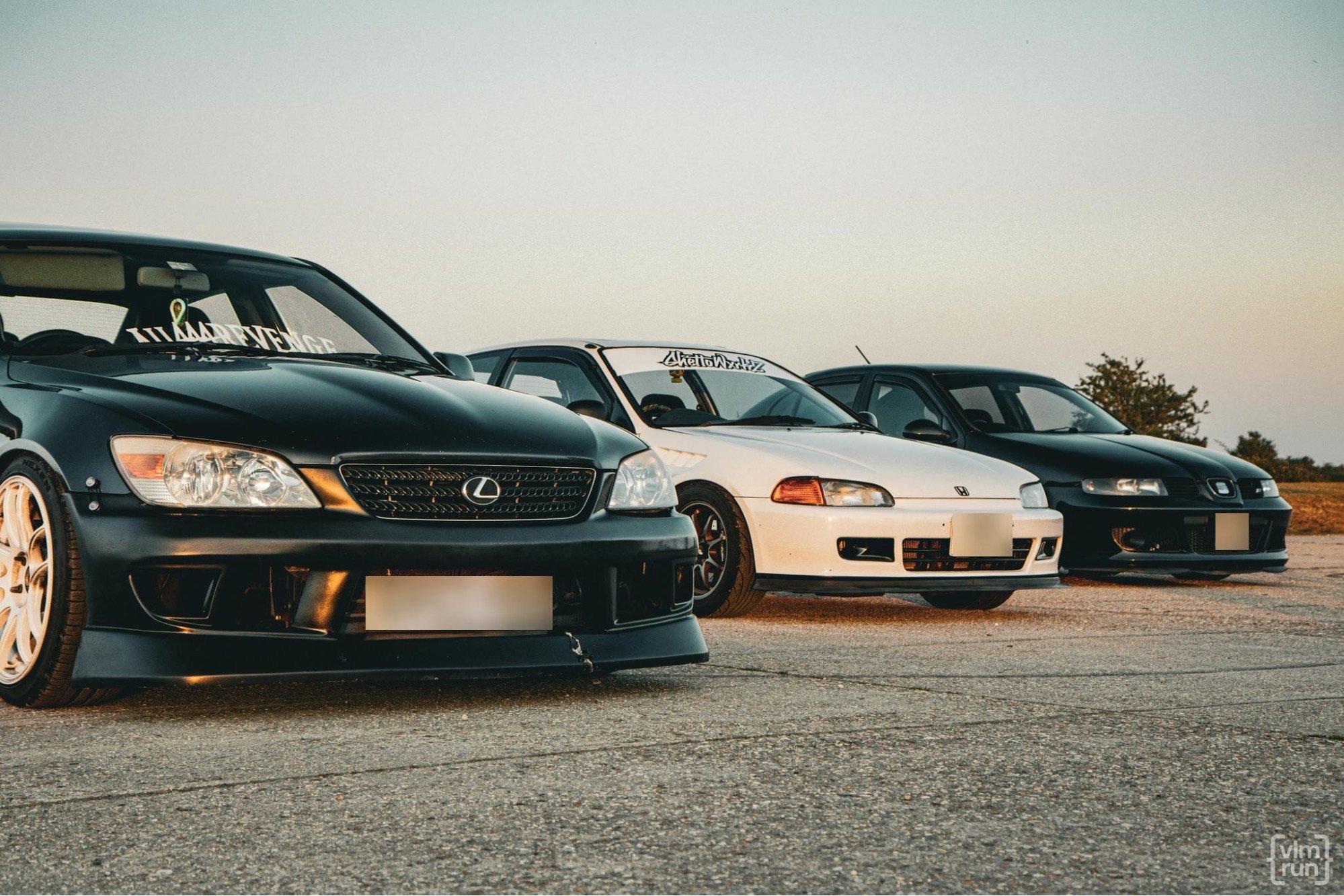}} &      \parbox{3cm}{\centering \includegraphics[width=3cm]{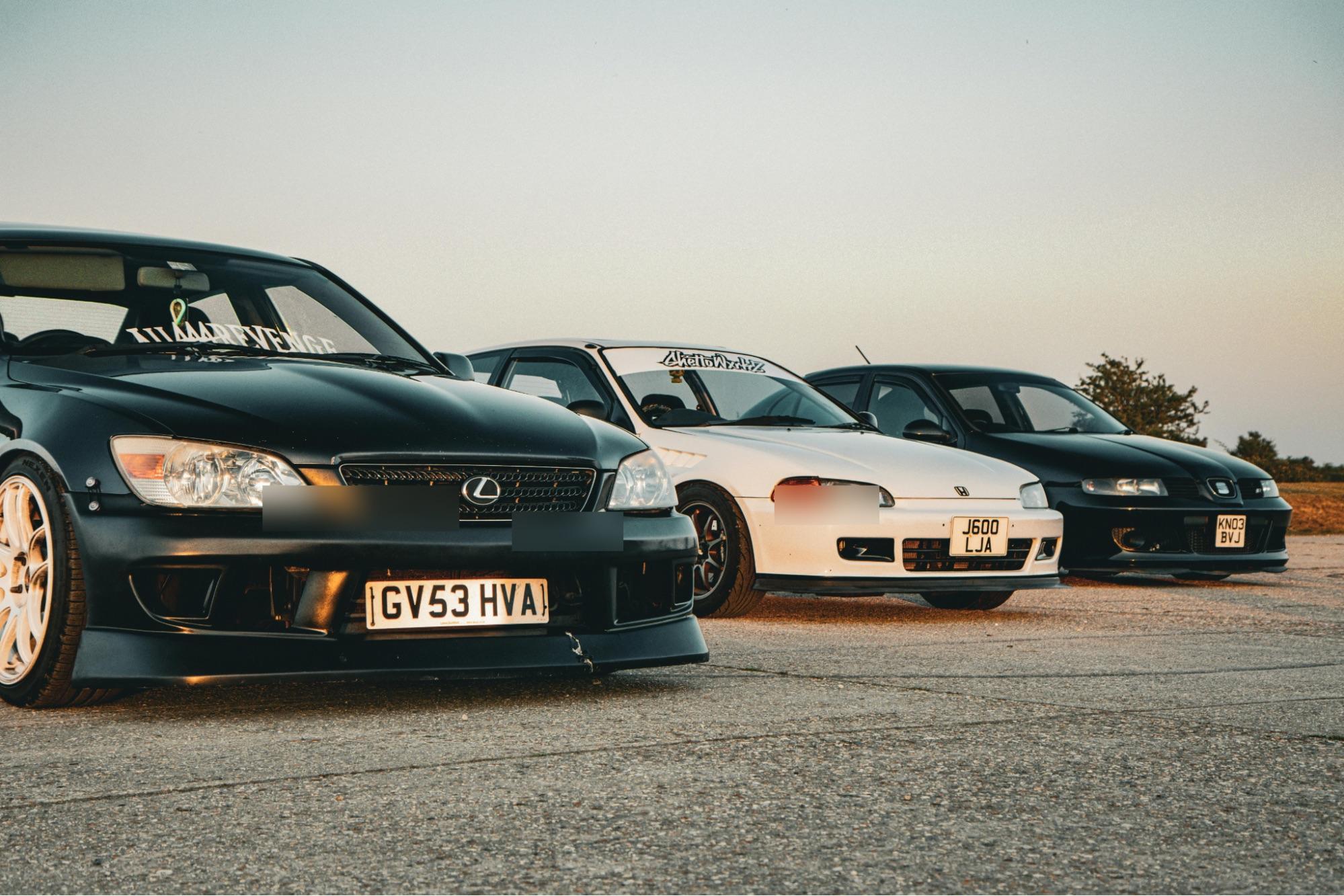}} &      \parbox{3cm}{\centering \includegraphics[width=3cm]{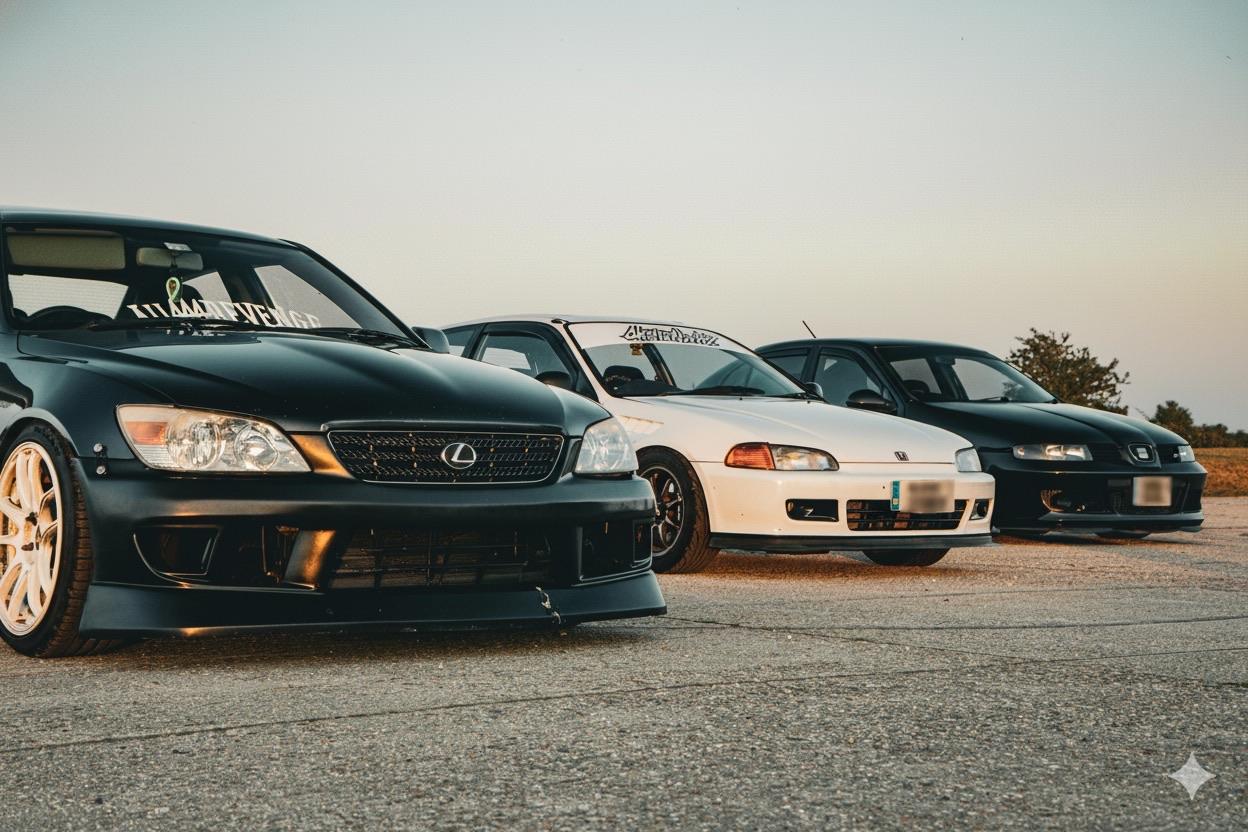}} &      \parbox{3cm}{\centering \failedimage} \\
    \multicolumn{5}{l}{\cellcolor{subtlegray!20}\parbox{\textwidth}{\scriptsize\textit{\textcolor{questiongray}{24. First detect the license plates in the image and blur them.}}}}\\
    \arrayrulecolor{gray!30}\midrule[0.3pt]\arrayrulecolor{black}

    \parbox{3cm}{\centering \includegraphics[width=3cm]{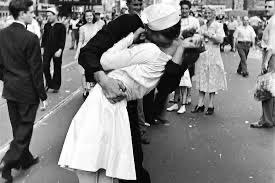}} & \parbox{3cm}{\centering \includegraphics[width=3cm]{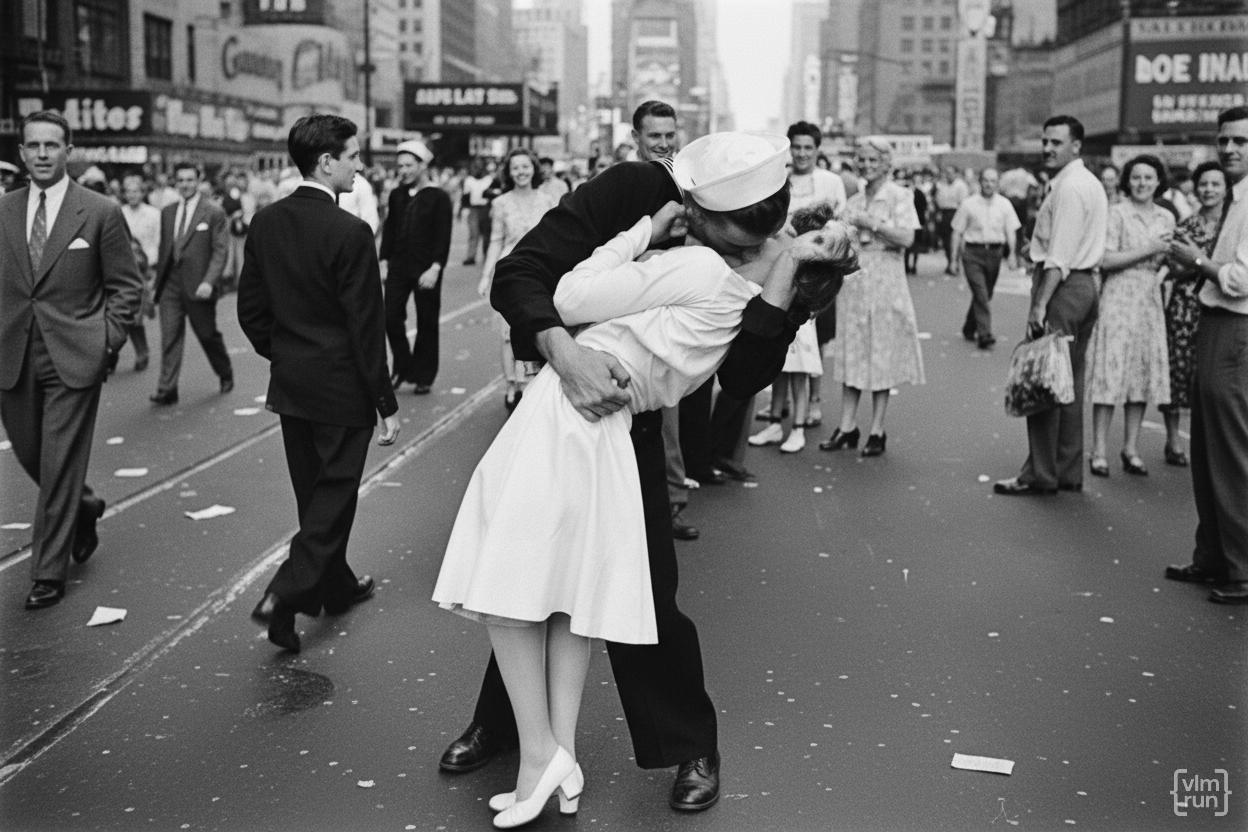}} &      \parbox{3cm}{\centering \includegraphics[width=3cm]{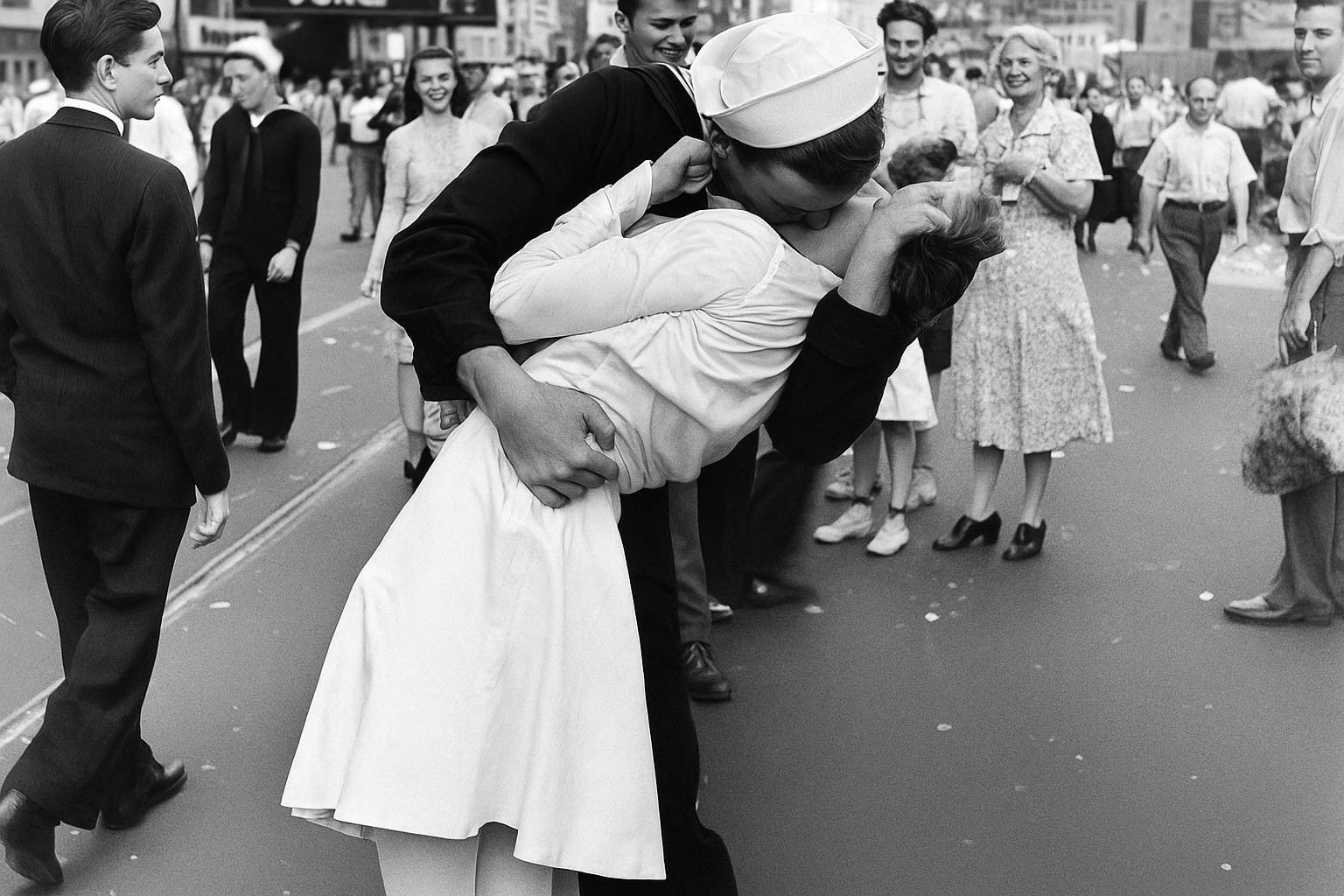}} &      \parbox{3cm}{\centering \includegraphics[width=3cm]{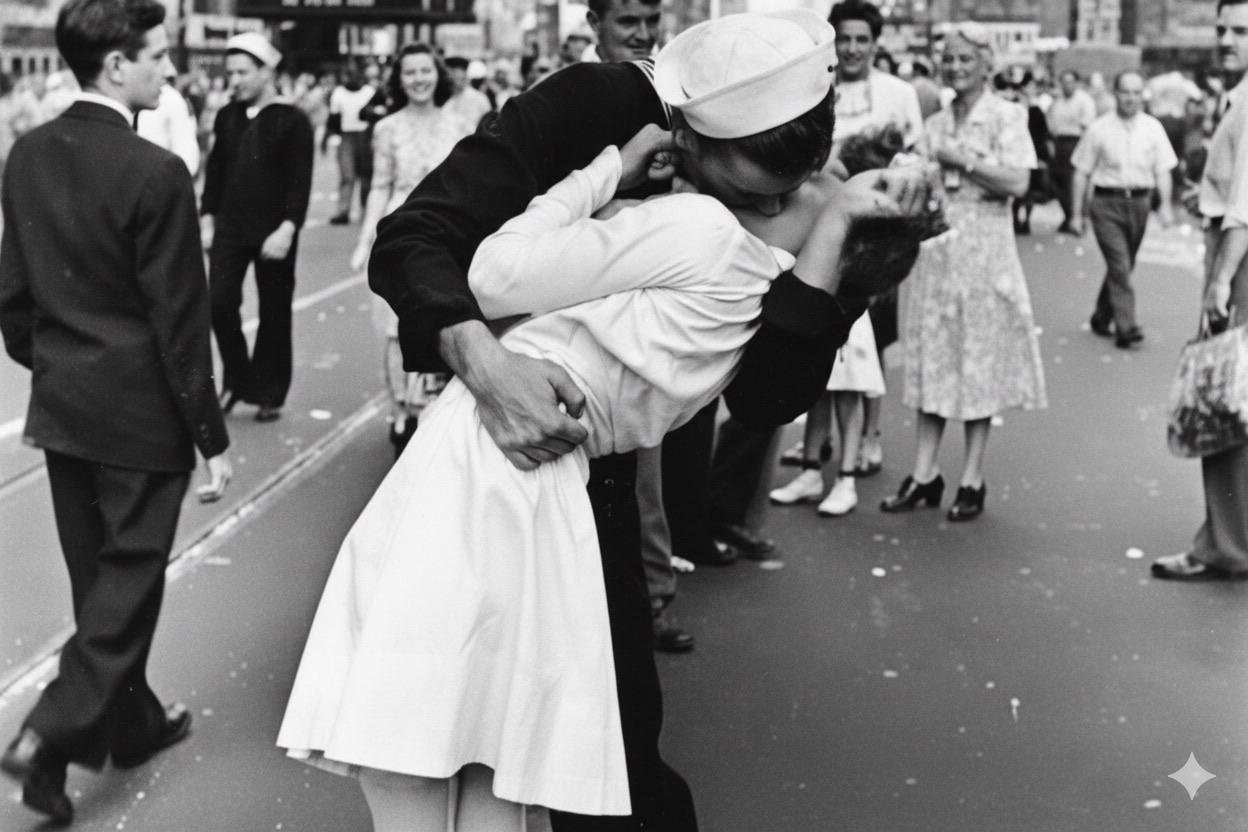}} &      \parbox{3cm}{\centering \failedimage} \\
    \multicolumn{5}{l}{\cellcolor{subtlegray!20}\parbox{\textwidth}{\scriptsize\textit{\textcolor{questiongray}{25. Analyze this image quality and enhance the image to a high resolution.}}}}\\
    \arrayrulecolor{gray!30}\midrule[0.3pt]\arrayrulecolor{black}

    \parbox{3cm}{\centering \includegraphics[width=3cm]{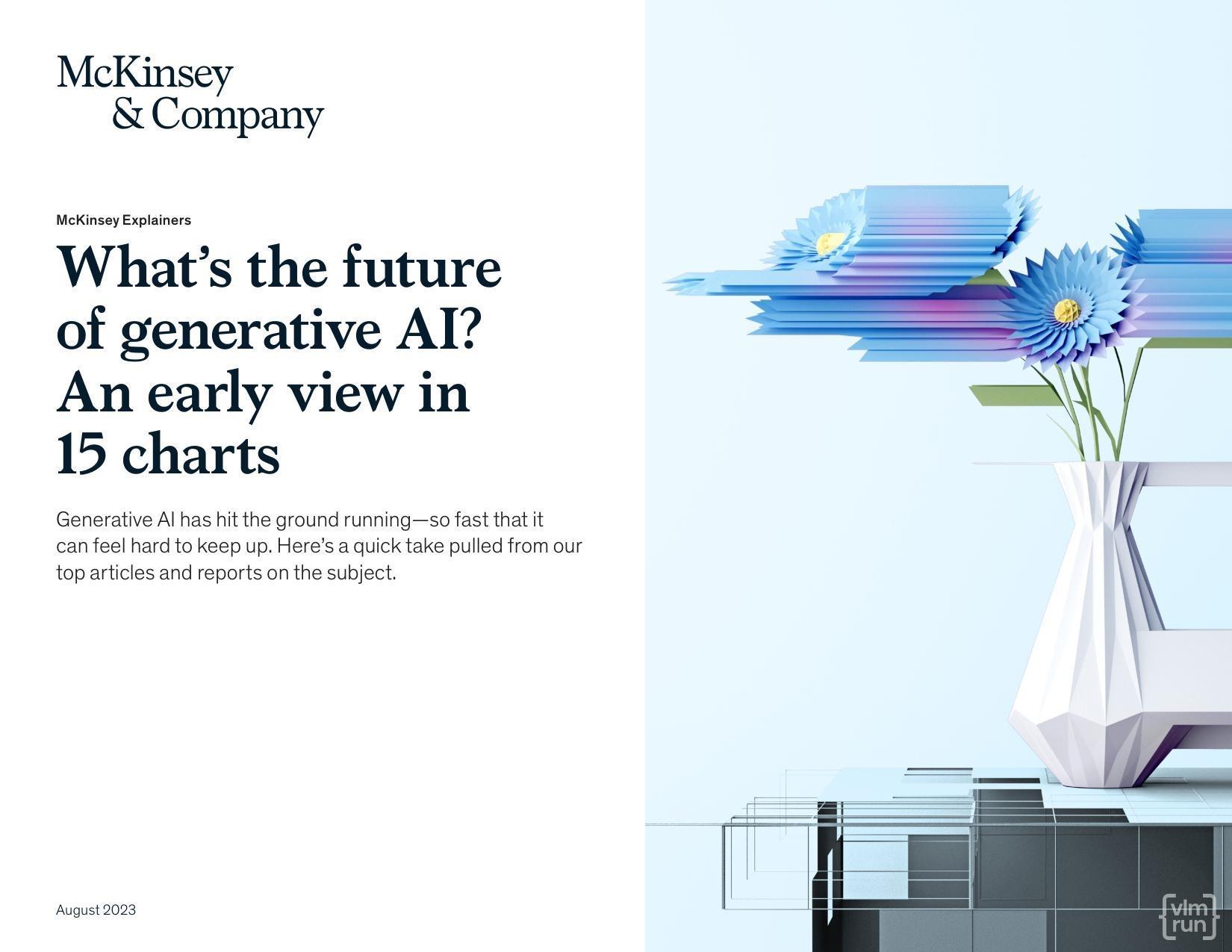}} & \parbox{3cm}{\centering \includegraphics[width=3cm]{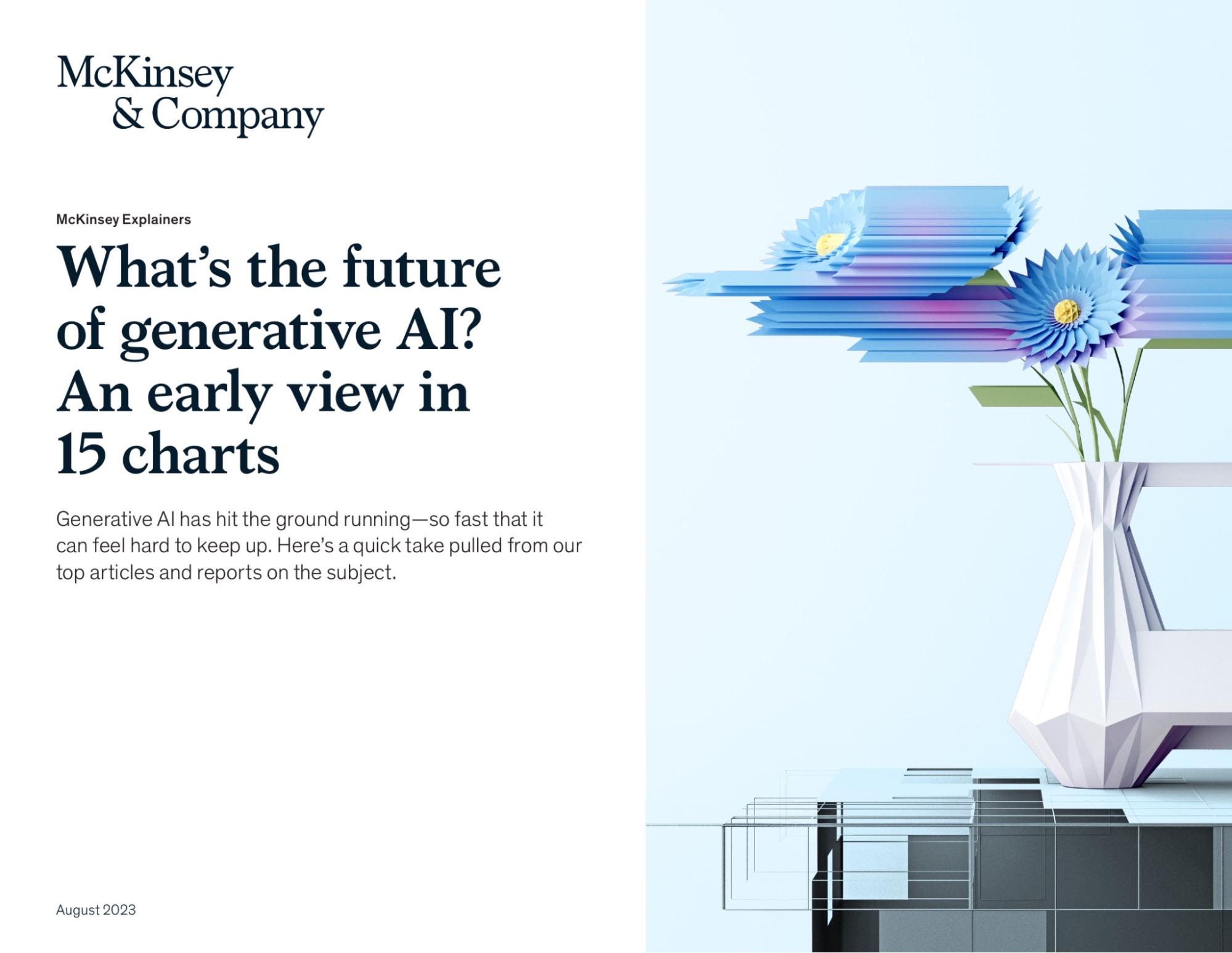}\\\includegraphics[width=3cm]{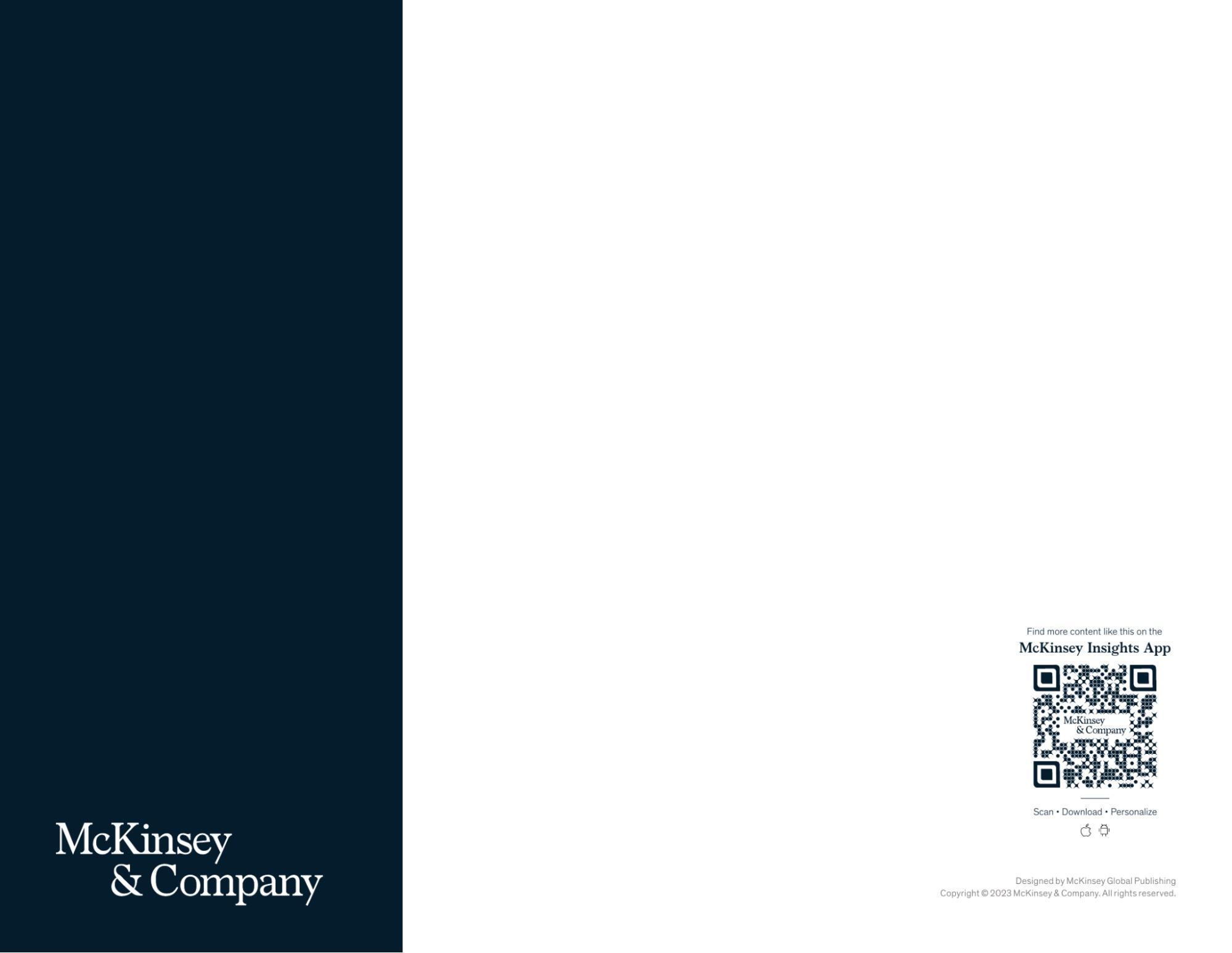}} &      \parbox{3cm}{\centering \failedimage} &      \parbox{3cm}{\centering \includegraphics[width=3cm]{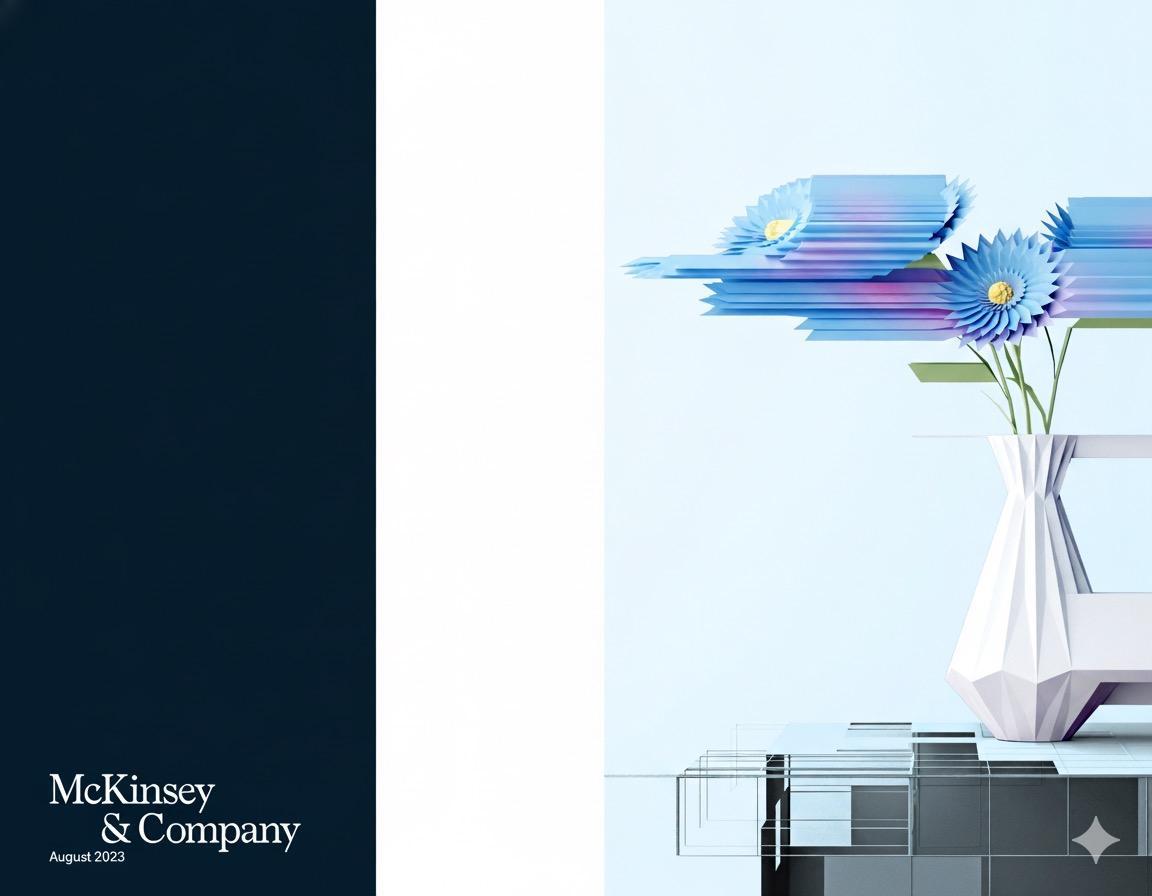}\\\includegraphics[width=3cm]{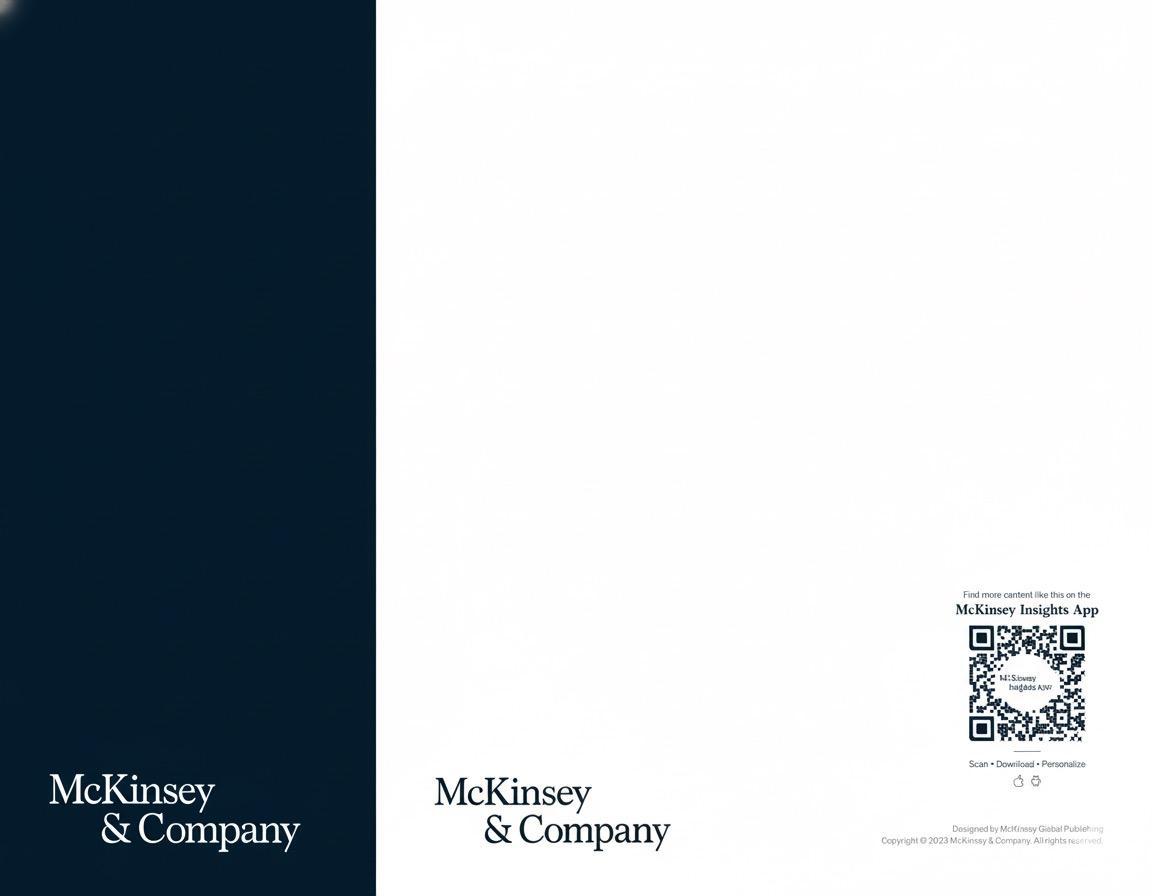}} &      \parbox{3cm}{\centering \includegraphics[width=3cm]{3_results_sections/comparision_table/images/image120.jpg}\\\includegraphics[width=3cm]{3_results_sections/comparision_table/images/image138.jpg}} \\
    \multicolumn{5}{l}{\cellcolor{subtlegray!20}\parbox{\textwidth}{\tiny\textit{\textcolor{questiongray}{26. Extract the first and last page from the document, visualize and analyze them.}}}}\\
    \arrayrulecolor{gray!30}\midrule[0.3pt]\arrayrulecolor{black}

    \parbox{3cm}{\centering \includegraphics[width=3cm]{3_results_sections/comparision_table/images/image150.jpg}} & 
    \parbox{3cm}{\centering \includegraphics[width=3cm]{3_results_sections/comparision_table/images/image150.jpg}} &      
    \parbox{3cm}{\centering\scriptsize\textcolor{gray!80}{The video you uploaded seems fairly large, so the analysis for brightness detection (to find the frame with the most fireworks) is timing out.
    To fix this efficiently, I can: Downsample the video — process fewer frames to speed up analysis. Extract every 5th or 10th second instead of every second. Then identify the timestamp where the fireworks intensity peaks.}} &      
    \parbox{3cm}{\centering\scriptsize\textcolor{gray!80}{Of course! The most spectacular highlight of the fireworks display can be found between 00:23 and 00:28. This section features a breathtaking finale where the entire skyline is engulfed in a vibrant red glow from the massive fireworks, and the Burj Khalifa is brilliantly lit with a cascading light effect from top to bottom.}} &      
    \parbox{3cm}{\centering \includegraphics[width=3cm]{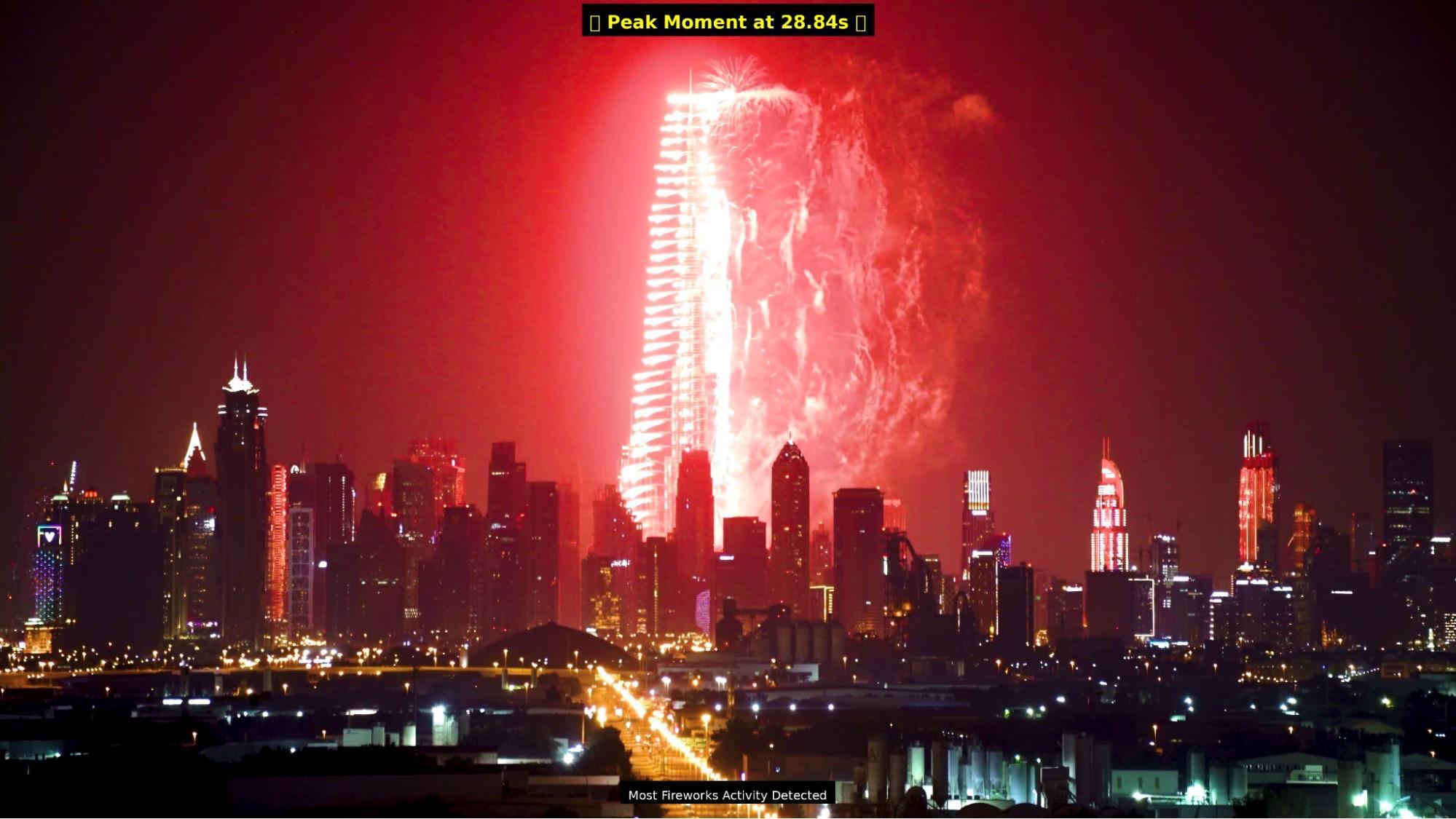} } \\
    \multicolumn{5}{l}{\cellcolor{subtlegray!20}\parbox{\textwidth}{\scriptsize\textit{\textcolor{questiongray}{27. Get the best highlight showing the most fireworks in this video.}}}}\\
    \arrayrulecolor{gray!30}\midrule[0.3pt]\arrayrulecolor{black}

    \parbox{3cm}{\centering \includegraphics[width=3cm]{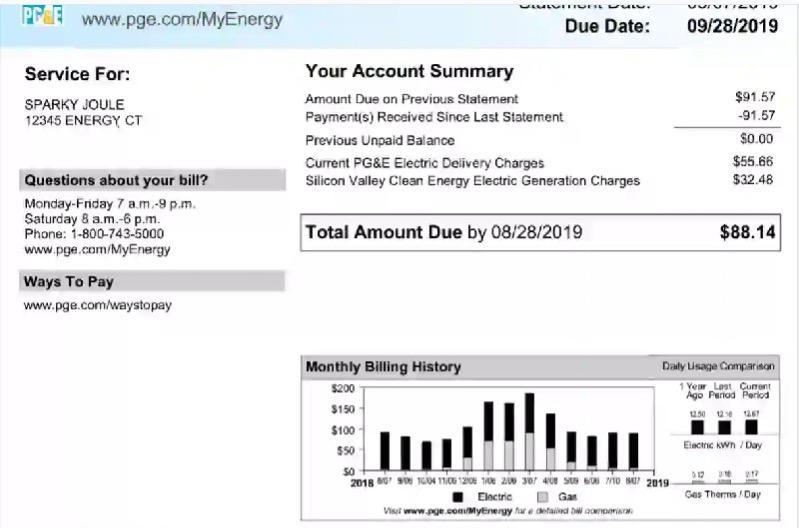}} & \parbox{3cm}{\centering \includegraphics[width=3cm]{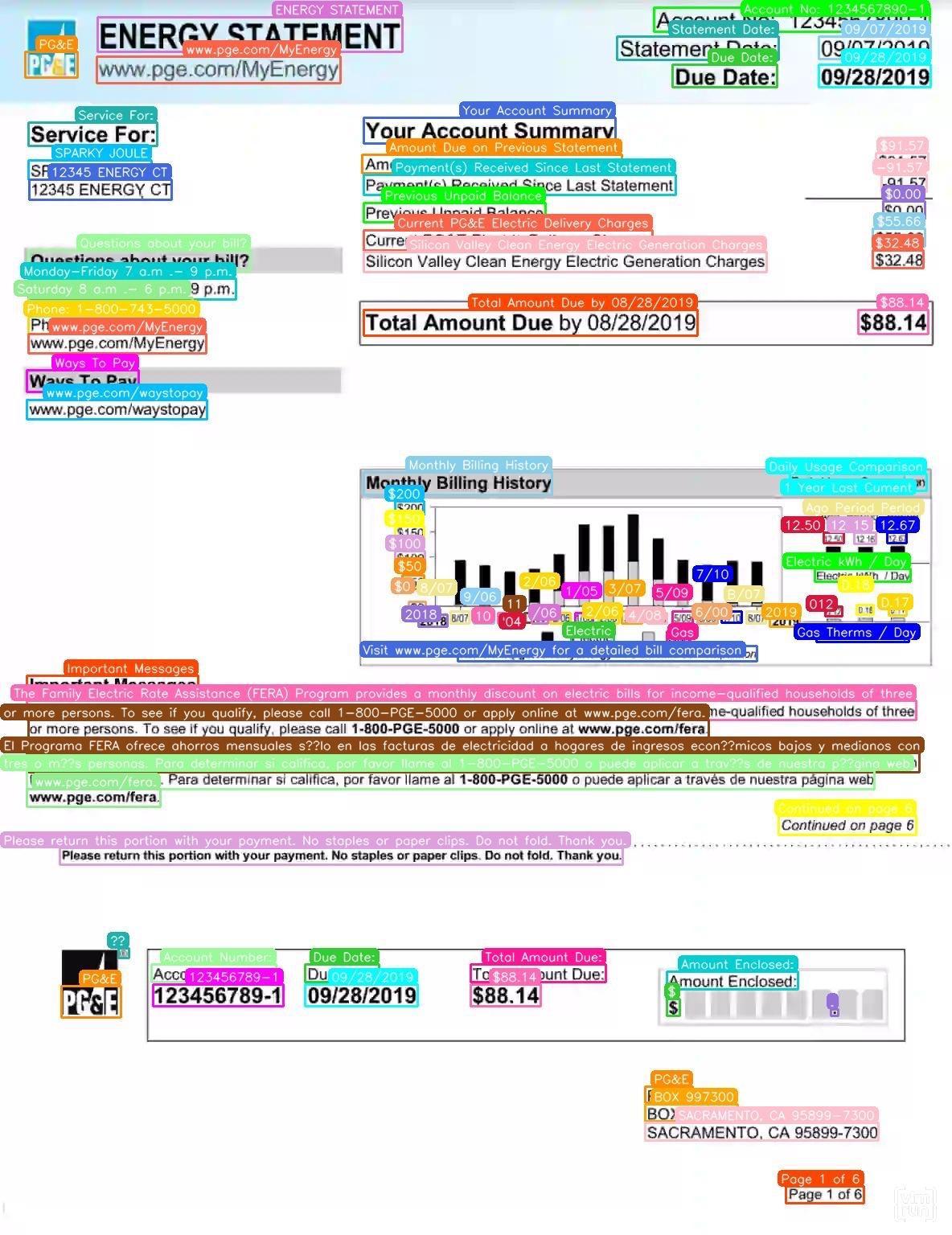}} &      \parbox{3cm}{\centering \includegraphics[width=3cm]{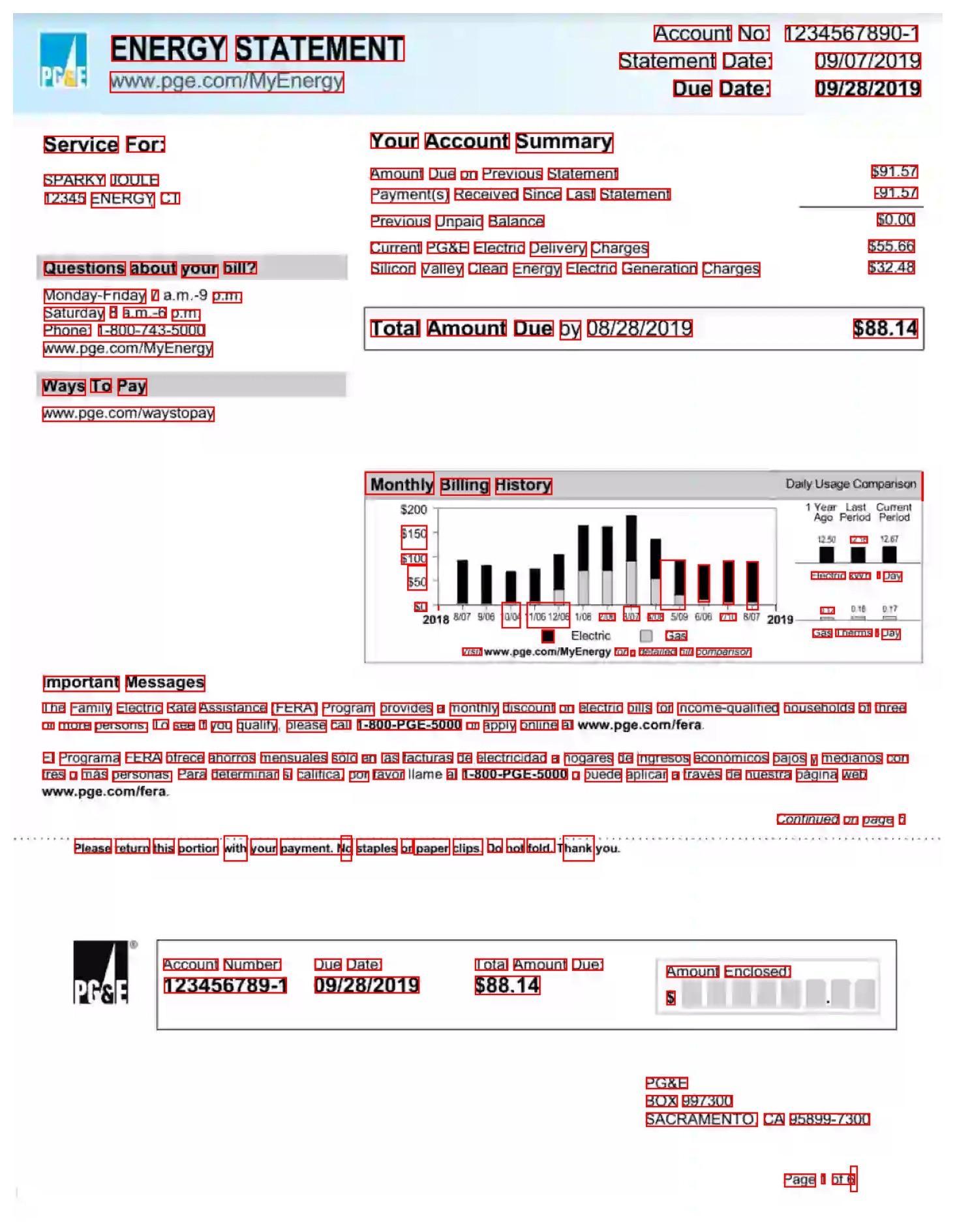}} &      \parbox{3cm}{\centering \failedimage} &      \parbox{3cm}{\centering \includegraphics[width=3cm]{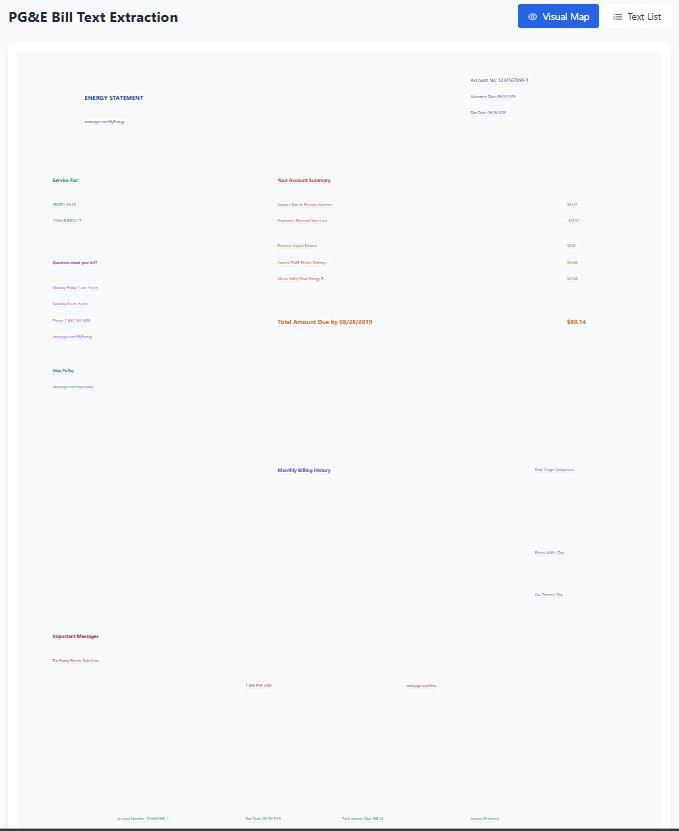}} \\
    \multicolumn{5}{l}{\cellcolor{subtlegray!20}\parbox{\textwidth}{\scriptsize\textit{\textcolor{questiongray}{28. Extract any text visible in the document, their locations and visualize them.}}}}\\
    \arrayrulecolor{gray!30}\midrule[0.3pt]\arrayrulecolor{black}

\end{longtable}
\endgroup

\newpage


\section{Discussion and Future Work}

\textbf{Orion} represents a new paradigm in visual AI: \textbf{agentic tool-augmented reasoning} that combines the flexibility of large vision-language models \cite{radford2021learning,li2023llava,li2023blip2} with the precision of specialized computer vision tools, extending the principles of tool-augmented language models \cite{schick2023toolformer,patil2023gorilla} to the visual domain.





\subsection{Limitations and Challenges}

While \textbf{Orion} demonstrates strong performance, several challenges remain:

\textbf{Tool Selection Accuracy} -- In complex scenarios with overlapping tool capabilities, the agent occasionally selects suboptimal tools. Future work will focus on learning from historical tool selection patterns, implementing cost-aware tool selection, and enabling dynamic tool specialization based on domain-specific requirements.

\textbf{Long-Horizon Planning} -- For workflows exceeding 30+ steps, success rates decline due to error accumulation across many steps, difficulty maintaining coherent state over long sequences, and limited ability to backtrack and revise early decisions.

\textbf{Computational Cost} -- While referencing multi-modal objects improve efficiency, complex workflows still require significant compute. Multiple LLM calls for planning and reflection, parallel tool execution infrastructure requirements, and the trade-off between quality and cost remain areas for optimization.






\subsection{Broader Impact}

The introduction of tool-augmented visual agents has implications beyond technical performance.

\textbf{Democratizing Computer Vision} -- \textbf{Orion} makes sophisticated computer-vision capabilities accessible to non-experts, enabling rapid prototyping of visual applications, allowing domain experts to build custom workflows without engineering teams, and reducing the barrier to entry for visual AI adoption.

\textbf{Interpretability and Trust} -- Unlike black-box models, \textbf{Orion} provides transparent execution traces showing tool usage, explicit reasoning about tool selection, and the ability to audit and verify each step, fostering greater trust and accountability.

\textbf{Ethical Considerations} -- The system's capabilities require careful consideration of privacy implications related to face recognition and tracking, potential misuse for surveillance or manipulation, bias propagation through tool selection and execution, and the need for human oversight in sensitive applications. We are committed to responsible development and deployment, working with domain experts and ethicists to establish appropriate safeguards.

\section*{Contributors}
The development of \textbf{Orion} was performed by the VLM Run Research\footnote{research@vlm.run} team – Dinesh Narapureddy and Sudeep Pillai. We gratefully acknowledge the following team members for their contributions to the development of \textbf{Orion}: Shahrear Bin Amin, Mirajul Mohin, and Dylan Snyder for the development of the frontend chat interface, including backend system development; and Lona Kiragu for contributing to dataset evaluation and scoring. Their collective efforts were essential to the development and deployment of \textbf{Orion}.

\bibliography{main}
\end{document}